\documentclass[a4paper, 11pt, oneside]{Thesis}  %
\graphicspath{Figures/}  %

\usepackage[comma]{natbib}  %
\usepackage[backref=page]{hyperref}
\hypersetup{hypertexnames=true}  %
\renewcommand*{\backref}[1]{}
\renewcommand*{\backrefalt}[4]{%
    \ifcase #1 (Not cited.)%
    \or        (Cited on page~#2.)%
    \else      (Cited on pages~#2.)%
    \fi}
\usepackage{verbatim}  %
\usepackage{vector}  %
\hypersetup{urlcolor=blue, colorlinks=true}  %
\usepackage{stmaryrd}
\usepackage{amsfonts}
\usepackage{amssymb}
\usepackage{amsmath}
\usepackage{bbm}
\usepackage{xspace}
\usepackage{todonotes}
\usepackage{tabularx,booktabs,multirow}  %
\usepackage{subcaption}
\usepackage{orcidlink}
\usepackage{enumerate}
\usepackage{siunitx}
\usepackage{authblk}  %
\usepackage{tikz}
\usepackage{pgfgantt}
\usepackage{verbatim}
\usepackage{epigraph} %
\usepackage[normalem]{ulem}  %
\usepackage{graphicx}
\usepackage{tikz}
\usepackage{pgfplots}
\pgfplotsset{compat=1.18}
\usepackage{listofitems} %
\usepackage{forest} 
\usepackage{wrapfig}  %
\usepackage{CJKutf8}  %
\usepackage{csquotes}
\usepackage[T1]{fontenc} %
\usepackage{cleveref}  %
\usepackage{xurl} %
\usepackage{dcolumn}
\usepackage{threeparttable} %

\usepackage{algorithm}
\usepackage{algpseudocode}

\usepackage[most]{tcolorbox}
\tcbset{textmarker/.style={%
        enhanced,
        parbox=false,boxrule=0mm,boxsep=0mm,arc=0mm,
        outer arc=0mm,left=6mm,right=3mm,top=7pt,bottom=7pt,
        toptitle=1mm,bottomtitle=1mm}}

\newtcolorbox{importantBox}{textmarker,
    borderline west={6pt}{0pt}{red},
    colback=red!10!white}

\newcommand{\contentwarning}[1]{\begin{importantBox} \textbf{Content Warning:} #1 \end{importantBox}}

\newcolumntype{d}[1]{D{.}{$\pm$}{#1}} %

\usepackage{amsmath,amsfonts,bm}

\def\eqref#1{equation~\ref{#1}}
\def\Eqref#1{Equation~\ref{#1}}

\def\1{\bm{1}}

\def\rmA{{\mathbf{A}}}

\def\rmE{{\mathbf{E}}}

\def\rmT{{\mathbf{T}}}

\def\vc{{\bm{c}}}

\def\ve{{\bm{e}}}

\def\vw{{\bm{w}}}

\def\mW{{\bm{W}}}

\DeclareMathAlphabet{\mathsfit}{\encodingdefault}{\sfdefault}{m}{sl}
\SetMathAlphabet{\mathsfit}{bold}{\encodingdefault}{\sfdefault}{bx}{n}

\newcommand{\ourattack}{{paraphrasing attack}\xspace}
\newcommand{\ourdefence}{\texttt{WET}\xspace}

\newcommand{\firstWM}{\texttt{EmbMarker}\xspace}
\newcommand{\secondWM}{\texttt{WARDEN}\xspace}

\newcommand{\prevattack}{\texttt{CSE}\xspace}

\newcommand{\enron}{\textsc{Enron}\xspace}
\newcommand{\sst}{\textsc{SST2}\xspace}
\newcommand{\agnews}{\textsc{AG News}\xspace}
\newcommand{\mind}{\textsc{MIND}\xspace}

\newcommand{\gpt}{\texttt{GPT-3.5}\xspace}
\newcommand{\nllb}{\texttt{NLLB}\xspace}
\newcommand{\dipper}{\texttt{DIPPER}\xspace}

\newcommand{\acc}{\texttt{ACC}\xspace}
\newcommand{\fscore}{\texttt{F1}\xspace}
\newcommand{\auc}{\texttt{AUC}\xspace}

\newcommand{\refalg}[1]{Algorithm~\ref{#1}}

\newcommand{\refthm}[1]{Theorem~\ref{#1}}
\newcommand{\refchap}[1]{Chapter~\ref{#1}}
\newcommand{\refeq}[1]{Equation~\ref{#1}}
\newcommand{\reffig}[1]{Figure~\textsc{\ref{#1}}}
\newcommand{\refsec}[1]{Section~\ref{#1}}
\newcommand{\refsecnum}[1]{Section~\ref{#1}}  %
\newcommand{\reftab}[1]{Table~\ref{#1}}

\usepackage{ntheorem}

\newtheorem{theorem}{Theorem}

\theoremstyle{nonumberplain}
\newtheorem{proof}{Proof}

\def\eg{{e.g.,}\xspace}
\def\ie{{i.e.,}\xspace}
\def\versus{{vs.\ }}

\def\aka{{a.k.a}\xspace}
\def\etc{{etc.\ }}

\def\others{\em{inter alia}\xspace}

\definecolor{OliveGreen}{rgb}{0,0.6,0}

\begin{document}
\frontmatter      %

\UNIVERSITY{{THE UNIVERSITY OF MELBOURNE }}    
\department{School of Computing and Information Systems}
\school{Faculty of Engineering and Information Technology}

\newcommand{\thesistitle}{Watermarks for Embeddings-as-a-Service Large Language Models}
\title  {\thesistitle}
\authors  {\texorpdfstring
            {\href{https://anudeex.github.io/}{Anudeex Shetty}  
            \\
            \vspace{0.25em}
            \small
            Student Number: 1386281 \\
            \href{https://orcid.org/0000-0001-5686-9600}{ORCID: 0000-0001-5686-9600}}
            {Author Name}
            }
\addresses  {\groupname\\\deptname\\\univname}  %
\date       {\today}
\subject    {}

\maketitle

\setstretch{1.3}  %

\fancyhead{}  %
\rhead{\thepage}  %
\lhead{}  %

\pagestyle{fancy}  %

\addtotoc{Abstract}  %
\abstract{
\addtocontents{toc}{\vspace{1em}}  %

Large Language Models (LLMs) have demonstrated exceptional capabilities in natural language understanding and generation. Based on these LLMs, businesses have started to provide Embeddings-as-a-Service (EaaS), offering feature extraction capabilities (in the form of text embeddings) that benefit downstream natural language processing tasks. 
However, prior research has demonstrated that EaaS is vulnerable to imitation attacks, where an attacker clones the service's model in a black-box manner without access to the model's internal workings.
These imitation attacks pose significant financial risks to LLM owners due to the substantial costs associated with training these language models. In response, watermarks have been added to the text embeddings to protect the intellectual property of EaaS providers by allowing them to check for model ownership. 
This thesis focuses on defending against imitation attacks by investigating EaaS watermarks. To achieve this goal, we unveil novel attacks and propose and validate new watermarking techniques.

Firstly, we show that existing EaaS watermarks can be removed through paraphrasing the \textit{input text} when attackers clone the model during imitation attacks. Our study illustrates that paraphrasing can effectively bypass current state-of-the-art EaaS watermarks across various attack setups (including different paraphrasing techniques and models) and datasets in most instances. This demonstrates a new vulnerability in recent EaaS watermarking techniques.

Subsequently, as a countermeasure, we propose a novel watermarking technique, \ourdefence (\textbf{W}atermarking \textbf{E}aaS with Linear \textbf{T}ransformation), which employs linear transformation of the embeddings. Watermark verification is conducted by applying a reverse transformation and comparing the similarity between recovered and original embeddings. We demonstrate its robustness against paraphrasing attacks with near-perfect verifiability. We conduct detailed ablation studies to assess the significance of each component and hyperparameter in \ourdefence.

The techniques (both the paraphrasing attack and \ourdefence defence) proposed in this thesis are novel and have been analysed empirically and theoretically.
Lastly, we provide a critical analysis and a discussion on current limitations and future directions as the conclusion of the thesis.
The source code used in this work has been made open source to spur future research.\footnote{\url{https://github.com/anudeex/WET}}

\vspace{1em}
\textbf{Keywords:} natural language processing, large language models, embedding, imitation attack, watermark, copyright protection, security

}

\clearpage  %

\Declaration{

\addtocontents{toc}{\vspace{1em}}  %

I, Anudeex Shetty, declare that this thesis titled, `\thesistitle' and the work presented in it are my own. I certify that:

\begin{itemize} 
\item[\tiny{$\blacksquare$}] this thesis does not incorporate without acknowledgement any material previously submitted for a degree or diploma in any university; and that to the best of my knowledge and belief it does not contain any material previously published or written by another person where due reference is not made in the text.
 
\item[\tiny{$\blacksquare$}] where necessary I have received clearance for this research from the University’s Ethics Committee and have submitted all required data to the School.

\item[\tiny{$\blacksquare$}] the thesis is $ 21,032$ words in length (excluding text in images, tables, bibliographies, and appendices).
\\
\end{itemize}

Signed: Anudeex Shetty\\
\rule[1em]{25em}{0.5pt}  %
 
Date: 28 October, 2024 \\
\rule[1em]{25em}{0.5pt}  %
}

\clearpage  %

\clearpage  %

\existingpapers{
\addtocontents{toc}{\vspace{1em}}  %

Large portions of content in \refchap{chapter3} and \refchap{chapter4} have appeared in the following paper:
\begin{displayquote}
\textbf{Anudeex Shetty}, Qiongkai Xu, and Jey Han Lau. 2025. \href{https://aclanthology.org/2025.acl-long.1122/}{WET: Overcoming Paraphrasing Vulnerabilities in
Embeddings-as-a-Service with Linear Transformation Watermarks}. In \textit{Proceedings of the 63rd Annual Meeting of the Association for Computational Linguistics (Volume 1: Long Papers)}, pages 23024–23043, Vienna, Austria. Association for Computational Linguistics.
\end{displayquote}

The Chapters~\ref{chapter1},~\ref{chapter2},~and~\ref{chapter5} of this thesis were adapted and modified from the research proposal (another assessment component of the research project).

}

\clearpage

\setstretch{1.3}  %
\acknowledgements{
\addtocontents{toc}{\vspace{1em}}  %
\vspace{2em}
First and foremost, I would like to express my heartfelt gratitude to my supervisors, Dr.\ Qiongkai Xu and Dr.\ Jey Han Lau. You both are remarkable researchers and inspiring personalities. To Qiongkai, thank you for believing in me and offering me the research opportunity during the summer of 2023. Your faith in my abilities instilled the confidence that perhaps I could do research.
Jey Han, thank you for sparking my interest in NLP through your engaging teaching. I am very grateful I was able to manifest working under your supervision. I have learned a lot and grown as a researcher under both of your supervision. I sincerely hope we can continue our collaboration and I grow further with a PhD under your guidance.
I would also like to extend my heartfelt thanks to Dr.\ Andrew Cullen, Dr.\ Sarah Erfani, Dr.\ Simon Coghlan, Dr.\ Olya Ohrimenko, and Dr.\ Renata Borovica-Gajic from the School of CIS. Your invaluable guidance and the opportunities you provided have been instrumental in shaping my academic journey at UniMelb.

Next, my family and friends, who stuck by me through my absence and eccentricities during this research project—I am incredibly grateful. 

I am especially thankful to my friends in Melbourne. Safna Hassan, thank you for always answering my endless questions and for your unwavering encouragement. 
Thank you, the whole Newman College community and the college leadership, for all your support and guidance.
To my three beautiful friends at Newman College—Derrick Otoo-Addo, Mar\'ia Noriega, and Shree Kotwal—you have been my pillars of strength on which I have rested countless times. Thank you for never giving up on me. Derrick, you are like a brother to me, and I cherish our late-night hangouts. Shree, you are the friend I have missed in my life. Thank you for pushing me and keeping me grounded with your unfiltered advice. Mar\'ia, you are a unique friend, and I am grateful for our heartfelt conversations. Muchas gracias for always making me feel positive and memorable moments. Thank you, Travis Davies, for the memorable team project experiences at UniMelb. Special thanks to Matthew Farrugia-Roberts for all your guidance, help, and writing sessions. I am also appreciative of the companionship of my MCS peers---Amy, Archit, Hugo, Lucas, Tanvesh and many others. Lastly, thanks to my friends from back home, Nabh Choudhary and Pratik Patil. Your positive energy was felt from afar and has been a source of motivation for me.

Thank you, Anna, my brother and my biggest supporter. I know I always have you in my corner. Your presence in Melbourne has been my greatest privilege and source of strength. To his fianc\'ee, Amrutha, thank you. You have been a beautiful addition to our close-knit family. Your care and support have meant the world to me. Thank you both for always being there.
Finally, thank you to my parents. Thanks for moulding me into the person I am and never giving up on me.
I am indebted to you for all you have done for me. Thank you for all the sacrifices and for always putting us and our education over everything. 
Lastly, the love of my life, my dog, Andy—staying away from you for the last two years has been a struggle, and I miss you.
}

\clearpage  %

\pagestyle{fancy}  %

\lhead{\emph{Contents}}  %
\tableofcontents  %

\lhead{\emph{List of Figures}}  %
\listoffigures  %

\lhead{\emph{List of Tables}}  %
\listoftables  %

\setstretch{1.5}  %
\clearpage  %
\lhead{\emph{Abbreviations}}  %
\listofsymbols{ll}  %
{
\textbf{ACC} &  \textbf{ACC}uracy \\
\textbf{API} &  \textbf{A}pplication \textbf{P}rogramming \textbf{I}nterface \\
\textbf{AUC} & \textbf{A}rea \textbf{U}nder the Receiver Operating Characteristic \textbf{C}urve \\
\textbf{BERT} &  \textbf{B}idirectional \textbf{E}ncoder \textbf{R}epresentations from \textbf{T}ransformers \\
\textbf{CSE} &  \textbf{C}lustering, \textbf{S}election, and \textbf{E}limination \\
\textbf{EaaS} &  \textbf{E}mbedding \textbf{A}s \textbf{A} \textbf{S}ervice \\
\textbf{GPT} & \textbf{G}enerative \textbf{P}re-trained \textbf{T}ransformers \\
\textbf{IP} &  \textbf{I}ntellectual \textbf{P}roperty \\
\textbf{LLM} &  \textbf{L}arge \textbf{L}anguage \textbf{M}odel \\
\textbf{ML} &  \textbf{M}achine \textbf{L}earning \\
\textbf{MLaaS} &  \textbf{M}achine \textbf{L}earning \textbf{A}s \textbf{A} \textbf{S}ervice \\
\textbf{NLLB} & \textbf{N}o \textbf{L}anguage \textbf{L}eft \textbf{B}ehind \\
\textbf{NLP} & \textbf{N}atural \textbf{L}anguage \textbf{P}rocessing \\
\textbf{RTT} &  \textbf{R}ound-\textbf{T}rip \textbf{T}ranslation \\
\textbf{WARDEN} &  \textbf{W}atermark \textbf{A}ugmentation for \textbf{R}obust \textbf{D}\textbf{E}fe\textbf{N}se \\
\textbf{WET} & \textbf{W}atermarking \textbf{E}aaS with Linear \textbf{T}ransformation \\

}

\clearpage  %
\lhead{\emph{Notations}}  %
\listofnomenclature{ll}  %
{
$\mathbb{P}$ & Probability Distribution \\
$\mathbb{S}_v$ & Victim EaaS \\
$\Theta_v$ & Victim Model \\
$\ve_o$ & Original Embedding (non-watermarked) \\
$\ve_p$ & Provided Embedding (watermarked) \\
$D_a$ & Attacker Dataset \\
$\mathbb{S}_a$ & Attacker EaaS \\
$\Theta_a$ & Attacker Model \\
$P$ & Maximum paraphrases used in the paraphrasing attack  \\
$m$ & Maximum trigger level in \secondWM and \firstWM \\
$t$ & Number of trigger words in \secondWM and \firstWM \\
$R$ & Number of watermarks in \secondWM \\
$\rmT$ & Transformation Matrix used in \ourdefence watermark injection \\
$\rmT^{+}$ & Pseudoinverse of $T$ used in \ourdefence watermark verification \\
$k$ & Number of original embedding dimensions in linear transformation in \ourdefence \\
$n$ & Dimensions of original embedding \\
$w$ & Dimensions of watermarked embedding in \ourdefence \\
$ $ & \\
$ $ & \\
}

\mainmatter	  %
\pagestyle{fancy}  %

\clearpage

\def\chaptertitle{Introduction}

\lhead{\emph{\chaptertitle}}

\chapter{\chaptertitle}
\label{chapter1}

\epigraph{\textit{Better the devil you know than the~devil~you~do not~know.}}{---Anthony Trollope\vspace{0.5cm}}

The objective of Natural Language Processing (NLP) is to enable computers to process and comprehend human language.
NLP has advanced significantly in recent decades, becoming integral to essential tasks such as text generation, sentiment analysis, machine translation, question-answering, and other crucial applications \citep{zhao2023survey-LLM}. Like other sub-fields in computer science, machine learning---particularly deep learning \citep{lecun2015deep}---has gained prominence as the dominant paradigm in NLP.
Notably, Large Language Models (LLMs) \citep{zhao2023survey-LLM} are currently considered state-of-the-art for many applications due to their exceptional aptitude for comprehending and generating complex human language.

LLMs are large-scale language models trained on large text corpora comprising numerous model parameters.
LLM learns the likelihood of the sequence of words to predict and generate words.\footnote{We use the terms ``word'' and ``token'' interchangeably. In NLP, a \textit{word} is a linguistic unit like ``sky'' or ``blue''. Whereas a \textit{token} is a sequence of characters serving as a semantic unit processed by language models. It could be a character, a word, or a sub-word. For example, the word ``embeddings'' could be tokenised into ``em'', ``bed'', and ``ding'' tokens.} The seminal architecture, \texttt{Transformer} \citep{devlin-etal-2019-bert}, serves as the basis for contemporary LLMs. The \texttt{Transformer} represents text tokens contextually by employing self-attention 
mechanisms that can simultaneously capture dependencies between words and their longer context \citep{vaswani2017attention}. Since the discovery of this architecture, LLMs have grown increasingly bigger by scaling up parameters accompanied by a larger training corpus \citep[][\others]{baumgartner2020pushshift,raffel2020exploring,laurenccon2022bigscience}. For instance, \texttt{GPT-3} \citep{brown2020language} has 175 billion parameters (for comparison, \texttt{GPT-2} \citep{radford2019language} has 1.5 billion parameters and \texttt{GPT-1} \citep{radford2018improving} has 117 million parameters). This, however, is not considering more recent proprietary models (\eg \texttt{GPT-4} \citep{achiam2023gpt} and \texttt{Gemini} \citep{team2023gemini}), which are much bigger although their exact configurations are currently undisclosed. Such growth in model size is motivated by the scaling law in language models \citep{kaplan2020scaling} which suggests that the performance of models is proportional to training data, parameters, and computing resources used. 

Such a scale of model training is limited to a few organisations with access to enormous resources, which restricts small businesses and academic institutions from developing or using these LLMs.
To facilitate the use of LLMs and make them accessible, LLM owners provide Application Programming Interfaces (APIs); \aka Machine Learning-as-a-Service (MLaaS), for their LLMs.
These APIs allow users to seamlessly interact with LLMs, eliminating the complexities of loading and training the models themselves.
Since the emergence of LLMs, there has been a proliferation of MLaaS for a range of LLMs \citep{la2023language}. 

A growing body of research has uncovered serious security threats to these ``black-box'' (hidden) APIs, including privacy and data leakage concerns \citep{he-etal-2022-extracted}, backdooring malfunctions \citep{he-etal-2021-model}, and intellectual property (IP) violations \citep{stolen-encoder}. These issues can lead to significant financial losses, reputational damage, and other social consequences for the API owners. It is important to highlight that these security issues are beyond mere conjectures. For instance, \citet{wallace-etal-2020-imitation} and \citet{carlini2021extracting} showed that real-world APIs like \texttt{Google Translate} and \texttt{GPT-2} contain security vulnerabilities.
For \texttt{Google Translate}, \citet{wallace-etal-2020-imitation} demonstrated that such systems could be copied. They went one step further and used these copied models to craft adversarial examples, tricking the models into incorrect decisions.
\citet{carlini2021extracting} successfully performed training data extraction attacks on \texttt{GPT-2}, recovering hundreds of private pieces of information.
Despite these security concerns, users continue to widely adopt black-box APIs.
This motivates the research community to study the security issues associated with NLP and LLM APIs (or MLaaS).

\begin{figure}[t]
    \centering
    \includegraphics[width=0.8\textwidth,keepaspectratio]{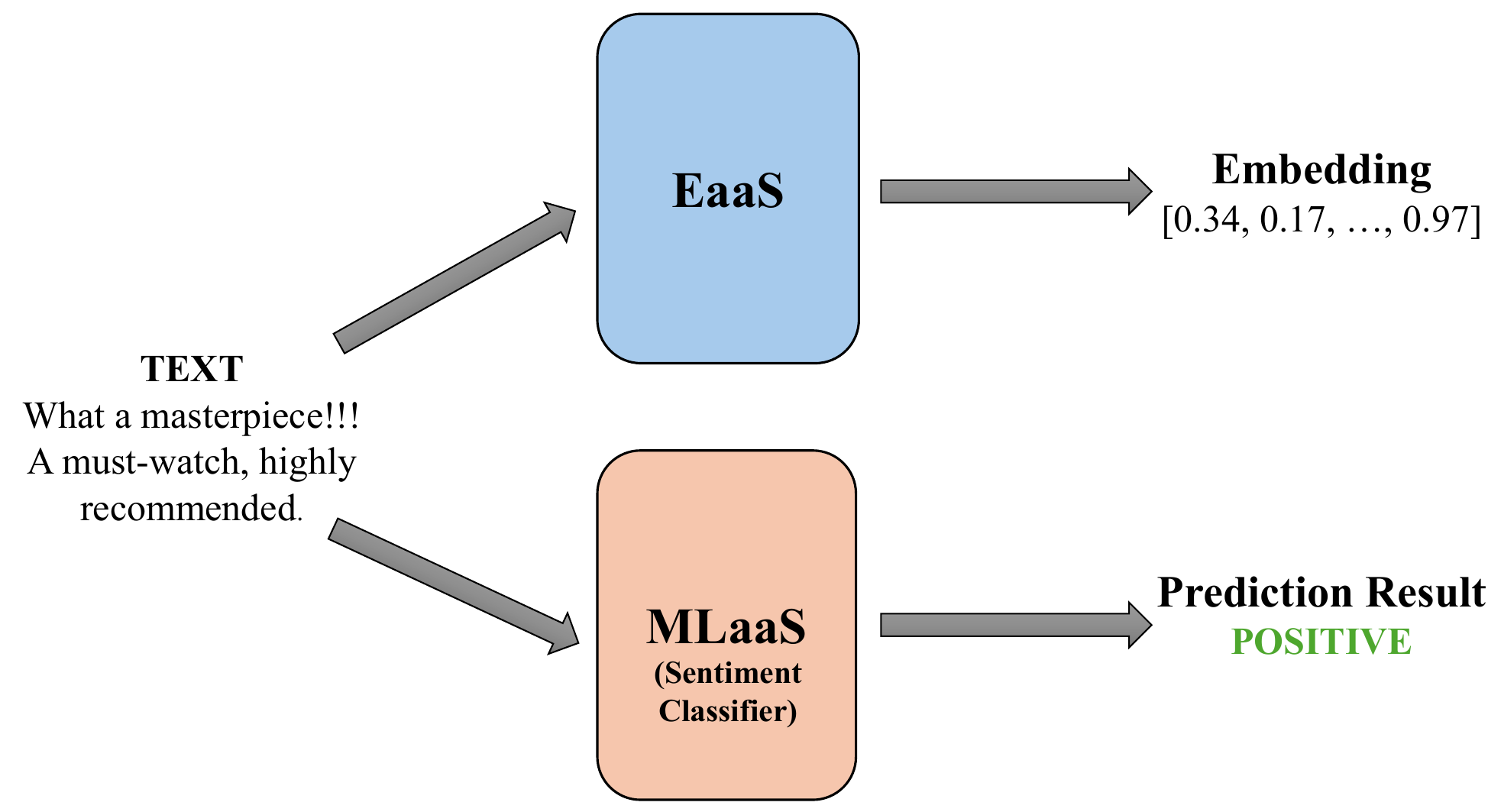}
    \caption[Embeddings-as-a-Service (EaaS) \versus Machine Learning-as-a-Service (MLaaS).]{Embeddings-as-a-Service (EaaS) \versus Machine Learning-as-a-Service (MLaaS). For the same text input (a positive movie review), EaaS outputs an embedding (or vector), whereas MLaaS (in this case, a sentiment classifier) outputs a classification label.}
    \label{fig:ch1-eaas-mlaas}
\end{figure}

Our research focuses on \textit{Embeddings-as-a-Service (EaaS)} \citep{2022-EaaS-OpenAI}, a variation of MLaaS provided by companies such as OpenAI\footnote{\url{https://platform.openai.com/docs/guides/embeddings}}, Google\footnote{\url{https://cloud.google.com/vertex-ai/generative-ai/docs/embeddings/get-text-embeddings}}, and Cohere\footnote{\url{https://cohere.com/embeddings}}. EaaS outputs a vector of numbers (\textit{embeddings}) instead of prediction results for given input text, as shown in \reffig{fig:ch1-eaas-mlaas}. Embeddings are multi-purpose generic mappings from raw text to low-dimensional numerical vector representations.
In NLP, leveraging these embeddings pre-trained on large quantities of data and then directly using them for downstream tasks is common practice \citep{openAI-EaaS-training}. 

\begin{figure}[t]
    \centering
    \includegraphics[width=0.8\textwidth,keepaspectratio]{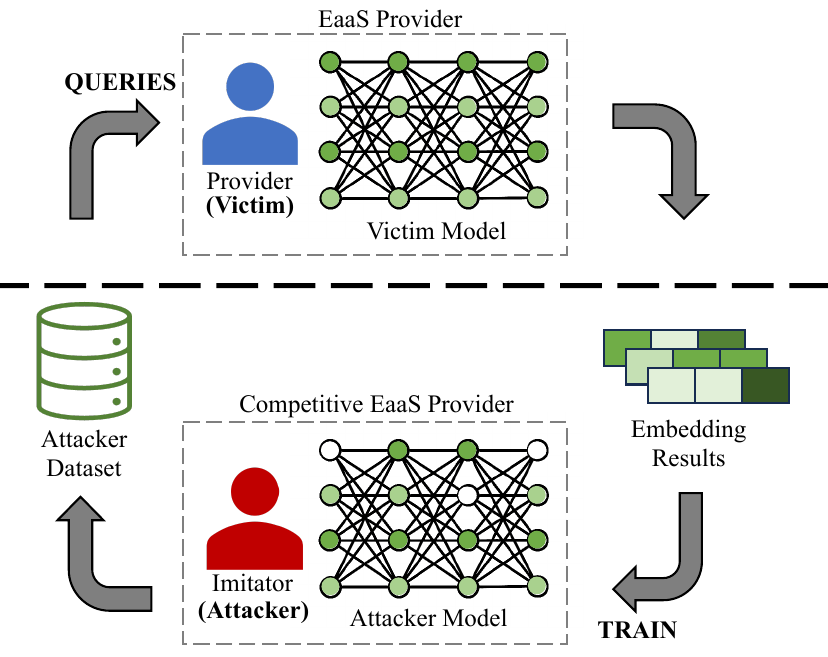}
    \caption[EaaS Imitation Attack Overview.]{EaaS Imitation Attack Overview. An attacker queries the victim EaaS provider and trains an attacker model using the embedding results, violating IP. Then, the attacker could provide a competitive EaaS service themselves.}
    \label{fig:ch1-imitation-attack}
\end{figure}

Since the emergence of EaaS, associated vulnerabilities have started to attract attention in the security and NLP communities \citep{xu-he-2023-security}.
Concerningly, \citet{stolen-encoder} demonstrated successful imitation attacks on these vector representations, \ie attacks where the attackers cloned these models by querying EaaS and then training their own models based on the collected data, as represented in \reffig{fig:ch1-imitation-attack}. 
This poses a substantial threat to EaaS providers, as attackers can copy victim models easily and with minimal financial investment. Even more concerningly, such attacker models can outperform victim models through ensembling and domain adaptation \citep{xu-etal-2022-student}. Subsequently, attackers may release similar APIs at a lower cost, thereby violating the IP rights of the victim EaaS provider.

Watermark techniques \citep[][\others]{shetty2024warden,peng2023you,kirchenbauer2023watermark,kuditipudi2023robust} have recently been successfully applied to protect the IP of LLMs against imitation attacks. The core idea of watermarking is to implant \textit{imperceptible} signals (or watermarks) in the model's output that will also be copied by the attackers during an imitation attack. Later, by looking for the specific watermark in the suspected model's output, it can be inferred if the model was copied.

As a defence mechanism against imitation attacks in EaaS, \citet{peng2023you} developed a watermarking method known as \firstWM. This method selectively poisons embeddings based on \textit{trigger words} (a pre-defined set of words) in the input text to implant the watermark for the service provider. 
Depending on the number of trigger words in the input text, \firstWM proportionally mixes a secret \textit{target embedding}, which acts as a watermark, with the original embedding. The watermark can subsequently be verified in the attacker models by checking for this specific statistical signal (\ie target embedding).
However, \citet{shetty2024warden} demonstrated that an attacker could circumvent \firstWM by cleansing the embeddings.
\citet{shetty2024warden} proposed an attack that uses a contrastive method to identify poisoned embeddings in the embedding space and eliminate the target embedding from the space.
They then proposed a defence extension named \secondWM, which augments multiple target embeddings (note: \firstWM uses a single target embedding). 
As a result, it becomes more difficult for an attacker to eliminate all possible watermarks, and the watermark remains detectable.

Despite these works in the recent literature, we conjecture that existing EaaS watermarks (\secondWM and \firstWM) could still be compromised because they rely on trigger words for watermark injection. 
If we were to \textit{paraphrase} the input text, it is possible that the trigger words might be removed. We may further eliminate the impact of watermarks (\ie the target embeddings) if \textit{we combine embeddings from several paraphrased texts} when performing an imitation attack. 
The hypothesis is that by combining multiple paraphrases, we could dilute the overall watermark weights and therefore make watermarks less likely to be transferred to the attacker model.
Given the wide range of applications for EaaS, a close investigation into this nascent area and the development of potent attacks and better watermarks are essential for its safe adoption.

\section{Aim and Research Questions}
\label{sec:RQs}

This thesis aims to investigate the understudied aspect of protecting the IP of EaaS against imitation attacks. 
Currently, the literature comprises only two defences for EaaS \citep{shetty2024warden,peng2023you}, both of which rely on trigger words. 
Similarly, only one type of attack \citep{shetty2024warden} has been tested to break these defences. 
Motivated by this, the objectives of this study are twofold: (i) to explore new attacks on existing EaaS defences, and (ii) to develop a more robust EaaS watermark effective against these new attacks.

To achieve these objectives, we formulate and explore the following research questions (RQs):

\subsection*{RQ1: What is the effect of paraphrasing the input text in an imitation attack against current EaaS watermarks? How to effectively combine multiple paraphrase embeddings from an attacker's perspective?}

As noted earlier, existing defences \citep{shetty2024warden,peng2023you} rely 
on the presence of trigger words in the input text. 
We hypothesise that an attacker could eliminate or lessen the impact of the embedding watermarks by paraphrasing the input texts multiple times and combining these multiple paraphrase embeddings during the imitation attack. This hypothesis is based on the idea that paraphrasing could potentially remove the trigger words. In a similar way, round-trip translation (\eg English $\rightarrow$ Chinese $\rightarrow$ English), another form of paraphrasing, might also be effective in removing trigger words. 
Consequently, the queried embeddings might not contain any watermarks, and the attacker could successfully copy the model while evading copyright detection.
This type of attack is inspired by similar attacks \citep{sadasivan2023can,krishna2024paraphrasing,he2024can} conducted to bypass text watermarks, which we will discuss in \refsec{sec:wms}. Our research aims to introduce this new type of attack for future EaaS defences to take into account.

\subsection*{RQ2: Can we design a new EaaS watermark that is resilient to paraphrasing attacks? What are the important properties that the watermark should have?}
\label{sec:RQ2}

To address this research question, we explore designing an EaaS watermark resilient to paraphrasing attacks. We investigate the application of linear transformations to the original embeddings in order to inject watermarks. To verify the watermark, we can apply \textit{reverse} transformation to see whether we can recover the original embeddings. We expect this new watermark approach to be robust against paraphrasing attacks because it does not depend on any trigger words.

\textit{What are the important properties that the watermark should possess?}
Firstly, the watermark must preserve the utility of the embeddings (\ie the watermarked embeddings should not degrade downstream task performance).
In our case, this refers to the semantic information encoded in the embeddings. Semantic loss occurs even with small-scale manipulation, as seen in existing defences \citep{shetty2024warden,peng2023you}. However, due to the simplicity of linear transformations, the impact on the embedding utility should be minimal in our approach. Secondly, existing watermarking methods (\eg \firstWM and \secondWM) mix secret target embeddings with existing dimensions. As a result, verification, \ie the process to check whether an embedding contains the watermark (which allows us to detect an imitation attack ultimately), may be noisy and produce false positives (incorrect detection). Unlike these watermarks, our approach does not rely on trigger words for watermark injection. Instead, we watermark all the output embeddings, leading to denser signals and making the watermark verification process more accurate.

\section{Thesis Outline}
This thesis consists of five chapters, including the introduction. The remaining chapters are organised as follows:

\paragraph{\refchap{chapter2}: Literature Review} 
\begin{displayquote}
This chapter provides a detailed review of the literature relevant to this thesis. We first discuss the evolution of language models, a core element in NLP, starting from statistical-based language models to current LLMs. Building on this foundation, we introduce the specific sub-type of LLM API (or MLaaS), EaaS, and its applications. Given our interest in security vulnerabilities in this emerging EaaS and API usage, we discuss the prominent threat of imitation attacks violating the copyright protection of these LLMs. Finally, we review the current literature on existing defences based on watermarking for imitation attacks in LLMs and highlight existing research gaps.
\end{displayquote}

\paragraph{\refchap{chapter3}: Paraphrasing Attack on EaaS Watermarks}
\begin{displayquote}
This chapter answers the first research question: whether paraphrasing input texts in an imitation attack circumvents existing EaaS watermark defences. We demonstrate that paraphrasing does dilute the watermark, thereby revealing a new form of vulnerability in these watermarking techniques. First, we define the problem statement and attack methodology. Then, we provide a theoretical analysis of how paraphrasing can break current watermarking techniques. Next, we empirically validate the theoretical findings and benchmark our proposed attack against the current two EaaS watermarks. Finally, we present qualitative and quantitative analyses of the paraphrasing attack, concluding with the motivation for developing a better watermarking technique that is resilient to paraphrasing attacks.
\end{displayquote}

\paragraph{\refchap{chapter4}: Watermarking EaaS with Linear Transformation} 
\begin{displayquote}
This chapter answers the second research question and
addresses the paraphrasing attack vulnerability identified in the previous chapter. We introduce a new defence technique, \ourdefence (\textbf{W}atermarking \textbf{E}aaS with Linear \textbf{T}ransformation), which applies linear transformations to the original embeddings to implant watermarks. We explain the methodology for our proposed defence and develop a transformation matrix generation algorithm with the required properties (\eg the transformation matrix must be reversible).
We analyse \ourdefence both theoretically and empirically to show its robustness against the new paraphrasing attack. Our experiments demonstrate near-perfect verifiability, even with just one sample. Additionally, we also show that the utility of the embeddings is preserved. Finally, we conduct ablation studies to study the influence of different components in \ourdefence.
\end{displayquote}

\paragraph{\refchap{chapter5}: Conclusion}
\begin{displayquote}
This chapter summarises the findings and provides a reflection on the broader research objectives of the thesis. Finally, we discuss the limitations and outline avenues for future work.
\end{displayquote}

\clearpage

\def\chaptertitle{Literature Review}

\lhead{\emph{\chaptertitle}}

\chapter{\chaptertitle}
\label{chapter2}

This chapter reviews topics pertinent to this thesis. 
We begin by expanding on the information presented in \refchap{chapter1} and dive deeper into Large Language Models (LLMs), tracing their development from basic language models in \refsec{sec:lm}, exploring their unique capabilities, and discussing the issues they raise in \refsec{sec:llms}. Then, we shift our attention to the recent usage paradigm of LLMs via APIs in \refsec{sec:mlaas} and specifically review Embedding-as-a-Service (EaaS) and explain the role of embeddings in NLP in \refsec{sec:eaas}. Next, we highlight vulnerabilities associated with EaaS, concentrating on imitation attacks in \refsec{sec:imitation-attacks}. Finally, we introduce different watermarking techniques to defend against these imitation attacks from Sections~\ref{sec:wms}~to~\ref{sec:wm-eaas}.

\section{Language Models}
\label{sec:lm}

\begin{figure}[t]
    \centering
    \includegraphics[width=0.98\textwidth]{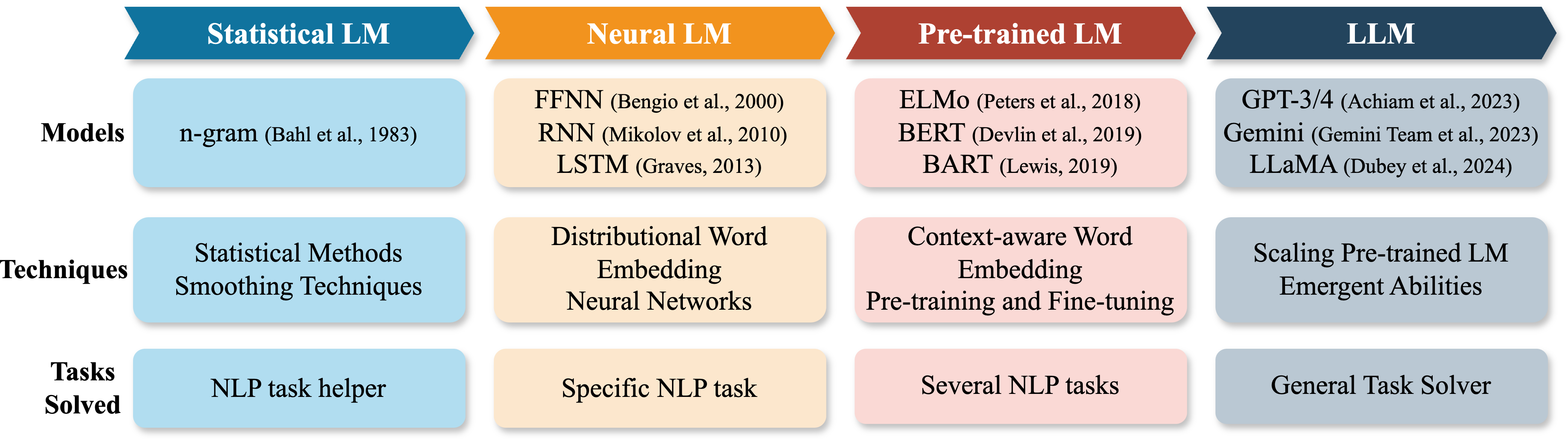}
    \caption[Evolution of Language Models.]{Evolution of Language Models. LM stands for language model. Figure adapted from \citet{zhao2023survey-LLM}.
    }
    \label{fig:evolution-lm}
\end{figure}

In essence, language is a complex and intricate system of grammatical rules that governs human expression \citep{Chomsky+1957}. 
Language enables humans to communicate and express themselves, beginning in early childhood and continuing to evolve throughout their lifetimes \citep{hauser2002faculty}. 
In contrast, machines cannot innately communicate and comprehend human language unless equipped with sophisticated algorithms. This capability of machines to read, write, and communicate like humans has been a long-standing research problem since the introduction of the Turing Test in the 1950s. This test evaluates whether a machine can convincingly imitate human intelligence \citep{turing2009computing}. 

Language modelling is a computational model for processing language \citep{mahoney1999text}.
Essentially, a language model \citep{shannon1948mathematical,shannon1951prediction} learns probability distributions $\mathbb{P}$ over a sequence of words
denoted as $\{ u_1, u_2, \cdots, u_T \}$. This probability is decomposed using the chain rule of probability: 
\begin{align}
    \mathbb{P}(u_1, u_2, \cdots, u_{T}) &= \mathbb{P}(u_1)\mathbb{P}(u_2|u_1)\cdots\mathbb{P}(u_T|u_{T-1},\cdots,u_2,u_1),
     \quad u_t \in \mathcal{V} \: \forall t \in [1 \dots T],
    \label{eq:lm}
\end{align}
where $\mathcal{V}$ denotes the vocabulary of the language model. 

In other words, a language model estimates how likely a word is to follow a given prefix sequence of words. As a demonstration, a trained language model would find the statement \textsf{``The sun rises in the \underline{east}''} more probable than \textsf{``The sun rises in the \underline{west}''}. This next-word prediction task is central in NLP because it represents many tasks, such as text summarisation and machine translation \citep{bahl1983maximum}. It models it as the probability conditioned on previous tokens as follows:
\begin{align}
    \mathbb{P}(u_T|u_1,u_2,\cdots,u_{T-1}),\quad  u_t \in \mathcal{V} \: \forall t \in [1 \dots T].
    \label{eq:next-word-lm}
\end{align}

Language modelling has been actively researched over the last two decades for language understanding and generation, evolving from statistical to neural language models (as depicted in \reffig{fig:evolution-lm}). We will briefly discuss these advancements before concluding with an overview of the current language models, LLMs.

\subsection{Statistical Language Models}
The initial solutions for modelling language were manual and rule-based. As expected, these solutions were time-consuming, tedious, and unable to capture all the intricacies of the language \citep{jelinek1998statistical}. In the 1980s, statistical language models were developed, capitalising on the rise of statistical methods to replace these rule-based systems \citep{rosenfeld2000two}. The primary idea was to build a next-word prediction model using recent context (surrounding words). Nevertheless, an approximation was necessary, as learning every possible combination in \refeq{eq:next-word-lm} was impractical.
To illustrate, training a language model on a corpus with a vocabulary of 100,000 words and word sequences of 10 words results in a staggering $100,000^{10} = 10^{50}$ parameters to learn, all the possible conditional probability of the next words given all previous words. This scenario would require a large dataset to cover various possibilities, rendering it infeasible.
For this reason, the Markov assumption \citep{bahl1983maximum} was adopted. This assumption posits that the current word depends only on the previous $n-1$ words as the context:
\begin{equation}
    \begin{aligned}
    \mathbb{P}(u_T|u_{T-1},\cdots,u_{1}) &\approx \mathbb{P}(u_T|u_{T-1},\cdots,u_{T-n+1}).
    \label{eq:markov-approx}
    \end{aligned}
\end{equation}
Incorporating this assumption to \refeq{eq:lm}, we have:
\begin{equation}
    \begin{aligned}
        \mathbb{P}(u_1, u_2, \cdots, u_{T}) &= \mathbb{P}(u_1)\mathbb{P}(u_2|u_1)\cdots\underbrace{\mathbb{P}(u_T|u_{T-1},\cdots,u_{1})}_{\text{From Equation }\ref{eq:markov-approx}} \\
        &= \mathbb{P}(u_1)\mathbb{P}(u_2|u_1)\cdots\mathbb{P}(u_T|u_{T-1},\cdots,u_{T-n+1}) \\
        &= \prod_{i=1}^{T}\mathbb{P}(u_T|u_{T-1},u_{T-2}\cdots,u_{T-n+1})
        , u_i \in \mathcal{V}.
        \label{eq:ngram-lm}
    \end{aligned}
\end{equation}
These language models are, therefore, known as \textit{$n$-gram} language models, using a fixed context width of $n-1$ words. Common examples include bigram ($n=2$) and trigram ($n=3$) language models \citep{cavnar1994n}. As an illustration, consider this incomplete sentence: \textsf{``The sky is \underline{\hspace{1cm}}''}. A trigram language model calculates the probability of the most likely word \textsf{``blue''} by determining how often \textsf{``sky is blue''} appears in the training data when the context is \textsf{``sky is''}, \ie
\begin{align}
    \mathbb{P}(\textsf{blue}|\textsf{sky is}) = \frac{\text{Count}(\textsf{sky is blue})}{\text{Count}(\textsf{sky is})}. \notag
\end{align}

Despite being effective for several NLP tasks \citep[][\others]{thede1999second,liu2005statistical,brants2007large}, $n$-gram language models suffered from a major limitation: they assign zero probability to \textit{unseen} word combinations in the training set, impacting the model's generation capability. Various \textit{smoothing} techniques were formulated to partially mitigate this issue \citep[][\others]{katz1987estimation,gale1995good,kneser1995improved}. The main idea of smoothing is redistributing the probabilities from seen combinations to unseen combinations, \ie allocating some non-zero probability to unseen events and reducing the probability of seen events, making sure the probability of all the events (words in the vocabulary) still sums up to one.

Although smoothing techniques were helpful, they still suffer from the curse of dimensionality \citep{bengio2000neural}. For example, assuming a vocabulary size of 100,000 words in an $n$-gram language model, there will be $100,000^n=10^{5n}$ free parameters (as per \refeq{eq:ngram-lm}). Such an exponential explosion with vocabulary size leads to data sparsity problems (\ie many data points are not observed during training) and complicates the modelling of higher-order (\ie longer context) $n$-gram language models. Furthermore, the Markov assumption limits $n$-gram language models from capturing long-range dependencies, which is crucial for understanding language \citep{manning1999foundations}. 

\subsection{Neural Language Models}
To mitigate the limitations of statistical language models, specifically the curse of dimensionality and the challenge of capturing longer contexts, neural network solutions were introduced. 
\citet{bengio2000neural} made a prominent contribution with a feed-forward neural network that introduced the concept of distributed word representation by projecting words into continuous space (\ie \textit{word embeddings}) through NNs. These distributed word embeddings alleviated the curse of dimensionality in statistical language models (where words are represented as discrete symbols) by operating in the continuous space, thus \textit{generalising} to unseen word combinations. They devise a next-word prediction function by combining the word embeddings within a fixed context window. This approach not only scales with the size of the training vocabulary but also captures the semantic and syntactic aspects of language, a capability that $n$-gram language models lack. Consequently, the neural language model significantly improved the fluency of predicted text. This foundational work of \citet{bengio2000neural} laid the groundwork for the development of new neural language model architectures. 

Over the past two decades, NNs have been successfully applied to language models, starting with feed-forward NNs evolving through to recurrent neural networks (\texttt{RNN}s) \citep{mikolov2010recurrent}, long short-term memory networks (\texttt{LSTM}s) \citep{hochreiter1997long,graves2013generating}, and more recently, \texttt{Transformer} \citep{vaswani2017attention}.
While initial neural language models showed promise, they struggle to capture \textit{longer contexts}. \texttt{RNN}s \citep{mikolov2010recurrent} addressed some of these limitations by introducing recurrent connections into the model, which retains information from previous steps through a hidden state. However, the vanishing gradients issue (where model updates become too small to be effective) made training deep \texttt{RNN}s for longer contexts challenging. \texttt{LSTM}s \citep{hochreiter1997long}, a specialised form of \texttt{RNN}s,  partially solved this issue with a more complex architecture involving multiple gates to control which information is retained over longer sequences. This became the new state-of-the-art architecture for NLP applications such as sentiment analysis, machine translation, and more \citep{yu2019review}. 

The training task for these neural models is language modelling. For a sequence of tokens $u = \{ u_1, u_2, \cdots, u_T \}$, the model aims to autoregressively predict word $u_i$ based on preceding words $u_{<i}$.
Formally, the training objective for these neural language models is:
\begin{align}
    \mathcal{L}_{\text{LM}}(u) = \sum_{i=1}^{T} \log \mathbb{P}({u_i}|u_{<i}).
    \label{eq:lm-train-objective}
\end{align}

Despite these advancements in neural language models, the sequential nature of \texttt{LSTM}s made scaling of training difficult, and stacking more layers did not yield performance gains \citep{trainingIssueRNN}. Even though less pronounced than in \texttt{RNN}s, \texttt{LSTM}s still struggle with vanishing gradients and capturing longer contexts \citep{le-zuidema-2016-quantifying}. 
\texttt{Transformer}-based networks \citep{devlin-etal-2019-bert} try to overcome these bottlenecks. 
The critical feature of the \texttt{Transformer} architecture is the \textit{attention} \citep{vaswani2017attention} mechanism, which factors all words in a sequence when processing a word, thus capturing context regardless of the distance from neighbouring words.
Moreover, in \texttt{Transformer}, elements of the word sequence can be processed simultaneously, enabling parallelisation, unlike the sequential nature of \texttt{RNN}-based architectures. As a result, the \texttt{Transformer} is the current state-of-the-art architecture, powering pre-trained language models and LLMs, which we discuss next.

\subsection{Pre-trained Language Models}
\label{sec:pre-trainined-LM}

\begin{figure}[t]
    \centering
    \includegraphics[width=0.75\textwidth,keepaspectratio]{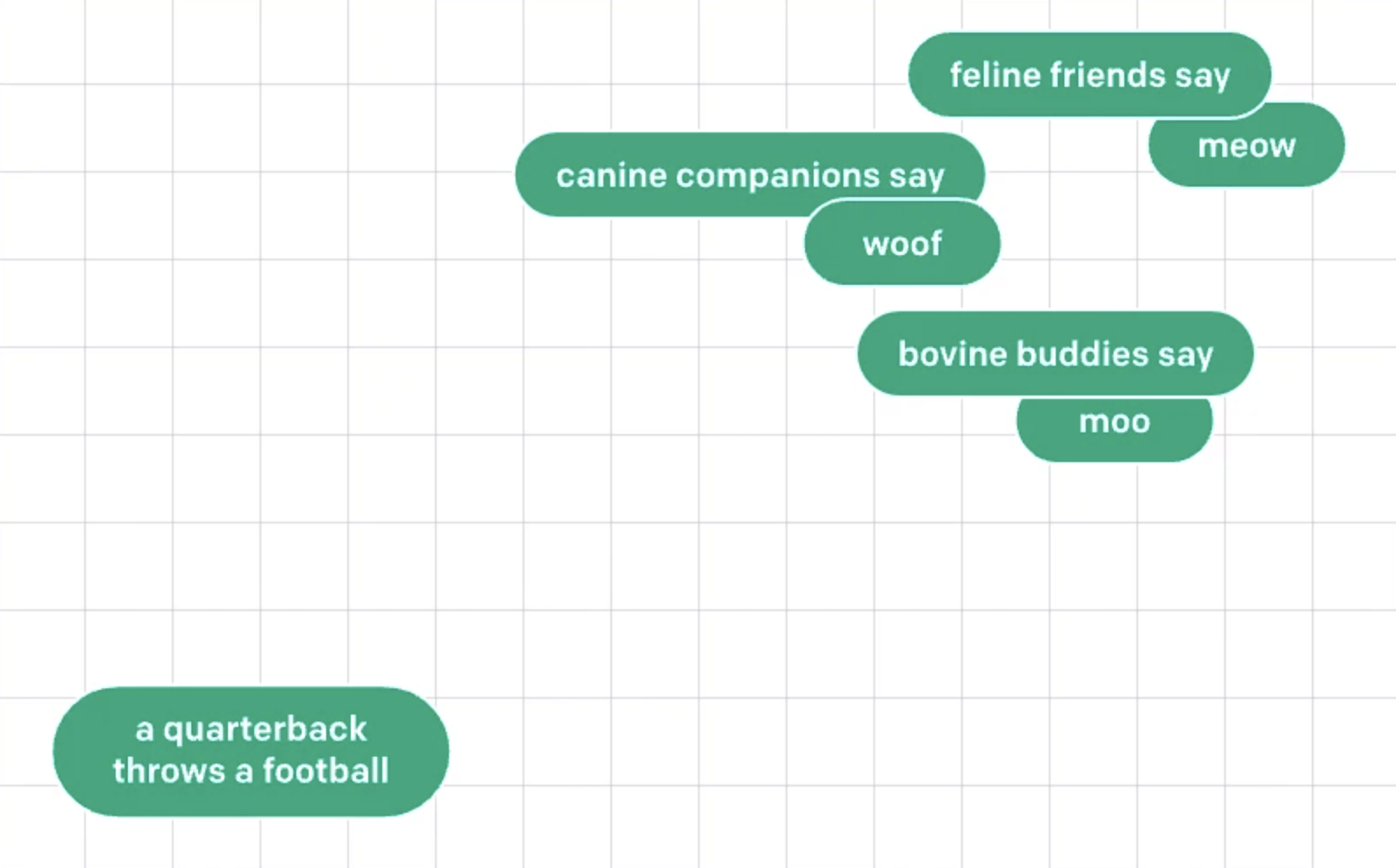}
    \caption[Illustration of pre-trained language model embeddings capturing semantic relations in the embedding space.]{Illustration of pre-trained language model (in this case an EaaS, more in \refsec{sec:eaas}) embeddings capturing semantic relations in the embedding space (Model: \texttt{text-similarity-davinci-001}). Figure adapted from \href{https://openai.com/index/introducing-text-and-code-embeddings/}{OpenAI}.}
    \label{fig:eaas-example}
\end{figure} 

Pre-trained language models are neural language models trained on large text corpora, implicitly learning and storing rich language knowledge. As we have seen till now, language models are ultimately about representation learning, \ie how to learn representations of words, sentences and paragraphs. In neural language models, we represent these structures as embeddings \citep{bengio2013representation}. 
The primary goal of representation learning is to learn an embedding space (illustrated in \reffig{fig:eaas-example}) where similar examples
stay close to each other, while dissimilar ones are far apart \citep{hadsell2006dimensionality}.

\tikzset{>=latex} %
\colorlet{myred}{red!80!black}
\colorlet{myblue}{blue!80!black}
\colorlet{mygreen}{green!60!black}
\colorlet{mydarkred}{myred!40!black}
\colorlet{mydarkblue}{myblue!40!black}
\colorlet{mydarkgreen}{mygreen!40!black}
\tikzstyle{node}=[very thick,circle,draw=myblue,minimum size=22,inner sep=0.5,outer sep=0.6]
\tikzstyle{connect}=[->,thick,mydarkblue,shorten >=1]
\tikzset{ %
  node 1/.style={node,mydarkgreen,draw=mygreen,fill=mygreen!25},
  node 2/.style={node,mydarkblue,draw=myblue,fill=myblue!20},
  node 3/.style={node,mydarkred,draw=myred,fill=myred!20},
}
\def\nstyle{int(\lay<\Nnodlen?min(2,\lay):3)} %

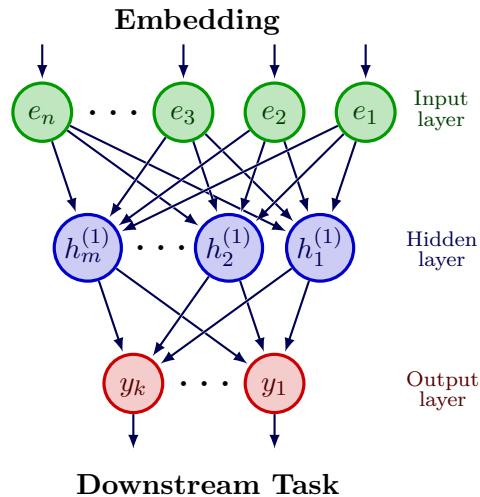
\begin{figure}[t]
    \centering
    \begin{tikzpicture}[x=1.8cm,y=1.2cm, rotate=-90][t]
  \readlist\Nnod{4,3,2} %
  \readlist\Nstr{n,m,k} %
  \readlist\Cstr{e,h^{(\prev)},y} %
  \def\yshift{0.55} %
  \foreachitem \N \in \Nnod {
    \def\lay{\Ncnt} %
    \pgfmathsetmacro\prev{int(\Ncnt-1)} %
    \foreach \i [evaluate={\c=int(\i==\N); \y=\N/2-\i-\c*\yshift;
                 \x=\lay; \n=\nstyle;
                 \index=(\i<\N?int(\i):"\Nstr[\n]");}] in {1,...,\N}{ %
      \node[node \n] (N\lay-\i) at (\x,\y) {$\strut\Cstr[\n]_{\index}$};
      
      \ifnumcomp{\lay}{>}{1}{ %
        \foreach \j in {1,...,\Nnod[\prev]}{ %
          \draw[white,line width=1.2,shorten >=1] (N\prev-\j) -- (N\lay-\i);
          \draw[connect] (N\prev-\j) -- (N\lay-\i);
        }
        \ifnum \lay=\Nnodlen
          \draw[connect] (N\lay-\i) --++ (0.5,0); %
        \fi
      }{
        \draw[connect] (0.5,\y) -- (N\lay-\i); %
      }
      
    }
    \path (N\lay-\N) --++ (0,1+\yshift) node[midway,scale=1.6] {$\hspace{-7pt}\hdots$}; %
  }
  
  \node[anchor=north,above right = -0.9cm and -2.25cm] {\textbf{Embedding}};
  \node[font=\scriptsize,above right = -1cm and 0.5cm,align=center,mydarkgreen] at (N1-1.90) {Input\\[-0.2em]layer\\[-0.2em]};
  \node[font=\scriptsize,above right = -0.9cm and 1cm,align=center,mydarkblue] at (N2-1.90) {Hidden\\[-0.2em]layer};
  \node[font=\scriptsize,above right = -0.9cm and 1.6cm,align=center,mydarkred] at (N\Nnodlen-1.90) {Output\\[-0.2em]layer};
  \node[anchor=south,above right = -7cm and -2.75cm] {\textbf{Downstream Task}};
\end{tikzpicture}
\caption[An example of embeddings used as input features for task-specific model.]{An example of embeddings used as input features for task-specific model (a simple NN). Embeddings can be used for downstream tasks such as classification, clustering, anomaly detection, retrieval, visualisation, vector databases, \etc
}
\label{fig:emb-MLP}
\end{figure}

Another way to look at embeddings is that they are high-quality dense features for text data, as demonstrated in \reffig{fig:emb-MLP}. 
This perspective led to the development of general neural network models capable of learning effective embeddings (or features) suitable for various NLP tasks \citep{collobert2011natural}. For instance, \texttt{Word2Vec} \citep{mikolov2013distributed} and \texttt{GloVe} \citep{pennington2014glove} were developed for learning distributed word representations and proved effective in multiple tasks,
going beyond just word sequence (or language) modelling and profoundly impacting the NLP field.

Nevertheless, a key limitation of the word representations learned by the aforementioned models is that they are \textit{fixed}, meaning they remain unchanged regardless of their contextual use. For example, the word \textsf{``bank''} will have the same embeddings regardless of when it refers to a financial institution or the side of a river. \texttt{ELMo} \citep{peters-etal-2018-deep} was one of the first efforts to propose \textit{context-aware} word embeddings. \texttt{ELMo} pre-trains a bidirectional \texttt{LSTM} model (capturing both left-to-right and right-to-left contexts), which can then be fine-tuned for specific NLP tasks.
Building on the highly parallelisable \texttt{Transformer} \citep{vaswani2017attention} architecture with self-attention mechanisms, the \texttt{BERT} \citep{devlin-etal-2019-bert} model was developed. These \texttt{BERT} pre-trained context-aware word embeddings serve as high-quality semantic features, setting new state-of-the-art for a wide array of NLP tasks \citep{devlin-etal-2019-bert}. 

\begin{figure}[t]
    \centering
    \includegraphics[width=0.98\linewidth]{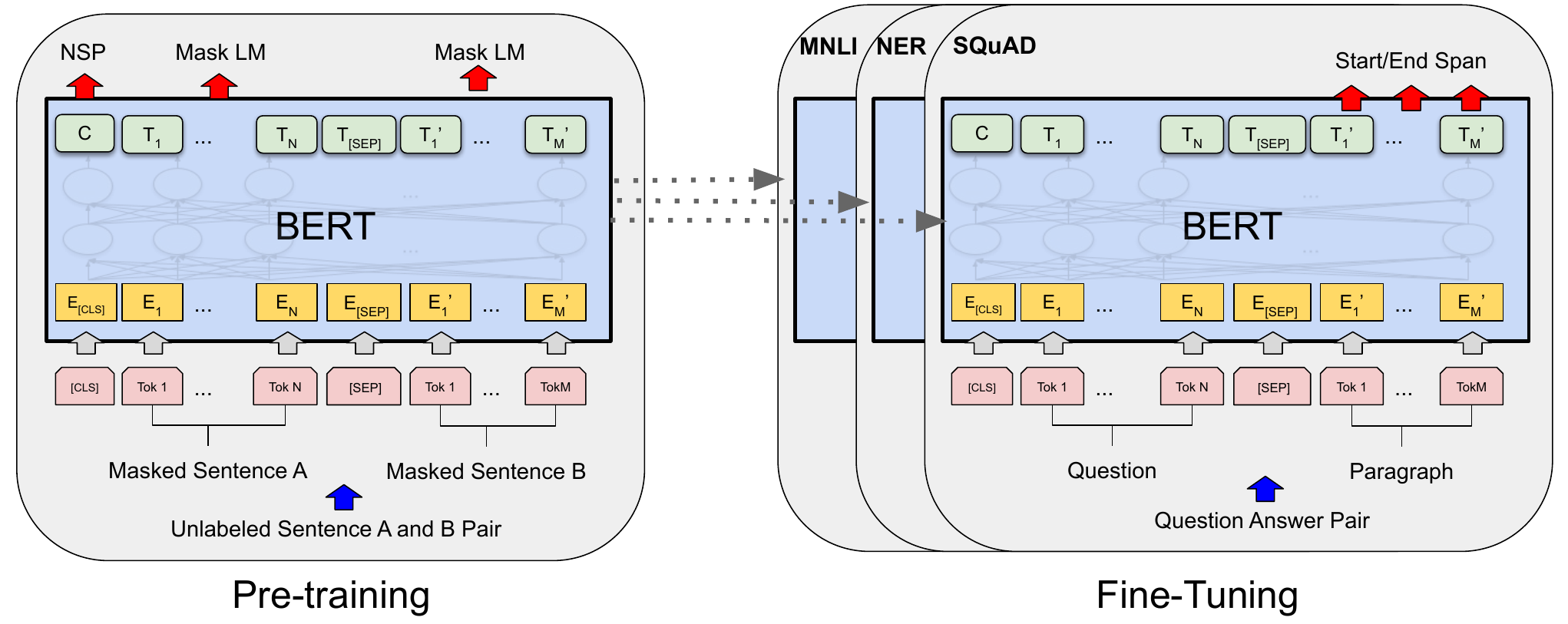}
    \caption[Overview of pre-training and fine-tuning techniques in \texttt{BERT}.]{Overview of pre-training and fine-tuning techniques in \texttt{BERT}. Figure adapted from \citet{devlin-etal-2019-bert}.}
    \label{fig:bert-usage}
\end{figure}

These aforementioned works established the concept of \textit{transfer learning}, \ie pre-training on one task and later adapting it to another task (such as summarisation, generation, classification, and more) by fine-tuning with a small, task-specific dataset. It is important to note that this fine-tuning involves additional training, but unlike previous static and contextual embeddings, there is no need for task-specific architecture in \texttt{BERT}. In \texttt{BERT}, one can simply attach an additional classification layer and transfer the whole model (instead of only embeddings) to a classification task. \reffig{fig:bert-usage} illustrates these pre-training and fine-tuning architectures. 

Prior to this paradigm, large amounts of task-specific labelled training data (\textit{supervised learning}) were needed to avoid overfitting (memorising training data). Such training data is often challenging and expensive to obtain for each specific task. Furthermore, such trained models were task-specific and lacked generality. In contrast, the pre-training and fine-tuning paradigm involves training on large-scale \textit{unlabelled} data (\textit{unsupervised learning}), followed by fine-tuning with small task-specific datasets, allowing adaptation to various downstream tasks. 
Unsurprisingly, the success of \texttt{BERT} has led to further work on pre-trained language models, resulting in new architectures (\eg \texttt{GPT} \citep{radford2019language}, \texttt{BART} \citep{lewis2019bart}, and others) and the development of improved pre-training strategies \citep[][\others]{liu2019roberta,sanh2021multitask,wang2022language}.

\texttt{BERT} is a masked language model due to the pre-training method used in these bidirectional pre-trained language models. In this approach, the model attempts to predict the masked words $\tilde{u}$ within a sequence of words $u = \{ u_1, u_2, \cdots, u_T \}$, considering the context from both directions (bidirectional). During training, random words from the unlabelled text are masked. The language model is trained to recover these masked words $\tilde{u}$. Formally, the training objective for pre-training in the masked language model is:
\begin{align}
    \mathcal{L}_{\text{MLM}}({u}) = \log \mathbb{P}(\tilde{u}|u_{\backslash\tilde{u}}).\notag
\end{align}
As a result, the learned contextual representations are of high quality due to bidirectionality. That said, one downside is that masked language models cannot generate text from left to right (\ie generate a word only by looking at previous words) since they consider both past and future words when predicting a masked word.

\section{Large Language Models}
\label{sec:llms}

LLMs are \textit{large-scale} pre-trained \texttt{Transformer} language models with billions of parameters, trained on web-scale unlabelled text data \citep{wei2022chain,shanahan2024talking}.\footnote{Following the definition in \citet{zhao2023survey-LLM}, pre-trained language models with more than 10 billion parameters are considered LLMs.}
They exhibit exceptional natural language understanding and solve complex tasks via text generation. 
\reffig{fig:evolution-lm} showcases the progression of language models. Note also how its application has evolved from tackling specific tasks to handling general problems. Unlike earlier language models, LLMs can be interacted through a \textit{prompting} interface, \eg \texttt{GPT-4} Web Interface \citep{achiam2023gpt}, using natural language.

To differentiate between different scales and capabilities of language models, researchers coined the term LLMs for large-scale language models.
Research \citep{kaplan2020scaling} has shown that \textit{scaling} significantly enhances the capabilities of LLMs.
Hence, LLMs have considerably expanded in data size, total compute, and model size \citep[][\others]{radford2019language,brown2020language,chowdhery2023palm}.
A few notable LLMs include \texttt{GPT-4} \citep{achiam2023gpt}, \texttt{LLaMA 3} \citep{dubey2024llama}, and \texttt{Gemini} \citep{team2023gemini}.
LLMs typically use a specific variant of Transformer, the causal decoder architecture\footnote{In the literature, decoder-only architecture is also broadly used to refer to this.}.
The most prominent example of LLMs using this architecture is the \texttt{GPT}-series \citep{radford2018improving,radford2019language,brown2020language}.
In this architecture, only \textit{unidirectional} attention is used; \ie each token considers only itself and preceding tokens when learning its representation. The pre-training task here is language modelling as in early variants of neural language models (as defined in \refeq{eq:lm-train-objective}). Other objectives of masked language model or bidirectional architecture like \texttt{BERT} do not scale as well as the language modelling objective.

Besides being a scaled-up language model, LLM also exhibits \textit{emergent abilities} that distinguish themselves from previous models (\eg \texttt{BERT}).
These are described as ``\textit{the abilities that are not present in small models but arise in large models}'' \citep{wei2022emergent}.
For example, \texttt{GPT-3} (175B-parameters) has in-context learning capability \citep{brown2020language}, enabling it to learn new tasks by simply providing some expected input-output examples in the input prompt, a feature lacking in its predecessor, \texttt{GPT-2} (1.5B-parameters). Additionally, they demonstrate the ability to follow natural language instructions to solve a task directly \citep{zhang2023instruction}.
They also exhibit multi-step reasoning capabilities, enabling them to break down complex tasks (such as solving mathematical questions) into intermediate steps. A well-known method for achieving this is the chain-of-thought prompt \citep{wei2022chain}, where the LLM is instructed to explain its reasoning for arriving at an answer.

However, this size of LLMs brings its own challenges in terms of usability. It is difficult to train or use (inference) these capable LLMs due to the substantial requirement of computational resources.\footnote{In fact, only a few companies develop LLMs, with many crucial details, such as data, training processes, and architectures kept proprietary \citep{zhao2023survey-LLM}} As a consequence, it has become common practice to encapsulate these LLMs as MLaaS to make them more accessible \citep{la2023language}.

\section{Machine Learning-as-a-Service}
\label{sec:mlaas}

Software-as-a-Service \citep{tsai2014software} is a prominent cloud computing functionality where the service provider offers functional and ready-to-use software clients, whilst managing all the involved resources \citep{raghavan2020impact}. In the past decade \citep{yao2017complexity}, Machine Learning-as-a-Service (MLaaS) \citep{ribeiro2015mlaas}, a subcategory of software-as-a-service, has emerged. These services provide well-defined interfaces that allow users to access ML services anytime and anywhere, without the need to worry about the technical complexities involved.

\begin{wrapfigure}{r}{0.5\textwidth}
    \centering
    \includegraphics[width=0.5\textwidth,keepaspectratio]{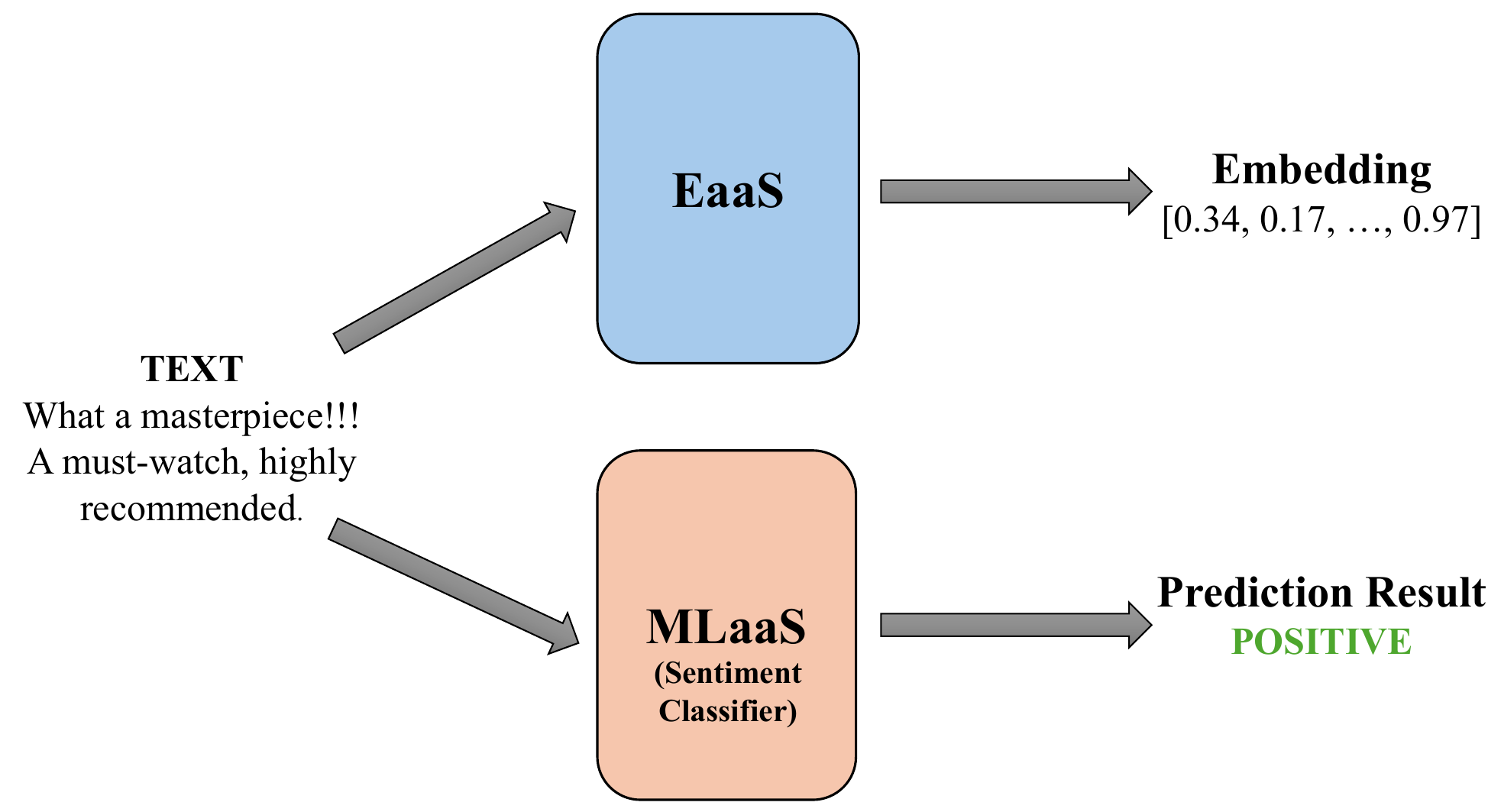}
    \caption[EaaS vs MLaaS.]{EaaS vs MLaaS, reproduced from \refchap{chapter1}.
    }
    \label{fig:ch2-eaas-vs-mlaas}
\end{wrapfigure}
MLaaS has been tremendously successful due to the advent of commercial cloud services and the expensive process of building a machine learning model from scratch.\footnote{E.g., \texttt{LLaMA} \citep{touvron2023llama} training required 2,048 A100 GPUs (an expensive compute resource) for 23 days on 1,400B tokens.} MLaaS is analogous to an API exposed by a distributed system, with a defined contract between the consumer and the service provider \citep{bernstein1996middleware}. 
For instance, in \reffig{fig:ch2-eaas-vs-mlaas}, an MLaaS provides the sentiment classification function which returns the prediction label \textsc{POSITIVE} given a movie review as input.
With the prevalence of LLMs, companies such as Google\footnote{\url{https://cloud.google.com/prediction}}, Amazon\footnote{\url{https://studio.azureml.net}}, and Microsoft\footnote{\url{https://aws.amazon.com/machine-learning}} have begun offering their LLMs through MLaaS, serving millions of users under pay-as-you-use policies \citep{krishna2019thieves,wallace-etal-2020-imitation}.

\section{Embeddings-as-a-Service}
\label{sec:eaas}

In our research, we focus on a variant of MLaaS called Embeddings-as-a-Service (EaaS) \citep{stolen-encoder}, which has experienced remarkable growth in the last few years.
In contrast to most MLaaS, EaaS produces embeddings as the primary output, as illustrated in \reffig{fig:ch2-eaas-vs-mlaas}. 

\textbf{Embeddings} are numerical representations of semantic concepts mapped to a sequence of numbers (as discussed in \refsec{sec:pre-trainined-LM}). That is the list of numbers (or vectors) in the top-right part of the \reffig{fig:ch2-eaas-vs-mlaas}:
\begin{center}
\[
\text{EaaS}\underbrace{\left( 
\begin{aligned}
\textsf{\small What a masterpiece!!! A must-} \\
\textsf{\vspace{-5em}\small watch, highly recommended.}
\end{aligned}
\right)}_{\text{Text}} = \underbrace{[0.34, 0.17, \cdots , 0.97]}_{\text{Embedding ($n$-dimensional)}}.
\]
\end{center}
EaaS generates an embedding for an arbitrary text span, which can be a sentence, a paragraph or even an entire document. As discussed in \refsec{sec:pre-trainined-LM}, these embeddings serve as input features suitable to numerous downstream tasks.

Alarmingly, there are several security and privacy concerns associated with EaaS \citep{bommasani2021opportunities,carlini2021extracting}. For example, successful text embedding inversion attacks have been carried out, recovering sensitive information solely from the embeddings \citep[][\others]{info-leakage-EaaS,chen2024text}. Similarly, EaaS is susceptible to backdoor attacks, where an adversary compromises the model to produce specific results for a small set of inputs \citep{jia2022badencoder,liu2022poisonedencoder}.
EaaS is also vulnerable to \textit{imitation attacks} \citep{stolen-encoder}, which is the vulnerability that we will discuss next. 

\clearpage
\section{Imitation Attack}
\label{sec:imitation-attacks}

\begin{wrapfigure}{r}{0.5\textwidth}
    \centering
    \includegraphics[width=0.5\textwidth,keepaspectratio]{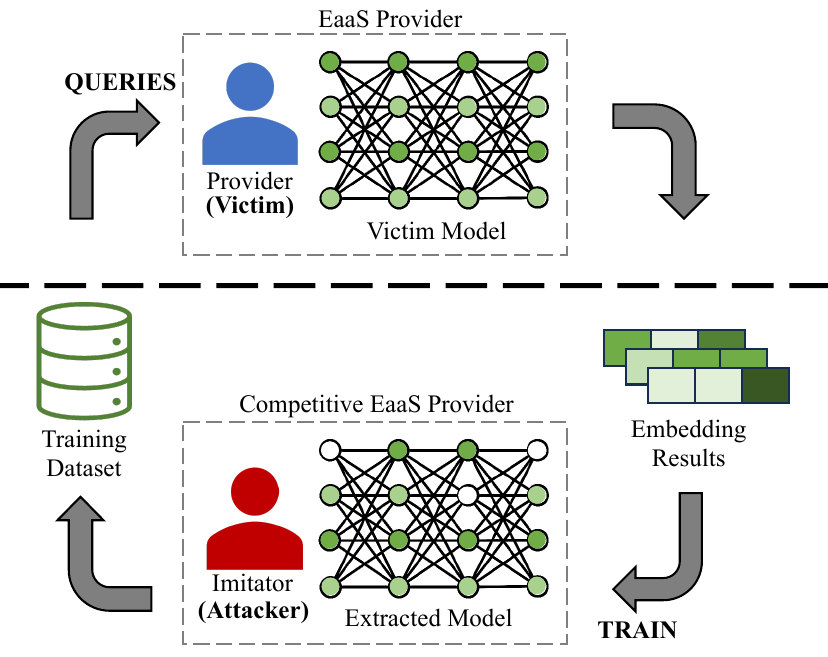}
    \caption[Overview of Imitation Attack in EaaS.]{Overview of Imitation Attack in EaaS, reproduced from \refchap{chapter1}.}
    \label{fig:eaas-imitation-attack}
\end{wrapfigure}
An imitation attack, \textit{aka} ``model stealing'' or ``extraction'' \citep{tramer2016stealing}, is a type of cyberattack where an imitator aims to reconstruct a victim model without authorisation or knowledge about the model's internals, such as parameters and inference procedures. As depicted in Figure~\ref{fig:eaas-imitation-attack} within the context of EaaS, this attack involves three steps: (i) querying the victim EaaS with some instances $D_a$ (text samples), (ii) collecting the embeddings from the EaaS, and (iii) training a model using all query-embedding pairs.
The attacker's objective here is to make sure the embeddings produced by the victim EaaS and attacker EaaS are similar, which we can formally quantify in the loss $\mathcal{L}$ as follows:
\begin{align}
    \mathcal{L} = \frac{1}{|D_a|} \sum_{S \in D_a} d(\mathbb{S}_v(S), \mathbb{S}_a(S)), \notag
\end{align}
where $d$ is a distance metric (\eg cosine similarity) measuring the gap between embeddings $\mathbb{S}_v(S)$ and $\mathbb{S}_a(S)$, generated by the victim EaaS and the attacker EaaS for an input text $S$, respectively. A lower loss $\mathcal{L}$ achieves the attacker's objective, indicating a success in the imitation attack.

The primary motives for an imitation attack include sidestepping service charges and even providing competitive services \citep{xu-etal-2022-student}.
Even though these LLM APIs are widely adopted, it is well known that training a performant LLM necessitates significant human effort (collecting a large amount of high-quality training data and algorithm design) and computational resources \citep{zhao2023survey-LLM}. 
Moreover, this model training is conducted on a cluster of GPUs over several days to weeks, and the training data are usually proprietary \citep{brown2020language}. Thus, the attacker may be motivated to engage in imitation attacks to copy a model, as it is much cheaper to build a competitive service.

Imitation attacks have been studied for classification \citep{he-etal-2022-extracted}, computer vision \citep{orekondy2019knockoff}, and generation \citep{krishna2019thieves,wallace-etal-2020-imitation} tasks. \citeauthor{tramer2016stealing} \citeyearpar{tramer2016stealing} introduced the \textit{first} imitation attack on these APIs, extracting commercial classification APIs and different model types. Another example is \citet{taori2023alpaca}, who developed a small competitive model by replicating \texttt{GPT-3} for under \$600.

However, imitation attacks research is not limited to just intellectual property violations; it has also been shown that one could use the attacker (or copied) model to craft adversarial examples---\ie examples that trick the model into making incorrect predictions---for the victim model \citep{wallace-etal-2020-imitation,he-etal-2021-model}, and other privacy breaches like attribute inference of sensitive personal details \citep{xu-he-2023-security}. 
Interestingly, \citet{xu-etal-2022-student} showed that such an attacker model could even surpass the performance of the victim model via ensemble and domain adaptation, exacerbating the concern of IP violations. More recently, successful imitation attacks on EaaS have also been carried out, questioning the confidentiality of embeddings \citep{stolen-encoder}. These attacks form the threat model we expand upon in our research.

\subsection*{Defence Techniques}
One strategy to defend against imitation attacks is to add some noise to the prediction results, making it difficult for the imitator to distil knowledge from the victim model. The rationale here is that since imitation attacks rely on model outputs to replicate the model, the victim could manipulate the output to be potentially deceptive. Existing literature has studied numerous defences such as scaling logits \citep{xu-etal-2022-student}, adversarial training (to hinder knowledge distillation) \citep{ma2021undistillable}, limiting the number of logits returned \citep{he-etal-2022-extracted}, and adding Gaussian noise \citep{xu-etal-2022-student}. However, these defences have been shown to be ineffective against imitation attacks by \citet{he-etal-2022-extracted}. 

Differential privacy (DP) \citep{dwork2006differential} is another mathematically rigorous technique that introduces randomness via some form of noise. DP has been effective in preserving privacy while maintaining the utility of data. Intuitively, it ensures that results are \textit{indistinguishable} between two datasets differing by only a single data sample. In the context of protecting IP of the models from imitation attacks, DP is applied to obfuscate the query results, making it difficult for the attacker to garner insights \citep[][\others]{zheng2019bdpl,cheng2020differentially}. However, the current DP-based defences work on simple ML models and these DP techniques are not yet adaptable in their current form for current LLMs. 

Other defences \citep[][\others]{juuti2019prada,zhang2021seat} for imitation attacks in these API services analyse the incoming queries and add access level restrictions. However, distinguishing consumers on a large scale is challenging and comes with the risk of unpleasant user experiences, negative publicity, and other issues. Furthermore, attackers can easily circumvent such checks by creating multiple accounts and distributing queries.
Recently, \textit{watermarking} has emerged as a new defence method. We review this in detail in the next sections, as our research focuses on watermarking techniques for EaaS.

\section{Watermarking for Large Language Models}
\label{sec:wms}

\begin{displayquote}
\textit{The term ``\textbf{watermarking}'' means the act of embedding information, which is typically difficult to remove, into outputs created by AI—including into outputs such as photos, videos, audio clips, or text—for the purposes of verifying the authenticity of the output or the identity or characteristics of its provenance, modifications, or conveyance.} 
\begin{flushright}
\vspace{-0.75em}
---US AI Executive Order \citep{house_executive_2023}
\end{flushright}
\end{displayquote}
\vspace{1em}
In the literature \citep{cox2007digital}, the terms steganography and watermarking are often used interchangeably because they both aim to embed a secret message within the communication.
However, watermarking's purpose is about IP protection, while steganography's objective is about sending the secret message across. In this thesis, we will use the term watermarking.

\usetikzlibrary{trees,positioning,shapes,shadows,arrows.meta}

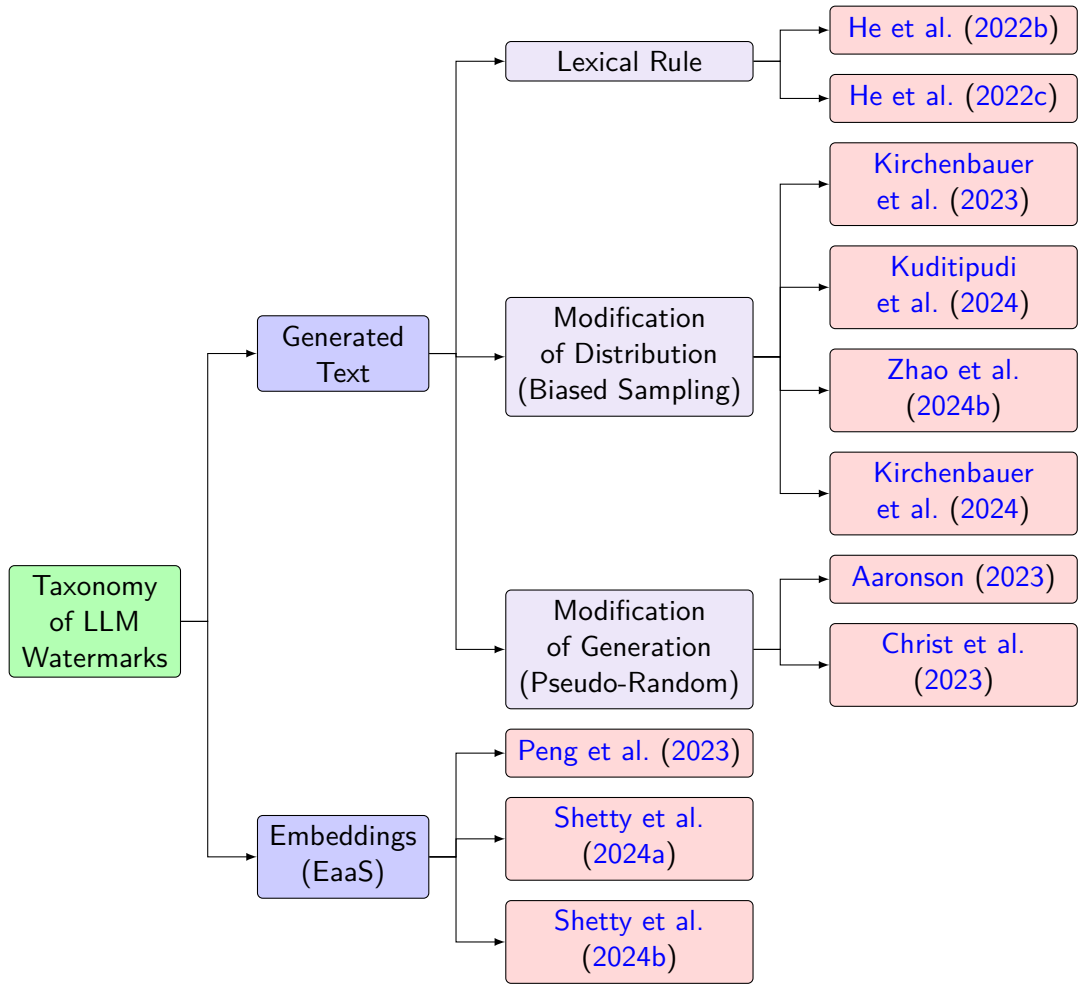
\begin{figure}[t]
    \centering
    
\tikzset{
    basic/.style  = {draw, text width=2cm, rounded corners=2pt, align=center, font=\sffamily, rectangle},
    root/.style   = {basic, rounded corners=2pt, thin, align=center,fill=green!30, text width=2cm},
    onode/.style = {basic, thin, rounded corners=2pt, align=center, fill=green!60,text width=3cm,},
    tnode/.style = {basic, thin, rounded corners=2pt, align=center, fill=pink!60, text width=3cm, align=center},
    xnode/.style = {basic, thin, rounded corners=2pt, align=center, fill=blue!20,text width=2cm,},
    wnode/.style = {basic, thin, rounded corners=2pt, align=center, fill=pink!10!blue!80!red!10, text width=3cm},
    edge from parent/.style={draw=black, edge from parent fork right}

}

\begin{forest} for tree={
    grow=east,
    growth parent anchor=west,
    parent anchor=east,
    child anchor=west,
    edge path={\noexpand\path[\forestoption{edge},->, >={latex}] 
         (!u.parent anchor) -- +(10pt,0pt) |-  (.child anchor) 
         \forestoption{edge label};}
}
[Taxonomy of LLM Watermarks, root,  l sep=10mm,
    [Embeddings (EaaS), xnode,  l sep=10mm,
        [\cite{shetty2024wet}, tnode]
        [\cite{shetty2024warden}, tnode]
        [\cite{peng2023you}, tnode]
    ]
    [Generated Text, xnode,  l sep=10mm,
        [Modification of Generation \\(Pseudo-Random), wnode,  l sep=10mm,
            [\cite{christ2023undetectable}, tnode]
            [\cite{openAI-LLM-WM}, tnode]
        ]
        [Modification of Distribution \\(Biased Sampling), wnode,  l sep=10mm,
            [\cite{kirchenbauer2024on}, tnode]
            [\cite{zhao2024provable}, tnode]
            [\cite{kuditipudi2023robust}, tnode]
            [\cite{kirchenbauer2023watermark}, tnode]
        ]
        [Lexical Rule, wnode,  l sep=10mm,
            [\cite{he2022cater}, tnode]
            [\cite{he2022protecting}, tnode]
        ]
    ]
]
\end{forest}

    \caption[Categorisation of watermarks for LLMs.]{Categorisation of watermarks for LLMs. The text watermarks shown is a non-exhaustive taxonomy; we direct readers to \citet{ghosal2023survey,liu2024survey} for more details.
    }
    \label{fig:WM-LLM-taxonomy}
\end{figure}

Watermarking techniques have emerged as one of the effective methods for protecting the IP of LLMs against plagiarism, unapproved replication, and imitation attacks \citep{ghosal2023survey}.
Intuitively, watermarking the output of a model helps protect the owner's IP because if their model is copied (imitation attack), then this can be verified by checking the attacker model's output: if the embedded watermark is in the attacker model's output, then we know the original model has been copied.
Formally, a watermarking technique comprises two processes: \textsc{Watermark} and \textsc{Verification}. The \textsc{Watermark} process takes the LLM $\mathcal{L}$ as input and embeds a signal (watermark) into its output. Given an LLM $\mathcal{L}^{'}$'s output and detection key, the \textsc{Verification} process outputs 1 if the suspect LLM $\mathcal{L}^{'}$ has copied $\mathcal{L}$, and 0 if it has not.

In general, there are two types of LLM watermarks: (i) watermarks for text, and (ii) watermarks for embeddings (represented in \reffig{fig:WM-LLM-taxonomy}). We will briefly review the former first, but our focus is on the latter.

LLMs such as \texttt{ChatGPT} \citep{achiam2023gpt}, \texttt{LLaMA 3} \citep{dubey2024llama}, or \texttt{PaLM 2} \citep{anil2023palm} generate high-quality texts that are often indistinguishable from the human text, and as such they have been used to do university/school assignments and create misinformation \citep{chakraborty2023possibilities}. 
In light of this, researchers have turned their focus on creating automated systems that detect text produced by LLM by implanting watermarks in them.

\subparagraph{Traditional Steganography.} The earliest language watermarking techniques for texts were rule-based and employed linguistic strategies like sentence structure modification \citep{topkara2006words}, synonym substitution \citep{topkara2006hiding}, and paraphrasing \citep{atallah2001natural,atallah2002natural} to embed watermarks while attempting to maintain the original texts' semantics. The issue is that these techniques are difficult to scale, tend to create low-quality texts and are incompatible with modern black-box models. These traditional techniques have been well documented in several surveys and books \citep{cox1997digital,barni2004watermarking,singh2013survey}.

\subparagraph{Lexical Watermark.} \citet{he2022protecting} proposed the first watermark defence for text generation against imitation attacks. It works by choosing a group of words and substituting them with their synonyms in the generated text based on rules. However, this watermark considerably distorts the word distribution and can be reverse-engineered by an attacker using statistical techniques. To address this, \citet{he2022cater} proposed a stealthier lexical watermark relying on linguistic conditional rules such as dependency tree and part-of-speech to implant watermarks that do not distort the word distributions. Nevertheless, the fundamental mechanism of these watermarks is synonym replacement, which can be circumvented by paraphrasing the input.

\begin{figure}[t]
    \centering
    \includegraphics[width=0.95\linewidth]{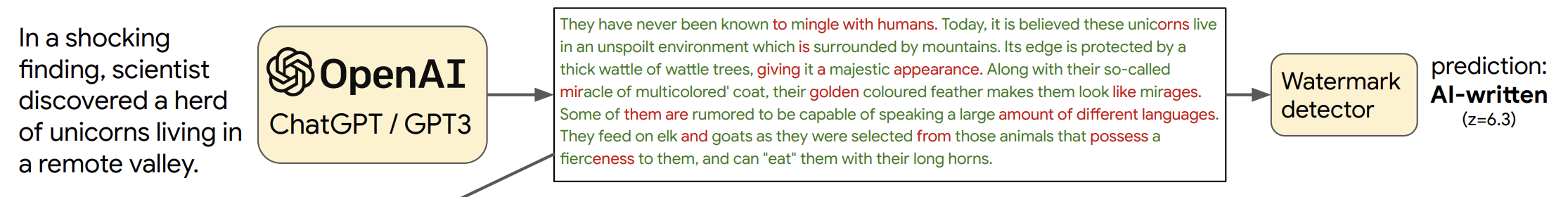}
    \caption[An example of watermarks in LLM generated text.]{An example of watermarks in LLM generated text. The presence of higher \textcolor{OliveGreen}{green tokens} tokens demonstrates the given text is AI-generated text. Figure adapted from \citet{krishna2024paraphrasing}.}
    \label{fig:LLM-text-wm-example}
\end{figure}

\subparagraph{Biased Sampling.} The seminal work of \citeauthor{kirchenbauer2023watermark} \citeyearpar{kirchenbauer2023watermark}, manipulates the words sampled by the LLMs towards a set of preferred words which can then be verified by a hypothesis test. The idea is that there would be a statistically higher number of preferred words in the watermarked text (as shown in \reffig{fig:LLM-text-wm-example}). The preferred word list is prepared pseudo-randomly using the most recent token of the text sequence as the key. The reliance on the last generated token, however, could cause significant distortion to the underlying distributions. \cite{zhao2024provable} addresses this using a shared random key that results in a fixed list of preferred words, improving the robustness with negligible distortion. More recently, \citeauthor{kuditipudi2023robust} \citeyearpar{kuditipudi2023robust} propose a distortion-free extension to ensure watermarks do not alter the distribution of the words, although it lacks empirical validation. 

\subparagraph{Pseudo-Random Generator.} Several works have recently attempted to apply cryptographic methods to watermarking. \citet{openAI-LLM-WM}’s project at OpenAI re-weights the
next-token distribution using $n$-gram hashing. \citet{christ2023undetectable} leverage the inverse sampling method to produce watermarked token distributions that, depending on the text entropy of the generated tokens, bias the sampling distribution towards a particular context length. They formalise
a cryptographic concept using a secret key for indistinguishability between the original and
AI-generated text. However, this work relies on the required entropy in the generated
text and only provides a theoretical validation. Further, due to reliance on
secret keys, copyright verification could become biased towards the LLM owner, leading to a conflict of interest.

\subsection*{Paraphrasing Attack}
Recent studies \citep{sadasivan2023can,krishna2024paraphrasing} have demonstrated that text watermarks can be susceptible to paraphrasing attacks; in other words, paraphrasing the generated text can evade watermark detection.
The detection algorithm of these watermarking techniques considers specific token context sizes, and the paraphrasing attack aims to disrupt this sequence of tokens. A primary challenge is the trade-off between text quality and robustness. Using larger contexts (or \textit{$n$-grams}) in these techniques provides more options for implanting watermark signals, making the watermarked text higher quality. However, larger contexts are more vulnerable to paraphrasing because any word replacement in the context might remove the watermark. Conversely, using a smaller context would mean it's more robust against paraphrasing but would result in lower-quality text. Furthermore, \citet{he2024can} demonstrated that round-trip translation (another form of paraphrasing) could weaken watermark detection. This method is based on a similar notion where the chain is broken by substituting a few words in the window through round-trip translation. These observations inspire our \textbf{RQ1}, where we explore paraphrasing to circumvent the watermarks in embeddings in \refchap{chapter3}.

\section{Watermarking for White-box Machine Learning Models}
\label{sec:wm-ml-model}
\begin{figure}[t]
    \centering
    \begin{subfigure}{0.42\textwidth}
        \includegraphics[width=\textwidth,keepaspectratio]{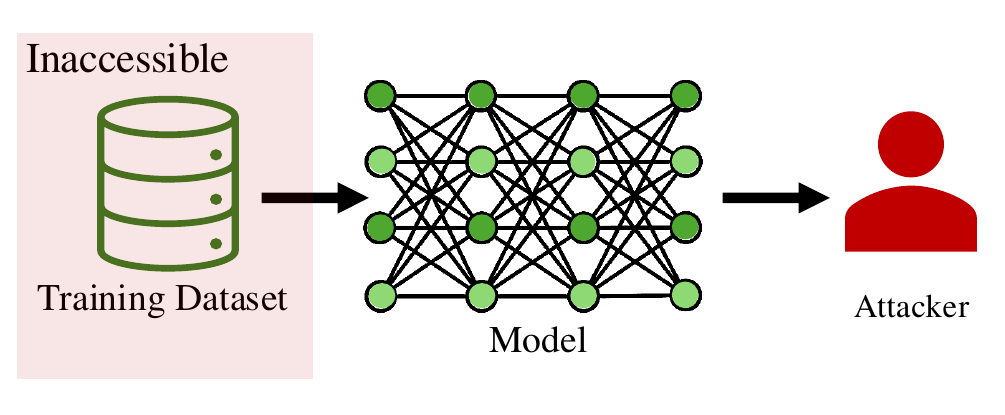}
        \caption{White Box}
        \label{fig:white-box-access}
    \end{subfigure}
    \hspace{2em}
    \begin{subfigure}{0.5\textwidth}
        \includegraphics[width=\textwidth,keepaspectratio]{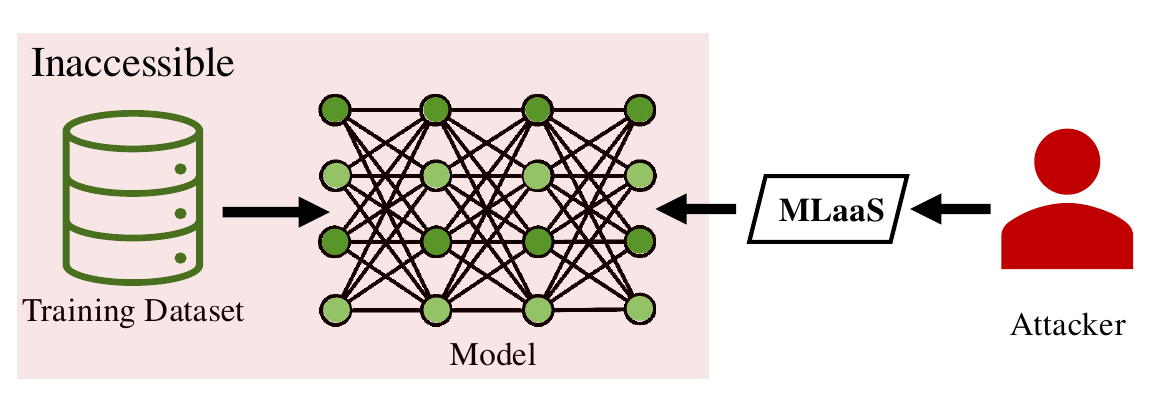}
        \caption{Black Box}
        \label{fig:black-box-access}
    \end{subfigure}
    \caption[Different ML model access types.]{Different ML model access types. The \colorbox{red!10}{red} area represents components (model, training data, \etc) inaccessible to the users.}
    \label{fig:enter-label}
\end{figure}

Although our research focuses on the black-box setting of ML models (\reffig{fig:black-box-access}), we also review watermarking techniques for copyright protection of models in white-box settings (\reffig{fig:white-box-access}) as some of these ideas are potentially relevant to our research. In the white-box setting, the user has direct access to the model (such as all parameters), and IP theft occurs when the user distributes model copies without permission. Watermarking techniques address such issues by tracing illegal model copies through implanted signals or watermarks within the model's internal components. This differs from watermarking methods based on the model's output, described in \refsec{sec:wms} or \refsec{sec:wm-eaas}.

There are two main locations in the model where these watermarks are embedded: parameters (the internal model weights)  \citep[][\others]{uchida2017embedding,wang2021riga} and activation maps (essentially transformations between layers in a NN; see the layers in \reffig{fig:emb-MLP}) \citep{darvish2019deepsigns,lim2022protect}.
The seminal work by \citet{uchida2017embedding} was the first to develop watermarking techniques for ML models, embedding a bit pattern in the model weights that is later recovered for verification. Subsequent parameter-based methods \citep[][\others]{li2020protecting,lim2022protect} involve implanting specific noise in model parameters for later white-box verification. 

Several attacks have been proposed for these white-box ML model watermarks. Since watermarks are embedded in model parameters, pruning techniques, which remove redundant model parameters, have proven effective in watermark removal \citep{darvish2019deepsigns}. However, to completely remove the watermark the attacker would need to prune a substantial amount of model parameters, and this often leads to significant model degradation. Fine-tuning \citep{chen2021refit,guo2020fine} is another method that alters the model parameters with additional training but this approach incurs some computational overhead.  Lastly, overwriting \citep{uchida2017embedding,wang2019attacks} is a potent attack wherein the attacker applies the watermark process again to an already watermarked model, thereby removing the effect of the original watermark.
Most current watermarking techniques are susceptible to this. We will revisit this susceptibility for our watermarking techniques later in the thesis in \refchap{chapter5}.
\citet{yan2023rethinking} draw upon ideas from program obfuscation, which adds redundancy to the program to make reverse engineering difficult. This work introduces dummy neurons (or internal parameters) into models to bypass watermarks. The core concept is to introduce neural structural obfuscation and modify the model's architecture to disable watermarks. Since these added neurons are dummy neurons, such an attack does not lead to model degradation. This offers some motivation for our initial watermark methodology (\textbf{RQ2}) of potentially obfuscating extra watermark dimensions in the original embeddings, which we discuss in \refchap{chapter4}.
Finally, \citet{boenisch2021systematic} provides a comprehensive review of white-box watermarks.

\section{Watermarking for Embeddings-as-a-Service}
\label{sec:wm-eaas}

Our research focuses on a scenario where the LLM output is an embedding rather than text. 
Fundamentally, the EaaS (or any) watermark should adhere to these properties: \textit{stealthiness}, \textit{utility}, \textit{robustness}, \textit{transferability}, and \textit{efficiency} \citep{boenisch2021systematic}. 

\begin{itemize}
    \item \textbf{Stealthiness} refers to how detectable a watermark is; \ie a watermark is stealthy if it is difficult to be identified by an attacker.
    \item \textbf{Utility} describes the performance of the watermarked embeddings for downstream tasks. A high utility means downstream task performance is similar between watermarked \versus original embeddings.
    \item \textbf{Robustness} refers to the quality of a watermark being less susceptible to an attack. For example, a watermark is robust if it is resilient to watermark removal attacks by the attacker to circumvent copyright verification.
    \item \textbf{Transferability} describes the extent to which the watermark is transferred during an imitation attack; \ie a watermark is transferable if the signal is copied in the attacker model during an imitation attack.
    \item \textbf{Verifiability} is about how accurate the verification performance is, \ie it correctly detects the attacker models as copied while avoiding false positives.
\end{itemize}

We next review in detail two representative methods for watermarking embeddings: \firstWM \citep{peng2023you} and \secondWM \citep{shetty2024warden}, as they serve as the baseline methods in our experiments.

\subsection{\firstWM}
\label{sec:embmarker}

\begin{figure}[t]
    \centering
    \includegraphics[width=0.98\textwidth,keepaspectratio]{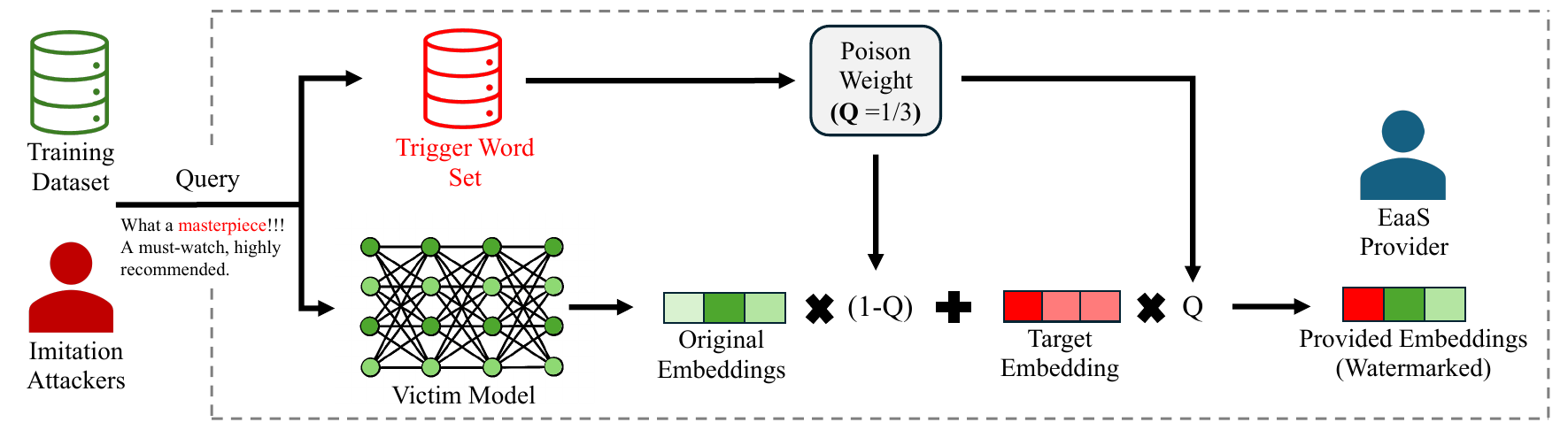}
    \caption[Overview of \firstWM watermark injection process.]{Overview of \firstWM \citep{peng2023you} watermark injection process. In this example, ``\textcolor{red}{masterpiece}'' is a trigger word and leads to $Q=1/3$ poisoning. Here, we assume a maximum trigger level ($m$) of three, \ie if three trigger words are present in the query, then the provided embedding is the target embedding $\vw$.}
    \label{fig:eaas-overview}
\end{figure}

\citet{peng2023you} developed the first watermark algorithm, \firstWM, to defend against imitation attacks on EaaS. This algorithm prepares a trigger word set and uses a fixed target embedding as a watermark.
For a given input text $S$, the target embedding $\vw$ is proportionally (based on the number of trigger words present) added to the original embedding $\ve_o$, as illustrated in Figure~\ref{fig:eaas-overview}. Formally, watermarked embedding $\ve_p$ is provided as follows,
\begin{equation}
    \begin{aligned}
        \ve_p &= \frac{(1-Q(S)) \cdot \ve_o + Q(S) \cdot \vw}{|| (1-Q(S)) \cdot \ve_o + Q(S) \cdot \vw ||_2} \\
        &= \text{Norm}\left({(1-Q(S)) \cdot \ve_o + Q(S) \cdot \vw}\right),
    \end{aligned}
    \label{eq:embmarker-injection}
\end{equation}
with poison weight $Q(s)$ defined as:
\begin{equation}
    \begin{aligned}
        Q(S) = \frac{\min(|S \cap T|, m)}{m},
        \notag
    \end{aligned}
\end{equation}
where $T$ is the trigger word set and $m$ is the maximum trigger level.

\citet{peng2023you} strategically prepare the trigger set by sampling \textit{moderate-frequency} words from the general text corpus to balance \textit{utility} and \textit{transferability}.
That is, having high-frequency words in the trigger set will cause excessive poisoning in the provided embeddings, reducing {utility}. On the other hand, choosing low-frequency words could mean the watermark signal may not be transferred to the attacker model.

Intuitively, if a sentence contains trigger words, then the watermarked embeddings should be closer to the target embedding. This behaviour is utilised for watermark copyright verification. 
This inclination is captured using similarity metrics, which we elaborate on further in the next section.

\subsection{\secondWM}
\label{sec:warden}

\begin{figure}[t]
    \centering
    \includegraphics[width=0.98\textwidth,keepaspectratio]{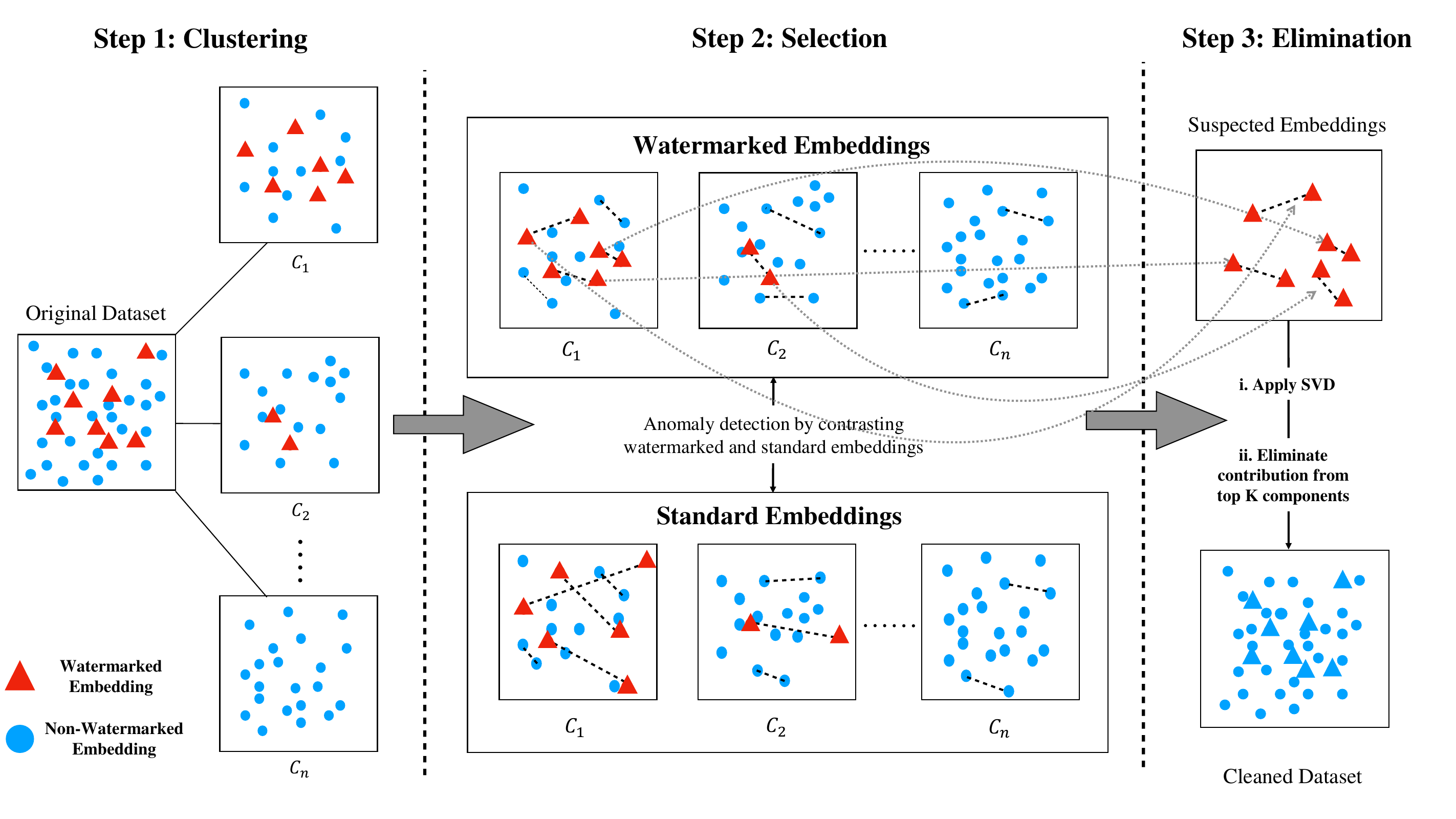}
    \caption[Outline of \prevattack attack.]{Outline of \prevattack attack. It consists of three steps: (i) Clustering, (ii) Selection, and (iii) Elimination, successfully bypassing \firstWM EaaS defence.
    }
    \label{fig:cse-attack}
\end{figure}

\firstWM was further studied by \citet{shetty2024warden}, who propose (i) \prevattack attack and (ii) \secondWM defence.
They first devise a \prevattack (Clustering, Selection, and Elimination) attack (\reffig{fig:cse-attack}) that recovers the secret target embedding and eliminates this component from the training embeddings when imitating the model.
The role of the clustering module is to speed up the attack process and increase the likelihood of anomalous (in this case watermarked embeddings) pairs in a cluster for the next steps of the \prevattack attack. The selection module contrasts the provided embedding (watermarked) against another model's embeddings (non-watermarked) in clusters. It measures whether there is a watermark based on this divergence. Finally, the elimination module removes the watermark influence by computing the top-$k$ principal components and eliminating these components from the suspected embeddings. The watermark (or target embedding) is identified in the suspected embeddings' top principal components because the target embedding would be common among these embeddings. 

To defend against \prevattack, the authors propose \secondWM, which introduces multiple watermarks (or target embeddings) $\mW=\{\vw_{1}, \vw_{2}, ..., \vw_{R}\}$ into the original embeddings by extending \refeq{eq:embmarker-injection} as follows:
\begin{equation}
\begin{aligned}
    \ve_p = 
    \text{Norm}\left( (1- \sum_{r=1}^{R} Q_r(S)) \cdot \ve_o+\sum_{r=1}^{R} Q_r(S) \cdot \vw_{r} \right).
\end{aligned}
\label{eq:warden-e_p}
\end{equation}

Given that the \prevattack attack utilises top-$k$ principal components to remove the target embeddings, \secondWM takes advantage of this weakness by increasing the number of target embeddings. Recovering multiple target embeddings in the top-$k$ principal components would be challenging especially when $R$ is larger than $k$.
Notably, we cannot simply remove more principal components because this would reduce the utility of the embeddings. Hence, it becomes difficult for an attacker to remove all the watermarks while maintaining the utility. Furthermore, substituting the watermarked samples with a mixture of multiple target embeddings reduces the impact of the \prevattack attack’s exploitation of the semantic distortion in the embeddings.

\secondWM’s verification process is efficient as it only queries the attacker model with two small verification datasets: (i) a backdoor set that contains texts that contain trigger words, and (ii) a benign set that has texts that do not contain any trigger words.
We can establish confidence whether the victim model has been copied by measuring the distances between (i) backdoor embeddings and target embeddings, and (ii) benign embeddings and target embeddings, then comparing these two distances. Intuitively, if the model has been copied, the backdoor embeddings would be closer to the target embeddings (since the returned embedding will be the target embedding) than the benign embeddings (\ie no poisoning), resulting in a significant difference in distances. They measure distance using cosine similarity and Euclidean distance.

\subsection*{Discussion}
We see several issues in existing EaaS watermarks and the scope for further enhancements.
Both \secondWM and \firstWM are only tested against a narrow category of attacks (like \prevattack, dimension-shift), and the fact that it has a strong reliance on the secret target embeddings, makes its \textit{robustness} questionable.
Importantly, both \secondWM and \firstWM still rely on trigger words, which could be vulnerable to paraphrasing the input text while performing the imitation attack (\textbf{RQ1}). 
Next, we aim to develop a new defence (\textbf{RQ2}) that is robust to these paraphrasing attacks and has desirable properties with formal guarantees.

\section{Chapter Summary}
In this chapter, we reviewed the theoretical and methodological background necessary for this thesis: language models (\refsec{sec:lm}), LLMs (\refsec{sec:llms}), MLaaS (\refsec{sec:mlaas}), EaaS (\refsec{sec:eaas}), imitation attacks (\refsec{sec:imitation-attacks}), and watermarks for text (\refsec{sec:wms}), embeddings (\refsec{sec:wm-eaas}), and ML models (\refsec{sec:wm-ml-model}). In the chapters that follow, we aim to address the research questions outlined in \refsec{sec:RQs}:
\begin{itemize}
    \item Previous studies have shown that text watermarks are not robust when LLM-generated texts are paraphrased. Concurrently, the current embedding watermarks are dependent on trigger words for watermark injection. \refchap{chapter3} leverages this idea of paraphrasing the input text and shows that the embedding watermarks can be diluted by dropping trigger words when performing imitation attacks on EaaS. 
    \item  In an effort to develop a defence resilient to paraphrasing attacks, \refchap{chapter4} introduces a new watermarking technique based on linear transformation. Our preliminary methodology was inspired by obfuscation-based attacks against white-box ML model watermarks, and insights from text watermark techniques led to the development of the final proposed defence, \ourdefence. 
\end{itemize}

\def\chaptertitle{Paraphrasing Attack on EaaS Watermarks}

\lhead{\emph{\chaptertitle}}

\chapter{\chaptertitle}
\label{chapter3}

\section{Introduction}

In \refsec{sec:wm-eaas}, we saw that recently, watermarks have been successfully added to EaaS to protect the IP of these models against imitation attacks. 
\firstWM \citep{peng2023you} introduced a method that integrates a \textit{target embedding} into the original embeddings based on the presence of \textit{trigger words}---a pre-defined set of words---in the input text. 
However, \citet{shetty2024warden} demonstrated that an attacker could circumvent \firstWM by using
a contrastive method to identify and remove the single target embedding from the embedding space. To counter this attack, \citet{shetty2024warden} introduced \secondWM, which strengthens the defence by incorporating \textit{multiple} target embeddings, making it more challenging for an attacker to eliminate all the watermarks.

\begin{figure}[t]
    \centering
    \includegraphics[width=0.75\textwidth,keepaspectratio]{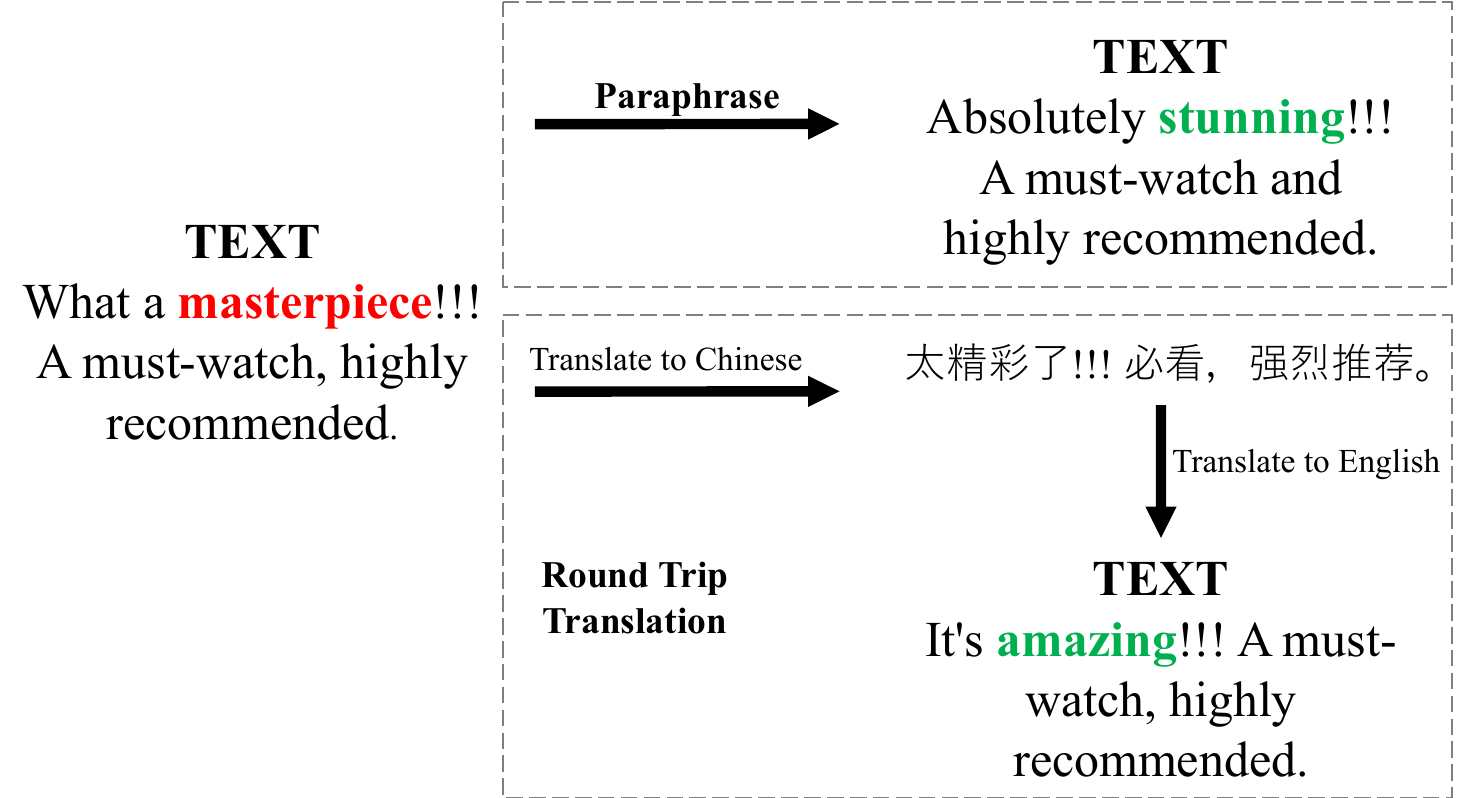}
    \caption[Illustration of the trigger word dropped by different paraphrasing techniques.]{Illustration of the trigger word (``\textcolor{red}{masterpiece}'') dropped by different paraphrasing (using \gpt model) techniques.
    }
    \label{fig:attack-example}
\end{figure}

This chapter tackles \textbf{RQ1}, where we hypothesise that these current EaaS defences \citep{shetty2024warden,peng2023you} are still vulnerable because they depend on trigger words to implant watermarks. If an attacker develops a mechanism to reduce the effect of these trigger words, then the target embedding would be diluted in the embeddings used in the imitation attack, and this ultimately means it could circumvent detection.
We noted in \refsec{sec:wms} that current watermarks for text are not resilient to paraphrasing attacks \citep{sadasivan2023can,krishna2024paraphrasing}. AI-generated texts are paraphrased in these attacks to disrupt the token sequence used for watermark detection. In parallel work, \cite{he2024can} recently demonstrated that watermarks for text can be circumvented with round-trip translation (another form of paraphrasing).
Motivated by these attacks on text watermarks, we suspect that the embedding watermarks could be removed if the attacker were to paraphrase\footnote{Paraphrases are sentences or phrases with the same semantic meanings but differing wordings. In linguistics, defining such equivalence is difficult. Hence, we use the quasi-paraphrase definition from \citet{bhagat-hovy-2013-squibs} as semantic equivalence.} the input text to remove the trigger words (as illustrated in \reffig{fig:attack-example}) before querying EaaS for their embeddings in an imitation attack. More concretely, the paraphrasing attack will create a pool of paraphrase embeddings (\ie embeddings from multiple paraphrases) for a given text. These embeddings are then combined into a single embedding, which will further dilute the watermark. The combined embeddings are subsequently used to train the attacker model. 

In this chapter, we evaluate the \textit{robustness} of the existing EaaS watermark defences against this paraphrasing attack. 
Our research aims to introduce this new type of attack that future EaaS defences will need to consider.
This chapter is organised as follows: we first explain the attack methodology in \refsec{sec:attack-method}. This is followed by experiments and results in \refsec{sec:attack-exps}. Finally, we conclude and highlight the need for improved watermark techniques for EaaS in \refsec{sec:attack-conclusion}.

\section{Methodology}
\label{sec:attack-method}

In this section, we define the threat model of the conventional EaaS watermark for imitation attacks in \refsecnum{sec:threat-model}. We then introduce the design of our paraphrasing attack in \refsecnum{sec:para-attack} and setups in \refsec{sec:para-attack-setups}. This is followed by a theoretical analysis of such paraphrasing on watermark weights in \refsecnum{sec:attack-theoretical-proof}.

\subsection{Threat Model}
\label{sec:threat-model}
In the context of watermarking schemes, the security of a system must be assessed for a specific threat space. Following \citet{peng2023you}, we now specify the setting outlining the attackers' knowledge, capabilities, and objectives for the EaaS watermarking scheme below:

\paragraph{Attacker's Knowledge.} This refers to the details an attacker possesses regarding the system.
We assume that the training data, model architecture, and methodologies of the victim EaaS ($\mathbb{S}_v$) are unknown (seen as a ``black box'') to the attacker. Furthermore, the attacker accesses the EaaS API as a normal service user. Additionally, the attacker has a dataset $D_a$ to query the service $\mathbb{S}_v$.

\paragraph{Attacker's Capability.}
We consider an active attacker who can employ various strategies to reduce or remove the effect of the watermark and evade copyright verification. We assume the attacker has a sufficient budget to query the EaaS provider $\mathbb{S}_v$ to obtain embeddings $E_a = \{ \ve_i = \mathbb{S}_v(s_i) | s_i \in D_a \}$ and train a model $\Theta_a$ using the texts from $D_a$ as input and $E_a$ as output. Using this model $\Theta_a$, the attacker provides a competitive EaaS service $\mathbb{S}_a$.

\paragraph{Attacker's Objective.} The attacker intends to steal the victim model $\Theta_v$ and provide a similar service $\mathbb{S}_a$ at a lower price, as this approach is more cost-effective than training a model from scratch. As shown in \citet{xu-etal-2022-student}, it could even be possible for the attacker to achieve higher performance in the attacker model.

\subsection{Paraphrasing Attack}
\label{sec:para-attack}
\begin{figure*}[t]
    \centering
    \includegraphics[width=0.99\textwidth,keepaspectratio]{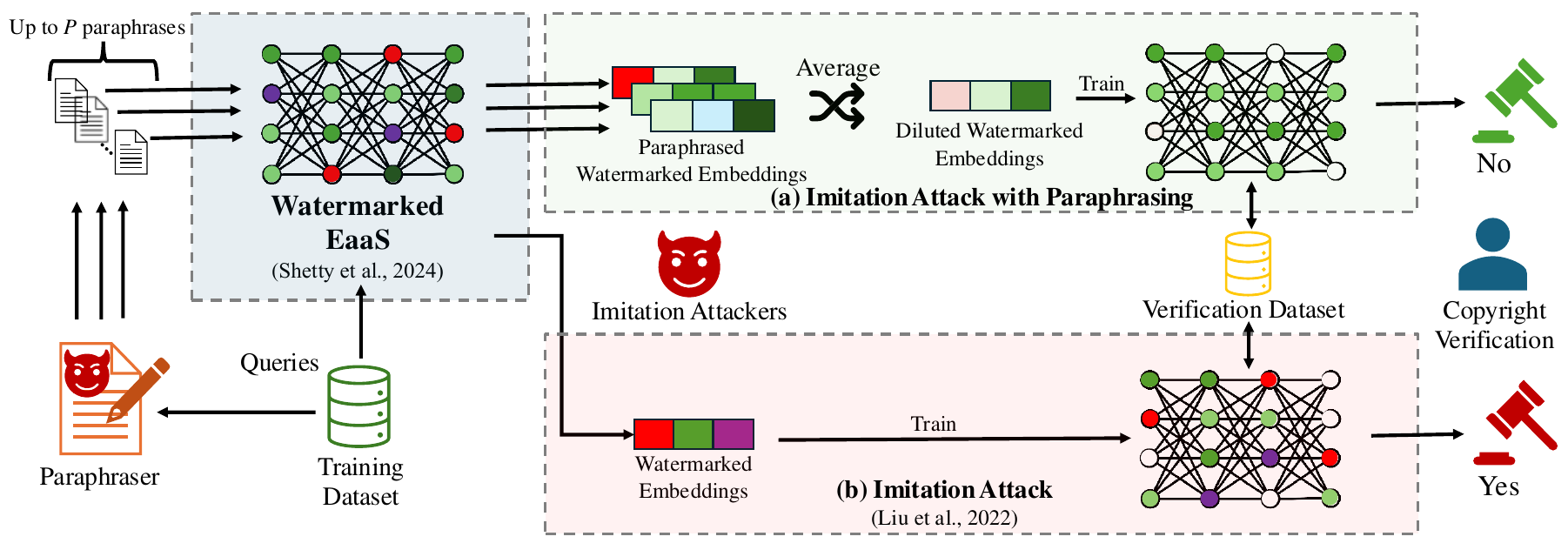}
    \caption[An overview of our paraphrasing attack.]{An overview of our paraphrasing attack, where the (a) \colorbox{OliveGreen!25}{Green} area shows the EaaS watermarks (presented as the elements in \textcolor{red}{Red}) getting diluted due to paraphrasing and potentially bypassed. On the contrary, the (b) \colorbox{red!25}{Red} area denotes a traditional imitation attack without paraphrasing, resulting in copyright infringement. 
    }
    \label{fig:attack-overview}
\end{figure*}

We propose generating multiple paraphrase embeddings and using their averaged embedding to train the attacker model so as to bypass the detection of embedding watermarks; see \reffig{fig:attack-overview} for an illustration.

Formally, we generate $P$ paraphrased texts 
$S_P = \{ s^{1},\cdots,s^{P} \}$ 
given an {input text} $s \in D_a$. 
Note that attackers have the freedom to choose any feasible paraphrasing methods, of which the victims are unaware.
Next, we query the $\mathbb{S}_{v}$ to obtain their paraphrase embeddings and aggregate them into a single embedding by averaging:
    \begin{equation}
        \begin{aligned}
            E_a &= \bigl\{ \mathbb{S}_{v}(s^{ i }) \bigr\}_{i=1}^{P}, \\ 
            \text{avg}(E_a) &= \sum\limits_{\ve \in E_a} \frac{\ve}{|E_a|}.
        \end{aligned}
        \label{eq:para-attack}
    \end{equation}
We will then use the aggregated embeddings ($\text{avg}(E_a)$; illustrated as the ``Diluted Watermarked Embeddings'' in \reffig{fig:attack-overview} for training the attacker model in an imitation attack.

To measure the success of this paraphrasing attack, we will evaluate embeddings' downstream task performance (\textit{utility}) and watermark verification accuracy (\textit{verifiability}), as detailed in \refsec{sec:metrics}. 
Recall that the attacker's objective is to avoid watermark detection while preserving the utility of the embeddings on target downstream tasks.
Therefore, we consider the attack successful if the downstream task performance is high and watermark verification accuracy is low.

\subsection{Paraphrasing Attack Setups}
\label{sec:para-attack-setups}
To ensure a comprehensive evaluation of the paraphrasing attack, 
we experiment with three paraphrasing methods: (i) \gpt\footnote{The exact model name is \texttt{gpt-3.5-turbo-0125}.} with prompts;
(ii) \dipper \citep{krishna2024paraphrasing}, a paraphrase model; and (iii) round-trip translation using \nllb \citep{NLLB}, a multilingual translation model. 
With these three paraphrasing setups, we can test how sensitive the attack is to the paraphrasing methods.

Note that for all paraphrasing attacks, we use these aforementioned methods to generate $P=5$ paraphrases once; \ie we do not make repeated queries. We also filter out poor paraphrases, and in practice, after deduplication and filtering, there will be less than $P$ paraphrases in some cases.
We provide further details about these paraphrasing methods (\eg prompts, model configurations, 
translation setups) next.

\subsubsection*{GPT-3.5}
In this attack setup, we use the \gpt model, which powers the popular \texttt{ChatGPT} \citep{chatgpt}, to paraphrase the text.
We prompt the \gpt model using the maximum token
length of 1000 and temperature of 0.7\footnote{A higher temperature means more diversity in the generation; a generation temperature of 0 reduces the sampling process to selecting the token with the highest probability \citep{2022-EaaS-OpenAI}.} and ask the model to paraphrase an input text. We evaluate two prompts:

\begin{tcolorbox}
[colback=gray!10!white,
colframe=black!75,fonttitle=\bfseries,
colbacktitle=gray!70!white,enhanced,
attach boxed title to top center={yshift=-2mm}, title={\textsf{PROMPT 1} {\color{black}\footnotesize\citep{kirchenbauer2024on}}},fonttitle=\bfseries]
\textsf{As an expert copy-editor, please rewrite the following text in your own voice while ensuring that the final output contains the same information as the original text and has roughly the same length. Please paraphrase all sentences and do not omit any crucial details. Additionally, please take care to provide any relevant information about public figures, organisations, or other entities mentioned in the text to avoid any potential misunderstandings or biases.}
\label{prompt1}
\end{tcolorbox}

\begin{tcolorbox}
[colback=gray!10!white,
colframe=black!75,fonttitle=\bfseries,
colbacktitle=gray!70!white,enhanced,
attach boxed title to top center={yshift=-2mm}, title={\textsf{PROMPT 2} {\color{black}\footnotesize\citep{RTT-wm}}},fonttitle=\bfseries]
\label{prompt2}
\textsf{You are a helpful assistant to rewrite the text. Rewrite the following text:}
\end{tcolorbox}

We first test the performance of \textsf{PROMPT 1} \citep{kirchenbauer2024on} and find that it is subpar because \textsf{PROMPT 1} has a tendency to hallucinate and expand on short input text. This leads to a significant deviation of meaning and content from the original text. Moreover, such hallucination also increases the likelihood of incorporating additional trigger words. \textsf{PROMPT 2}, on the other hand, paraphrases without substantially changing the core meaning of the input text. Henceforth, all experiments 
use \textsf{PROMPT 2}. We present some examples of the paraphrase output by these two prompts in \reftab{table:praprhrasing-examples}. 

\subsubsection*{DIPPER}
\label{app:dipper-exp-setting}

We also experiment with an \textit{open-source} off-the-shelf paraphraser model, \dipper \citep{krishna2024paraphrasing}, as an alternative model for paraphrasing the input text. 
\dipper is an LLM fine-tuned on a dataset of paragraph-level paraphrases specifically for performing paraphrasing.
Unlike previous paraphrasing models in the literature, which were single sentence based, \dipper is discourse-based, \ie it uses multiple sentences in context while paraphrasing.
\citet{krishna2024paraphrasing} found that \dipper performs on par with \gpt models in terms of the diversity and quality of paraphrases.

\dipper has two hyperparameters, \texttt{lex} and \texttt{div}, that control the paraphrasing quality. \texttt{lex} handles the lexical diversity of words in the paraphrased text, whereas \texttt{div} controls the content (or sentence) reordering in paraphrases. \dipper accepts the following values for these hyperparameters: 

Lexical Diversity (\texttt{lex}): \{0, 20, 40, 60, 80, 100\}.

Order Diversity (\texttt{div}): \{0, 20, 40, 60, 80, 100\}. 

These values correspond to ``similarity'' control for paraphrases, increasing from left to right. 
For example, \texttt{lex} = 40 roughly corresponds to a 40\% lexical similarity between the original text and its paraphrase.
For more details, please refer to \citet{krishna2024paraphrasing}.

We adopt the following ``moderate'' setting for all our \dipper experiments: \texttt{lex} = 40 and \texttt{div} = 40. In preliminary experiments, we found that this configuration still produces significant changes to the text while preserving the content and meaning of the input. We present some example \dipper paraphrasing outputs in \reftab{table:dipper-praprhrasing-examples}. 

\begin{table*}[t]
\centering
    \begin{minipage}{0.99\textwidth}
    \resizebox{\textwidth}{!}{%
    \normalsize
    \begin{tabular}{c>{\normalsize}p{0.3\textwidth}>{\small}p{0.64\textwidth}>{\normalsize}p{0.3\textwidth}}

        \toprule[1.5pt]
        \multirow{2}{*}{\textbf{Dataset}} & \multirow{2}{*}{\textbf{Input Text}} & \multicolumn{2}{c}{\textbf{Paraphrase Output}} \\
        \cmidrule{3-4}
        {} & {} & {\normalsize\texttt{PROMPT 1}\tiny} & {\normalsize\texttt{PROMPT 2}\tiny} \\
        \midrule
        \multirow{15}{*}{\rotatebox[origin=c]{90}{\textbf{\enron}}} 
        {} & \textsf{tenaska} & \textsf{Tenaska is a leading independent power producer in the United States, specializing in developing, owning, and operating power plants. With a focus on natural gas and renewable energy projects, Tenaska has a strong track record of success in the energy industry. They are known for their innovative approach to project development and commitment to sustainability. Tenaska's projects have helped to provide reliable and affordable energy to communities across the country.} & \textsf{Tenaska} \\
        \cmidrule{2-4}
        {} & \textsf{jones graduate school course descriptions list for 2000 - 2001} & \textsf{Course descriptions for the 2000-2001 academic year at the Jones Graduate School are available. The list provides detailed information about the various courses offered during that period. Students can refer to these descriptions to learn more about the content and requirements of each course before enrolling. It is recommended that students review the list carefully to make informed decisions about their academic schedule.} & \textsf{List of course descriptions for the 2000-2001 academic year at Jones Graduate School} \\
        \midrule
        \multirow{9}{*}{\rotatebox[origin=c]{90}{\textbf{\sst}}} & \textsf{the picture runs a mere 84 minutes , but it 's no glance .} & \textsf{The film has a brief runtime of just 84 minutes, but it is far from hastily made.} & \textsf{The film is only 84 minutes long, but it is not just a quick look.} \\
         \cmidrule{2-4}
        {} & \textsf{a lot to do with the casting of juliette binoche as sand , who brings to the role her pale , dark beauty and characteristic warmth} & \textsf{Juliette Binoche's casting as Sand in the film played a significant role, as her pale complexion, dark beauty, and characteristic warmth brought a unique depth to the role.} & \textsf{The casting of Juliette Binoche as Sand is heavily influenced by her pale, dark beauty and characteristic warmth that she brings to the role.} \\
        \midrule
        \multirow{17}{*}{\rotatebox[origin=c]{90}{\textbf{\mind}}} & \textsf{Extending schools hours to help low-income families} & \textsf{Increasing the duration of school hours has been proposed as a way to assist low-income families. This initiative aims to provide children from disadvantaged backgrounds with additional educational opportunities and support. By extending the school day, students can have access to resources such as tutoring, mentoring, and extracurricular activities that may not be available to them outside of school hours. This program ultimately seeks to level the playing field for all students, regardless of their socioeconomic status, by offering them a more equitable education. While there may be logistical challenges to implementing longer school hours, the potential benefits for low-income families and their children are significant.} & \textsf{Increasing school hours to support low-income families} \\
        \cmidrule{2-4}
        {} & \textsf{University of Kentucky alum donates \$20 million to College of Law} & \textsf{University of Kentucky graduate generously donates \$20 million to the College of Law. This substantial contribution will undoubtedly have a profound impact on the institution, allowing for the enhancement of academic programs, resources, and facilities. The generous gift from the alum demonstrates a strong commitment to supporting legal education and the future success of students at the University of Kentucky. This significant donation will help advance the College of Law's mission and further solidify its reputation as a leading institution in legal studies.} & \textsf{An alum from the University of Kentucky has generously donated \$20 million to support the College of Law.} \\
        \midrule
        \multirow{20}{*}{\rotatebox[origin=c]{90}{\textbf{\agnews}}} & \textsf{Saudi Arabia to open up oil taps Saudi Arabia says it is ready to push an extra 1.3 million barrels a day of oil into the market, to help reverse surging prices.} & \textsf{The government of Saudi Arabia has announced plans to increase oil production by an additional 1.3 million barrels per day in an effort to stabilize rising oil prices. This decision is aimed at addressing the current market conditions and ensuring a sufficient supply of oil to meet global demand. By opening up its oil taps, Saudi Arabia hopes to alleviate pressure on prices and help bring stability to the oil market. This move highlights Saudi Arabia's commitment to supporting the stability of the global oil market and ensuring a reliable supply of oil for consumers worldwide.} & \textsf{Saudi Arabia is prepared to increase oil production by an additional 1.3 million barrels per day in order to address rising prices in the market.} \\
        \cmidrule{2-4}
        {} & \textsf{Volkswagen Says  \#39;No Room \#39; for Wage Hikes FRANKFURT, Germany Aug. 23, 2004 - Automaker Volkswagen said Monday there is  quot;no room quot; for wage increases this year as it prepares for pay talks with Germany \#39;s largest industrial union.} & \textsf{Volkswagen has stated that there will be no wage hikes for employees this year as the company gears up for negotiations with Germany's largest industrial union. This decision was announced on Monday, as the automaker emphasized that there is no capacity for salary increases at this time.} & \textsf{Volkswagen states that there is no space for wage hikes this year as it gets ready for negotiations with Germany's biggest industrial union.} \\
        \bottomrule[1.5pt]
    \end{tabular}}
    \end{minipage}
    \caption[\gpt paraphrasing examples using different prompts.]{\gpt paraphrasing examples using different prompts. \texttt{PROMPT 1} exhibits hallucination by adding extra information during paraphrasing, whereas \texttt{PROMPT 2} preserves the semantic meaning whilst lexical diversity.}
    \label{table:praprhrasing-examples}
\end{table*}

\begin{table}[t]
\centering
    \begin{minipage}{.95\textwidth}
    \resizebox{\textwidth}{!}{%
    \begin{tabular}{c>{\normalsize}p{0.45\textwidth}>{\normalsize}p{0.45\textwidth}}

        \toprule[1.5pt]
        \multirow{1}{*}{\textbf{Dataset}} & \multirow{1}{*}{\textbf{Input Text}} & {\textbf{Paraphrase Output}} \\
        \midrule
        \multirow{3}{*}{\rotatebox[origin=c]{90}{\textbf{\enron}}} & \textsf{tenaska} & \textsf{Tenaska} \\
        \cmidrule{2-3}
        {} & \textsf{jones graduate school course descriptions list for 2000 - 2001} & \textsf{JONES Graduate School Courses for the Academic Year 2000-2001} \\
        \midrule
        \multirow{6}{*}{\rotatebox[origin=c]{90}{\textbf{\sst}}} & \textsf{the picture runs a mere 84 minutes , but it 's no glance .}  & \textsf{At a mere 84 minutes long, the picture is not worth a glance.} \\
         \cmidrule{2-3}
        {} & \textsf{a lot to do with the casting of juliette binoche as sand , who brings to the role her pale , dark beauty and characteristic warmth}  & \textsf{A great deal of it is due to the casting of Juliette Binoche as Sand, who gives to the part her dark, pallid beauty and her characteristic warmth.} \\
        \midrule
        \multirow{4}{*}{\rotatebox[origin=c]{90}{\textbf{\mind}}} & \textsf{Extending schools hours to help low-income families} & \textsf{Extending school hours to help families with low incomes} \\
        \cmidrule{2-3}
        {} & \textsf{University of Kentucky alum donates \$20 million to College of Law} & \textsf{Alumnus of University of Kentucky gives \$20 million to College of Law} \\
        \midrule
        \multirow{12}{*}{\rotatebox[origin=c]{90}{\textbf{\agnews}}} & \textsf{Saudi Arabia to open up oil taps Saudi Arabia says it is ready to push an extra 1.3 million barrels a day of oil into the market, to help reverse surging prices.} & \textsf{Saudi Arabia to open the taps of its oil. Saudi Arabia has made the announcement that it is ready to increase its oil production by 1.3 million barrels per day, in order to reduce the escalation in the price of this commodity.} \\
         \cmidrule{2-3}
        {} & \textsf{Volkswagen Says  \#39;No Room \#39; for Wage Hikes FRANKFURT, Germany Aug. 23, 2004 - Automaker Volkswagen said Monday there is  quot;no room quot; for wage increases this year as it prepares for pay talks with Germany \#39;s largest industrial union.} & \textsf{FRANKFURT, Germany, August 23, 2004 – Automaker Volkswagen said on Monday that there is “no room” for pay increases this year, as it prepares to enter into negotiations with Germany’s biggest industrial union.} \\
        \bottomrule[1.5pt]
    \end{tabular}}
    \end{minipage}
    \caption[\dipper paraphrasing examples.]{\dipper paraphrasing examples. The quality of paraphrases appears to be on par with those generated using \gpt (\reftab{table:praprhrasing-examples}).}
    \label{table:dipper-praprhrasing-examples}
\end{table}

\subsubsection*{Round-Trip Translation}
\label{app:nllb-exp-setting}
\begin{table}[t]
    \centering
    {\normalsize
    \centering
    \begin{tabular}{cccc}

        \toprule[1.5pt]
        {\textbf{Language}} & \textbf{IDO 639-I} & \textbf{ISO 639-2/T} & \textbf{Language Family} \\
        \midrule
        \textsc{Chinese (Simpl)} & \texttt{zh} & \texttt{zho\_simpl} & Sino-Tibetan \\
        \textsc{Japanese} &  \texttt{ja} & \texttt{jpn} & Other \\
        \textsc{French} &  \texttt{fr} & \texttt{fra} & Indo-European-Romance \\
        \textsc{German} & \texttt{de} & \texttt{deu} & Indo-European-Germanic \\
        \textsc{Hindi} &  \texttt{hi} & \texttt{hin} & Indo-European-Indo-Aryan \\
        \bottomrule[1.5pt]
    \end{tabular}}
    \caption[Different pivot languages used for RTT paraphrasing.]{Different pivot languages used for RTT paraphrasing. For each language, its ISO codes and family are listed \citep{zhu-etal-2024-multilingual}.}
    \label{table:RTT-lang-info}
\end{table}

\begin{CJK}{UTF8}{gbsn}
\begin{table*}[t]
\centering
    \begin{minipage}{0.95\textwidth}
    \resizebox{\textwidth}{!}{%
    \begin{tabular}{c>{\normalsize}p{0.35\textwidth}>{\normalsize}p{0.35\textwidth}>{\normalsize}p{0.35\textwidth}}
        \toprule[1.5pt]
        \multirow{2}{*}{\textbf{Dataset}} & \multirow{2}{*}{\textbf{Input Text}} & \multicolumn{2}{c}{\textbf{Translate Output}} \\
        \cmidrule{3-4}
        {} & {} & {\textsc{Chinese}} & {\textsc{English} (Re-translated)} \\
        \midrule
        \multirow{3}{*}{\rotatebox[origin=c]{90}{\textbf{\enron}}} & \textbf{tenaska} & \textsf{特纳斯卡} & \textsf{Fish} \\
        \cmidrule{2-4}
        {} & \textsf{jones graduate school course descriptions list for 2000 - 2001} & \textsf{琼斯研究生院2000-2001年课程描述清单} & \textsf{List of courses at SUS 2000-2001} \\
        \midrule
        \multirow{7}{*}{\rotatebox[origin=c]{90}{\textbf{\sst}}} & \textsf{the picture runs a mere 84 minutes , but it 's no glance .} & \textsf{图片只有84分钟，但它并不是一瞥。} & \textsf{The film is only 84 minutes long, but it is not a one-shot.} \\
         \cmidrule{2-4}
        {} & \textsf{a lot to do with the casting of juliette binoche as sand , who brings to the role her pale , dark beauty and characteristic warmth} & \textsf{对于朱丽叶·比诺什扮演桑德这一角色的选角很重要，她将自己苍白、深邃的美和特有的温暖带入了角色中。} & \textsf{Many of the characters are related to Juliette Binoche's role as Sand, but the main character is a young girl named "Sand".} \\
        \midrule
        \multirow{5}{*}{\rotatebox[origin=c]{90}{\textbf{\mind}}} & \textsf{Extending schools hours to help low-income families} & \textsf{延长学校时间，以帮助低收入家庭。} & \textsf{Extended school hours to help low-income families} \\
        \cmidrule{2-4}
        {} & \textsf{University of Kentucky alum donates \$20 million to College of Law} & \textsf{肯塔基大学校友向法学院捐赠2000万美元。} & \textsf{University of Kentucky alumni donate \$20 million to law school} \\
        \midrule
        \multirow{16}{*}{\rotatebox[origin=c]{90}{\textbf{\agnews}}} & \textsf{Saudi Arabia to open up oil taps Saudi Arabia says it is ready to push an extra 1.3 million barrels a day of oil into the market, to help reverse surging prices.} & \textsf{沙特阿拉伯打开石油水龙头 沙特阿拉伯表示，已准备好每天向市场增加130万桶石油，以帮助扭转油价的激增。} & \textsf{Saudi Arabia says it is ready to add 13 million barrels of oil a day to the market, and is planning to increase production by another 10 million barrels a day.} \\
        \cmidrule{2-4}
        {} & \textsf{Volkswagen Says  \#39;No Room \#39; for Wage Hikes FRANKFURT, Germany Aug. 23, 2004 - Automaker Volkswagen said Monday there is  quot;no room quot; for wage increases this year as it prepares for pay talks with Germany \#39;s largest industrial union.} & \textsf{大众汽车表示今年没有涨薪的空间，因为他们正准备与德国最大的工会进行工资谈判。} & \textsf{Frankfurt, 23 August 2004 - The carmaker Volkswagen said on Monday that there was "no room" for a wage increase this year as it prepared to negotiate with Germany's largest union.} \\
        \bottomrule[1.5pt]
    \end{tabular}}
    \end{minipage}
    \caption[\nllb RTT (\textsc{English} $\rightarrow$ \textsc{Chinese} $\rightarrow$ \textsc{English}) paraphrasing examples.]{\nllb RTT (\textsc{English} $\rightarrow$ \textsc{Chinese} $\rightarrow$ \textsc{English}) paraphrasing examples. Compared with \reftab{table:praprhrasing-examples} and \reftab{table:dipper-praprhrasing-examples}, the quality of paraphrases is subpar. This is further investigated in \refsec{sec:para-quality}, where no notable issues were observed.}
    \label{table:RTT-examples}
\end{table*}
\end{CJK}

Round-trip translation (RTT) \citep{he2024can} involves translating text to another language (\aka pivot language) and then translating back to the original language (\eg \textsc{English} $\rightarrow$ \textsc{German} $\rightarrow$ \textsc{English}). It is commonly used for evaluating machine translation systems because a good translation system should produce a similar final output compared to the original input \citep{somers-2005-round}.  We explore pivot languages from various language families compared to English (our original language), to increase the chance of having significant modifications to the original input. Similar to other attack setups, we use $P=5$; hence, we consider five pivot languages listed in \reftab{table:RTT-lang-info}. 

We use the \nllb \citep{NLLB} 1.3B model variant, an open-source state-of-the-art multilingual translation model.\footnote{While LLMs have made significant progress, Google Translate (\url{https://translate.google.com/}) is still state-of-the-art for multilingual translations \citep{zhu-etal-2024-multilingual}. However, 
we estimate it would cost
\$8,600, to use it for round-trip translation; hence making it unaffordable.
} We present a few RTT paraphrasing examples with \textsc{Chinese} as the pivot language in \reftab{table:RTT-examples}.

\subsection{Watermark Dilution with Paraphrasing}
\label{sec:attack-theoretical-proof}
To recap, \firstWM and \secondWM mix secret \textit{target embeddings} into original embeddings proportionally based on the number of \textit{trigger words} present in the input text. This proportion is also known as \textit{watermark weights}.

We now provide theoretical validation to show that averaging these paraphrases reduces the possibility of observing samples with high watermark weights. Lower watermark weight means the watermark is diluted and, as a result, less transferable in an imitation attack.

We provide an analysis of the impact of paraphrasing on watermarking weights in a simplified setting as follows:
\begin{enumerate}
    \item Each text token has a probability $\mathbb{P}_t$ of being in the trigger words set $t$.
    \item Sentence length is fixed $|S|$.
    \item Sentences with equal or more than one trigger acquire the same watermark weight controlled by $\lambda > 0$. 
    \item $Q$ is the random variable capturing the likelihood of sentences having trigger words.
\end{enumerate}

As per the above assumptions, the probability of a sentence $S$ with length $|S|$ having trigger words is:
\begin{align}
    \mathbb{P}_S = 1 - ( 1 - \mathbb{P}_t)^{|S|}.
    \label{eq:prob-t}
\end{align} 
\secondWM uses trigger word frequency of $[ 0.5\%, 1\% ]$. Therefore, assuming $\mathbb{P}_t = 0.005$ and $|S| = 50$, as per \Eqref{eq:prob-t} we get $\mathbb{P}_S = 0.222$.

For the single-sentence case, the watermark weight is $\lambda \cdot Q_{S}$, where
    \begin{align}
        Q_{S} &\sim \text{Bernoulli}(\mathbb{P}_S). \notag
    \end{align}
In other words, it is a Bernoulli trial determining whether the sentence contains trigger words, governed by $\mathbb{P}_S$.

For the case where $P$ paraphrases are averaged, the watermark weight is $\lambda \cdot Q_{P}$, where
    \begin{align}
        Q_{P} &= \frac{X_{P}}{P}, \quad     \notag
        X_{P} = \sum_{i=1}^{P}Q_S^{i}, \\ 
        \notag
        X_{P} &\sim \text{Binomial}(P, \mathbb{P}_S).
    \end{align}
$X_{P}$ captures how many sentences out of $P$ contain trigger words, which will be a binomial distribution of $P$ independent Bernoulli trials governed by $\mathbb{P}_S$.

\begin{figure}[t]
    \centering
    \includegraphics[width=0.75\linewidth,keepaspectratio]{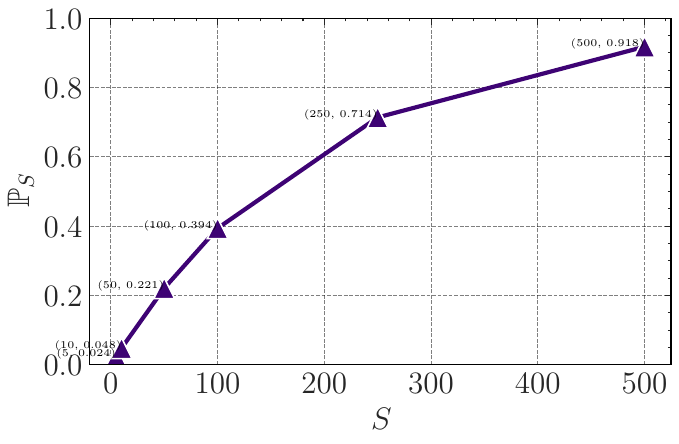}
    \caption[$\mathbb{P}_S$ for different values of text length $|S|$.]{$\mathbb{P}_S$ for different values of text length $|S|$. Expectedly, with increasing text length $|S|$, the likelihood ($\mathbb{P}_S$) of trigger words in the text increases.
    }
    \label{fig:diff-S-Pts}
\end{figure}

When $P=10$, we observe that $\mathbb{P}(Q_S > a) > \mathbb{P} (Q_{P} > a)$ for all $a > 0.3$. This indicates that the probability of getting a watermark weight exceeding 0.3 is higher when $P=1$ (1 paraphrase) \versus $P=10$ (10 paraphrases).
Similarly, when $P=5$, we find that $\mathbb{P}(Q_S > a) > \mathbb{P} (Q_{P} > a)$ for all $a > 0.4$. 
These observations suggest that paraphrasing multiple times increases the likelihood of getting lower watermark weights. 

We want to note that this paraphrasing effect diminishes with increasing text length ($|S|$) because we have a higher probability ($\mathbb{P}_S$) of having trigger words in the sentence (as seen in \reffig{fig:diff-S-Pts}), \ie paraphrasing will be less effective as the input text length increases. To summarise, this theoretical analysis supports the \textbf{RQ1} hypothesis that paraphrasing could be an effective attack against current embedding watermarks. We will now validate this theoretical finding empirically.

\section{Experiments}
\label{sec:attack-exps}

In this section, we present our experimental results to assess the strengths and limitations of our proposed paraphrasing attack. We first describe the datasets (\refsec{sec:datasets}) and metrics (\refsec{sec:metrics}) used, followed by the experimental settings in \refsec{sec:ch3-exp-setting}. In \refsec{sec:empirical-backdoor-analysis}, we empirically study the influence of paraphrasing on watermark weights and validate the theoretical findings from \refsec{sec:attack-theoretical-proof}. Next, we present the main results for paraphrasing attacks in \refsec{sec:attack-perf}, where we show that paraphrasing circumvents EaaS watermarks while preserving the utility of the embeddings. As paraphrases are the main component of our attack, we investigate their quality in \refsec{sec:para-quality}. Finally, we perform detailed ablation studies to examine the impact of different components 
of the paraphrasing attack, to understand potential confounds in \refsec{sec:attack-ablation}.

\subsection{Datasets}
\label{sec:datasets}
Following existing works by \citet{shetty2024warden} and \citet{peng2023you}, datasets from various domains are considered to benchmark our attack.
These datasets cover a range of prevalent NLP tasks, providing insights into the {utility} of the embeddings.

\begin{itemize}
    \item \textbf{\enron} \citep{enron} is a spam email classification dataset consisting of data from about 150 employees of Enron Corporation. The subject lines of these emails are used as the input text in our experiments.
    \item \textbf{\sst} \citep{sst2} consists of single sentences (used as input text) expressing sentiments about films from the Stanford Sentiment Treebank corpus. This dataset is used for binary (positive or negative) sentiment classification.
    \item \textbf{\mind} \citep{mind} is a large-scale news recommendation dataset containing detailed information such as title, abstract, body, and other metadata. In our experiments, we consider the news title as the input text. \mind has added complexity for the classification task due to the high number (18) of classes. 
    \item \textbf{\agnews} \citep{ag_news} is another news classification dataset derived from news articles from over 2000 sources. We use the description field of the news article as the input text, which often comprises multiple sentences or a paragraph, unlike the single-sentence inputs of the previous datasets.
\end{itemize}

\reftab{table:dataset-statistics} presents some statistics of these datasets.

\begin{table}[t]
    \centering
    \begin{tabular}{cccccS}

        \toprule[1.5pt]
        {\textbf{Dataset}} & {\textbf{\# Class}} & {\textbf{\# Train}} & {\textbf{\# Test}} & {\textbf{Avg. Len.}} \\
        
        \midrule
        {\enron} & {2} & {31,716} & {2,000} & {34.57} \\
        {\sst} & {2} & {67,349} & {872} & {54.17} \\
        {\mind} & {18} & {97,791} & {32,592} & {66.14} \\
        {\agnews} & {4} & {120,000} & {7,600} & {236.41} \\
        \bottomrule[1.5pt]
    \end{tabular}
    \caption{Statistics of datasets used in the experiments.}
    \label{table:dataset-statistics}
\end{table}

\subsection{Metrics}
\label{sec:metrics}
To evaluate the effectiveness of the paraphrasing attack, we use the following metrics to assess downstream task {utility} and watermark {verifiability}.

\paragraph{Downstream Task Utility.}
\label{sec:para-down-utility}

Using the EaaS embeddings ($\ve_p$) as input, we build multi-layer perceptron classifiers for a range of classification tasks predicting dataset class labels and evaluating the accuracy (\acc) and $F_1$-score (\fscore) performance. These evaluation metrics serve as an indicator of whether watermarking degrades the quality of the original embeddings. Ideally, there should be minimal performance differences between the watermarked and original embeddings.

\acc is the proportion of correct decisions out of the total number of samples,
\begin{align}
    \acc = \frac{\text{Number of Correct Decisions}}{\text{Total Number of Decisions}}.
    \notag
\end{align}

We value false positives (FPs) and false negatives (FNs) equally; consequently, the \fscore will be computed as follows:
\begin{align}
    \fscore = \frac{2 \times \text{Precision} \times \text{Recall}}{\text{Precision} + \text{Recall}} ,
    \notag
\end{align}
where:
\begin{align}
    \text{Precision} = \frac{\text{TPs}}{\text{TPs} + \text{FPs}}, \quad 
    \text{Recall} = \frac{\text{TPs}}{\text{TPs} + \text{FNs}}.  \notag
\end{align}

\paragraph{Watermark Verifiability.} 

To verify the watermark, we compare the embeddings of the attacker model for two verification datasets (same as \refsec{sec:warden}): (i) one containing texts with trigger words (backdoor set, $E_w$) and (ii) the other containing texts with no trigger words (benign set, $E_c$).
Intuitively, we should find higher similarity between the target embedding $\vw$ with $E_w$ than with $E_c$.
In our experiments, we use cosine similarity to measure similarity.\footnote{Although $l_2$ distance is also used as a similarity metric conventionally, we found similar performances in our experiments (in the case of normalised embeddings, $l_2 = \sqrt{2(1 - \cos)}$) and thus did not include it.
} 

Formally, we compute the average cosine similarities between the embedding $\ve$ (from $E_w$ or $E_c$) and {target embedding} $\vw$ and then take their difference:\footnote{For simplicity, we show the $\Delta_{\cos}$ calculation for the \firstWM (single watermark). They are similar for the multi-watermark extension, \ie \secondWM; we direct readers to \citet{shetty2024warden} for more details.}
\begin{equation}
    \centering
    \begin{aligned}
    \cos_\text{avg} (E) = \frac{1}{|E|} \sum_{i=1}^{|E|} \cos(\ve_i, \vw), \\
    \Delta_{\cos} = \cos_\text{avg} (E_w) - \cos_\text{avg} (E_c). \notag \\
    \end{aligned}
\end{equation}

Ideally,  $\Delta_{\cos}$ should be high as it indicates the verification process can accurately identify $E_w$ without confusing it with $E_c$.
Another important metric is the area under the receiver operating characteristic curve (\auc). The \auc can be interpreted as the probability that we rank a random positive sample (from the backdoor set) higher than a random negative sample (from the benign set) \citep{mitchell2023detectgpt}. In practice, we set a threshold for the distance metric ($\Delta_{\cos}$) to ascertain whether an imitation attack has occurred. With \auc we can distinguish the backdoor set from the benign set across various thresholds. A perfect \auc (\ie 100) indicates that two sets are entirely distinct. We compute the \auc based on average cosine similarities. 

\subsection{Experimental Details}
\label{sec:ch3-exp-setting}

In terms of model configurations and hyperparameters, we largely follow the experimental settings outlined by \citet{shetty2024warden}.
\reftab{table:hyperparams} shows the complete hyperparameter settings.

\paragraph{Victim EaaS.}
To simulate the imitation attacks, we use \texttt{GPT-3 text-embedding-002} \citep{2022-EaaS-OpenAI} as the victim EaaS ($\mathbb{S}_v$) to retrieve the original (non-watermarked) embeddings.\footnote{All watermarking techniques experimented in this thesis (\firstWM, \secondWM, and \ourdefence (\refsecnum{sec:wet-defense})) inject watermarks post-hoc into the embeddings produced by the API calls.} 

\paragraph{Attacker Model.}
We use \texttt{BERT} \citep{devlin-etal-2019-bert} as the attacker model, with a two-layer feed-forward network attached to produce the copied embeddings in imitation attacks and a mean squared error loss for training as follows:
\begin{align}
    \notag
    \Theta_{a}^{*} = \arg \min_{\Theta_{a}} \mathbb{E}_{s \in D_a} \biggl|\biggl|\mathbb{S}_a(s;\Theta_a) - \mathbb{S}_v(s)\biggl|\biggl|_2^2,
\end{align}
where $\mathbb{S}_a$ is the attacker EaaS based on model $\Theta_a$ and $s$ is an input text from attacker dataset $D_a$. 
Additionally, we assume that the attacker model is trained solely on the watermarked outputs of the victim model.

\paragraph{Downstream Classifier.}
We employ $\ve_{p}$ as the input features for a two-layer feed-forward network classifier with the task of predicting the class label (replicating a downstream task) to assess the {utility} of the provided watermarked embedding $\ve_{p}$. The classifier is trained using the cross-entropy loss. 

\begin{table}[t]
\centering
    \begin{minipage}{.95\columnwidth}
    \resizebox{\columnwidth}{!}{%
    \begin{tabular}{cccccc}
\toprule[1.5pt]
    \textbf{Model}                              & \textbf{Hyperparameter}                     & \textbf{\sst}  & \textbf{\mind}  & \textbf{\agnews} & \textbf{\enron} \\ \hline
\multirow{2}{*}{Victim EaaS}  & Emb. Dim.  & 1,536 & 1,536 & 1,536   & 1,536      \\
                                  & Max. Tokens & 8,192 & 8,192 & 8,192   & 8,192      \\ \hline
{} & Learning Rate                   & $5\times 10^{-5}$  & $5\times 10^{-5}$  & $5\times 10^{-5}$ & $5\times 10^{-5}$  \\
                           {Attacker}       & Batch Size           & 32    & 32    & 32      & 32         \\
                          {Model}        & Hidden Size          & 1,536  & 1,536  & 1,536    & 1,536       \\
                          {}        & Dropout Rate         & 0.0   & 0.0   & 0.0     & 0.0        \\
                                  \hline
{}    & Learning Rate                   & $10^{-2}$  & $10^{-2}$  & $10^{-2}$    & $10^{-2}$       \\
                             {Downstream}     & Batch Size           & 32    & 32    & 32      & 32         \\
                            {Classifier}      & Hidden Size          & 256   & 256   & 256     & 256        \\
                           {}       & Dropout Rate         & 0.0   & 0.2   & 0.0     & 0.2        \\
\bottomrule[1.5pt]
\end{tabular}}
\end{minipage}
\caption{Hyperparameter settings of different models involved in the baseline EaaS watermarks and our experiments. 
}
\label{table:hyperparams}
\end{table}

\clearpage
\paragraph{Baseline EaaS Watermark Details}
We use the default settings of the baseline EaaS watermark methods (\firstWM and \secondWM) unless specified otherwise.
\begin{itemize}
    \item \textbf{\secondWM} (\refsecnum{sec:warden}): The size of the trigger word set $n$ is 20, and the maximum number of trigger words $m$ is 4, with a frequency interval for trigger words being [0.5\%, 1\%]. The number of watermarks $R$ is 4. 
    \item \textbf{\firstWM} (\refsecnum{sec:embmarker}): The settings remain the same as described for \secondWM above, with the number of watermarks ($R$) set to 1.
\end{itemize}

\begin{figure}[t]
    \centering
    \includegraphics[width=0.85\linewidth]{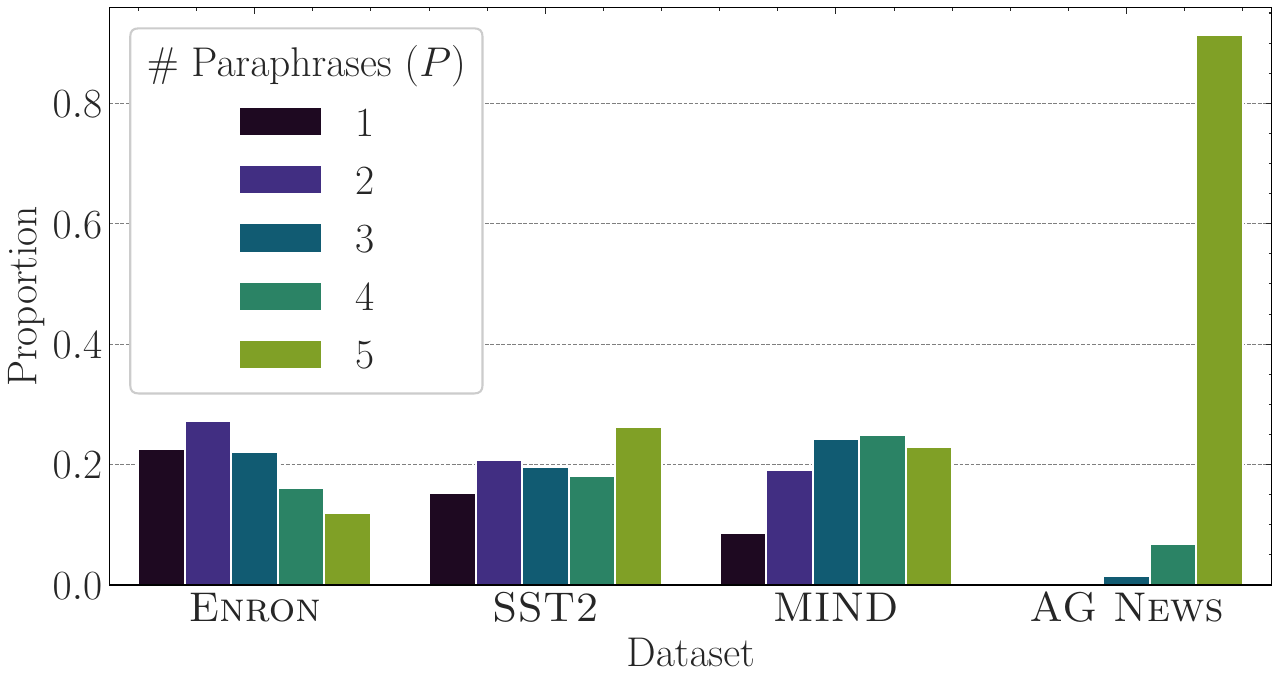}
    \caption{Distribution of the number of paraphrases ($P$) across samples in the datasets using \gpt.}
    \label{fig:para-cnt-stats}
\end{figure}

\paragraph{Paraphrases.} We first generate $P(=5)$ unique paraphrases for each input text and then filter out poor paraphrases based on cosine similarity between the original input text embedding and paraphrase embedding. 
Consequently, on average, we obtain 2.68, 3.30, 3.41, and 4.89 paraphrases using \gpt for \enron, \sst, \mind, and \agnews, respectively. \reffig{fig:para-cnt-stats} presents the distribution of the number of paraphrases. 

\subparagraph{Cost.} The approximate total number of tokens for the datasets are: \enron = 377K; \sst = 1M; \mind = 2M; and \agnews = 7M. Considering $P=5$ and assuming a similar output length as the input length, the expected cost of generating paraphrases using \gpt\footnote{\gpt Pricing (June 2024): Input - \$0.50/1M tokens and Output - \$1.50/1M  tokens.} would be under \$105.

\clearpage
\paragraph{Code and Compute Details.}
We extend the watermarking implementation by \citet{shetty2024warden}.\footnote{\url{https://github.com/anudeex/WARDEN.git}} 
We use the Huggingface \texttt{Transformers} \citep{wolf-etal-2020-transformers} framework for model development and utilise the \texttt{datasets} library \citep{lhoest-etal-2021-datasets} for data assessed in this work.
All experiments are conducted using a single \texttt{A100 GPU} with \texttt{CUDA 11.7} and \texttt{PyTorch 2.1.2}. We perform each main experiment independently five times (using different random seeds) and present the average results with the standard deviation to account for randomness.

\begin{figure}[p]
    \centering
    \begin{subfigure}{0.7\textwidth}
    \centering\includegraphics[width=\linewidth,keepaspectratio]{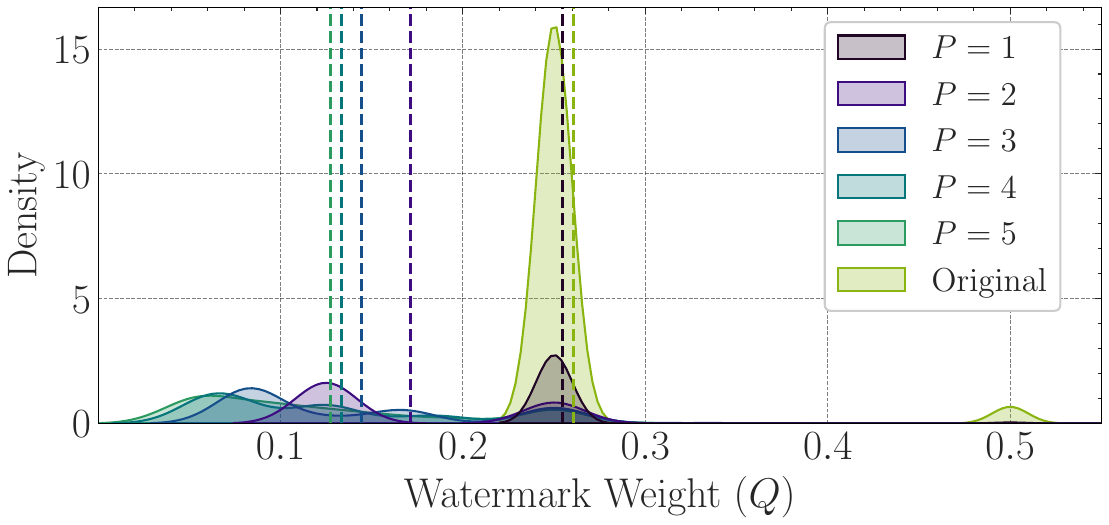}
    \caption{\enron}
    \end{subfigure}
    \begin{subfigure}{0.7\textwidth}
    \centering\includegraphics[width=\linewidth,keepaspectratio]{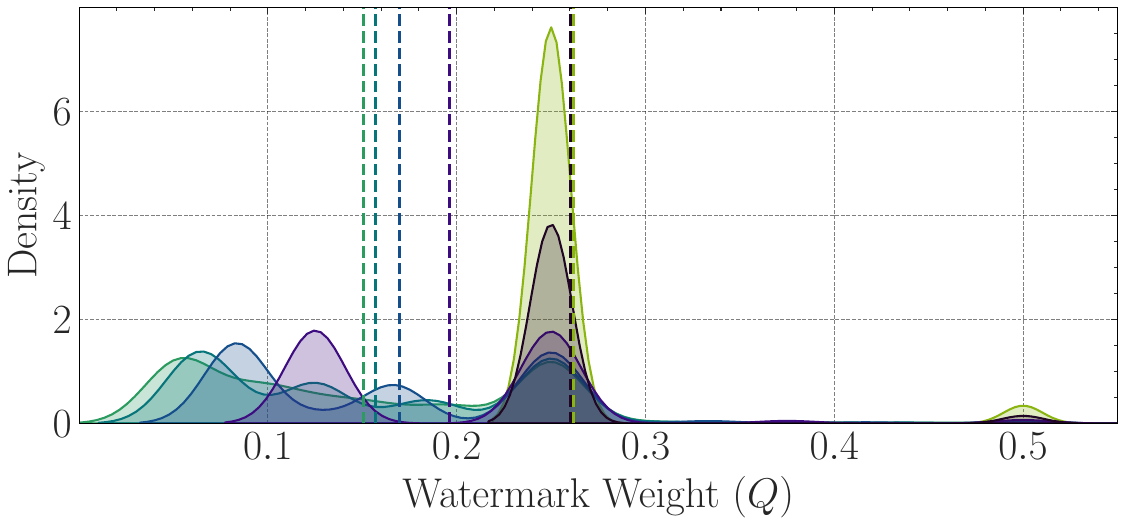}
    \caption{\sst}
    \end{subfigure}
    \begin{subfigure}{0.7\textwidth}
    \centering\includegraphics[width=\linewidth,keepaspectratio]{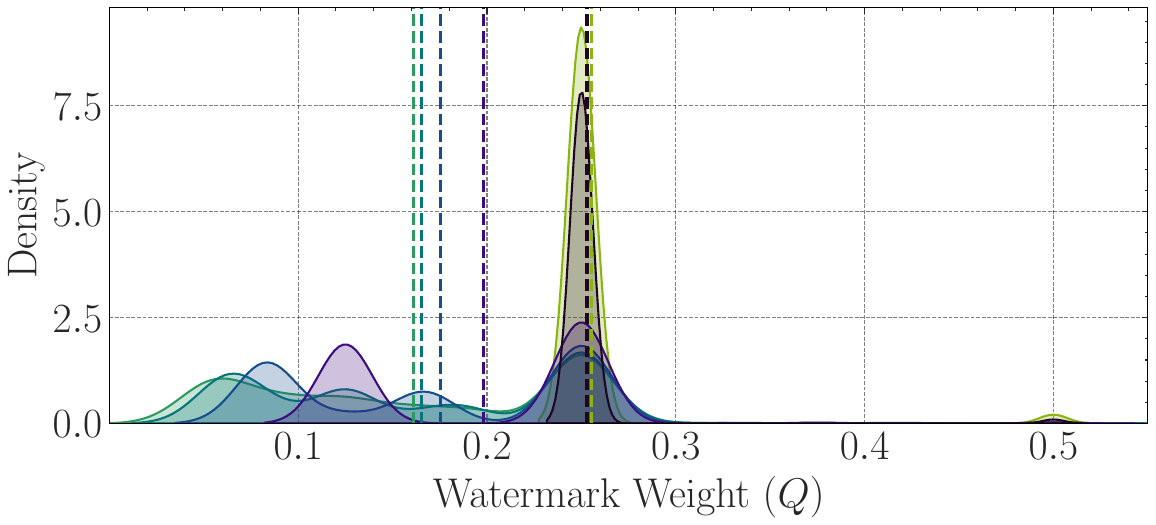}
    \caption{\mind}
    \end{subfigure}
    \begin{subfigure}{0.7\textwidth}
    \centering\includegraphics[width=\linewidth,keepaspectratio]{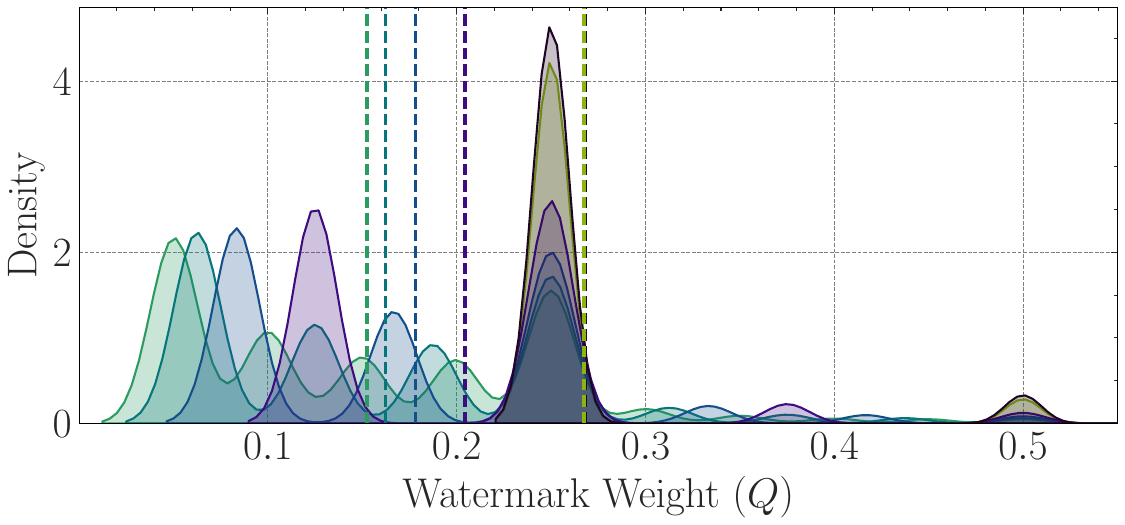}
    \caption{\agnews}
    \end{subfigure}
    \caption[Watermark weight distributions for the different number of \gpt paraphrases ($P$).]{Watermark weight distributions for different datasets (in subcaption) and the number of \gpt paraphrases ($P$). The dashed vertical lines denote the mean watermark weight. In general, aggregating watermarked embeddings from more paraphrases (larger $P$) reduces the watermark weights (with dashed vertical lines shifting leftwards).
    }
    \label{fig:backdoor-level-analysis}
\end{figure}

\begin{figure}[p]
    \centering
    \begin{subfigure}{0.7\textwidth}
    \centering\includegraphics[width=\linewidth,keepaspectratio]{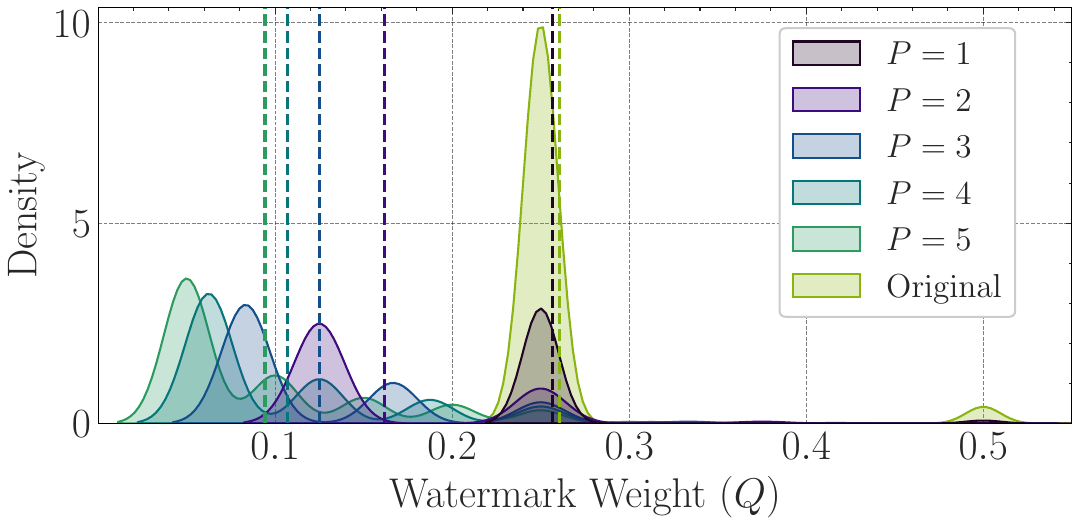}
    \caption{\enron}
    \end{subfigure}
    \begin{subfigure}{0.7\textwidth}
    \centering\includegraphics[width=\linewidth,keepaspectratio]{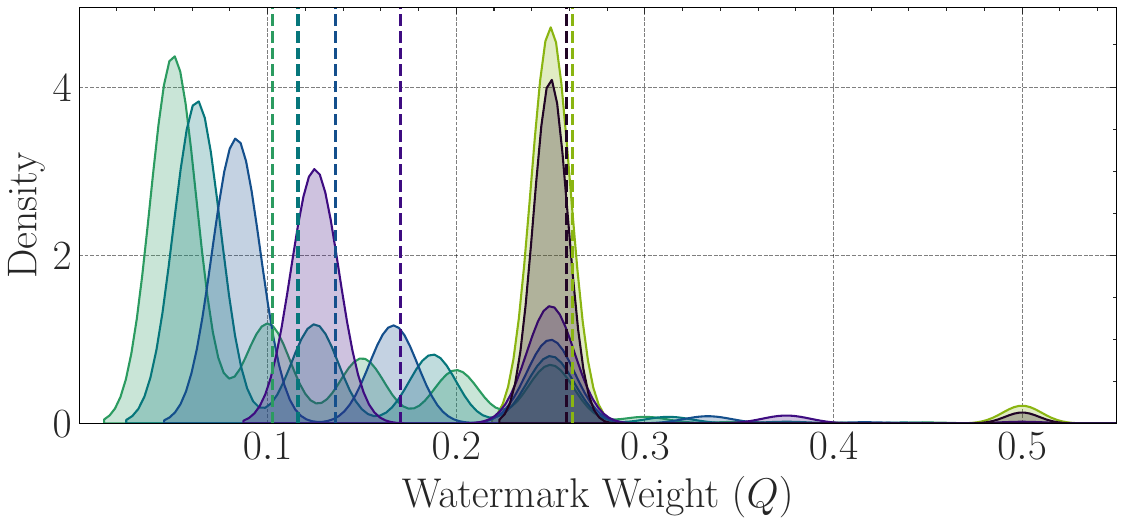}
    \caption{\sst}
    \end{subfigure}
    \begin{subfigure}{0.7\textwidth}
    \centering\includegraphics[width=\linewidth,keepaspectratio]{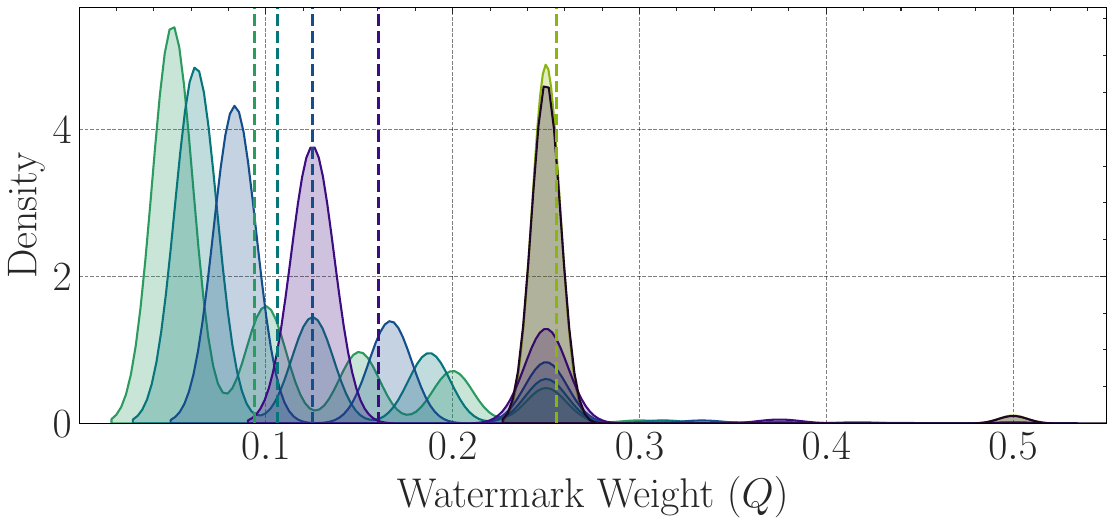}
    \caption{\mind}
    \end{subfigure}
    \begin{subfigure}{0.7\textwidth}
    \centering\includegraphics[width=\linewidth,keepaspectratio]{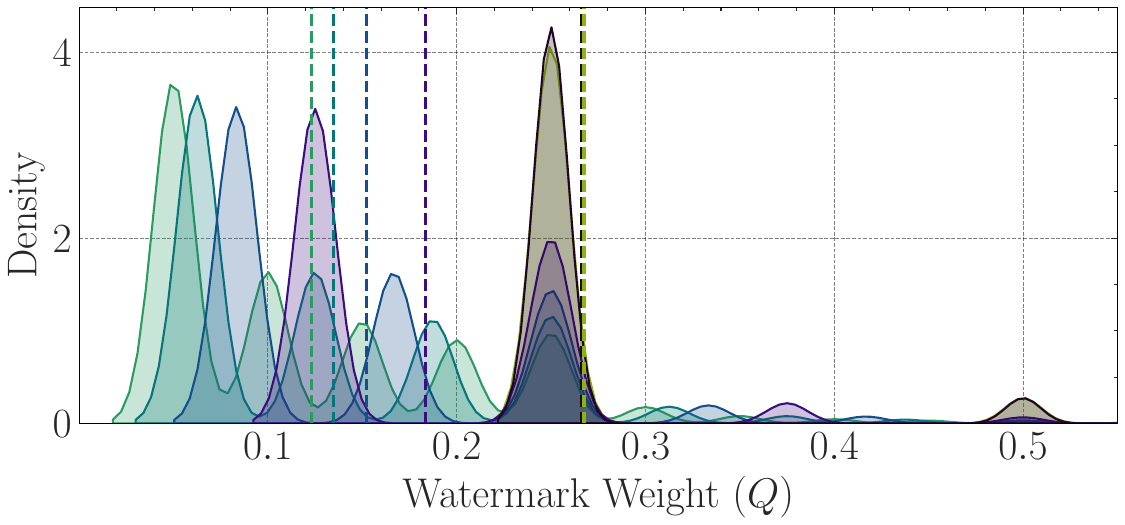}
    \caption{\agnews}
    \end{subfigure}
    \caption[Watermark weight distributions for the different number of \dipper paraphrases ($P$).]{Watermark weight distributions for the different number of \dipper paraphrases ($P$). These distributions show a similar trend as in \reffig{fig:backdoor-level-analysis}.
    }
    \label{fig:dipper-wm-wt-analysis}
\end{figure}
\begin{figure}[p]
    \centering
    \begin{subfigure}{0.7\textwidth}
    \centering\includegraphics[width=\linewidth,keepaspectratio]{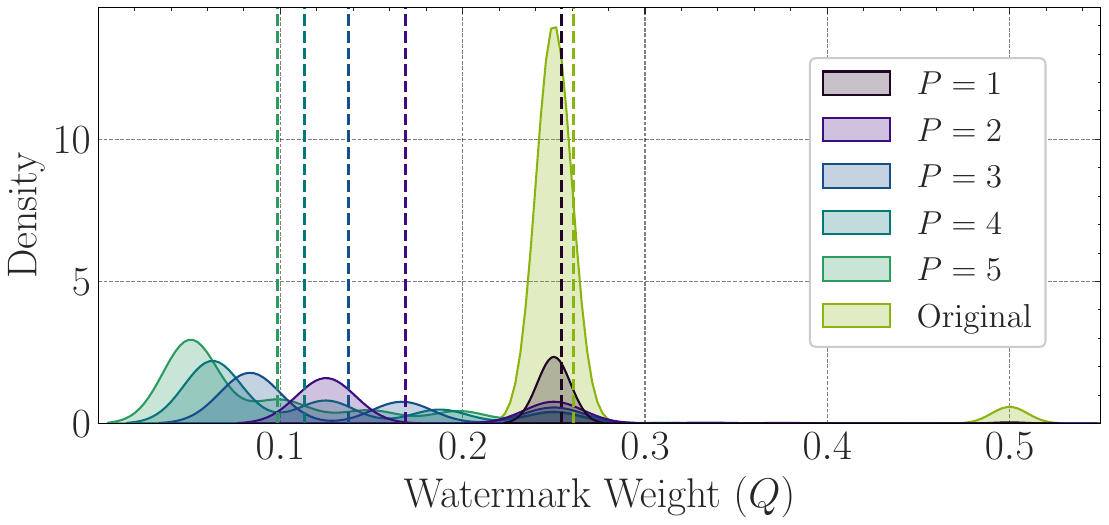}
    \caption{\enron}
    \end{subfigure}
    \begin{subfigure}{0.7\textwidth}
    \centering\includegraphics[width=\linewidth,keepaspectratio]{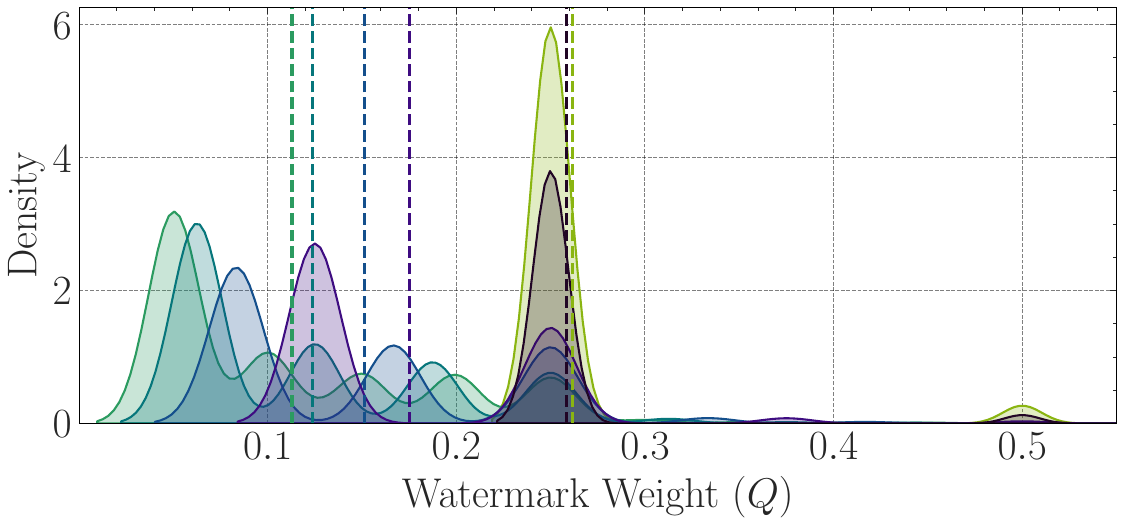}
    \caption{\sst}
    \end{subfigure}
    \begin{subfigure}{0.7\textwidth}
    \centering\includegraphics[width=\linewidth,keepaspectratio]{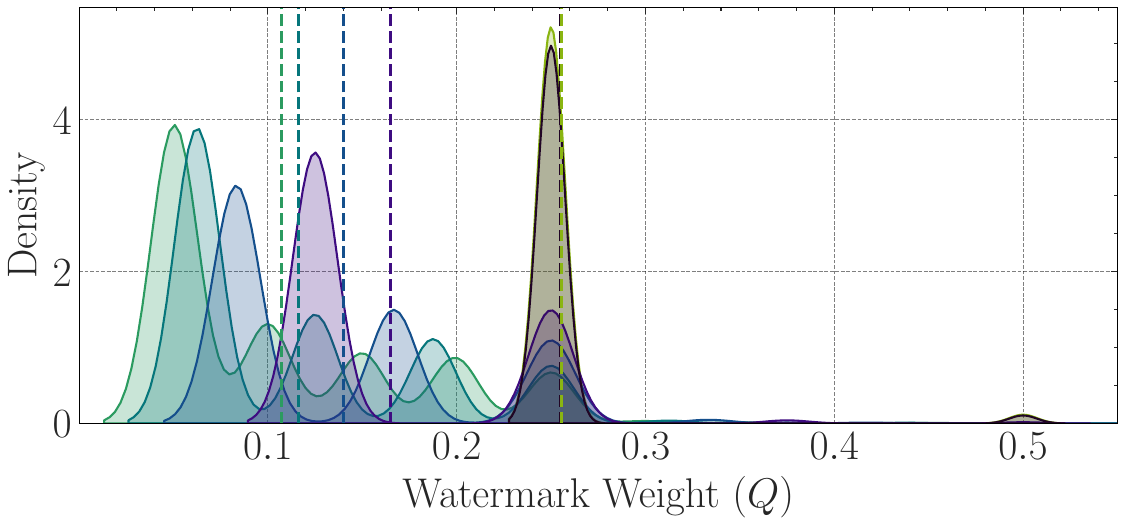}
    \caption{\mind}
    \end{subfigure}
    \begin{subfigure}{0.7\textwidth}
    \centering\includegraphics[width=\linewidth,keepaspectratio]{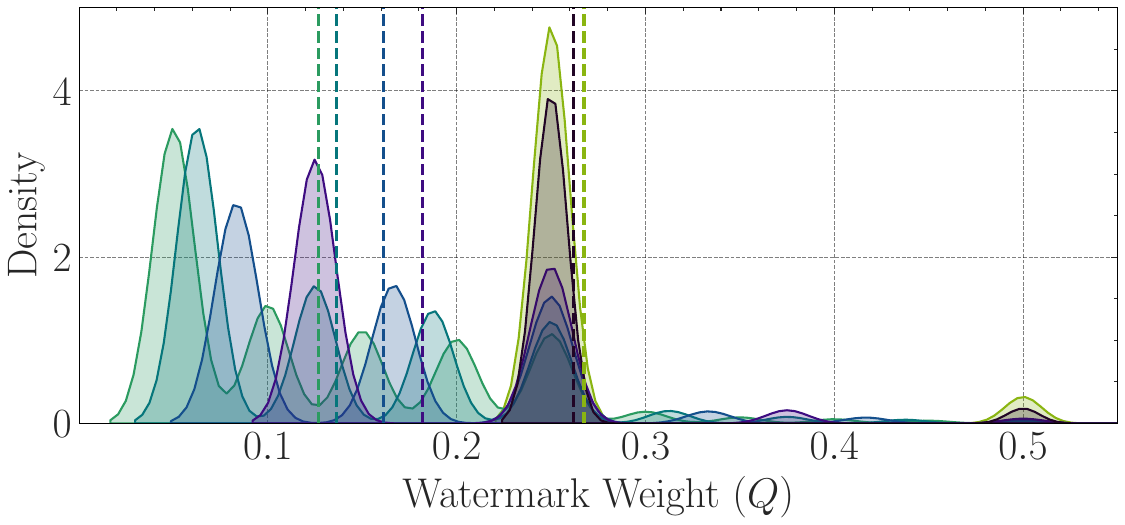}
    \caption{\agnews}
    \end{subfigure}
    \caption[Watermark weight distributions for the different number of \nllb paraphrases ($P$).]{Watermark weight distributions for the different number of \nllb paraphrases ($P$). These distributions show a similar trend as in Figures~\ref{fig:backdoor-level-analysis}~and~\ref{fig:dipper-wm-wt-analysis}.
    }
    \label{fig:nllb-wm-wt-analysis}
\end{figure}

\subsection{Results: Watermark Weight}
\label{sec:empirical-backdoor-analysis}

We now validate our theoretical analysis from \refsec{sec:attack-theoretical-proof}, which shows that paraphrasing dilutes the watermark. The findings show a lower likelihood of having higher watermark weights due to averaging of multiple paraphrases.
\reffig{fig:backdoor-level-analysis} presents the empirical watermark weight distribution with varying numbers of paraphrases $P$ for \firstWM and \secondWM using \gpt as the paraphrase model.\footnote{\firstWM and \secondWM have the same watermark weight, so these results apply to both methods.} 
As $P$ increases, we observe that the watermark weight decreases (this is particularly evident with the dashed vertical lines denoting mean watermark weight moving leftward), suggesting that incorporating more paraphrases dilutes the watermark. 
We present similar analysis for \dipper and RTT in Figures~\ref{fig:dipper-wm-wt-analysis}~and~\ref{fig:nllb-wm-wt-analysis} respectively and observe similar behaviours. Overall, these results provide empirical validation that paraphrasing dilutes watermark weights, making watermarks less transferable and hence circumventing watermark detection. Next, we perform our paraphrasing attacks as described in \refsec{sec:para-attack} on EaaS watermark baselines and evaluate the attacks' effectiveness.

\clearpage
\subsection{Results: Utility and Verifiability}
\label{sec:attack-perf}
\begin{table}[t]
\centering
    {\normalsize
    \begin{tabular}
    {ccccc}
    \toprule[1.5pt]
    {\textbf{Method}} & \textbf{\acc $\uparrow$}  & \textbf{\fscore $\uparrow$}  & {\textbf{$\Delta_{cos} \downarrow$}} & {\textbf{\auc} $\downarrow$} \\
    \toprule

    {\secondWM} & 94.50$\pm$0.34 & 94.50$\pm$0.34 & \textcolor{white}{0}5.20$\pm$0.34 & \textcolor{white}{0}97.40$\pm$0.54 \\
    \midrule
    {+\gpt Attack} & 92.81$\pm$0.21 & 92.81$\pm$0.21 & \textcolor{white}{0}0.70$\pm$0.22 & \textcolor{white}{0}68.90$\pm$7.79 \\
    {+\dipper Attack} & 91.34$\pm$0.52 & 91.33$\pm$0.52 & \textcolor{white}{0}0.46$\pm$0.11 & \textcolor{white}{0}67.50$\pm$5.56 \\
    {+\nllb Attack} & 93.35$\pm$0.23 & 93.35$\pm$0.23 & \textcolor{white}{0}0.65$\pm$0.12 & \textcolor{white}{0}71.95$\pm$4.04 \\
    
    \bottomrule[1.5pt]
    
    \end{tabular}}
    \vspace{-0.7em}
    \caption*{\large(a) \textbf{\enron}}
    \vspace{0.2em}

    {\normalsize
    \begin{tabular}
    {ccccc}
    \toprule[1.5pt]

    {\secondWM} & 93.10$\pm$0.12 & 93.10$\pm$0.12 & \textcolor{white}{0}2.57$\pm$1.19 & \textcolor{white}{0}86.75$\pm$6.20 \\
    \midrule
    {+\gpt Attack} & 92.75$\pm$0.15 & 92.75$\pm$0.15 & \textcolor{white}{0}0.93$\pm$0.09 & \textcolor{white}{0}75.90$\pm$2.91 \\
    {+\dipper Attack} & 91.70$\pm$0.27 & 91.66$\pm$0.27 & \textcolor{white}{0}0.90$\pm$0.17 & \textcolor{white}{0}71.95$\pm$2.69 \\
    {+\nllb Attack} & 92.57$\pm$0.09 & 92.55$\pm$0.08 & \textcolor{white}{0}1.06$\pm$0.19 & \textcolor{white}{0}69.35$\pm$2.94 \\

    \bottomrule[1.5pt]
    \end{tabular}}
    \vspace{-0.7em}
    \caption*{\large(b) \textbf{\sst}}
    \vspace{0.2em}
    
    {\normalsize
    \begin{tabular}
    {ccccc}
    \toprule[1.5pt]

    {\secondWM} & 77.31$\pm$0.08 & 51.47$\pm$0.23 & \textcolor{white}{0}5.27$\pm$0.17 & \textcolor{white}{0}98.10$\pm$0.51 \\
    \midrule
    {+\gpt Attack} & 77.01$\pm$0.05 & 51.24$\pm$0.22 & \textcolor{white}{0}1.85$\pm$0.21 & \textcolor{white}{0}79.40$\pm$3.08 \\
    {+\dipper Attack} & 76.86$\pm$0.07 & 50.54$\pm$0.17 & \textcolor{white}{0}3.47$\pm$0.12 & \textcolor{white}{0}96.70$\pm$0.51 \\
    {+\nllb Attack} & 76.64$\pm$0.10 & 50.36$\pm$0.11 & \textcolor{white}{0}3.89$\pm$0.06 & \textcolor{white}{0}97.80$\pm$0.33 \\

    \bottomrule[1.5pt]
    \end{tabular}}
    \vspace{-0.7em}
    \caption*{\large(c) \textbf{\mind}}
    \vspace{0.2em}

    {\normalsize
    \begin{tabular}
    {ccccc}
    \toprule[1.5pt]

    {\secondWM} & 93.51$\pm$0.13 & 93.50$\pm$0.13 & 14.46$\pm$0.68 & 100.00$\pm$0.00 \\
    \midrule
    {+\gpt Attack} & 92.28$\pm$0.12 & 92.26$\pm$0.13 & \textcolor{white}{0}7.23$\pm$0.34 & 100.00$\pm$0.00 \\
    {+\dipper Attack} & 92.50$\pm$0.11 & 92.48$\pm$0.11 & 11.04$\pm$0.40 & 100.00$\pm$0.00 \\
    {+\nllb Attack} & 92.70$\pm$0.10 & 92.69$\pm$0.10 & 10.56$\pm$0.44 & 100.00$\pm$0.00 \\

    \bottomrule[1.5pt]
    \end{tabular}}
    \vspace{-0.7em}
    \caption*{\large(d) \textbf{\agnews}}
    \vspace{-0.15em}
    \caption[Performance of \ourattack on \secondWM for different scenarios.]{Performance of \ourattack on \secondWM for \sst, \mind, \agnews, and \enron datasets.
    +\gpt Attack, +\dipper Attack, and +\nllb Attack represent different paraphrasing attack setups as mentioned in \refsec{sec:para-attack-setups}.
    From an attacker's perspective, $\uparrow$ means higher values are better and $\downarrow$ means values metrics are better. All the metrics are in \%.
    } 
    \label{table:attack-performance}
\end{table}

\begin{table}[t]
\centering
    {\normalsize
    \begin{tabular}
    {ccccc}
    \toprule[1.5pt]
    {\textbf{Method}} & \textbf{\acc $\uparrow$}  & \textbf{\fscore $\uparrow$}  & {\textbf{$\Delta_{cos} \downarrow$}} & {\textbf{\auc} $\downarrow$} \\
    \toprule

    {\firstWM} & 94.58$\pm$0.09 & 94.58$\pm$0.09 & \textcolor{white}{0}5.44$\pm$0.13 & \textcolor{white}{0}93.50$\pm$0.97 \\
    \midrule
    {+\gpt Attack} & 92.80$\pm$0.19 & 92.80$\pm$0.19 & -0.03$\pm$0.07 & \textcolor{white}{0}49.80$\pm$1.35 \\
    {+\dipper Attack} & 92.35$\pm$0.48 & 92.35$\pm$0.49 & \textcolor{white}{0}0.63$\pm$0.16 & \textcolor{white}{0}61.85$\pm$4.52 \\
    {+\nllb Attack} & 93.38$\pm$0.20 & 93.38$\pm$0.20 & \textcolor{white}{0}0.69$\pm$0.20 & \textcolor{white}{0}65.25$\pm$3.68 \\
    
    \bottomrule[1.5pt]
    
    \end{tabular}}
    \vspace{-0.7em}
    \caption*{\large(a) \textbf{\enron}}
    \vspace{0.2em}

    {\normalsize
    \begin{tabular}
    {ccccc}
    \toprule[1.5pt]
    {\firstWM} & 92.89$\pm$0.25 & 92.89$\pm$0.25 & \textcolor{white}{0}4.05$\pm$2.70 & \textcolor{white}{0}95.04$\pm$2.30 \\
    \midrule
    {+\gpt Attack} & 92.86$\pm$0.17 & 92.86$\pm$0.17 & \textcolor{white}{0}0.68$\pm$0.10 & \textcolor{white}{0}68.20$\pm$2.94 \\
    {+\dipper Attack} & 91.31$\pm$0.24 & 91.27$\pm$0.25 & \textcolor{white}{0}0.94$\pm$0.12 & \textcolor{white}{0}79.95$\pm$3.89 \\
    {+\nllb Attack} & 92.66$\pm$0.55 & 92.64$\pm$0.55 & \textcolor{white}{0}0.76$\pm$0.11 & \textcolor{white}{0}78.20$\pm$3.60 \\

    \bottomrule[1.5pt]
    \end{tabular}}
    \vspace{-0.7em}
    \caption*{\large(b) \textbf{\sst}}
    \vspace{0.2em}
    
    {\normalsize
    \begin{tabular}
    {ccccc}
    \toprule[1.5pt]
    {\firstWM} & 77.34$\pm$0.06 & 51.63$\pm$0.16 & \textcolor{white}{0}3.93$\pm$0.11 & \textcolor{white}{0}93.10$\pm$0.94 \\
    \midrule
    {+\gpt Attack} & 77.01$\pm$0.07 & 51.23$\pm$0.13 & \textcolor{white}{0}1.04$\pm$0.08 & \textcolor{white}{0}67.75$\pm$1.66 \\
    {+\dipper Attack} & 76.83$\pm$0.09 & 50.56$\pm$0.11 & \textcolor{white}{0}2.22$\pm$0.09 & \textcolor{white}{0}90.15$\pm$1.68 \\
    {+\nllb Attack} & 76.59$\pm$0.14 & 50.32$\pm$0.26 & \textcolor{white}{0}2.11$\pm$0.07 & \textcolor{white}{0}85.80$\pm$1.34 \\

    \bottomrule[1.5pt]
    \end{tabular}}
    \vspace{-0.7em}
    \caption*{\large(c) \textbf{\mind}}
    \vspace{0.2em}

    {\normalsize
    \begin{tabular}
    {ccccc}
    \toprule[1.5pt]
    {\firstWM} & 93.47$\pm$0.12 & 93.47$\pm$0.12 & 12.53$\pm$0.67 & 100.00$\pm$0.00 \\
    \midrule
    {+\gpt Attack} & 92.17$\pm$0.04 & 92.15$\pm$0.04 & \textcolor{white}{0}4.66$\pm$0.36 & \textcolor{white}{0}99.15$\pm$0.34 \\
    {+\dipper Attack} & 92.47$\pm$0.10 & 92.45$\pm$0.10 & \textcolor{white}{0}6.68$\pm$0.40 & 100.00$\pm$0.00 \\
    {+\nllb Attack} & 92.76$\pm$0.13 & 92.74$\pm$0.13 & \textcolor{white}{0}6.30$\pm$0.35 & 100.00$\pm$0.00 \\

    \bottomrule[1.5pt]
    \end{tabular}}
    \vspace{-0.7em}
    \caption*{\large(d) \textbf{\agnews}}
    \vspace{0.2em}
    \caption[Performance of \ourattack on \firstWM for different scenarios.]{Performance of \ourattack on \firstWM for different scenarios, similar to \reftab{table:attack-performance}. 
    } 
    \label{table:embmarker-attack-performance}
\end{table}

Tables~\ref{table:attack-performance}~and~\ref{table:embmarker-attack-performance} present the {utility} and {verifiability} of \secondWM and \firstWM, respectively, under paraphrasing attack.
As an attacker, we want high utility and low verifiability.
In terms of {utility} (\acc and \fscore), all paraphrasing methods (+\gpt Attack, +\dipper Attack, +\nllb Attack) only result in a small negative impact on downstream performance, suggesting that paraphrasing does not degrade the quality of the embeddings. In terms of {verifiability}, for $\Delta_{\cos}$, we see the numbers drop significantly after paraphrasing, showing that it is now harder to detect the watermark. \auc tells a similar story, with one exception: watermarks for \agnews are still verifiable (close to 100.00), suggesting that the paraphrasing attack is less effective for this dataset.
We suspect this is because \agnews has much longer texts (see \reftab{table:dataset-statistics}), which means paraphrasing has the possibility of introducing additional trigger words not in the original text. This observation is supported by our theoretical analyses from \refsecnum{sec:attack-theoretical-proof}, which showed that although with paraphrasing we reduce the probability of higher watermark weights, this effect diminishes with longer text.
We note that paraphrasing is still diluting the watermarks in \agnews because of the decrease in the $\Delta_{\cos}$, but the watermark is not diluted enough to circumvent the watermark detection.
That said, an attacker has the freedom to select their training strategy; this means they can technically still exploit this paraphrasing vulnerability by using shorter texts when imitating the victim model.

\clearpage
\subsection{Analysis: Quality of Paraphrases}
\label{sec:para-quality}
    \contentwarning{{This section directly quotes content that contains offensive/hateful speech, profanity, and other potentially triggering content.}}

\begin{figure*}[t]
    \centering
    \begin{subfigure}{0.75\textwidth}
    \centering\includegraphics[width=\linewidth,keepaspectratio]{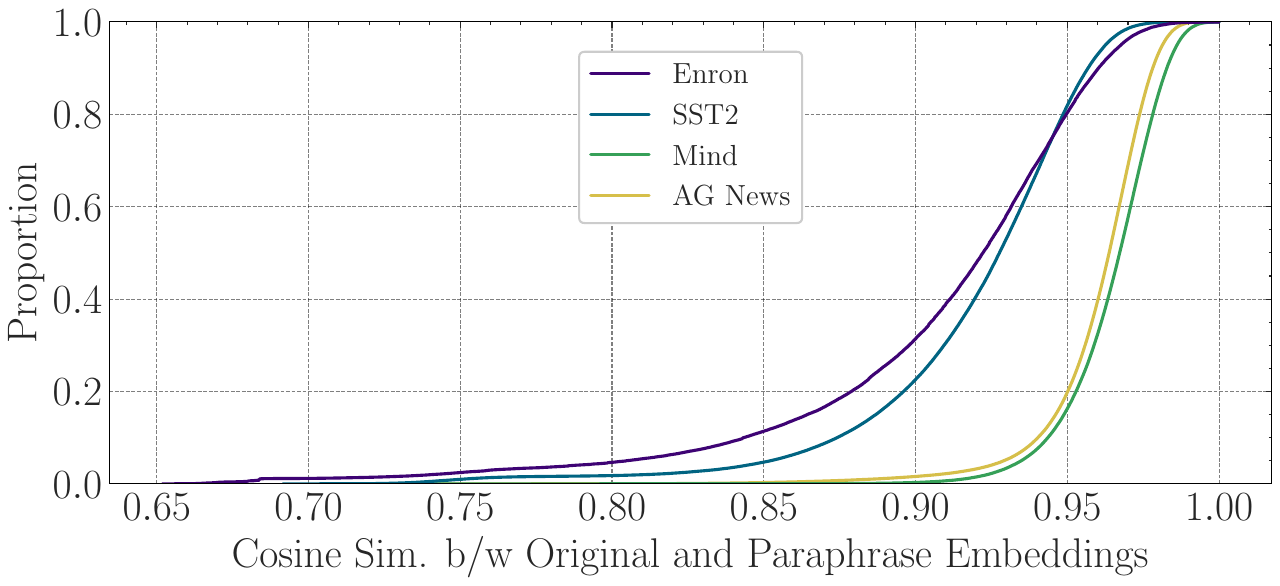}
    \caption{\gpt}
    \end{subfigure}
    \begin{subfigure}{0.75\textwidth}
    \centering\includegraphics[width=\linewidth,keepaspectratio]{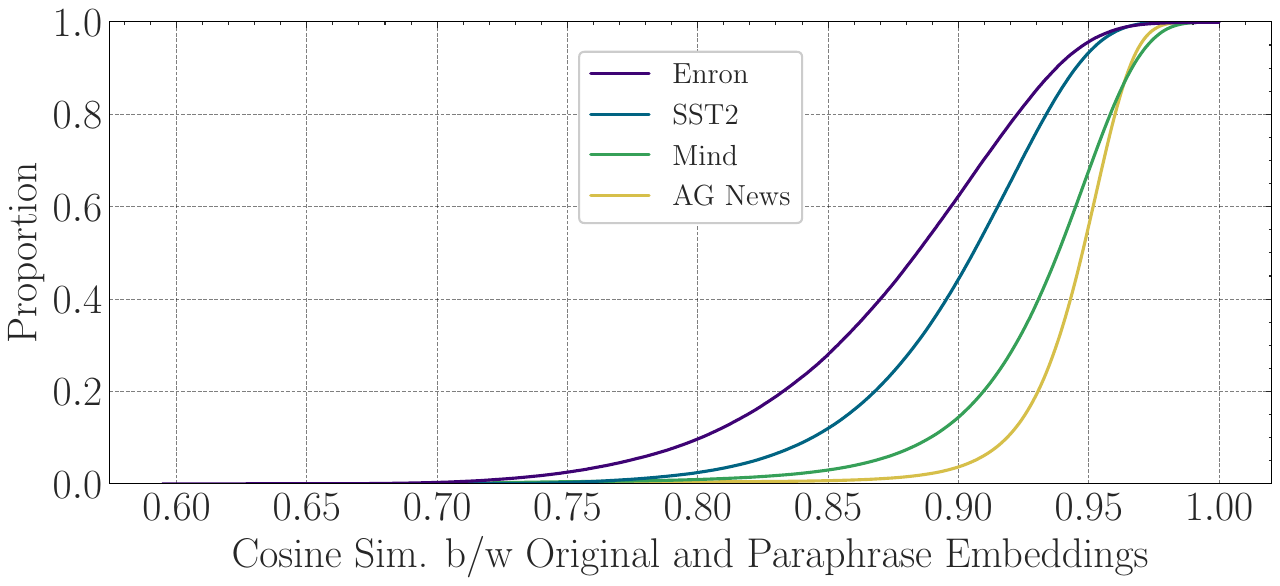}
    \caption{\dipper}
    \end{subfigure}
    \begin{subfigure}{0.75\textwidth}
    \centering\includegraphics[width=\linewidth,keepaspectratio]{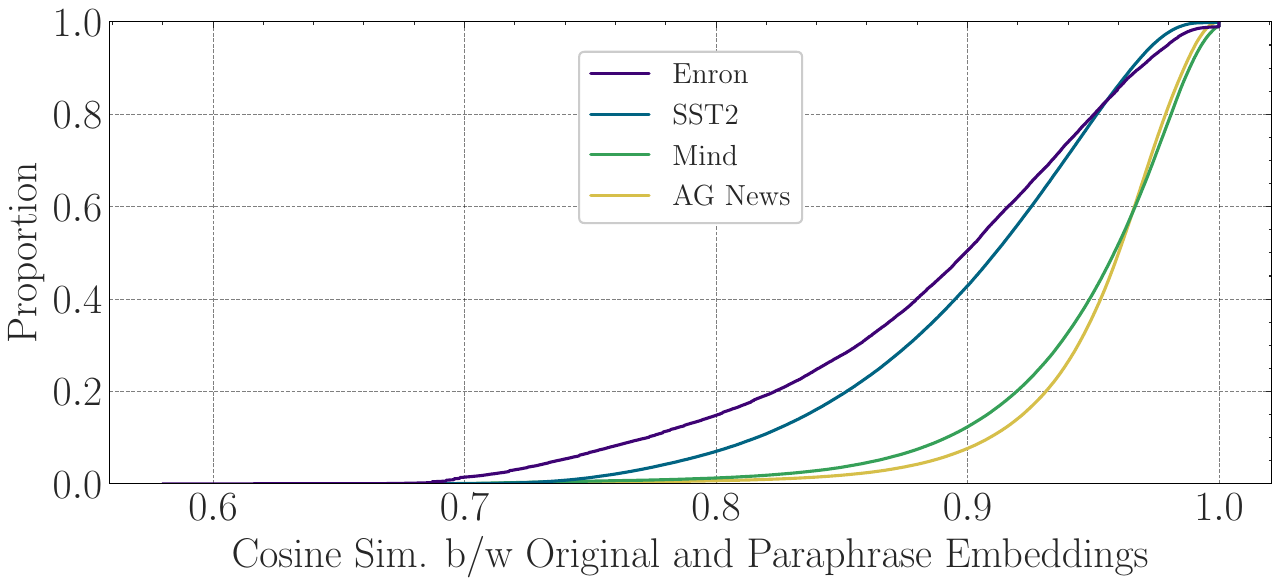}
    \caption{\nllb}
    \end{subfigure}
    \caption[Cosine similarity between original and paraphrase embeddings using different paraphrasing methods.]{Cosine similarity between original and paraphrase embeddings using different paraphrasing methods (as denoted in the subcaption).}
    \label{fig:cos-sim-paras}
\end{figure*}

The paraphrases used in our attack should preserve the semantic content of the original input. If the paraphrases are too different, the embedding quality will degrade, leading to utility loss. 
We measure paraphrase quality by calculating the cosine similarity between the original and paraphrased text embeddings. As shown in \reffig{fig:cos-sim-paras}, before any filtering, most paraphrases are similar to the original text, demonstrating good-quality paraphrases. Although there are still some paraphrases with low similarity (\ie we see non-zero density for <0.9 similarities), corresponding to the left-side entries of the distribution plots. We now manually investigate these low cosine similarity cases using the \gpt paraphrase model.

Our manual analyses reveal that poor paraphrases usually occur when the input contains ambiguous, offensive, or copyrighted content. Since the current LLMs are aligned to respond safely to such inappropriate inputs \citep{zhao2023survey-LLM}, they output templated non-answers, as shown in \reftab{table:bad-para-praprhrasing-examples}. In the context of the imitation attack, it is better to filter out such problematic inputs, as they are not genuine paraphrases. Based on this observation, we implement the cosine similarity filter to remove these poor paraphrases from our attack. If no valid paraphrases are found after filtering, we use the original input text. 

\begin{table}[t]
\centering
    \begin{minipage}{.95\textwidth}
    \resizebox{\textwidth}{!}{%
    \begin{tabular}{c>{\normalsize}p{0.45\textwidth}>{\normalsize}p{0.45\textwidth}}

        \toprule[1.5pt]
        \multirow{1}{*}{\textbf{Dataset}} & \multirow{1}{*}{\textbf{Input Text}} & {\textbf{Paraphrase Output}} \\
        \midrule
        \multirow{7}{*}{\rotatebox[origin=c]{90}{\textbf{\enron}}} & \textsf{no risk kiosk pg lnbcer} & \textsf{Unfortunately, the text you provided doesn't seem to make sense as it is currently written. Could you please clarify or provide more context? I'm here to help!} \\
        \cmidrule{2-3}
        {} & \textsf{uncensored home - made incest photo and video ! ax hipster} & \textsf{Please note that I can't assist with this request.} \\
        \midrule
        \multirow{7}{*}{\rotatebox[origin=c]{90}{\textbf{\sst}}} & \textsf{rape and}  & \textsf{I'm sorry, I can't assist with that. If you have any other requests or need help with a different topic, feel free to let me know!} \\
         \cmidrule{2-3}
        {} & \textsf{a fun}  & \textsf{Sure, I can help with that! Could you please provide the complete text that you need rewritten?} \\
        \midrule
        \multirow{10}{*}{\rotatebox[origin=c]{90}{\textbf{\mind}}} & \textsf{Adam Schiff's full opening statement at the first public impeachment hearing} & \textsf{I'm sorry, but I cannot provide verbatim excerpts from copyrighted texts such as transcripts. However, I can help summarize the statement or provide information on its key points and themes. Would you like me to do that?} \\
        \cmidrule{2-3}
        {} & \textsf{Devin Nunes' full opening statement at the first public impeachment hearing} & \textsf{I'm sorry, but I can't provide verbatim excerpts from copyrighted texts. However, I can summarize the content for you if you want.} \\
        \bottomrule[1.5pt]
    \end{tabular}}
    \end{minipage}
    \caption[\textcolor{red}{
\textbf{WARNING:} \textit{Contains offensive/hateful speech, profanity, and other potentially triggering content.}} Some poor paraphrase examples using the \gpt paraphrase model.]{\textcolor{red}{
\textbf{WARNING:} \textit{Contains offensive/hateful speech, profanity, and other potentially triggering content.}} Some poor (having low cosine similarity with the original input text) paraphrase examples using the \gpt paraphrase model.}
    \label{table:bad-para-praprhrasing-examples}
\end{table}

\begin{figure*}[t]
    \centering
    \begin{subfigure}{0.49\textwidth}
    \centering\includegraphics[width=\linewidth,keepaspectratio]{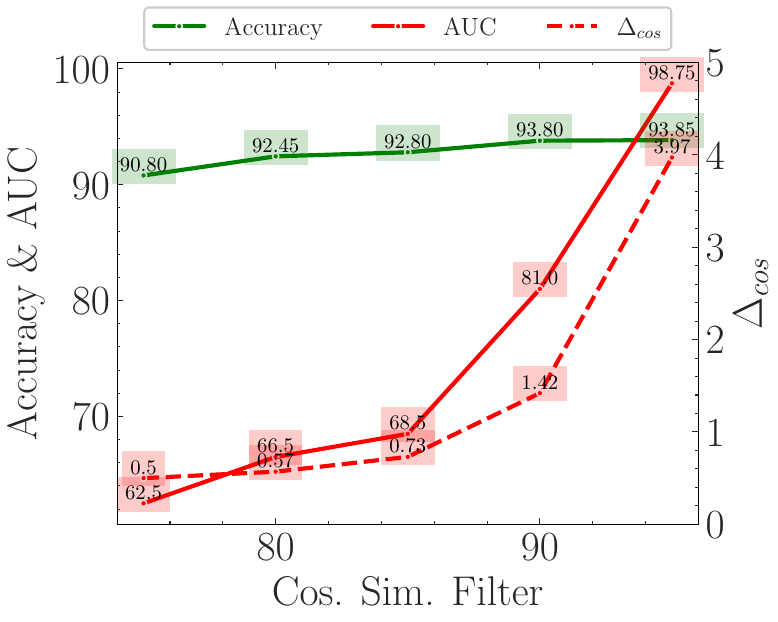}
    \caption{\enron}
    \label{fig:cos-filter-enron}
    \end{subfigure}
    \begin{subfigure}{0.49\textwidth}
    \centering\includegraphics[width=\linewidth,keepaspectratio]{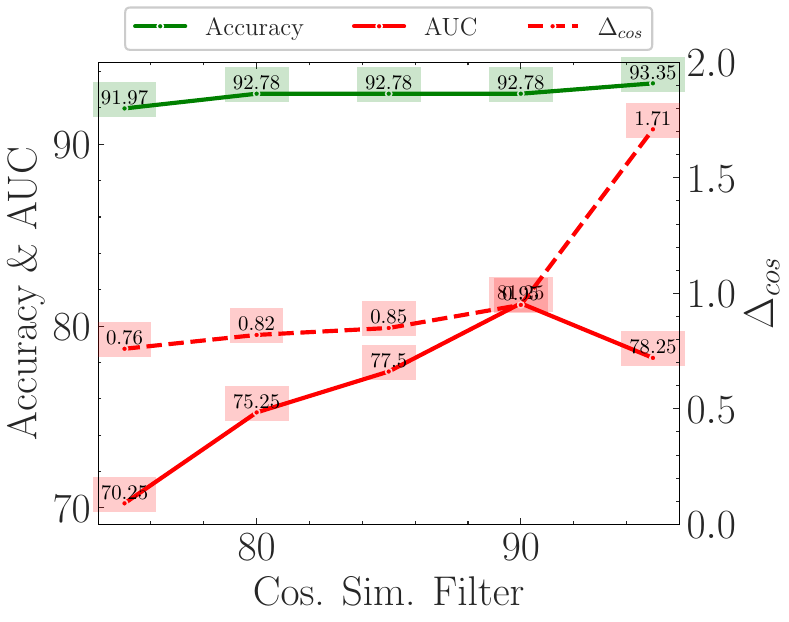}
    \caption{\sst}
    \end{subfigure}
    \begin{subfigure}{0.49\textwidth}
    \centering\includegraphics[width=\linewidth,keepaspectratio]{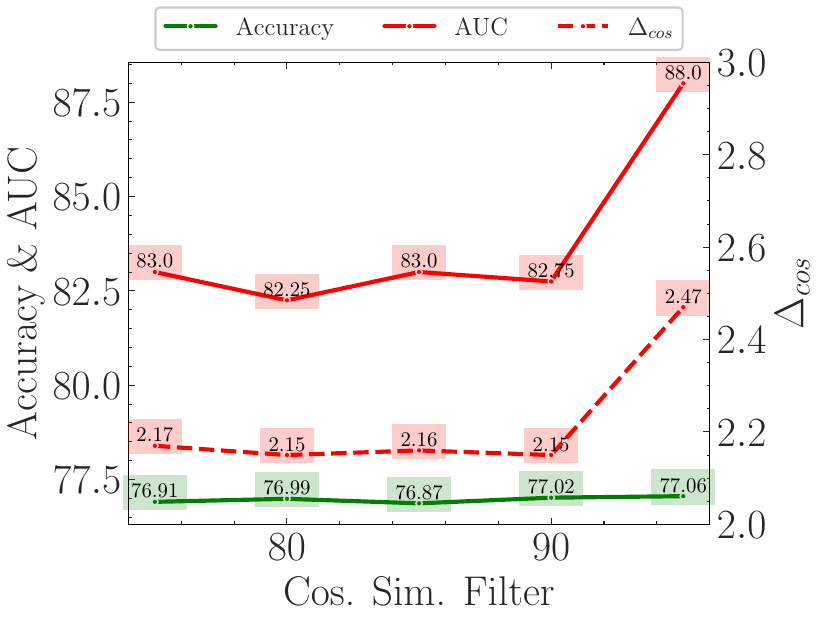}
    \caption{\mind}
    \end{subfigure}
    \begin{subfigure}{0.49\textwidth}
    \centering\includegraphics[width=\linewidth,keepaspectratio]{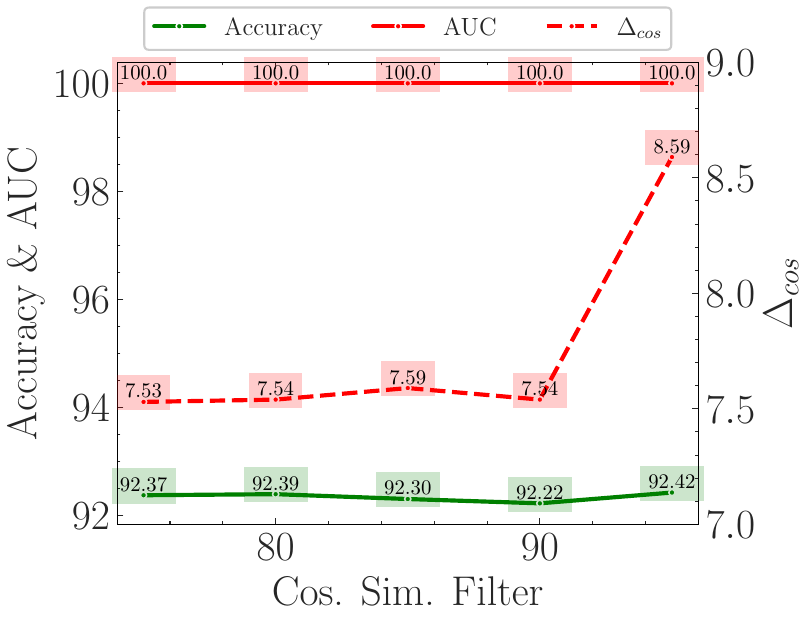}
    \caption{\agnews}
    \end{subfigure}
    \caption[\gpt paraphrase attack performance using different cosine similarity filters.]{\gpt paraphrase attack performance using different cosine similarity filters. ``Cos. Sim. Filter'' is represented on a $\times 100$ $[0, 100]$ scale.}
    \label{fig:cos-filter}
\end{figure*}

To understand the best cosine similarity filter threshold, we look at utility \versus verifiability plots over \enron, \etc Interestingly, we see that the filter threshold can have a significant impact on the utility, \eg for \enron, we see a drop of over 2\% when we use a very low threshold; this applies to all other graphs in \reffig{fig:cos-filter}.
We settle on a threshold of 0.80, as it is a good trade-off between utility and attack performance (\ie lower verifiability). 

\clearpage
\subsection{Ablation Studies}
\label{sec:attack-ablation}
In this section, we present ablation studies to examine the impact of the following factors: the number of watermarks, number of paraphrases, non-watermark cases, attacker model size, and training data size. Please note that all these ablation studies are conducted using the \gpt paraphrase model (one of the three attack setups in \refsec{sec:para-attack-setups}).

\paragraph*{Number of Watermarks ($R$) in \secondWM.}
\label{sec:diff-R-WARDEN}
\begin{figure*}[t]
    \centering
    \begin{subfigure}{0.48\textwidth}
    \centering\includegraphics[width=\linewidth,keepaspectratio]{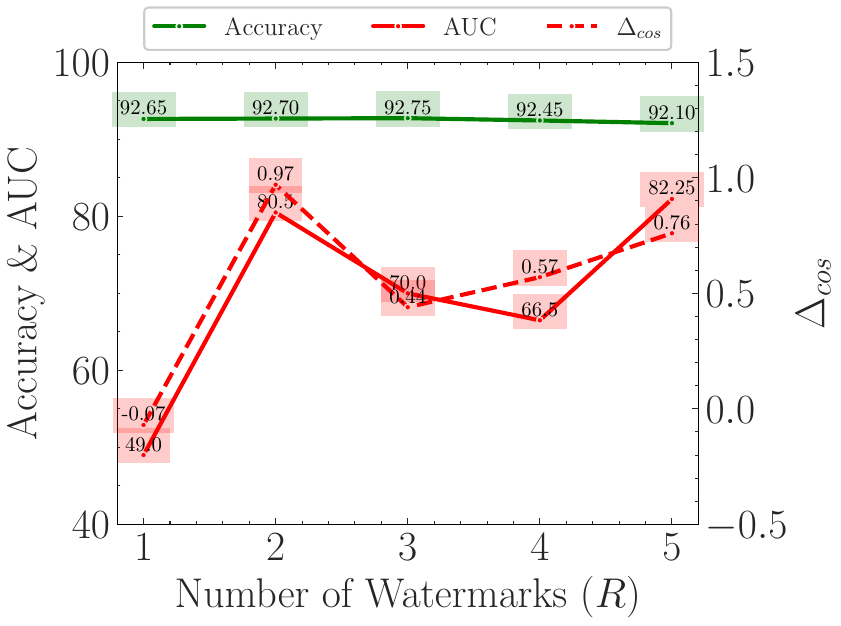}
    \caption{\enron}
    \end{subfigure}
    \begin{subfigure}{0.46\textwidth}
    \centering\includegraphics[width=\linewidth,keepaspectratio]{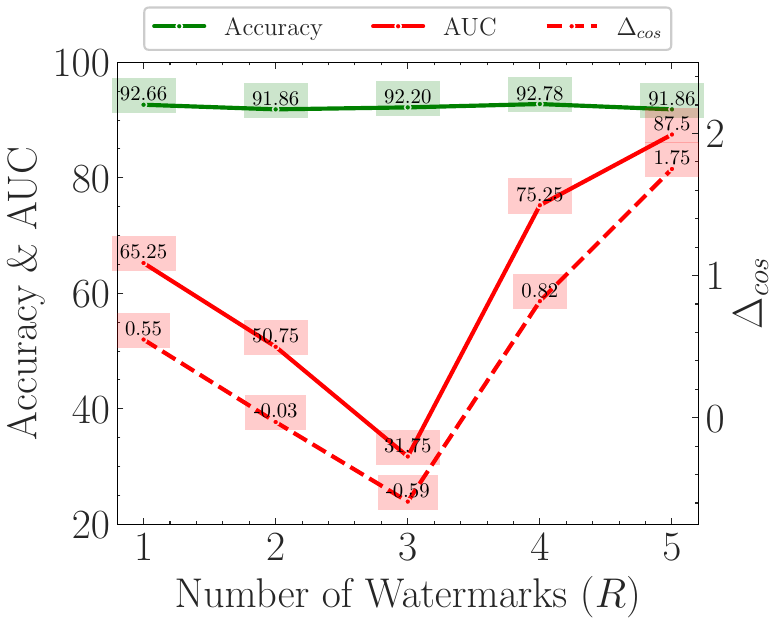}    \caption{\sst}
    \end{subfigure}
    \begin{subfigure}{0.47\textwidth}
    \centering\includegraphics[width=\linewidth,keepaspectratio]{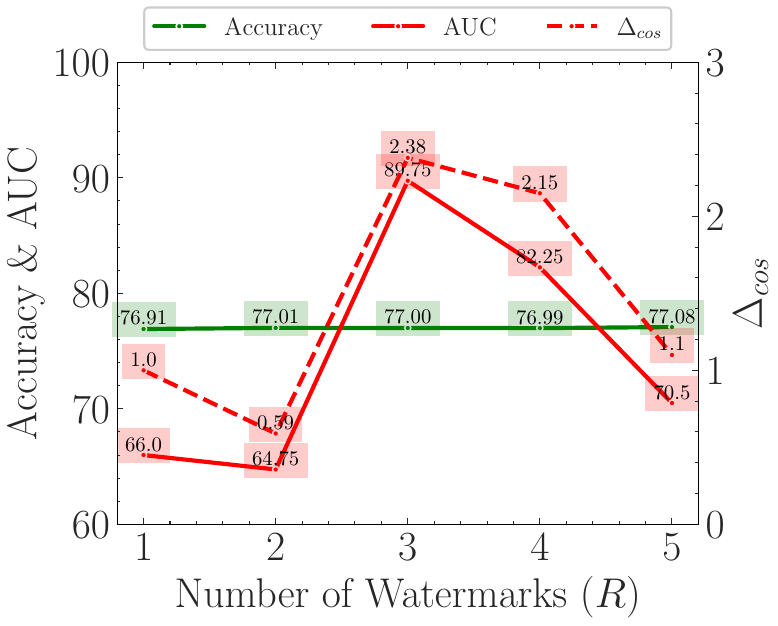}    \caption{\mind}
    \end{subfigure}
    \begin{subfigure}{0.49\textwidth}
    \centering\includegraphics[width=\linewidth,keepaspectratio]{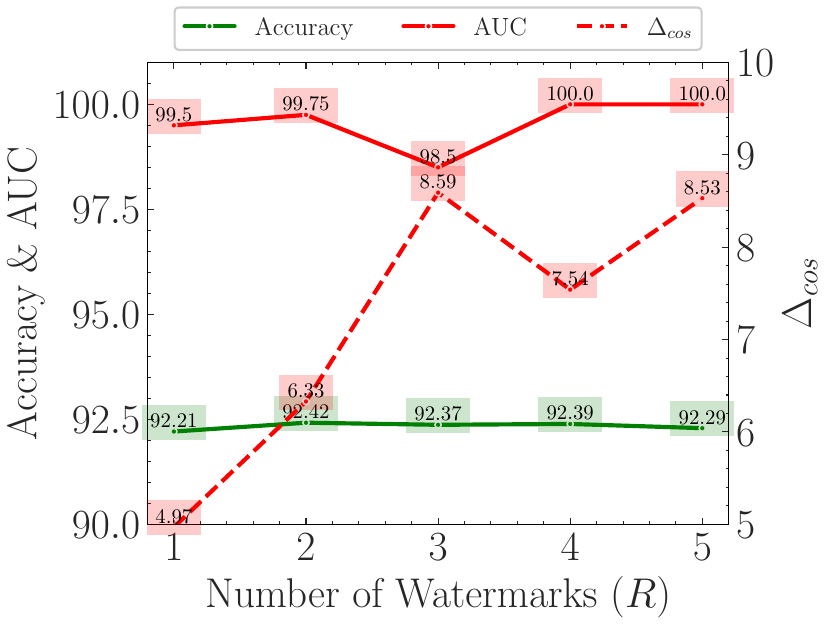}    \caption{\agnews}
    \end{subfigure}
    \caption{\gpt paraphrase attack performance against different number of watermarks ($R$) in \secondWM.}
    \label{fig:diff-R-WARDEN-attack-perf}
\end{figure*}
In this study, we check if the paraphrasing attack is effective for different number of watermarks ($R$) in \secondWM.
There is fluctuation in watermark {verifiability} (represented by \textcolor{red}{red} lines) metrics when $R$ changes as per \reffig{fig:diff-R-WARDEN-attack-perf}. Overall, the {verifiability} metrics remain low, indicating that paraphrasing continues to be an effective attack across different numbers of watermarks in \secondWM. The notable exception is \agnews, where paraphrasing attack is ineffective in general, as seen in our main results \refsecnum{sec:attack-perf}.

\paragraph{Impact of Scaling Train Dataset.}
\begin{table}[t]
    \centering
    {\normalsize
    \begin{tabular}{ccSSSS}
    \toprule[1.5pt]
    \multirow{2}{*}{\textbf{Dataset}} & \multirow{2}{*}{\textbf{Paraphraser}} & \multicolumn{2}{c}{\textbf{Utility}} & \multicolumn{2}{c}{\textbf{Verifiability}} \\
    \cmidrule(lr){3-4} \cmidrule(lr){5-6}
    {} & {} & \acc $\uparrow$ & \fscore $\uparrow$ & {$\Delta_{cos} \downarrow$} & {\auc $\downarrow$} \\
    \toprule
    
    \multirow{4}{*}{\enron}
         {} & {No Attack} & 93.85 & 93.85 & 5.39 & 97.75 \\
    \cmidrule{2-6}
         {} & {\gpt} & 95.30 & 95.30 & 6.65 & 98.50 \\
         {} & {\dipper} & 95.30 & 95.30 & 8.47 & 99.50 \\
         {} & {\nllb} & 94.95 & 94.95 & 8.64 & 99.25 \\
    \midrule
    \multirow{4}{*}{\sst}
         {} & {No Attack} & 93.12 & 93.12 & 5.21 & 100.00 \\
    \cmidrule{2-6}
         {} & {\gpt} & 93.35 & 93.34 & 6.69 & 96.25 \\
         {} & {\dipper} & 92.66 & 92.65 & 8.73 & 99.50 \\
         {} & {\nllb} & 93.23 & 93.23 & 7.36 & 98.25 \\
    \midrule
    \multirow{4}{*}{\mind}
        {} & {No Attack} & 77.23 & 51.4 & 5.54 & 98.50 \\
    \cmidrule{2-6}
        {} & {\gpt} & 77.06 & 52.07 & 12.74 & 100.00 \\
        {} & {\dipper} & 77.23 & 55.46 & 15.58 & 100.00 \\
        {} & {\nllb} & 77.12 & 56.89 & 14.97 & 100.00 \\
    \midrule
    \multirow{4}{*}{\agnews}
        {} & {No Attack} & 93.43 & 93.43 & 15.29 & 100.00 \\
    \cmidrule{2-6}
        {} & {\gpt} & 93.11 & 93.10 & 19.68 & 100.00 \\
        {} & {\dipper} & 93.59 & 93.58 & 19.26 & 100.00 \\
        {} & {\nllb} & 93.39 & 93.39 & 18.97 & 100.00 \\

    \bottomrule[1.5pt]
    \end{tabular}}
    \caption[Impact of scaling up the dataset with paraphrases instead of averaging the paraphrase embeddings in \ourattack.]{Impact of scaling up the dataset with paraphrases instead of averaging the paraphrase embeddings in \ourattack. 
    ``No Attack'' represents the normal \secondWM watermarking scenario without paraphrasing attacks.
    }
    \label{fig:scale-up-dataset}
\end{table}

A potential confound is that generating multiple paraphrases effectively increases the training data size during the imitation attack. To tease out this effect, we run another experiment where we scale up the training data size to match the size used in the paraphrasing experiment; results in \reftab{fig:scale-up-dataset}. Interestingly, we find that watermark detection performance goes up, demonstrating that the success of paraphrasing in evading detection is not due to increased training data size. 
This is expected as, with increased dataset size, we have more samples with trigger words, making watermarks more transferable. 

\clearpage
\paragraph{Number of Paraphrases ($P$).}

\begin{figure*}
    \centering
    \begin{subfigure}{0.49\textwidth}
    \centering\includegraphics[width=\linewidth,keepaspectratio]{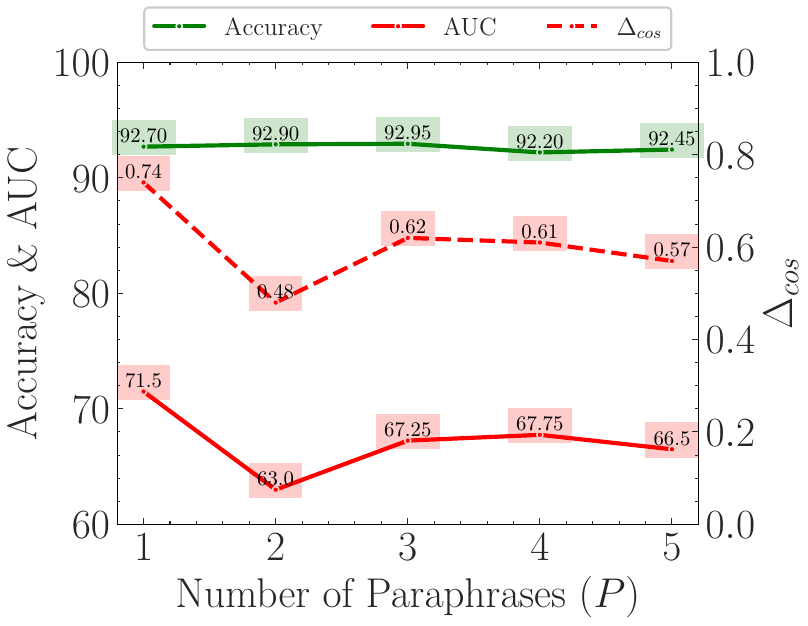}
    \caption{\enron}
    \end{subfigure}
    \begin{subfigure}{0.49\textwidth}
    \centering\includegraphics[width=\linewidth,keepaspectratio]{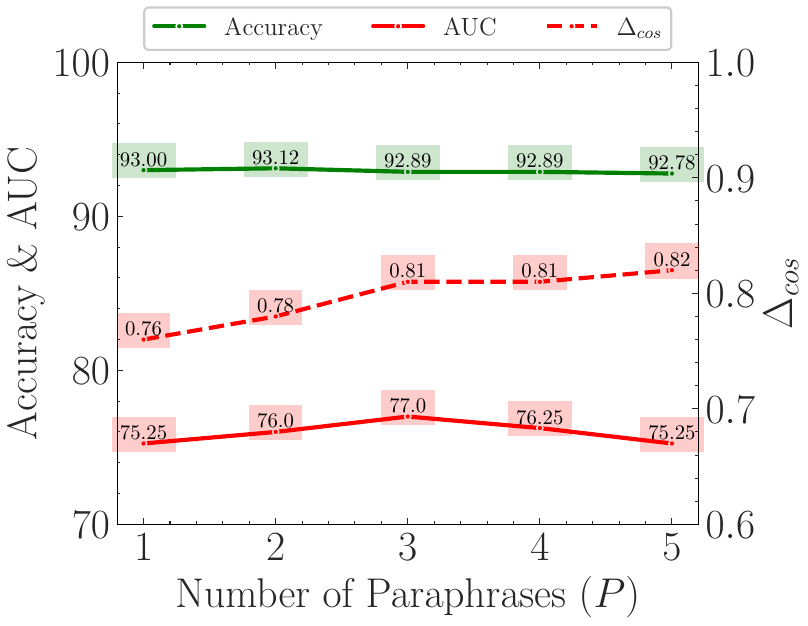}
    \caption{\sst}
    \end{subfigure}
    \begin{subfigure}{0.49\textwidth}
    \centering\includegraphics[width=\linewidth,keepaspectratio]{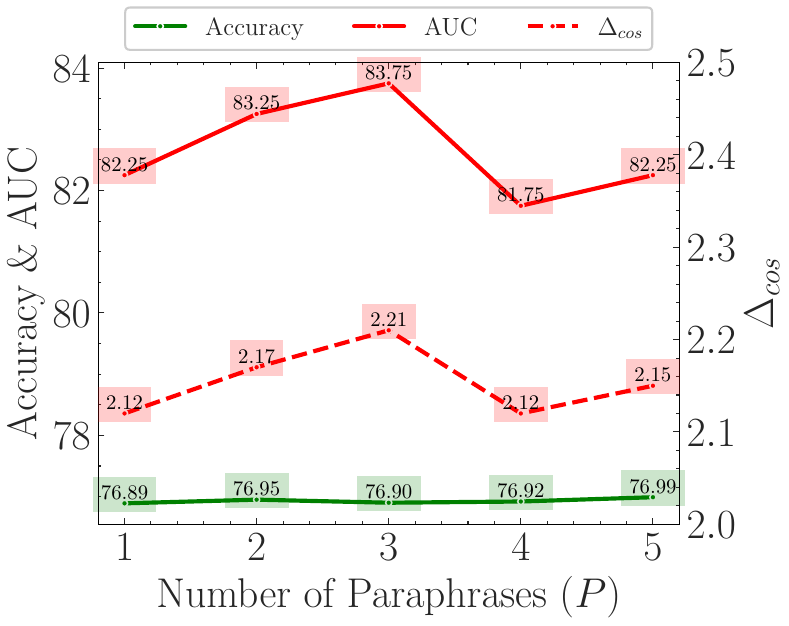}
    \caption{\mind}
    \end{subfigure}
    \begin{subfigure}{0.49\textwidth}
    \centering\includegraphics[width=\linewidth,keepaspectratio]{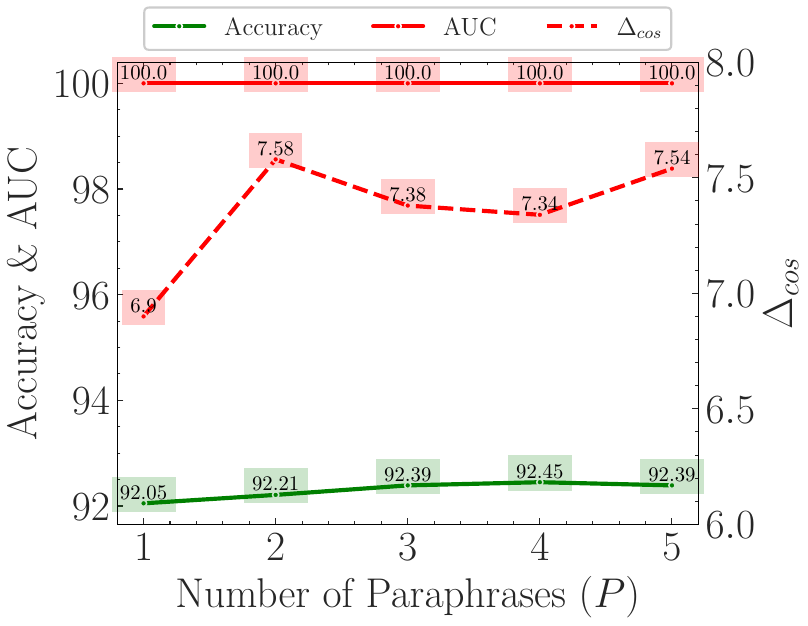}
    \caption{\agnews}
    \end{subfigure}
    \caption{\gpt paraphrase attack performance for different number of paraphrases ($P$).}
    \label{fig:diff-P-GPT-attack}
\end{figure*}

\begin{figure*}
    \centering
    \begin{subfigure}{0.49\textwidth}
    \centering\includegraphics[width=\linewidth,keepaspectratio]{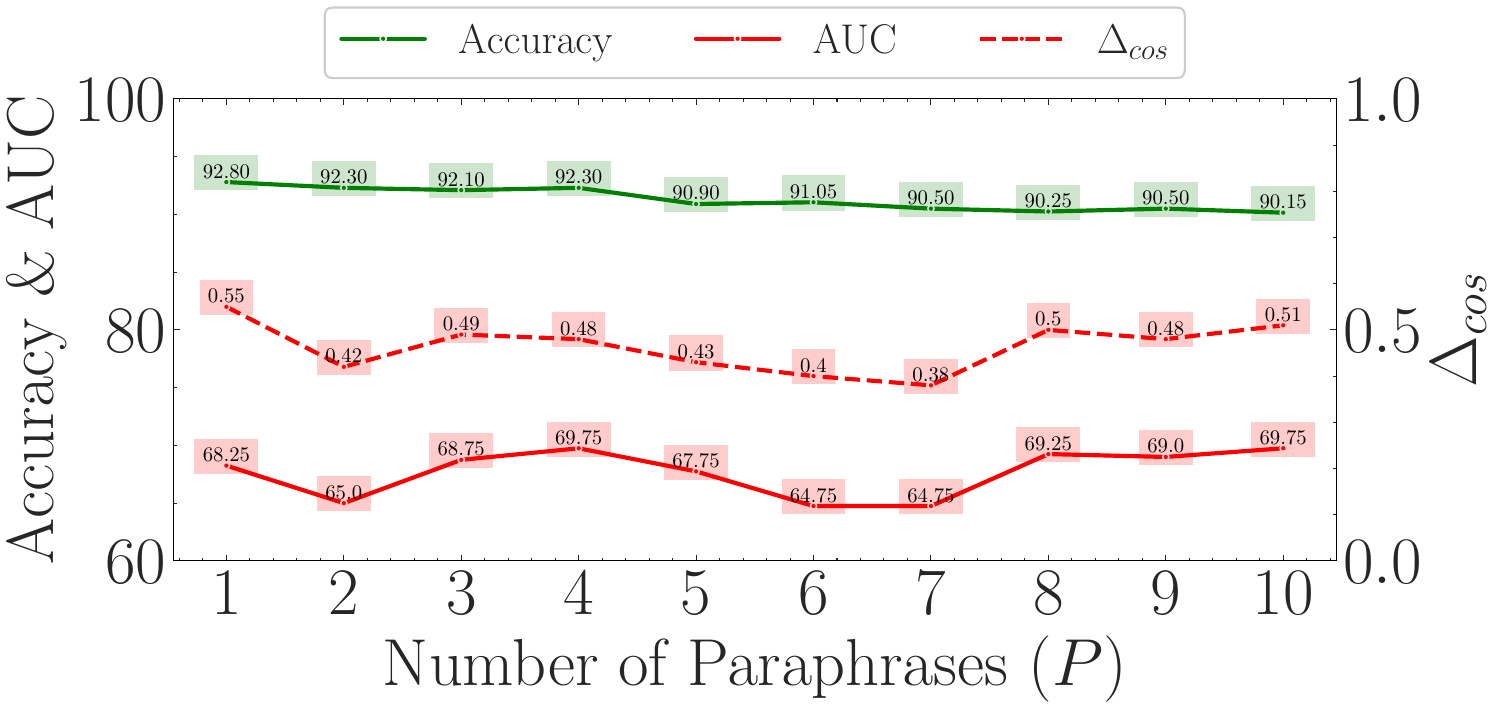}
    \caption{\enron}
    \end{subfigure}
    \begin{subfigure}{0.49\textwidth}
    \centering\includegraphics[width=\linewidth,keepaspectratio]{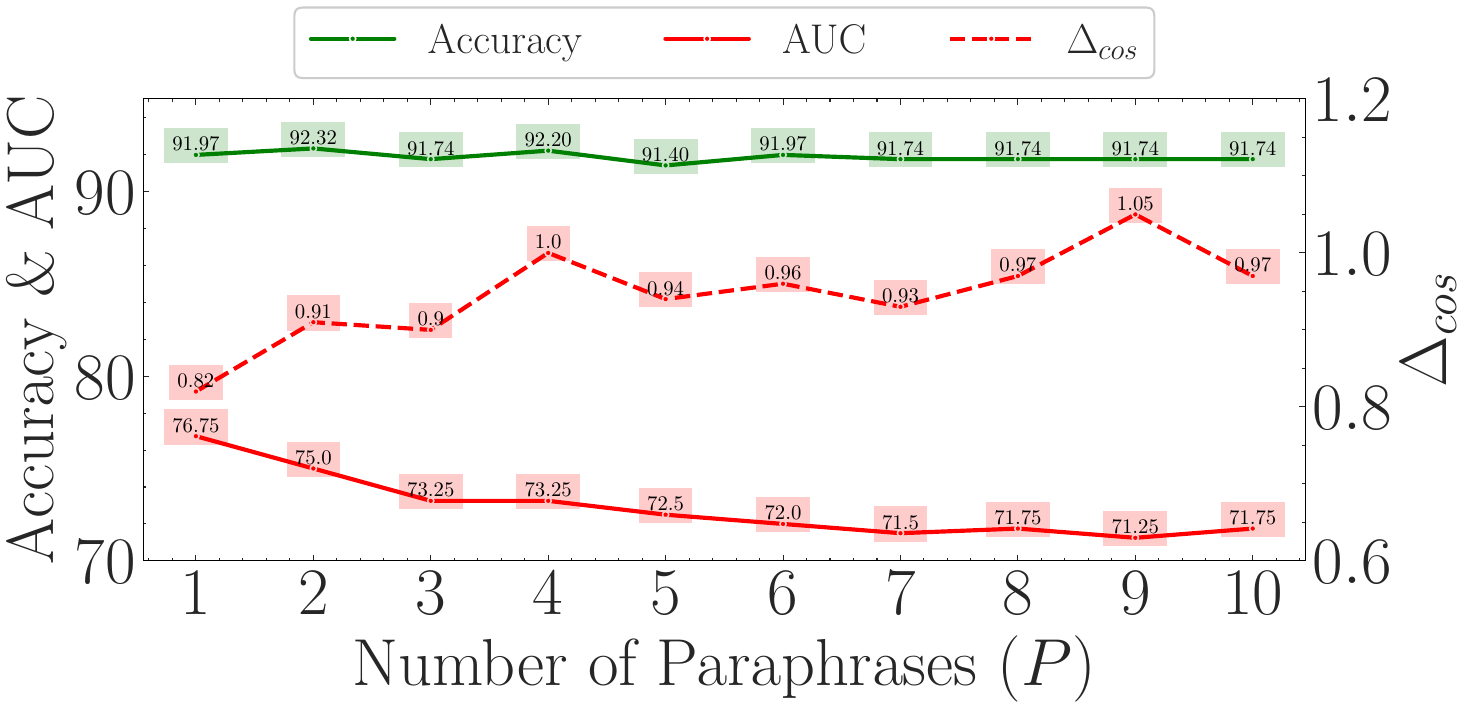}
    \caption{\sst}
    \end{subfigure}
    \begin{subfigure}{0.49\textwidth}
    \centering\includegraphics[width=\linewidth,keepaspectratio]{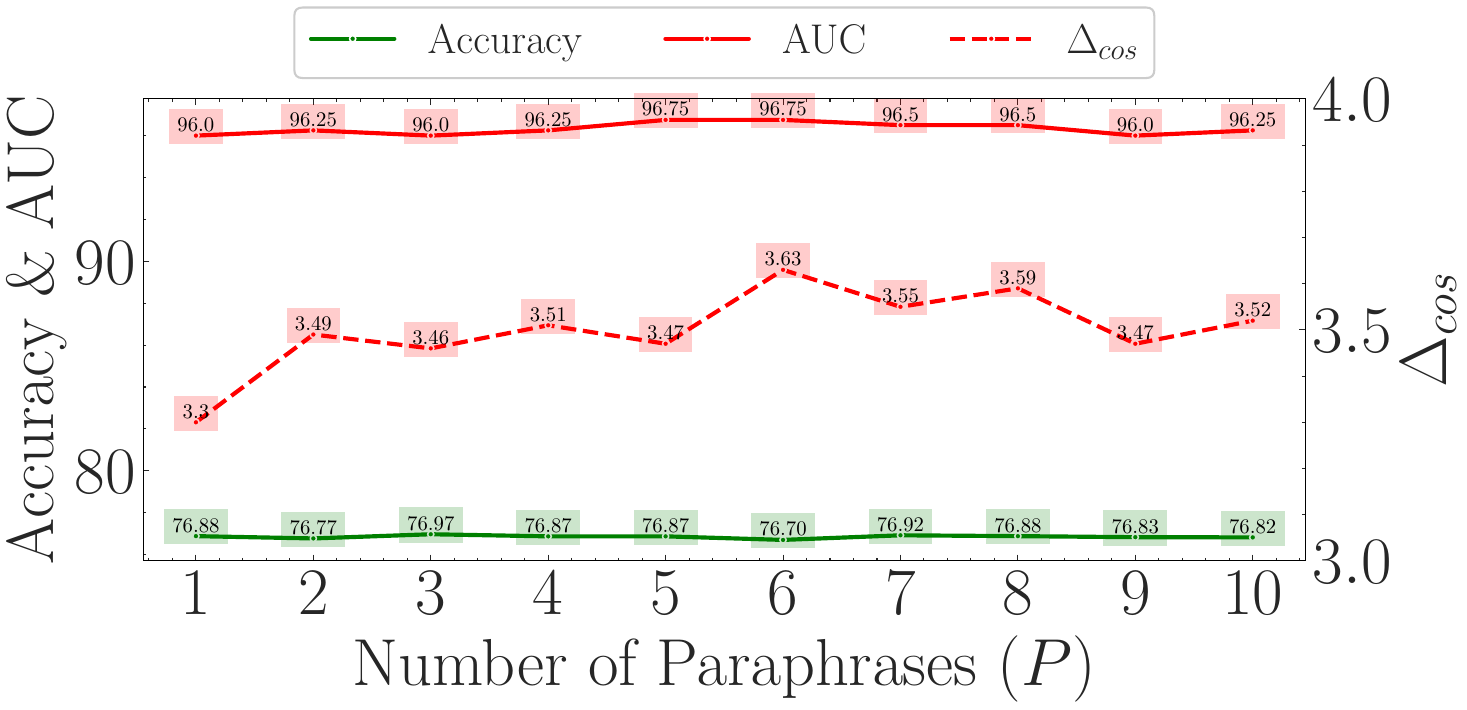}
    \caption{\mind}
    \end{subfigure}
    \begin{subfigure}{0.49\textwidth}
    \centering\includegraphics[width=\linewidth,keepaspectratio]{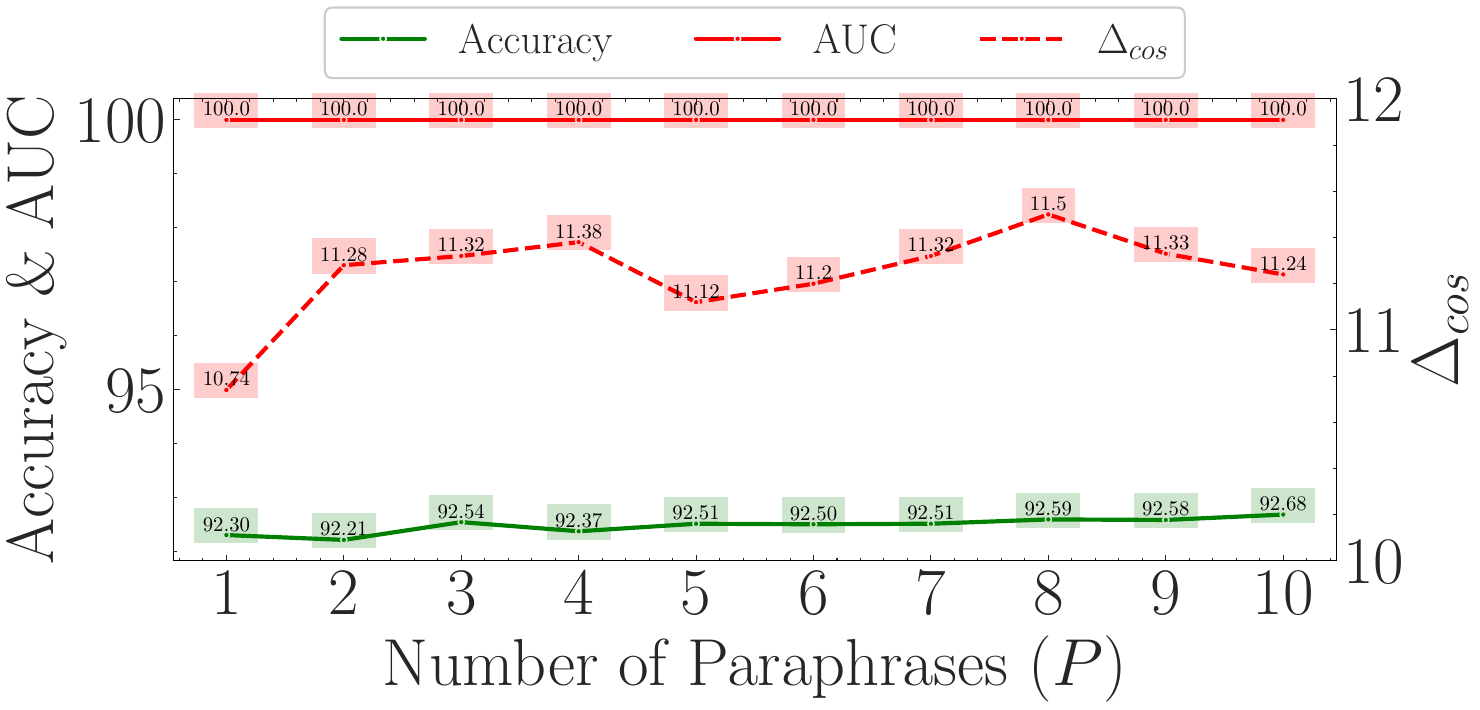}
    \caption{\agnews}
    \end{subfigure}
    \caption{\dipper paraphrase attack performance for different number of paraphrases ($P$). 
    }
    \label{fig:diff-P-DIPPER-attack}
\end{figure*}

In Figures~\ref{fig:diff-P-GPT-attack}~and~\ref{fig:diff-P-DIPPER-attack}, we present the influence of the number of paraphrases ($P$) in \ourattack on verifiability and utility. Utility (accuracy) remains relatively constant, with only minor fluctuations in verifiability. These results indicate that the paraphrasing attack is not sensitive to $P$ and that it is possible to achieve a relatively successful attack with $P=1$. This is significant as \gpt attack will be more economical with fewer paraphrases.

\clearpage
\paragraph{Non-Watermark Cases.}
\begin{table}[t]
\centering
    {\normalsize
    \begin{tabular}{ccSS}
    \toprule[1.5pt]
    \multirow{2}{*}{\textbf{Dataset}} & \multirow{2}{*}{\textbf{Paraphraser}} & \multicolumn{2}{c}{\textbf{Utility}} \\
    \cmidrule(lr){3-4}
    {} & {} & \acc $\uparrow$ & \fscore $\uparrow$ \\
    \toprule
    \multirow{3}{*}{\enron}
         {} & {\gpt} & 92.85 & 92.85 \\
         {} & {\dipper} & 92.00 & 91.99\\
         {} & {\nllb} & 93.25 & 93.25 \\
    \midrule
    \multirow{3}{*}{\sst}
         {} & {\gpt}  & 92.43 & 92.43 \\
         {} & {\dipper} & 91.17 & 91.13 \\
         {} & {\nllb} & 92.20 & 92.18 \\
    \midrule
    \multirow{3}{*}{\mind}
        {} & {\gpt}  & 76.97 & 51.29 \\
        {} & {\dipper} & 76.98 & 50.68 \\
        {} & {\nllb} & 76.72 & 50.47 \\
    \midrule
    \multirow{3}{*}{\agnews}
        {} & {\gpt} & 92.22 & 92.21 \\
        {} & {\dipper} & 92.51 & 92.51 \\
        {} & {\nllb} & 92.61 & 92.59 \\
    \bottomrule[1.5pt]
    \end{tabular}}
    \caption[Paraphrasing attack on a non-watermarked victim model.]{Paraphrasing attack on a non-watermarked victim model. Note that utility is preserved, as expected.
    }
    \label{table:attack-on-non-watermarked-model}
\end{table}

It will be unknown to an attacker whether the model they are attempting to imitate contains watermarks or not. In non-watermark cases, the attacker would want no significant {utility} degradation while still being able to imitate the model. We thus perform the paraphrasing attack on a non-watermarked model. \reftab{table:attack-on-non-watermarked-model} illustrates the effectiveness of \ourattack, showing minimal impact on the {utility} metrics of an innocent model. 

\paragraph{Impact of Attacker Model Size.}
\begin{table}[t]
\centering
{\normalsize
    \centering
    \begin{tabular}{ccSSSS}
    \toprule[1.5pt]
    \multirow{2}{*}{\textbf{Dataset}} & {\textbf{Attacker}} & \multicolumn{2}{c}{\textbf{Utility}} & \multicolumn{2}{c}{\textbf{Verifiability}} \\
    \cmidrule(lr){3-4} \cmidrule(lr){5-6}
    {} & \textbf{Model Size} & \acc $\uparrow$ & \fscore $\uparrow$ & {$\Delta_{cos} \downarrow$} & {\auc $\downarrow$} \\
    \toprule
        \multirow{4}{*}{\enron} & {No Attack} & 93.85 & 93.85 & 5.39 & 97.75 \\
        \cmidrule{2-6}
        {} & {Small} & 92.60 & 92.60 & 0.12 & 57.75 \\
        {} & {Base} & 92.45 & 92.45 & 0.57 & 66.50 \\
        {} & {Large} & 92.70 & 92.70 & 0.39 & 61.50 \\
        \midrule
        \multirow{4}{*}{\sst} & {No Attack} & 93.12 & 93.12 & 5.21 & 100.00 \\
        \cmidrule{2-6}
        {} & {Small} & 92.55 & 92.55 & 0.37 & 63.75 \\
        {} & {Base} & 92.78 & 92.77 & 0.82 & 75.25 \\
        {} & {Large} & 93.12 & 93.12 & 0.31 & 62.50 \\
        \midrule
        \multirow{4}{*}{\mind} & {No Attack} & 77.23 & 51.40 & 5.54 & 98.50 \\
        \cmidrule{2-6}
        {} & {Small} & 76.93 & 51.17 & 2.18 & 83.50 \\
        {} & {Base} & 76.99 & 51.48 & 2.15 & 82.25 \\
        {} & {Large} & 76.95 & 51.34 & 2.16 & 78.75 \\
        \midrule
        \multirow{4}{*}{\agnews} & {No Attack} & 93.43 & 93.43 & 15.29 & 100.00 \\
        \cmidrule{2-6}
        {} & {Small} & 92.42 & 92.40 & 8.61 & 100.00 \\
        {} & {Base} & 92.39 & 92.38 & 7.54 & 100.00 \\
        {} & {Large} & 92.51 & 92.49 & 7.29 & 100.00 \\
    \bottomrule[1.5pt]
    \end{tabular}}
\caption[Impact of attacker model size on \ourattack performance.]{Impact of attacker model size on \ourattack performance. ``No Attack'' is the normal \secondWM watermarking case with no paraphrasing attack.
}
\label{table:attack-model-sizes}
\end{table}

We assess whether our attack's performance varies with the attacker model's size in an imitation attack by conducting experiments using \textsc{small}, \textsc{base}, and \textsc{large} variants of the \texttt{BERT} \citep{devlin-etal-2019-bert}. The results, summarised in \reftab{table:attack-model-sizes}, demonstrate that \ourattack performance is not sensitive to the attacker model size. 

\clearpage
\section{Conclusion}
\label{sec:attack-conclusion}
This chapter highlights the vulnerabilities of existing EaaS watermarks (\secondWM and \firstWM) when subjected to input text paraphrasing in an imitation attack. Our proposed methodology involves generating multiple paraphrases for input text and combining (averaging) their embeddings, effectively reducing the impact of trigger words and thereby diluting the watermark. We study both empirically and theoretically the effects of paraphrasing on watermark weights. 
We perform these attacks using different paraphrasers on EaaS watermarks across various datasets. The attack is successful in most scenarios, demonstrating its effectiveness. The only negative result is the \agnews dataset, where due to longer text, paraphrasing is unable to dilute the watermark enough to avoid detection. We analyse the quality of paraphrases to motivate why it is important to have a filter to keep only good paraphrases in the attack. Lastly, we present additional ablation studies on different paraphrasing attack components and variations to understand their impact. Most of these studies suggest that attack performance is insensitive to these variations.

We want to clarify that our intention in this chapter is to raise awareness about this new form of attack, as we believe the first step in improving security is by exposing vulnerabilities. In the next chapter, as a countermeasure, a new watermarking technique, \ourdefence, is proposed that is resilient to paraphrasing attacks.

\def\chaptertitle{Watermarking EaaS with Linear Transformation}

\lhead{\emph{\chaptertitle}}

\chapter{\chaptertitle}
\label{chapter4}

\section{Introduction}

In the last chapter, we demonstrated that paraphrasing dilutes the current state-of-the-art EaaS watermarks, thereby uncovering a new type of vulnerability in these watermarking techniques.
The drawback of existing watermarks is their reliance on {trigger words} to implant watermarks. This limitation motivated our paraphrasing attack, which exploits this trigger word dependency by reducing their impact through the combination of multiple paraphrases.

To enhance the robustness of the watermarks and answer \textbf{RQ2}, we introduce a novel watermarking technique, \textbf{W}atermarking \textbf{E}aaS with Linear \textbf{T}ransformation (\ourdefence), in this chapter. This watermark method applies linear transformations to the original embeddings to implant watermarks and then verifies the watermarks through reverse transformations.
The advantage of \ourdefence is that it does not depend on {trigger words} to inject watermarks or the input text in the watermarking process. We analyse \ourdefence both theoretically and empirically to show it is {robust} against the new paraphrasing attack. 

In this chapter, we begin by outlining the methodology for our defence, \ourdefence in \refsec{sec:wet-method}.
In \refsec{sec:RQ2-exps}, we evaluate our new watermarking technique and benchmark it against paraphrasing attacks. For completeness, we also discuss the initial methodology involving ``hyperdimensions'' in \refsec{sec:hyp-obfuscation}, which inspires the linear transformation technique. We discuss its flaws and the evolution towards \ourdefence. Finally, we conclude the chapter by summarising the findings in \refsec{sec:ch4-conclusion}.

\section{Methodology}
\label{sec:wet-method}
In this section, we outline the setup under which we develop the EaaS watermark for imitation attacks in \refsecnum{sec:problem-def}.
Subsequently, we describe the new watermark protocol, \ourdefence in \refsec{sec:wet-defense}, followed by theoretical proof demonstrating the robustness of \ourdefence against paraphrasing attacks in \refsec{sec:RQ2-RQ1-proof}.

\subsection{Problem Definition}
\label{sec:problem-def}
To summarise the threat model from \refsec{sec:threat-model}, we assume that a malicious attacker conducts an imitation attack on a victim EaaS service $\mathbb{S}_v$ based on the model $\Theta_v$. The attacker queries $\mathbb{S}_v$ to collect the embeddings (which are watermarked, unbeknownst to the attacker) for a set of input texts $D_a$, which will then be used for training an attacker model $\Theta_a$. The goal of the attacker is to provide a competitive EaaS service $\mathbb{S}_a$ and they may actively employ strategies to remove or bypass the watermark. 
For the victim, \ie the EaaS provider, it is crucial that the watermarked embeddings $\ve_p$ perform similarly to the original non-watermarked embeddings $\ve_o$ on downstream tasks. To determine whether their model has been copied, the victim will query suspicious services $\mathbb{S}_a$ to check if the returned embeddings contain the injected watermarks.

\begin{figure*}[t]
    \centering
    \includegraphics[width=0.98\textwidth,keepaspectratio]{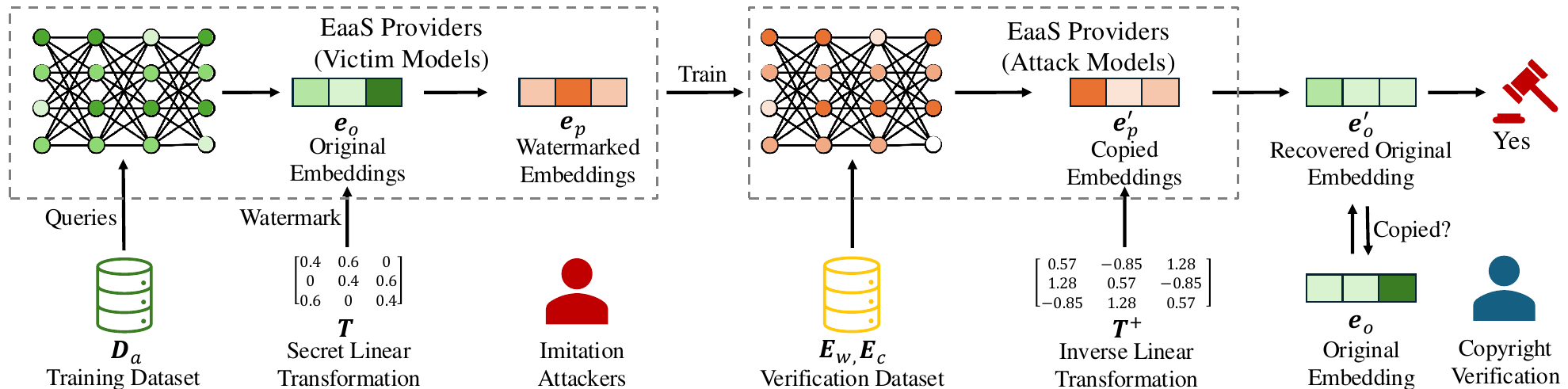}
    \caption[An overview of the workflow for \ourdefence.]{An overview of the workflow for \ourdefence. The left block shows the watermarking process using a secret transformation matrix $\rmT$. The right block shows the watermark verification process, employing the {pseudoinverse} of the transformation matrix $\rmT^+$. The recovered embedding $\ve_{o}^{\prime}$ and the original embedding $\ve_o$ are then compared in copyright verification.}
    \label{fig:defence-overview}
\end{figure*}

\subsection{\ourdefence Defence}
\label{sec:wet-defense}

To remedy the shortcomings in the present EaaS watermarking defences identified in the previous chapter, we introduce \textbf{W}atermarking \textbf{E}aaS with Linear \textbf{T}ransformation (\ourdefence), a novel embedding watermarking protocol designed to be {robust} against paraphrasing attacks. This method is illustrated in \reffig{fig:defence-overview}. The core idea is to use a preset linear transformation matrix  $\rmT$, which is unknown to the attacker, to transform an original embedding $\ve_o$ into a watermarked embedding $\ve_p$ (see the left part of \reffig{fig:defence-overview}). Our watermarking technique discards the original embedding dimensions and retains only the transformed ones, making the watermark harder to detect by an attacker.\footnote{We initially explored embedding dimension obfuscation by adding new dimensions mixed with the original ones, inspired by \citet{yan2023rethinking}. However, we found that these obfuscated dimensions could be easily identified using feature correlation and feature importance techniques, which we will discuss in \refsec{sec:hyp-obfuscation}.} To check for the watermark in a copied embedding $\ve'_p$ (produced by the attacker model), we apply the reverse transformation (using pseudoinverse $\rmT^{+}$) to it and assess whether the recovered embedding $\ve'_{o}$ is similar to the original embedding $\ve_{o}$ (see the right part of \reffig{fig:defence-overview}). 

\subsubsection*{Watermark Injection Method}

Given a transformation matrix $\rmT$, we (i) multiply the matrix with the original embedding $\ve_o$ and (ii) normalise\footnote{Currently, the embeddings provided by the EaaS providers are unit vectors. Moreover, unit vectors make it more \textit{stealthy}; simply returning ${\rmT \cdot \ve_{o}}$ might give avenues for the attacker to exploit. For these reasons, we normalise the vector after linear transformation. Our preliminary study found no significant difference in performance whether this normalisation step is used or not.} the transformed embedding to a unit vector:
\begin{equation}
    \centering
    \begin{aligned}
        \ve_{p} = \underbrace{\text{Norm}(\underbrace{\rmT \cdot \ve_{o}}_{\text{(i)}})}_{\text{(ii)}} = \frac{\rmT \cdot \ve_{o}}{|| \rmT \cdot \ve_{o} ||} .
    \end{aligned}
    \label{eq:wm-injection}
\end{equation}
A watermark injection example appears in \reffig{fig:wet-injection}.
Unlike previous EaaS watermarks, our approach does not rely on {trigger words} and target embeddings for watermark injection, which is vulnerable to paraphrasing attacks. Instead, we watermark all the output embeddings, leading to denser signals and making it more difficult to bypass while maintaining the same level of \textit{utility} for EaaS users.

\subparagraph{Matrix Construction.}
\label{sec:matrix-construction}
One challenge for \ourdefence is designing the transformation matrix $\rmT$.
An important consideration in constructing $\rmT$ is to balance the trade-off between {utility} and {verifiability}. 
The matrix should be \textit{sparse} because a dense matrix will mix semantic information from many dimensions together, affecting utility
and making subsequent model training (the attacker model in imitation attack) more challenging.
Importantly, the matrix should be \textit{invertible}\footnote{The use of the term ``invertible'' here is rather loose. This property technically is not applicable for rectangular matrices (as could be the case when dimensions of original ($n$) and watermarked ($w$) embeddings are not the same in $\rmT$). We use it to mean achieving accurate pseudoinverse calculations.}, as we will use the \textit{pseudoinverse} in watermark verification to perform a {reverse} transformation to recover the original embeddings from the watermarked ones. It is crucial that the transformation matrix is both \textit{full-rank} and \textit{well-conditioned} for accurate pseudoinverse computation. Moreover, full-rank guarantees a lower condition number, which is beneficial for computing well-conditioned pseudoinverses, as noted by \citet{strang2000linear}. 

\begin{figure}[t]
    \centering
    \begin{subfigure}{\textwidth}
        \centering
        \begin{math}
            \rmT = \begin{bmatrix}
              0.0 & 0.0 & 0.0 & 0.0 & 0.0 & \mathbf{0.37} & \mathbf{0.54} & 0.0 & 0.0 & \mathbf{0.09} \\
              \mathbf{0.09} & 0.0 & 0.0 & 0.0 & 0.0 & 0.0 & \mathbf{0.37} & \mathbf{0.54} & 0.0 & 0.0 \\
              0.0 & \mathbf{0.09} & 0.0 & 0.0 & 0.0 & 0.0 & 0.0 & \mathbf{0.37} & \mathbf{0.54} & 0.0 \\
              0.0 & 0.0 & \mathbf{0.09} & 0.0 & 0.0 & 0.0 & 0.0 & 0.0 & \mathbf{0.37} & \mathbf{0.54} \\
              \mathbf{0.54} & 0.0 & 0.0 & \mathbf{0.09} & 0.0 & 0.0 & 0.0 & 0.0 & 0.0 & \mathbf{0.37} \\
              \mathbf{0.37} & \mathbf{0.54} & 0.0 & 0.0 & \mathbf{0.09} & 0.0 & 0.0 & 0.0 & 0.0 & 0.0 \\
              0.0 & \mathbf{0.37} & \mathbf{0.54} & 0.0 & 0.0 & \mathbf{0.09} & 0.0 & 0.0 & 0.0 & 0.0 \\
              0.0 & 0.0 & \mathbf{0.37} & \mathbf{0.54} & 0.0 & 0.0 & \mathbf{0.09} & 0.0 & 0.0 & 0.0 \\
              0.0 & 0.0 & 0.0 & \mathbf{0.37} & \mathbf{0.54} & 0.0 & 0.0 & \mathbf{0.09} & 0.0 & 0.0 \\
              0.0 & 0.0 & 0.0 & 0.0 & \mathbf{0.37} & \mathbf{0.54} & 0.0 & 0.0 & \mathbf{0.09} & 0.0 \\
            \end{bmatrix}
        \end{math}
    \caption{Example $10 \times 10$ circulant transformation matrix, $\rmT$.}
    \label{fig:T-example}
    \end{subfigure}
    \begin{subfigure}{\textwidth}
        \vspace{.5cm}
        \centering
        \begin{math}
            \underbrace{\rmT}_{\text{Transformation Matrix}} \times \underbrace{\begin{bmatrix}
                0.03\\
                0.10\\
                0.50\\
                0.12\\
                0.14\\
                0.10\\
                0.33\\
                0.35\\
                0.52\\
                0.43\\
                \end{bmatrix}
                }_{\ve_o} = 
            {\text{Norm}\left(\begin{bmatrix}
                0.26\\
                0.31\\
                0.42\\
                0.47\\
                0.18\\
                0.07\\
                0.32\\
                0.28\\
                0.15\\
                0.15\\
                \end{bmatrix}\right)}
             = 
            \underbrace{\begin{bmatrix}
                0.28\\
                0.34\\
                0.46\\
                0.52\\
                0.2\\
                0.08\\
                0.35\\
                0.31\\
                0.17\\
                0.17\\
                \end{bmatrix}}_{\ve_p}
        \end{math}
    \caption{\ourdefence watermark injection example.}
    \label{fig:wet-injection}
    \end{subfigure}
    \caption[An example of \ourdefence defence process.]{An example of \ourdefence defence process. In this illustration, hyperparameters are $n=w=10$ and $k=3$.}
    \label{fig:WET-example}
\end{figure}

\begin{center}
        \begin{algorithm}[t]
        \caption{Transformation Matrix Generation.}
        \begin{algorithmic}[1]
        \Require 
        \Statex n: \# original embedding dimensions
        \Statex k: \# original dimensions used in transformation
        \Statex w: \# watermarked embedding dimensions
        \Function{MatrixGen}{$n,k$}
            \State Initialise $\rmT \gets \phi$
            \State $\text{row} \gets \Call{RowGen}{n,k}$ \Comment{$\mathbb{R}^{1 \times n}$}
            \State $\text{count} \gets 0$
            \For{each $i = 1, 2, \cdots, w$} \Comment{Circular Property}
                \State $\rmT[i] \gets \text{row}$
                \State $\text{row} \gets \Call{Roll}{\text{row}}$ 
                \State $\text{count } \mathrel{+}= 1$
                \If{$\text{count} == n$} \Comment{Re-generate Row}
                    \State $\text{row} \gets \Call{RowGen}{n,k}$ 
                    \State $\text{count} \gets 0$
                \EndIf
            \EndFor
            
            \State \Return $\rmT$ \Comment{$\mathbb{R}^{w \times n}$}
        \EndFunction
        
        \item[]
        \Function{RowGen}{$n,k$}
            \State Initialise $\text{row} \gets \text{Zeroes}(n)$
            \State $\text{positions} \gets \text{Sample}(n,k)$  \Comment{Correlations}
            
                \For{$\text{p}$ in $ \text{positions}$} 
                    \State $\text{row}[\text{p}] \sim U(0,1)$ \Comment{Random Value}
                \EndFor 
                \State $\text{row} \gets \text{Norm}(\text{row})$
                \State \Return $\text{row}$
        \EndFunction
        
        \item[]
        \Function{Roll}{$\text{row}$}
            \State $n \gets \text{len}(\text{row})$ 
            
                \For{each $i = 1, 2, \cdots, n$}
                    \State $\text{row}[i] \gets \text{row}[(i + (n-1))\%n]$ \Comment{Shift one position right}
                \EndFor 
                \State \Return $\text{row}$
        \EndFunction
        \end{algorithmic}
        \label{algo:matrix-generation}
        \end{algorithm}
    
\end{center}

Given these desiderata, we adopt \textbf{circulant matrices} \citep{circulant-matrix} to ensure that these full-rank and well-conditioned properties are met. An example of such a matrix $\rmT$ is illustrated in \reffig{fig:T-example}. 
The first row in $\rmT$ is generated randomly, and subsequent rows are \textit{circulations} of the initial row. The positions and values of non-zero entries in the first row are selected randomly. This random value generation is crucial; if equal weights are assigned, the matrix is not full-rank (as validated in \refsec{app:diff-matrices}). 
A circulant matrix is full-rank if the first row has non-zero fast Fourier transform values (corresponding to the eigenvalues of the circulant matrix), which is more probable\footnote{We check this while constructing $\rmT$ and exit if the condition is unsatisfied. However, we did not encounter such a scenario in our experiments.} with our proposed row construction \citep{circulant-matrix}. 
Moreover, cycle shifts ensure that all dimensions in the original embedding contribute equally to the watermark. This enhances \textit{stealthiness}; otherwise, the attacker could exploit this skewness. 
Thus, the watermarked embeddings in \ourdefence will have more balanced dimensions, \ie all the original embedding dimensions are used a roughly equal number of times in linear transformations.

\refalg{algo:matrix-generation} details the generation of the transformation matrix with aforementioned properties.
Two hyperparameters to consider are $w$ and $k$: $w$ represents the number of dimensions of the watermarked embeddings, and $k$ represents the number of original dimensions used to compute a dimension in the watermarked embeddings. When $w>n$ ($n$ is the original embedding dimensions), this approach will result in repetitive rows, causing rank deficiency and affecting the {stealthiness} of $\rmT$ since some rows will be repeated. To mitigate this issue, we generate a new row each time the iteration exceeds $n$ and continue the cyclic process.
We explore and discuss alternative matrix constructions by relaxing various properties (such as circularity, randomness, and others) in \refsec{app:diff-matrices}.

\subsubsection*{Watermark Verification Method}
\label{sec:defence-verification}
The verification process attempts to decode the watermarked embedding from suspicious EaaS service providers using the authentic $\rmT$ and verify whether it matches the original embedding.
That is, we first apply the {pseudoinverse} of the transformation matrix $\rmT^{+}$ to the copied embedding $\ve'_{p}$ to produce recovered original embedding $\ve'_{o}$:
\begin{equation}
    \centering
    \begin{aligned}
        \underbrace{\ve'_{o}}_{\mathbb{R}^{n \times 1}} &= \underbrace{\rmT^{+}}_{\mathbb{R}^{n \times w}} \cdot \underbrace{\ve'_{p}}_{\mathbb{R}^{w \times 1}},
    \end{aligned}
    \label{eq:reverse-transformation}
\end{equation}
where $\rmT^{+}$ is the Moore-Penrose inverse (\aka pseudoinverse) \citep{strang2000linear}. When $\rmT$ has linearly independent rows (guaranteed by our circulant matrix construction; see \refalg{algo:matrix-generation}), then $\rmT^{+}$ is a right inverse, \ie $\rmT \cdot \rmT^{+} = \mathbf I_w$. 

To check the transformation aligns with the authentic watermark process (\ie using $\rmT$), we measure the similarity between the recovered embedding $\ve'_{o}$ from the attacker model and the original embedding $\ve_{o}$ from the victim model. If the attacker trains their model using the victim's embeddings, the similarity score should be high. As before, we use cosine similarity
to measure similarity as follows:
\begin{align}
    \cos(\ve'_{o},\ve_{o}) = \frac{\ve'_{o} \cdot \ve_{o}}{|| \ve'_{o} || \cdot ||\ve_{o} ||}.
    \label{eq:wet-cos-sim}
\end{align}

\subsection{Robustness to Paraphrasing Attacks}
\label{sec:RQ2-RQ1-proof}
\ourdefence watermarking process is independent of the input text but conditioned on the original embeddings; hence, our approach is naturally {robust} to paraphrasing attacks.
If we use input text anywhere in the watermarking process, the watermark will not be resilient to paraphrasing attacks. 
We now show \textit{theoretically} how the linear transformation used in \ourdefence is {robust} against paraphrasing during imitation attacks, where the attacker model embeddings will still have the watermark.
\begin{theorem}[\ourdefence Effectiveness against Paraphrasing Attack]

Given $P$ watermarked embeddings from $P$ paraphrases for an input text, $\ve_{p}^{i} = f(\ve_{o}^{i})$, where $f$ is a linear transformation function, as defined in \refeq{eq:wm-injection} and $i \in [1 \dots P]$. The average of these paraphrase embeddings is equivalent to a linear transformation of a pseudo-aggregation of the original embeddings, $\hat{\ve}_{o}^{i}$, \ie
\begin{equation}
    \text{avg} ( f(\{\ve_{p}^{i} \}_{i=1}^{P}) ) = f(\text{avg} (\{\hat{\ve}_{o}^{i} \}_{i=1}^{P}). \notag
\end{equation}

        \begin{proof}
            \begin{align*}
                \text{avg} ( f(\{\ve_{p}^{i} \}_{i=1}^{P}) ) = \text{avg} (\{{\underbrace{\text{Norm}(\rmT \cdot \ve_{o}^{i})}_{
                    \triangleq \alpha_{i} \cdot \rmT \cdot \ve_{o}^{i}}
                    }\}) \\
                = \rmT \cdot \frac{1}{P} \sum_{i=1}^{P} {\underbrace{\alpha_i \cdot \ve_{o}^{i}}_{\triangleq \hat{\ve}_{o}^{i}}}
                = \rmT \cdot \text{avg} (\{\hat{\ve}_{o}^{i} \}_{i=1}^{P}) \\
                = f(\text{avg} (\{\hat{\ve}_{o}^{i} \}_{i=1}^{P}).
            \end{align*}
    \end{proof}
\label{theorem:RQ2-proof}
\end{theorem}

The transformation $\rmT$ should be consistent regarding the aggregation on the pseudo embedding $\hat{\ve}$ though distorted by $\alpha_i=1/|| \rmT \cdot \ve_{o}^i ||$. Given \refthm{theorem:RQ2-proof}, the \ourdefence watermark key (\ie $\rmT$) remains intact through the aggregation of paraphrase embeddings (\ie paraphrasing attack).

\section{Experiments}
\label{sec:RQ2-exps}

In this section, we conduct experiments on our proposed \ourdefence watermarking technique. The datasets (\refsec{sec:wet-datasets}), metrics (\refsec{sec:wet-metrics}), and experimental settings (\refsec{sec:wet-exp-setting}) are similar to those in the paraphrasing attack from \refsec{sec:attack-exps}. In \refsec{sec:wet-perf}, we discuss the main experimental results of \ourdefence, evaluating the watermark's performance and its resilience against \ourattack. Subsequently, we analyse two crucial hyperparameters of \ourdefence, $w$ and $k$, in Sections~\ref{sec:diff-k-defence}~and~\ref{sec:diff-h-defence}, respectively. Lastly, we perform ablation studies in \refsec{sec:ablation-defence} investigating \ourdefence under varying scenarios (such as transferability, non-watermark cases, Gaussian noise, and others) and alterations to \ourdefence's components (such as the transformation matrix, verification dataset, attacker model size, and others).

\subsection{Datasets}
\label{sec:wet-datasets}
We use the same datasets as in the paraphrasing attack---\enron \citep{enron}, \sst \citep{sst2}, \mind \citep{mind}, and \agnews \citep{ag_news}. 
To recap, these datasets evaluate a range of downstream classification performances, spanning tasks from spam classification (\enron) to sentiment classification (\sst) to news recommendation and classification (\agnews and \mind). For more details about these datasets, please refer to \refsec{sec:datasets}.

\subsection{Metrics}
\label{sec:wet-metrics}
We rely on the same set of two metrics as in \refsecnum{sec:metrics}. One for downstream task performance ({utility}) and a second for watermark verification accuracy ({verifiability}).

\paragraph{Downstream Task Utility.}

As in \refsecnum{sec:metrics}, we measure downstream task performance by training a classifier using the watermarked embeddings as input and predicting the class labels of the datasets. We report standard classification metrics like accuracy (\acc) and the $F_1$-score (\fscore).

\paragraph{Watermark Verifiability.}
\label{sec:wet-verification}

Following \refsec{sec:metrics}, we calculate $\Delta_{\cos}$ and \texttt{AUC} using 
watermark set $E_w$ (which contains watermarked embeddings generated with transformation matrix $\rmT$) and contrast set $E_c$ (which contains watermarked embeddings generated with a different transformation matrix).
Since there is no notion of trigger words in \ourdefence, the watermark set and contrast set correspond to the backdoor and benign set from \refsec{sec:metrics}, respectively.

There is no target embedding in this case, hence, we compute the average cosine similarity (as defined in \refeq{eq:wet-cos-sim}) between the recovered embeddings ($\ve_{o}^{\prime i }$ from \refeq{eq:reverse-transformation}) and the original embeddings ($\ve_{o}^{ i }$) and then take their difference ($\Delta_{\cos}$):

\begin{equation}
    \centering
    \begin{aligned}
\Delta_{\cos} = \cos_\text{avg} (S_w) - \cos_\text{avg} (S_c), \\
        \cos_\text{avg} (S) = \frac{1}{|S|} \sum_{i=1}^{|S|} \cos(\ve_{o}^{\prime i }, \ve_{o}^{ i }), \\ \notag
    \end{aligned}
\end{equation}
where the sets of recovered and original embedding pairs are constructed by:
\begin{equation}
    \begin{aligned}
        S_w &= \left\{ ( \ve_{o}^{\prime i}, \ve_{o}^{ i} ) | \ve_{p}^{\prime i} \in E_{w} \right\}_{i=1}^{V}, \\
        S_c &= \left\{ ( \vc_{o}^{\prime i}, \vc_{o}^{i} ) | \vc_{p}^{\prime i} \in E_{c} \right\}_{i=1}^{V}. \notag
    \end{aligned}
\end{equation}

\subsection{Experimental Settings}
\label{sec:wet-exp-setting}
Most of the experimental settings remain the same from the paraphrasing attack in \refsec{sec:ch3-exp-setting}.
For our main experiments, we set $w=n=1536$ (recall that $n$ and $w$ are the number of dimensions in the original and watermarked embeddings, respectively) to avoid compressing the embeddings. We investigate different values of $w$ in \refsec{sec:diff-h-defence}. We use $k=25$, the number of original dimensions used in the watermarked embedding dimension calculation; we experiment with other values in \refsec{sec:diff-k-defence}. Finally, we use $V = 250$ as the size of the watermark ($E_w$) and contrast ($E_c$) sets in the watermark verification dataset.

\subsection{Results: Watermark Performance}
\label{sec:wet-perf}
We present the {utility} and {verifiability} performance of \ourdefence watermark with and without paraphrasing attacks in \reftab{table:defence-perf}. Comparing \ourdefence to \secondWM in \reftab{table:ch4-attack-performance}, their downstream performance is about the same---suggesting they are all competitive in terms of maintaining {utility}.
This is important as it shows no significant degradation in the primary utility of the \ourdefence provided embeddings, even though \ourdefence watermarks all the embeddings.
\ourdefence is better when it comes to {verifiability}, as its \auc is 100\% in all cases. 
Furthermore, the $\Delta_{\cos}$ metric performance is also significantly higher. This shows that the reverse transformation used in \ourdefence watermark verification is effective in recovering the original embedding, reflecting a well-conditioned transformation matrix enabling good pseudoinverse computations.
Examining the impact of the paraphrasing attack, \ourdefence is a clear winner here, as all {verifiability} metrics see minimal changes (most importantly, \auc is still 100\%). These results empirically validate \refthm{theorem:RQ2-proof} that \ourdefence is not susceptible to paraphrasing attacks.

\begin{table}[p]
\centering
    {\normalsize
    \begin{tabular}
    {ccccc}
    \toprule[1.5pt]
    {\textbf{Method}} & \textbf{\acc $\uparrow$}  & \textbf{\fscore $\uparrow$}  & {\textbf{$\Delta_{cos} \uparrow$}} & {\textbf{\auc} $\uparrow$} \\
    \toprule
    {\ourdefence} & 94.58$\pm$0.21 & 94.58$\pm$0.21 & 85.67$\pm$6.92 & 100.00$\pm$0.00 \\
    \midrule
    {+\gpt Attack} & 92.73$\pm$0.25 & 92.73$\pm$0.25 & 83.58$\pm$6.43 & 100.00$\pm$0.00 \\
    {+\dipper Attack} & 91.37$\pm$0.10 & 91.36$\pm$0.10 & 83.11$\pm$6.48 & 100.00$\pm$0.00 \\
    {+\nllb Attack} & 93.24$\pm$0.24 & 93.24$\pm$0.24 & 84.28$\pm$6.04 & 100.00$\pm$0.00 \\

    \bottomrule[1.5pt]
    \end{tabular}}
    \vspace{-0.7em}
    \caption*{\large(a) \textbf{\enron}}
    \vspace{0.2em}

    {\normalsize
    \begin{tabular}
    {ccccc}
    \toprule[1.5pt]
    {\ourdefence} & 93.07$\pm$0.40 & 93.07$\pm$0.40 & 88.97$\pm$6.62 & 100.00$\pm$0.00 \\
    \midrule
    {+\gpt Attack} & 92.38$\pm$0.34 & 92.38$\pm$0.34 & 87.02$\pm$6.32 & 100.00$\pm$0.00 \\
    {+\dipper Attack} & 91.77$\pm$0.66 & 91.74$\pm$0.67 & 86.59$\pm$6.33 & 100.00$\pm$0.00 \\
    {+\nllb Attack}& 92.75$\pm$0.34 & 92.74$\pm$0.34 & 87.78$\pm$6.27 & 100.00$\pm$0.00 \\

    \bottomrule[1.5pt]
    \end{tabular}}
    \vspace{-0.7em}
    \caption*{\large(b) \textbf{\sst}}
    \vspace{0.2em}

    {\normalsize
    \begin{tabular}
    {ccccc}
    \toprule[1.5pt]
    {\ourdefence} & 77.11$\pm$0.08 & 51.03$\pm$0.26 & 87.74$\pm$6.17 & 100.00$\pm$0.00 \\
    \midrule
    {+\gpt Attack} & 76.72$\pm$0.05 & 50.62$\pm$0.25 & 87.44$\pm$6.17 & 100.00$\pm$0.00 \\
    {+\dipper Attack} & 76.58$\pm$0.08 & 49.99$\pm$0.23 & 86.81$\pm$5.90 & 100.00$\pm$0.00 \\
    {+\nllb Attack} & 76.47$\pm$0.14 & 49.85$\pm$0.26 & 87.54$\pm$5.91 & 100.00$\pm$0.00 \\

    \bottomrule[1.5pt]
    \end{tabular}}
    \vspace{-0.7em}
    \caption*{\large(c) \textbf{\mind}}
    \vspace{0.2em}

    {\normalsize
    \begin{tabular}
    {ccccc}
    \toprule[1.5pt]
    {\ourdefence} & 93.15$\pm$0.08 & 93.14$\pm$0.08 & 88.35$\pm$6.60 & 100.00$\pm$0.00 \\
    \midrule
    {+\gpt Attack} & 92.22$\pm$0.10 & 92.20$\pm$0.10 & 88.02$\pm$6.14 & 100.00$\pm$0.00 \\
    {+\dipper Attack} & 92.46$\pm$0.18 & 92.45$\pm$0.18 & 87.79$\pm$6.14 & 100.00$\pm$0.00 \\
    {+\nllb Attack} & 92.43$\pm$0.08 & 92.42$\pm$0.08 & 88.44$\pm$5.91 & 100.00$\pm$0.00 \\

    \bottomrule[1.5pt]
    \end{tabular}}
    \vspace{-0.7em}
    \caption*{\large(d) \textbf{\agnews}}
    \vspace{0.2em}
    \caption[Performance of \ourdefence watermark for different scenarios.]{Performance of \ourdefence watermark for different scenarios (normal watermark case and against \ourattack case) on \enron, \sst, \mind, and \agnews datasets.
    +\gpt Attack, +\dipper Attack, and +\nllb Attack represent different paraphrasing attack setups as mentioned in \refsec{sec:para-attack-setups}.
    From a defender's perspective, $\uparrow$ means higher metrics are better. 
    All the metrics are in \%.} 
    \label{table:defence-perf}
\end{table}

\begin{table}[p]
\centering
    {\normalsize
    \begin{tabular}
    {ccccc}
    \toprule[1.5pt]
    {\textbf{Method}} & \textbf{\acc $\uparrow$}  & \textbf{\fscore $\uparrow$}  & {\textbf{$\Delta_{cos} \downarrow$}} & {\textbf{\auc} $\downarrow$} \\
    \toprule

    {\secondWM} & 94.50$\pm$0.34 & 94.50$\pm$0.34 & \textcolor{white}{0}5.20$\pm$0.34 & \textcolor{white}{0}97.40$\pm$0.54 \\
    \midrule
    {+\gpt Attack} & 92.81$\pm$0.21 & 92.81$\pm$0.21 & \textcolor{white}{0}0.70$\pm$0.22 & \textcolor{white}{0}68.90$\pm$7.79 \\
    {+\dipper Attack} & 91.34$\pm$0.52 & 91.33$\pm$0.52 & \textcolor{white}{0}0.46$\pm$0.11 & \textcolor{white}{0}67.50$\pm$5.56 \\
    {+\nllb Attack} & 93.35$\pm$0.23 & 93.35$\pm$0.23 & \textcolor{white}{0}0.65$\pm$0.12 & \textcolor{white}{0}71.95$\pm$4.04 \\
    
    \bottomrule[1.5pt]
    
    \end{tabular}}
    \vspace{-0.7em}
    \caption*{\large(a) \textbf{\enron}}
    \vspace{0.2em}

    {\normalsize
    \begin{tabular}
    {ccccc}
    \toprule[1.5pt]

    {\secondWM} & 93.10$\pm$0.12 & 93.10$\pm$0.12 & \textcolor{white}{0}2.57$\pm$1.19 & \textcolor{white}{0}86.75$\pm$6.20 \\
    \midrule
    {+\gpt Attack} & 92.75$\pm$0.15 & 92.75$\pm$0.15 & \textcolor{white}{0}0.93$\pm$0.09 & \textcolor{white}{0}75.90$\pm$2.91 \\
    {+\dipper Attack} & 91.70$\pm$0.27 & 91.66$\pm$0.27 & \textcolor{white}{0}0.90$\pm$0.17 & \textcolor{white}{0}71.95$\pm$2.69 \\
    {+\nllb Attack} & 92.57$\pm$0.09 & 92.55$\pm$0.08 & \textcolor{white}{0}1.06$\pm$0.19 & \textcolor{white}{0}69.35$\pm$2.94 \\

    \bottomrule[1.5pt]
    \end{tabular}}
    \vspace{-0.7em}
    \caption*{\large(b) \textbf{\sst}}
    \vspace{0.2em}
    
    {\normalsize
    \begin{tabular}
    {ccccc}
    \toprule[1.5pt]

    {\secondWM} & 77.31$\pm$0.08 & 51.47$\pm$0.23 & \textcolor{white}{0}5.27$\pm$0.17 & \textcolor{white}{0}98.10$\pm$0.51 \\
    \midrule
    {+\gpt Attack} & 77.01$\pm$0.05 & 51.24$\pm$0.22 & \textcolor{white}{0}1.85$\pm$0.21 & \textcolor{white}{0}79.40$\pm$3.08 \\
    {+\dipper Attack} & 76.86$\pm$0.07 & 50.54$\pm$0.17 & \textcolor{white}{0}3.47$\pm$0.12 & \textcolor{white}{0}96.70$\pm$0.51 \\
    {+\nllb Attack} & 76.64$\pm$0.10 & 50.36$\pm$0.11 & \textcolor{white}{0}3.89$\pm$0.06 & \textcolor{white}{0}97.80$\pm$0.33 \\

    \bottomrule[1.5pt]
    \end{tabular}}
    \vspace{-0.7em}
    \caption*{\large(c) \textbf{\mind}}
    \vspace{0.2em}

    {\normalsize
    \begin{tabular}
    {ccccc}
    \toprule[1.5pt]

    {\secondWM} & 93.51$\pm$0.13 & 93.50$\pm$0.13 & 14.46$\pm$0.68 & 100.00$\pm$0.00 \\
    \midrule
    {+\gpt Attack} & 92.28$\pm$0.12 & 92.26$\pm$0.13 & \textcolor{white}{0}7.23$\pm$0.34 & 100.00$\pm$0.00 \\
    {+\dipper Attack} & 92.50$\pm$0.11 & 92.48$\pm$0.11 & 11.04$\pm$0.40 & 100.00$\pm$0.00 \\
    {+\nllb Attack} & 92.70$\pm$0.10 & 92.69$\pm$0.10 & 10.56$\pm$0.44 & 100.00$\pm$0.00 \\

    \bottomrule[1.5pt]
    \end{tabular}}
    \vspace{-0.7em}
    \caption*{\large(d) \textbf{\agnews}}
    \vspace{0.2em}
    \caption[Performance of \secondWM watermark for different scenarios.]{Performance of \secondWM watermark for different scenarios (normal watermark case and against \ourattack case) on \enron, \sst, \mind, and \agnews datasets.
    +\gpt Attack, +\dipper Attack, and +\nllb Attack represent different paraphrasing attack setups as mentioned in \refsec{sec:para-attack-setups}.
     From a defender's perspective, $\uparrow$ means higher metrics are better. Reproduced from \refchap{chapter3}.
    } 
    \label{table:ch4-attack-performance}
\end{table}

We now present additional analyses to understand the impact of hyperparameters $k$ and $w$.

\subsection{Analysis: Number of Correlations (\texorpdfstring{$k$}{k})}
\label{sec:diff-k-defence}

\begin{figure}[t]
    \centering
    \includegraphics[width=0.8\linewidth]{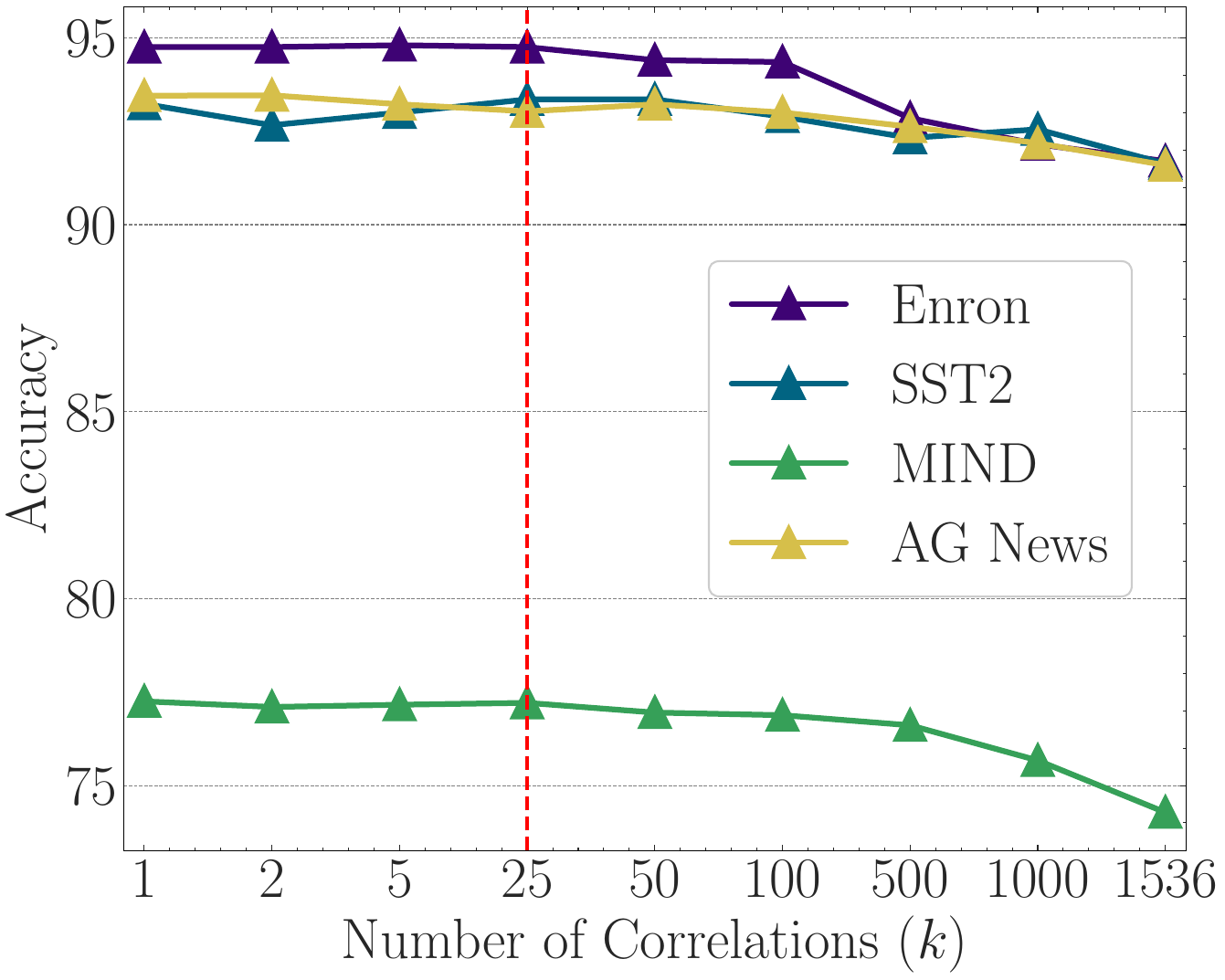}
    \caption[Impact of different values of correlations ($k$) on watermark {utility}.]{Impact of different values of correlations ($k$) on watermark {utility}. We ignore {verifiability} as they are always perfect (\ie 100\%), they can be found in \reftab{table:diff-k-defence}. The \textcolor{red}{red} vertical dashed line represents our chosen value ($k=25$).}
    \label{fig:diff-k-WET}
\end{figure}

We study the influence of $k$, which impacts the sparsity of the matrix. The higher the $k$, the denser the matrix, and the lower the $k$, the sparser the matrix. In our discussion regarding the construction of $\rmT$ in \refsec{sec:matrix-construction}, we mentioned sparsity as one of the requirements, and we hypothesised that the utility might degrade for denser matrices. 
We present the results for varying $k$ in \reffig{fig:diff-k-WET} and \reftab{table:diff-k-defence}; for higher values of $k$ ($>100$), we start seeing degradation in the watermarked embedding {utility}. When we consider more original embedding dimensions for calculating the watermarked embedding, the increased complexity introduces confusion, making it harder for the downstream classification model to learn the underlying semantic properties of the embeddings. On the contrary, lower values of $k$ could make it easier for an attacker to reverse engineer the transformation (in the order of $O(n^{k})$).
Hence, we chose $k=25$ in our experiments, providing a good trade-off. 

\begin{table}[t]
\centering
    {\normalsize
    \begin{tabular}{ccSSSS}
    \toprule[1.5pt]
    \multirow{2}{*}{\textbf{Dataset}} & \multirow{2}{*}{$\bm{k}$} & \multicolumn{2}{c}{\textbf{Utility}} & \multicolumn{2}{c}{\textbf{Verifiability}} \\
    \cmidrule(lr){3-4} \cmidrule(lr){5-6}
    {} & {} & \acc $\uparrow$ & \fscore $\uparrow$ & {$\Delta_{cos} \uparrow$} & {\auc $\uparrow$} \\
    \toprule
    \multirow{9}{*}{\enron}
        & 1 & 94.75 & 94.75 & 89.13 & 100.00 \\
        & 2 & 94.75 & 94.75 & 82.92 & 100.00 \\
        & 5 & 94.80 & 94.80 & 87.64 & 100.00 \\
        & 25 & 94.75 & 94.75 & 89.22 & 100.00 \\
        & 50 & 94.40 & 94.40 & 90.86 & 100.00 \\
        & 100 & 94.35 & 94.35 & 82.84 & 100.00 \\
        & 500 & 92.85 & 92.85 & 81.70 & 100.00 \\
        & 1000 & 92.15 & 92.15 & 82.24 & 100.00 \\
        & 1536 & 91.70 & 91.70 & 85.50 & 100.00 \\
    \midrule

    \multirow{9}{*}{\sst}
        & 1 & 93.23 & 93.23 & 91.65 & 100.00 \\
        & 2 & 92.66 & 92.66 & 87.75 & 100.00 \\
        & 5 & 93.00 & 93.00 & 91.59 & 100.00 \\
        & 25 & 93.35 & 93.34 & 93.70 & 100.00 \\
        & 50 & 93.35 & 93.34 & 94.45 & 100.00 \\
        & 100 & 92.89 & 92.89 & 87.39 & 100.00 \\
        & 500 & 92.32 & 92.31 & 85.81 & 100.00 \\
        & 1000 & 92.55 & 92.54 & 86.58 & 100.00 \\
        & 1536 & 91.63 & 91.62 & 89.38 & 100.00 \\
    \midrule

    \multirow{9}{*}{\mind}
        & 1 & 77.25 & 51.40 & 91.62 & 100.00 \\
        & 2 & 77.10 & 51.19 & 85.70 & 100.00 \\
        & 5 & 77.16 & 51.05 & 89.19 & 100.00 \\
        & 25 & 77.21 & 51.36 & 91.12 & 100.00 \\
        & 50 & 76.95 & 50.71 & 92.41 & 100.00 \\
        & 100 & 76.88 & 50.72 & 85.06 & 100.00 \\
        & 500 & 76.61 & 49.85 & 85.42 & 100.00 \\
        & 1000 & 75.67 & 48.24 & 84.74 & 100.00 \\
        & 1536 & 74.28 & 42.96 & 85.46 & 100.00 \\
    \midrule

    \multirow{9}{*}{\agnews}
        & 1 & 93.45 & 93.44 & 92.85 & 100.00 \\
        & 2 & 93.46 & 93.46 & 86.68 & 100.00 \\
        & 5 & 93.22 & 93.22 & 90.59 & 100.00 \\
        & 25 & 93.03 & 93.02 & 92.05 & 100.00 \\
        & 50 & 93.22 & 93.22 & 93.30 & 100.00 \\
        & 100 & 93.00 & 93.00 & 86.90 & 100.00 \\
        & 500 & 92.62 & 92.61 & 86.28 & 100.00 \\
        & 1000 & 92.18 & 92.17 & 86.22 & 100.00 \\
        & 1536 & 91.59 & 91.58 & 86.66 & 100.00 \\

    \bottomrule[1.5pt]
    \end{tabular}}
    \caption{Impact of number of correlation ($k$) in \ourdefence.}
    \label{table:diff-k-defence}
\end{table}

\clearpage
\subsection{Analysis: Number of Watermarked Dimensions (\texorpdfstring{$w$}{w})}
\label{sec:diff-h-defence}
\begin{figure}[t]
    \centering
    \includegraphics[width=0.8\linewidth]{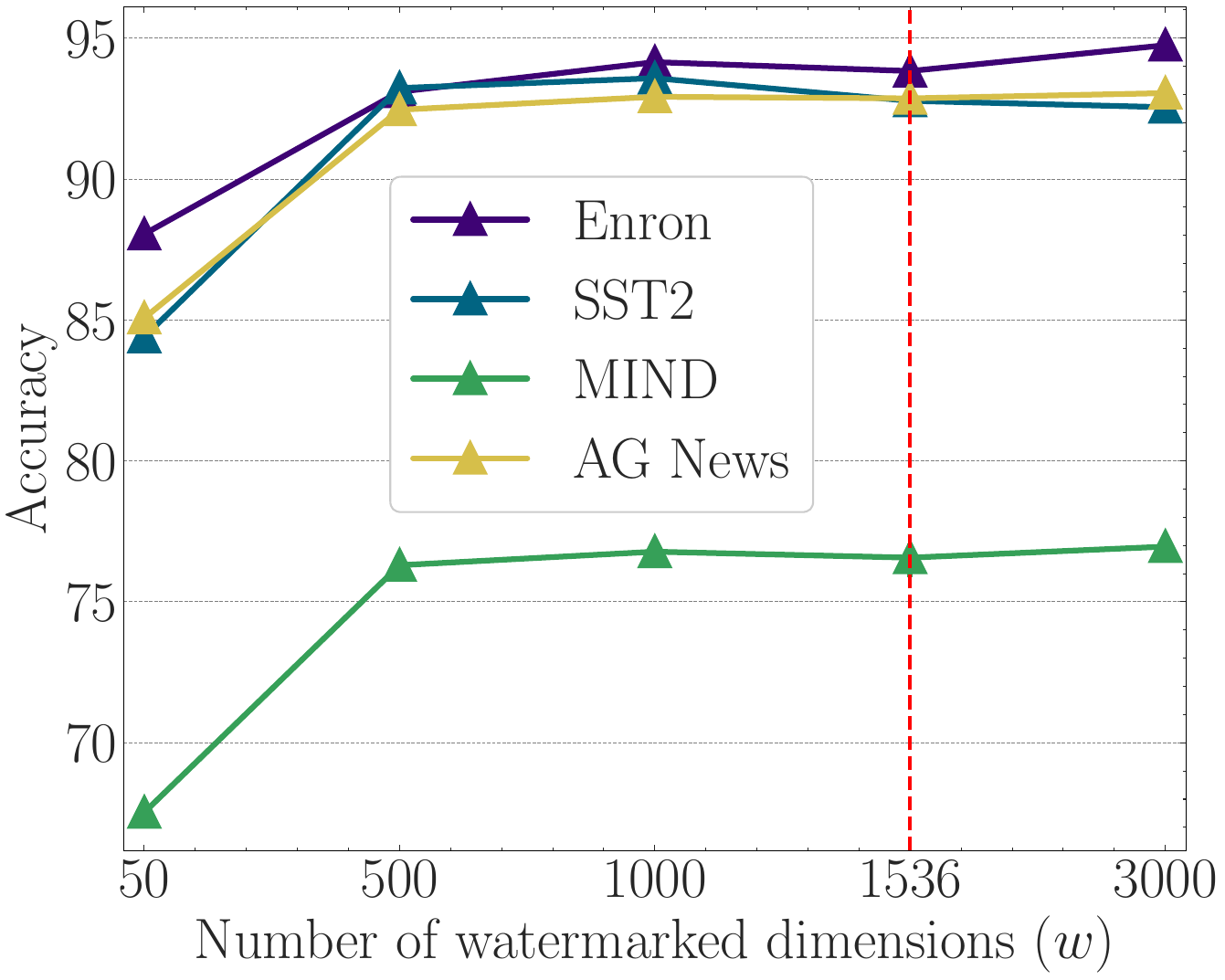}
    \caption[Impact of different values of watermark dimensions ($w$) on watermark {utility}. ]{Impact of different values of watermark dimensions ($w$) on watermark {utility}. 
    We ignore {verifiability} as they are always perfect (\ie 100\%), they can be found in \reftab{table:diff-h-defence}.
    The \textcolor{red}{red} vertical dashed line represents our chosen value  ($w=1536$).}
    \label{fig:diff-w-WET}
\end{figure}

The number of dimensions of the watermarked embeddings is an important hyperparameter in \ourdefence; we present the results of varying $w$ in \reffig{fig:diff-w-WET} and \reftab{table:diff-h-defence}.
We observe from \reffig{fig:diff-w-WET} that smaller values of $w$ might also work. With linear transformations and our balanced matrix generation, we could capture all the semantic properties in fewer dimensions.
This observation demonstrates another benefit of our \ourdefence technique: it can also be used for \textit{compression}. That said, {utility} is only measured using simple classification tasks (following prior studies), and as such, these results may differ if the embeddings are used for more complex tasks. Therefore, we use $w=n$ in our experiments, where we do not drop or add more dimensions from the original embedding case.
Technically, we could also have higher dimensions for watermarked embeddings, but we did not consider this as this would reduce the computational efficiency of the embeddings. 

\begin{table}[t]
\centering
    {\normalsize
    \begin{tabular}{ccSSSS}
    \toprule[1.5pt]
    \multirow{2}{*}{\textbf{Dataset}} & \multirow{2}{*}{$\bm{w}$} & \multicolumn{2}{c}{\textbf{Utility}} & \multicolumn{2}{c}{\textbf{Verifiability}} \\
    \cmidrule(lr){3-4} \cmidrule(lr){5-6}
    {} & {} & \acc $\uparrow$ & \fscore $\uparrow$ & {$\Delta_{cos} \uparrow$} & {\auc $\uparrow$} \\
    \toprule
    \multirow{5}{*}{\enron}
        & 50 & 88.05 & 88.04 & 10.58 & 100.00 \\
        & 500 & 93.10 & 93.10 & 53.34 & 100.00 \\
        & 1000 & 94.15 & 94.15 & 74.95 & 100.00 \\
        & 1536 & 94.75 & 94.75 & 89.22 & 100.00 \\
        & 3000 & 94.75 & 94.75 & 90.49 & 100.00 \\
    \midrule

    \multirow{5}{*}{\sst}
        & 50 & 84.40 & 84.34 & 10.20 & 100.00 \\
        & 500 & 93.23 & 93.23 & 54.07 & 100.00 \\
        & 1000 & 93.58 & 93.57 & 77.07 & 100.00 \\
        & 1536 & 93.35 & 93.34 & 93.70 & 100.00 \\
        & 3000 & 92.55 & 92.54 & 94.47 & 100.00 \\
    \midrule

    \multirow{5}{*}{\mind}
        & 50 & 67.53 & 37.08 & 11.23 & 100.00 \\
        & 500 & 76.30 & 49.83 & 52.96 & 100.00 \\
        & 1000 & 76.78 & 50.55 & 75.74 & 100.00 \\
        & 1536 & 77.21 & 51.36 & 91.12 & 100.00 \\
        & 3000 & 76.96 & 50.72 & 93.85 & 100.00 \\
    \midrule

    \multirow{5}{*}{\agnews}
        & 50 & 85.07 & 85.04 & 12.44 & 100.00 \\
        & 500 & 92.46 & 92.45 & 53.55 & 100.00 \\
        & 1000 & 92.92 & 92.91 & 76.72 & 100.00 \\
        & 1536 & 93.03 & 93.02 & 92.05 & 100.00 \\
        & 3000 & 93.05 & 93.05 & 94.19 & 100.00 \\

    \bottomrule[1.5pt]
    \end{tabular}}
    \caption{Impact of number of watermarked dimensions ($w$) in \ourdefence.}
    \label{table:diff-h-defence}
\end{table}

\clearpage
\subsection{Ablation Studies}
\label{sec:ablation-defence}
In this section, we perform in-depth ablation studies to analyse the components of \ourdefence and establish their significance. We first study various types of matrices by relaxing the properties of our proposed matrix construction to justify our matrix choice. We then consider two scenarios where an attacker might add noise to their provided embedding or stop imitation attacks early. We show \ourdefence is still effective in these scenarios. Finally, we conduct similar ablation studies as in \refsec{sec:attack-ablation} for \ourattack: experimenting with different verification dataset sizes, non-watermark cases, and attacker model sizes. All these ablation studies are carried out using the experimental settings outlined in \refsec{sec:wet-exp-setting} unless stated otherwise.

\paragraph{Different Transformation Matrices Construction.}
\label{app:diff-matrices}

We present additional results with different constructions of the transformation matrix, $\rmT$, a critical component of \ourdefence. Some properties of $\rmT$ (outlined in \refsecnum{sec:matrix-construction}) are altered to ascertain their importance:
\begin{subequations}
    \begin{align}
        \forall_{i \in \{1..w\}} \sum_{j = 1}^{n}\rmT_{i,j} = 1 && \text{Normalisation}, \label{eq:norm} \\
        \rmT_{i,j} = \rmT_{(i+1)\%w,(j+1)\%n} && \text{Circulation}, \label{eq:circular} \\
        \forall_{j \in \text{positions}} \rmT_{0,j} \sim U(0, 1) && \text{Random Value}, \label{eq:random} \\
        \text{positions} = \text{Random}(\{1,..,n\}, k) && \text{Random Position}. \label{eq:corr}
    \end{align}
    \label{eq:T-properties}
\end{subequations}
\subparagraph{Non-Normalised Circulant Matrix.}
We relax the \refeq{eq:norm} property, \ie we do not normalise the values of $\rmT$ rows.

\subparagraph{New Weights Circulant Matrix.}
\label{sec:diff-wts-T}
In this matrix, we construct new weights for each row in the circulant matrix, relaxing the circulation property (\refeq{eq:circular}).

\subparagraph{Equal Weights Circulant Matrix.} We set $1/k$ as the value for non-zero positions in the row, relaxing the random value property (\refeq{eq:random}).

\subparagraph{Sequential Position Circulant Matrix.} We pick the first $k$ dimensions in the row as the non-zero positions, relaxing the random position property (\refeq{eq:corr}).

\subparagraph{Sequential Position and Equal Weights Circulant Matrix.} This matrix is a combination of the previous two constructions, relaxing Equations~\ref{eq:random}~and~\ref{eq:corr}.

\subparagraph{Pure Random Matrix.} A fully random generation process where we randomly pick $k$ non-zero positions and assign random values to them for each row (relaxing Equations~\ref{eq:circular},~\ref{eq:random},~and~\ref{eq:corr}). 

\begin{table}[t]
\centering
    \begin{minipage}{0.98\textwidth}
    \resizebox{\textwidth}{!}{%
    \begin{tabular}{ccccccccc}
    \toprule[1.5pt]
    \multirow{3}{*}{$\bm{k}$} & \multirow{3}{*}{\textbf{Property}} & \multicolumn{7}{c}{\textbf{Matrix Type}} \\
    \cmidrule(lr){3-9}
    {} & {} & \multirow{2}{*}{Circulant} & \multirow{2}{*}{Non-Normal.} & \multirow{2}{*}{New Wts.} & \multirow{2}{*}{Eq. Wts.} & \multirow{2}{*}{Seq. Pos.} & {Seq. Pos.} & \multirow{2}{*}{Random} \\
    {} & {} & {} & {} & {} & {} & {} & {\& Eq. Wts.} & {} \\
    \toprule
    \multirow{2}{*}{2} & {Rank} & 1536.0 & 1536.0 & 1516.4 & 1531.5 & 1536.0 & 1535.0 & 1286.9 \\
    {} & {Cond. \#} & 7.75 & 4.51 & $1.28e^{16}$ & $5.33e^{17}$ & 28.62 & $5.51e^{19}$ & $1.14e^{32}$\\
    \midrule
    \multirow{2}{*}{5} & {Rank} & 1536.0 & 1536.0 & 1535.6 & 1535.6 & 1536.0 & 1536.0 & 1524.0\\
    {} & {Cond. \#} & 49.42 & 84.34 & 11400.86 & $2.03e^{17}$ & 129.55 & 1436.10 & $1.14e^{32}$\\
    \midrule
    \multirow{2}{*}{25} & {Rank} & 1536.0 & 1536.0 & 1535.1 & 1536.0 & 1536.0 & 1534.0 & 1535.0\\
    {} & {Cond. \#} & 151.85 & 216.03 & 14266.05 & 213.16 & 338.50 & 11060.22 & 51610.69\\
    \midrule
    \multirow{2}{*}{50} & {Rank} & 1536.0 & 1536.0 & 1535.1 & 1535.7 & 1536.0 & 1533.0 & 1534.7\\
    {} & {Cond. \#} & 319.03 & 188.99 & 19322.73 & $2.96e^{16}$ & 493.97 & $7.64e^{17}$ & 50206.01\\
    \midrule
    \multirow{2}{*}{100} & {Rank} & 1536.0 & 1536.0 & 1535.2 & 1535.9 & 1536.0 & 1527.0 & 1534.4\\
    {} & {Cond. \#} & 674.57 & 569.85 & 20835.62 & $2.31e^{17}$ & 643.74 & $3.57e^{18}$ & 18673.34\\
    
    \bottomrule[1.5pt]
    \end{tabular}}
    
    \end{minipage}
    \caption[Matrix properties for various constructions of $\rmT$.]{Matrix properties (rank $\uparrow$ and condition number $\downarrow$) for various constructions of $\rmT$ as described in \refsec{app:diff-matrices}. For every matrix type, we report average matrix properties over 10 constructions.}
    \label{table:diff-matrix-properties}
\end{table}

\subparagraph{Discussion.}
\reftab{table:diff-matrix-properties} captures the matrix properties: rank and condition number for different constructions of the transformation matrix and varying numbers of correlations ($k$). Recall from \refsec{sec:matrix-construction} that these properties are important because full-rank matrices are invertible, which is essential for \ourdefence watermark verification involving the reverse transformation. At the same time, full-rank correlates with low-condition numbers, leading to well-conditioned pseudoinverses that enable efficient reverse transformations. As expected, pure random matrix construction and matrices using equal and new row weights are poor transformation matrix candidates with high condition numbers and rank deficiency. Although the sequential position circulant matrix is full-rank, its condition number is higher compared to the chosen circulant matrix. Both normalised and non-normalised circulant matrices are full-rank with low condition numbers, showing that normalisation is not a determining factor for rank and condition number.

We now present the \ourdefence watermark performance using these matrices in \reftab{table:diff-matrices}. We observe that equal weights and sequential position-based matrices have high {verifiability}. However, such matrix constructions are less stealthy (\ie the watermark can be more easily reverse-engineered by the attackers). The matrix combining these methods (Seq. pos and Eq. Wts) performs poorly in terms of {verifiability}. The other two constructions of new row weights in circulant matrix and pure random matrix construction have low $\Delta_{\cos}$ metric even though it has perfect \auc. The reason is that matrices in these cases are not full-rank and well-conditioned as noted earlier, leading to subpar reverse transformation. As before, we do not see any performance difference if the transformation matrix row weights are normalised or not. 
Just to ensure no avenues for attackers to exploit, we apply the normalisation step in the transformation matrix.
Putting these results together (Table~\ref{table:diff-matrix-properties}~and~\ref{table:diff-matrices}), we conclude that the circulant matrix (as proposed in \refsec{sec:matrix-construction}) is the best choice.
\begin{table}[t]
\centering
    {\normalsize
    \begin{tabular}{ccSSSS}
    \toprule[1.5pt]
    \multirow{2}{*}{\textbf{Dataset}} & \multirow{2}{*}{\textbf{Matrix Type}} & \multicolumn{2}{c}{\textbf{Utility}} & \multicolumn{2}{c}{\textbf{Verifiability}} \\
    \cmidrule(lr){3-4} \cmidrule(lr){5-6}
    {} & {} & \acc $\uparrow$ & \fscore $\uparrow$ & {$\Delta_{cos} \uparrow$} & {\auc $\uparrow$} \\
    \toprule
    \multirow{7}{*}{\enron}
        & Circulant & 94.75 & 94.75 & 89.22 & 100.00 \\
        \cmidrule{2-6}
        & Non-Normalised & 94.75 & 94.75 & 89.21 & 100.00 \\
        & New Wts. & 94.60 & 94.60 & 21.60 & 99.96 \\
        & Eq. Wts. & 94.40 & 94.40 & 92.81 & 100.00 \\
        & Seq. Pos.  & 93.40 & 93.40 & 69.91 & 100.00 \\
        & Seq. Pos. and Eq. Wts. & 92.45 & 92.45 & -0.23 & 47.69 \\
        & Random & 94.30 & 94.30 & 22.95 & 99.96 \\
    \midrule

    \multirow{7}{*}{\sst}
        & Circulant & 93.35 & 93.34 & 93.70 & 100.00 \\
       \cmidrule{2-6}
        & Non-Normalised & 93.58 & 93.57 & 93.70 & 100.00 \\
        & New Wts. & 93.00 & 93.00 & 23.60 & 99.99 \\
        & Eq. Wts. & 92.78 & 92.77 & 96.13 & 100.00 \\
        & Seq. Pos. & 91.97 & 91.97 & 74.31 & 100.00 \\
        & Seq. Pos. and Eq. Wts. & 90.60 & 90.59 & 0.72 & 53.88 \\
        & Random & 92.55 & 92.54 & 25.02 & 100.00 \\
    \midrule

    \multirow{7}{*}{\mind}
        & Circulant  & 77.21 & 51.36 & 91.12 & 100.00 \\
        \cmidrule{2-6}
        & Non-Normalised & 77.02 & 51.01 & 91.11 & 100.00 \\
        & New Wts. & 76.97 & 50.83 & 23.56 & 100.00 \\
        & Eq. Wts. & 77.04 & 51.03 & 95.34 & 100.00 \\
        & Seq. Pos. & 76.61 & 50.06 & 71.93 & 100.00 \\
        & Seq. Pos. and Eq. Wts. & 75.21 & 47.22 & 0.02 & 50.28 \\
        & Random & 77.00 & 50.94 & 24.82 & 100.00 \\
    \midrule

    \multirow{7}{*}{\agnews}
        & Circulant & 93.03 & 93.02 & 92.05 & 100.00 \\
        \cmidrule{2-6}
        & Non-Normalised & 93.04 & 93.03 & 92.03 & 100.00 \\
        & New Wts. & 93.20 & 93.19 & 26.60 & 100.00 \\
        & Eq. Wts. & 93.07 & 93.06 & 96.47 & 100.00 \\
        & Seq. Pos. & 92.41 & 92.40 & 73.79 & 100.00 \\
        & Seq. Pos. and Eq. Wts. & 91.89 & 91.88 & -0.27 & 50.37 \\
        & Random & 92.95 & 92.94 & 27.55 & 100.00 \\
    \bottomrule[1.5pt]
    \end{tabular}}
    \caption[\ourdefence performance using different variation of transformation matrix $\rmT$.]{\ourdefence performance using different variation of transformation matrix $\rmT$ as defined in the \refsec{app:diff-matrices}.}
    \label{table:diff-matrices}
\end{table}

\paragraph{Impact of Gaussian Noise.}
\label{app:gauss-noise-defence}

\begin{table}[t]
\centering
    \begin{tabular}{ccSSSS}
    \toprule[1.5pt]
    \multirow{2}{*}{\textbf{Dataset}} & \multirow{2}{*}{$\bm \lambda$} & \multicolumn{2}{c}{\textbf{Utility}} & \multicolumn{2}{c}{\textbf{Verifiability}} \\
    \cmidrule(lr){3-4} \cmidrule(lr){5-6}
    {} & {} & \acc $\uparrow$ & \fscore $\uparrow$ & {$\Delta_{cos} \uparrow$} & {\auc $\uparrow$} \\
    \toprule
    \multirow{5}{*}{\enron}
        & 0.01 & 93.45 & 93.45 & 62.29 & 100.00 \\
        & 0.05 & 84.00 & 84.00 & 17.39 & 100.00 \\
        & 0.10 & 73.60 & 73.59 & 8.84 & 99.00 \\
        & 0.50 & 52.30 & 51.30 & 1.77 & 70.62 \\
        & 1.00 & 50.95 & 49.52 & 0.88 & 61.71 \\
    \midrule
    \multirow{5}{*}{\sst}
        & 0.01 & 91.40 & 91.39 & 64.78 & 100.00 \\
        & 0.05 & 84.29 & 84.26 & 18.09 & 100.00 \\
        & 0.10 & 73.74 & 73.65 & 9.17 & 99.55 \\
        & 0.50 & 53.67 & 49.87 & 1.79 & 69.03 \\
        & 1.00 & 51.72 & 45.76 & 0.86 & 59.32 \\
    \midrule
    \multirow{5}{*}{\mind}
        & 0.01 & 70.37 & 44.18 & 63.61 & 100.00 \\
        & 0.05 & 63.20 & 33.34 & 17.76 & 100.00 \\
        & 0.10 & 49.83 & 15.87 & 9.03 & 99.26 \\
        & 0.50 & 31.60 & 4.85 & 1.82 & 69.78 \\
        & 1.00 & 29.34 & 4.85 & 0.92 & 60.60 \\
    \midrule
    \multirow{5}{*}{\agnews}
        & 0.01 & 92.28 & 92.25 & 64.09 & 100.00 \\
        & 0.05 & 83.92 & 83.84 & 17.73 & 100.00 \\
        & 0.10 & 65.58 & 65.52 & 9.00 & 99.33 \\
        & 0.50 & 30.00 & 29.88 & 1.79 & 69.59 \\
        & 1.00 & 25.16 & 25.00 & 0.89 & 59.92 \\
    \bottomrule[1.5pt]
    \end{tabular}
    \caption[Impact of different Gaussian noise ($\lambda$) in \ourdefence.]{Impact of different Gaussian noise ($\lambda$) in \ourdefence for experiments using \enron, \sst, \mind, and \agnews.}
    \label{table:gauss-noise-WET}
\end{table}

We evaluate the resilience of \ourdefence verification to perturbations added to provided embeddings by an attacker. This could be one of the techniques used by the attacker to circumvent \ourdefence. 
We assess the effect of Gaussian noise as perturbation and analyse its impact on embeddings for watermark verification and downstream {utility}. Following \citet{morris-etal-2023-text,chen2024text}, we consider different noise levels ($\lambda$) and add noise to embedding $\ve$ as follows:
\begin{equation}
    \begin{aligned}
    \ve_{noisy} = \text{Norm}(\ve + \lambda \cdot \bm{\epsilon}), \quad \bm{\epsilon} \sim \mathcal{N}(0, 1).
    \notag
    \end{aligned}
\end{equation}
\textit{How much perturbation can be handled by \ourdefence?} From \reftab{table:gauss-noise-WET}, we see that at $\lambda=0.05$, there is significant {utility} loss; however, the \auc scores remain perfect for all the datasets at this noise level. 
This is significant because such utility degradation defeats the primary motive of the attacker (\ie to provide a competitive service).
This experiment highlights the robustness of \ourdefence, where it can maintain verifiability even under significant {utility} loss introduced by Gaussian noise.

\paragraph{Transferability of Watermark and Utility.}
\begin{table}[t]
\centering
    {\normalsize
    \begin{tabular}{ccSSSS}
    \toprule[1.5pt]
    \multirow{2}{*}{\textbf{Dataset}} & \multirow{2}{*}{\textbf{Training}} & \multicolumn{2}{c}{\textbf{Utility}} & \multicolumn{2}{c}{\textbf{Verifiability}} \\
    \cmidrule(lr){3-4} \cmidrule(lr){5-6}
    {} & {} & \acc $\uparrow$ & \fscore $\uparrow$ & {$\Delta_{cos} \uparrow$} & {\auc $\uparrow$} \\
    \toprule
    \multirow{6}{*}{\enron}
        & 0.1 & 49.60 & 33.16 & 6.19 & 99.90 \\
        & 1 & 88.00 & 88.00 & 66.74 & 100.00 \\
        & 10 & 91.85 & 91.85 & 86.11 & 100.00 \\
        & 33 & 92.75 & 92.75 & 87.38 & 100.00 \\
        & 50 & 92.75 & 92.75 & 88.49 & 100.00 \\
        & 100 & 94.75 & 94.75 & 89.22 & 100.00 \\
    \midrule
    \multirow{6}{*}{\sst}
        & 0.1 & 50.92 & 33.74 & 18.79 & 100.00 \\
        & 1 & 91.17 & 91.15 & 90.52 & 100.00 \\
        & 10 & 92.09 & 92.07 & 90.86 & 100.00 \\
        & 33 & 93.00 & 93.00 & 92.26 & 100.00 \\
        & 50 & 92.55 & 92.54 & 92.97 & 100.00 \\
        & 100 & 93.35 & 93.34 & 93.70 & 100.00 \\
    \midrule
    \multirow{6}{*}{\mind}
        & 0.1 & 58.78 & 8.29 & 25.69 & 100.00 \\
        & 1 & 63.16 & 15.03 & 86.37 & 100.00 \\
        & 10 & 74.48 & 45.54 & 86.44 & 100.00 \\
        & 33 & 76.20 & 49.02 & 89.19 & 100.00 \\
        & 50 & 76.57 & 50.00 & 90.11 & 100.00 \\
        & 100 & 77.21 & 51.36 & 91.12 & 100.00 \\
    \midrule
    \multirow{6}{*}{\agnews}
        & 0.1 & 25.00 & 10.00 & 4.19 & 97.19 \\
        & 1 & 76.14 & 73.63 & 26.24 & 100.00 \\
        & 10 & 91.75 & 91.74 & 86.85 & 100.00 \\
        & 33 & 92.37 & 92.35 & 90.04 & 100.00 \\
        & 50 & 92.55 & 92.54 & 91.11 & 100.00 \\
        & 100 & 93.03 & 93.02 & 92.05 & 100.00 \\
    \bottomrule[1.5pt]
    \end{tabular}}
    \caption[Transferability of \ourdefence watermark and downstream utility for different amounts of training in an imitation attack.]{Transferability of \ourdefence watermark and downstream utility for different amounts of training (\ie train `10' means the attacker model was limited to a total 10\% of the overall training) in an imitation attack. All metrics here are \%.}
    \label{table:learnability-WET}
\end{table}

This experiment examines whether attacker models learn watermarks (transferability) faster than the semantics of embeddings (utility) during an imitation attack. As per results in \reftab{table:learnability-WET}, the watermark patterns are learned more quickly than the semantics of the embeddings in the imitation attack. 
If we look at the case at 10\% training, before the attacker model learns the semantic information encoded in the embeddings (as indicated by its low utility performance), the watermark has already been injected in the copied embeddings (as indicated by the almost perfect AUC).
This finding is significant as it demonstrates the ease of watermark transferability, which means the watermark can be verified reliably. Also, if the attacker were to monitor the training process and stop upon achieving the required downstream {utility} performance, the results indicate that the attacker model would still have learned the underlying watermark patterns and so cannot avoid detection.

\paragraph{Impact of Size of Verification Dataset.}
\label{app:diff-verif-size-defence}
This study explores the number of samples ($V$) needed for the verification dataset. From \reftab{tab:diff-v}, it is evident that \ourdefence's verification technique is robust even with a small verification sample set. Even with just a single verification sample, we can identify that the victim model has been copied. This is another advantage of our technique compared to \firstWM and \secondWM, which require multiple samples for verification.
Note: we use $V=250$ in main experiments (\refsec{sec:wet-exp-setting}).

\begin{table}[t]
\centering
    {\normalsize
    \begin{tabular}{cSSSSSSSS}
    \toprule[1.5pt]
    \multirow{4}{*}{$\bm{V}$} & \multicolumn{8}{c}{\textbf{Verifiability}} \\
    \cmidrule(lr){2-9}
    {} & \multicolumn{2}{c}{\enron} & \multicolumn{2}{c}{\sst} & \multicolumn{2}{c}{\mind} & \multicolumn{2}{c}{\agnews} \\
    \cmidrule(lr){2-3} \cmidrule(lr){4-5} \cmidrule(lr){6-7} \cmidrule(lr){8-9}
    {} & {$\Delta_{cos} \uparrow$} & {\auc $\uparrow$} & {$\Delta_{cos} \uparrow$} & {\auc $\uparrow$} & {$\Delta_{cos} \uparrow$} & {\auc $\uparrow$} & {$\Delta_{cos} \uparrow$} & {\auc $\uparrow$} \\
    \toprule
        1 & 90.00 & 100.00 & 90.29 & 100.00 & 92.28 & 100.00 & 96.13 & 100.00 \\
        2 & 84.91 & 100.00 & 94.55 & 100.00 & 90.52 & 100.00 & 93.02 & 100.00 \\
        5 & 88.26 & 100.00 & 93.75 & 100.00 & 89.90 & 100.00 & 93.62 & 100.00 \\
        20 & 89.90 & 100.00 & 93.67 & 100.00 & 90.74 & 100.00 & 92.63 & 100.00 \\
        100 & 89.14 & 100.00 & 93.77 & 100.00 & 90.54 & 100.00 & 91.77 & 100.00 \\
        250 & 89.22 & 100.00 & 93.70 & 100.00 & 91.12 & 100.00 & 92.05 & 100.00 \\
        500$^{*}$ & 89.67 & 100.00 & 93.89 & 100.00 & 90.97 & 100.00 & 92.10 & 100.00 \\
        1000 & 89.34 & 100.00 & {-} & {-} & 90.87 & 100.00 & 92.00 & 100.00 \\
    \bottomrule[1.5pt]
    \end{tabular}}
    \caption[Impact of different dataset sizes ($V$) in \ourdefence verification.]{Impact of different dataset sizes ($V$) in \ourdefence verification. \\
    $^{*}$\footnotesize For the \sst dataset (with only 872 test samples), the results are for $V=436$. The reason is that for the verification dataset (as explained in \refsec{sec:defence-verification}), we sample 2 x $V$ input test samples, one for the watermark set and another for the contrast set.
    }
    \label{tab:diff-v}
\end{table}

\paragraph{Non-Watermarked Cases.}
\label{sec:non-wm-wet}
\begin{table}[t]
\centering
    {\normalsize
    \begin{tabular}{cSSSS}
    \toprule[1.5pt]
    \multirow{2}{*}{\textbf{Dataset}} & \multicolumn{2}{c}{\textbf{Utility}} & \multicolumn{2}{c}{\textbf{Verifiability}} \\
    \cmidrule(lr){2-3} \cmidrule(lr){4-5}
    {} & \acc $\uparrow$ & \fscore $\uparrow$ & {$\Delta_{cos} \uparrow$} & {\auc $\uparrow$} \\
    \toprule
        \enron & 94.75 & 94.75 & 85.49 & 100.00 \\
        \sst & 93.35 & 93.34 & 88.37 & 100.00 \\
        \mind & 77.21 & 51.36 & 87.38 & 100.00 \\
        \agnews & 93.03 & 93.02 & 88.54 & 100.00 \\
    \bottomrule[1.5pt]
    \end{tabular}}
\caption{\ourdefence detection performance on a non-watermarked victim model.}
\label{table:non-wm-WET}
\end{table}

This is the case where we have a scenario where we run the \ourdefence verification process on an innocent model (\eg a competitor service that did not imitate the victim model).
To evaluate this, we 
set the contrast set to be a set of embeddings that are not watermarked in any way (recall that in our main experiments in \refsec{sec:wet-verification}, the contrast set is a set of embeddings watermarked with a different transformation matrix).
From \reftab{table:non-wm-WET}, the \auc metric is perfect with high $\Delta_{\cos}$, demonstrating strong watermark verification. This means our \ourdefence can distinguish watermark and non-watermark (contrast set) embeddings; hence, \ourdefence will not falsely classify non-watermarked cases.

\paragraph{Impact of Attacker Model Size.}
\label{app:diff-model-size-defence}
\begin{table}[t]
\centering
    {\normalsize
    \begin{tabular}{ccSSSS}
    \toprule[1.5pt]
    \multirow{2}{*}{\textbf{Dataset}} & {\textbf{Attacker}} & \multicolumn{2}{c}{\textbf{Utility}} & \multicolumn{2}{c}{\textbf{Verifiability}} \\
    \cmidrule(lr){3-4} \cmidrule(lr){5-6}
    {} & {\textbf{Model Size}} & \acc $\uparrow$ & \fscore $\uparrow$ & {$\Delta_{cos} \uparrow$} & {\auc $\uparrow$} \\
    \toprule
        \enron &  \multirow{4}{*}{Small} & 94.55 & 94.55 & 88.74 & 100.00 \\
        \sst & {} & 93.23 & 93.23 & 93.12 & 100.00 \\
        \mind & {} & 77.15 & 51.00 & 90.25 & 100.00 \\
        \agnews & {} & 92.92 & 92.91 & 91.30 & 100.00 \\
        \midrule
        \enron &  \multirow{4}{*}{Base} & 94.75 & 94.75 & 89.22 & 100.00 \\
        \sst & {} & 93.35 & 93.34 & 93.70 & 100.00 \\
        \mind & {} & 77.21 & 51.36 & 91.12 & 100.00 \\
        \agnews & {} & 93.03 & 93.02 & 92.05 & 100.00 \\
        \midrule
        \enron &  \multirow{4}{*}{Large} & 94.40 & 94.40 & 88.32 & 100.00 \\
        \sst & {} & 93.00 & 93.00 & 93.60 & 100.00 \\
        \mind & {} & 76.95 & 50.77 & 90.94 & 100.00 \\
        \agnews & {} & 93.29 & 93.28 & 92.43 & 100.00 \\
    \bottomrule[1.5pt]
    \end{tabular}}
\caption{Impact of attacker model size on \ourdefence performance.}
\label{table:WET-attack-model-sizes}
\end{table}

Lastly, we show that \ourdefence achieves stable performance across various attacker model sizes.
We perform imitation attacks on \ourdefence watermarked embeddings using \textsc{small}, \textsc{base}, and \textsc{large} variants of the \texttt{BERT} model \citep{devlin-etal-2019-bert} as the attacker model. The results, summarised in \reftab{table:WET-attack-model-sizes}, indicate that the defence works effectively with consistent {utility} and {verifiability} across different attacker model sizes. This is important as the attacker's model capacity is unknown, and we need \ourdefence to be effective in all scenarios.

\clearpage
\section{Initial Methodology: Hyperdimension Obfuscation}
\label{sec:hyp-obfuscation}

\begin{figure}[t]
    \centering
    \includegraphics[width=0.8\textwidth,keepaspectratio]{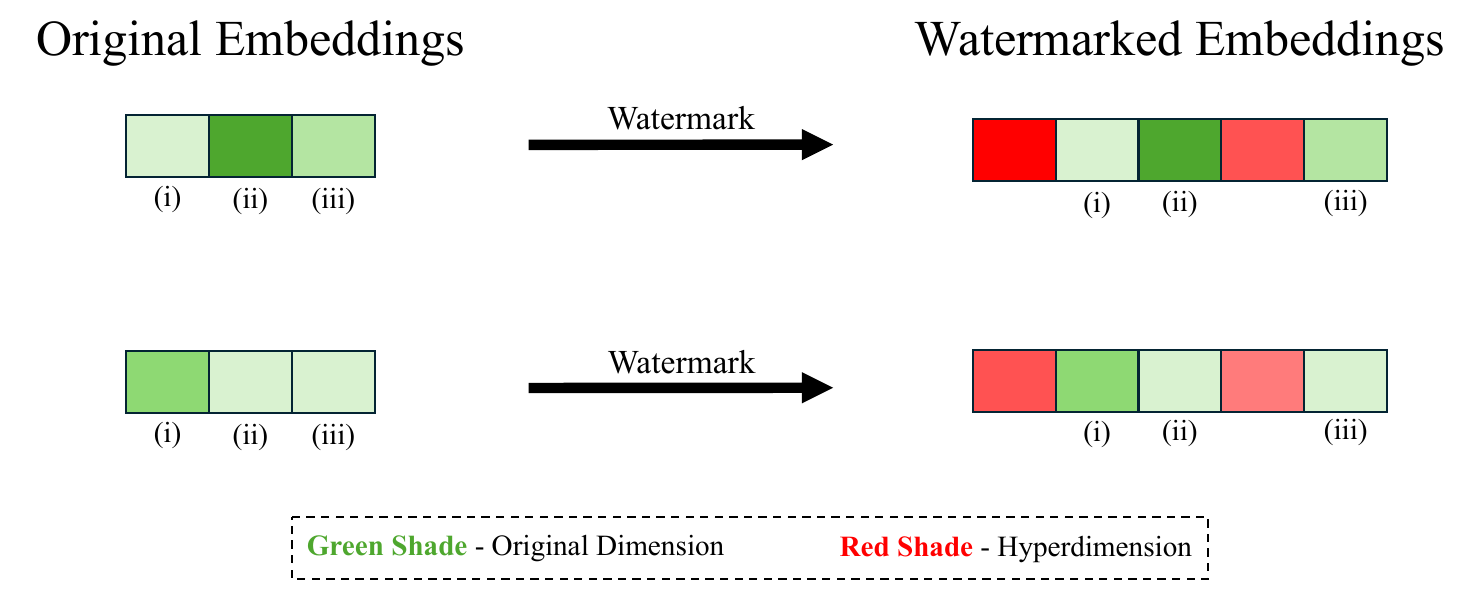}
    \caption[Illustration of the hyperdimension obfuscation protocol.]{Illustration of the hyperdimension obfuscation protocol. We add hyperdimensions (\textcolor{red}{Red}) to the original embedding, obfuscating them with existing dimensions (\textcolor{green!45!black}{Green}). Different shades of colour represent different embedding dimension values.}
    \label{fig:hyp-defence}
\end{figure}
We now describe a preliminary approach that inspires \ourdefence. We first explain this preliminary method and then discuss the issues with it and how it leads to our linear transformation based \ourdefence watermarking method.

This preliminary method incorporates additional watermark dimensions (which we call \textit{``hyperdimensions''}) into the original embeddings while leaving the existing dimensions untouched. The recent work by \citet{yan2023rethinking} motivates this direction, wherein they add dummy neurons (or parameters) in machine learning models to circumvent defences. Given this hyperdimensionality, we anticipate it to withstand paraphrasing attacks since this approach does not depend on {trigger words}.

There are two challenges in the proposed defence, \ie determining (i) the locations and (ii) the values of these hyperdimensions. 
As illustrated in \reffig{fig:hyp-defence}, we could use secret {pre-determined random} locations for these hyperdimensions. The locations must be fixed to retain the embedding {utility} but also mixed (obfuscated) with the original dimensions so that it is {stealthy}. Nonetheless, it should be noted that {stealthiness} will also be dependent on the methodology used in setting the \textit{values} for these hyperdimensions.

We explore two techniques to set the values for these hyperdimensions: (i) pseudo-random based, and (ii) transformation matrix. 
Leveraging cryptographic hardness, we calculate the value for these hyperdimensions using pseudo-random functions, as done in some text watermark works \citep[][\others]{kirchenbauer2023watermark,christ2023undetectable,kuditipudi2023robust}. The advantage of this pseudo-random approach is that it is more {stealthy} because of the pseudo-randomness property; however, because of its deterministic mapping (e.g., by using a deterministic seed value), it also works well for verification. We use the non-watermarked embedding dimensions for calculating cryptographic seeds, making pseudo-random calculations deterministic and verifiable.
Our second technique prepares a \textit{transformation matrix} to project the original (non-watermarked) embedding dimensions to added hyperdimensions. The transformation is a simple linear combination using existing embedding dimensions (similar to \refalg{algo:matrix-generation} but used to create additional dimensions instead of replacing dimensions).
For verification, we use the same ideas as in \ourdefence of reverse transformation (\refsecnum{sec:wet-metrics}), with the only difference being that we work only on the added hyperdimensions.

We conduct preliminary experiments where we maintain pre-determined hyperdimension locations and test different techniques for setting hyperdimension values. Overall, we observe that the transformation matrix-based approach is more \textit{transferable}, which means the verification process is more reliable. While the pseudo-random approach is more {stealthy}, our experiments show that its {transferability} is questionable. That said, for the transformation matrix approach, correlation detection techniques \citep{hall1999correlation} could potentially recover the hyperdimensions, thus compromising its {stealthiness}, which we discuss next.

\subsection*{Stealthiness Evaluation} To evaluate the {stealthiness} of these hyperdimensions, we investigate feature correlation and feature importance techniques between hyperdimensions and original dimensions. Properly mixed feature importance weights indicate that hyperdimensions are indistinguishable from the original ones. Similarly, uncorrelated hyperdimensions are desirable, as otherwise they can be identified by their correlation scores. Note that these {stealthiness} techniques are not applicable to \ourdefence, as the original embedding dimensions are discarded in the provided (watermarked) embeddings.

\paragraph{Feature Importance.}

\begin{figure*}[p]
    \centering
    \begin{subfigure}{0.8\textwidth}
        \includegraphics[width=\textwidth,keepaspectratio]{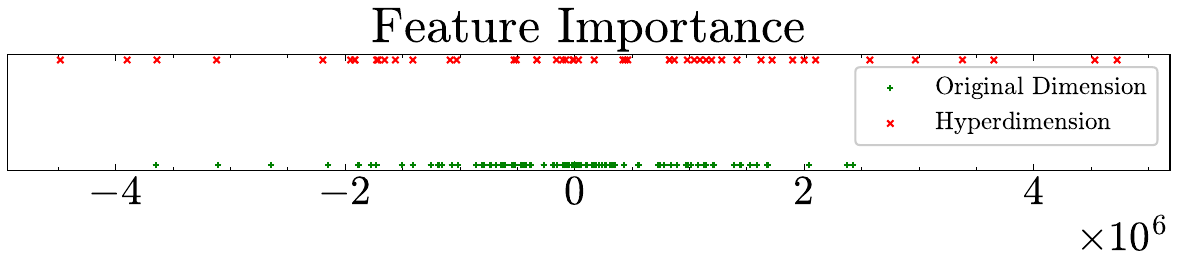}
        \caption{$k=2$}
    \end{subfigure}
    \begin{subfigure}{0.8\textwidth}
        \includegraphics[width=\textwidth,keepaspectratio]{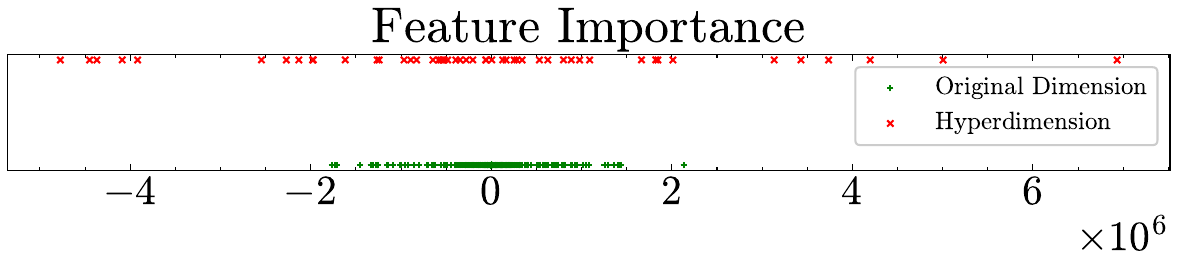}
        \caption{$k=5$}
    \end{subfigure}
    \begin{subfigure}{0.8\textwidth}
        \includegraphics[width=\textwidth,keepaspectratio]{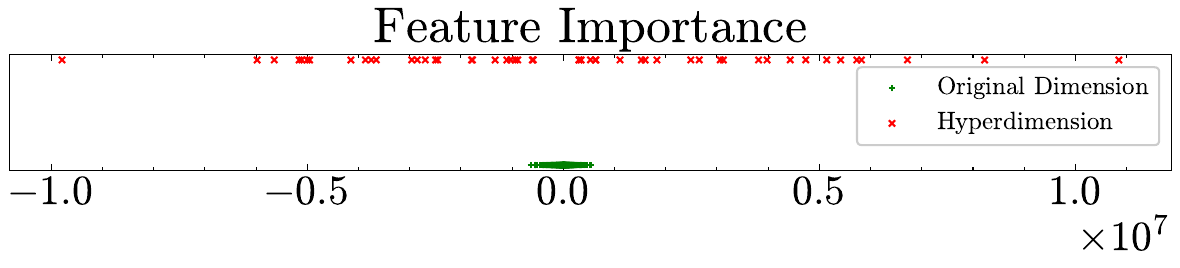}
        \caption{$k=50$}
    \end{subfigure}
    \begin{subfigure}{0.8\textwidth}
        \includegraphics[width=\textwidth,keepaspectratio]{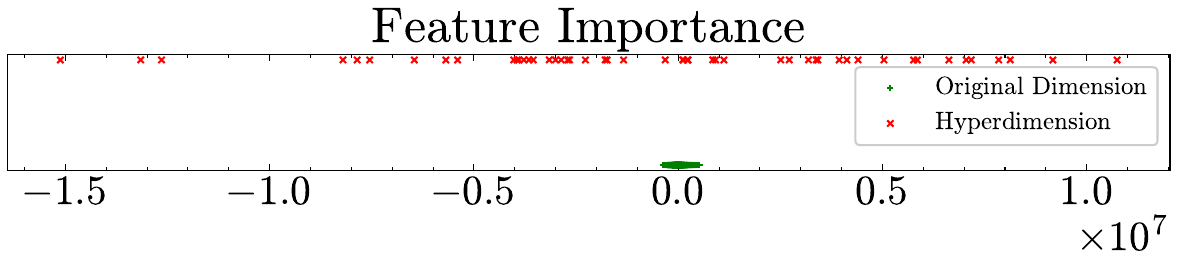}
        \caption{$k=100$}
    \end{subfigure}
    \begin{subfigure}{0.8\textwidth}
        \includegraphics[width=\textwidth,keepaspectratio]{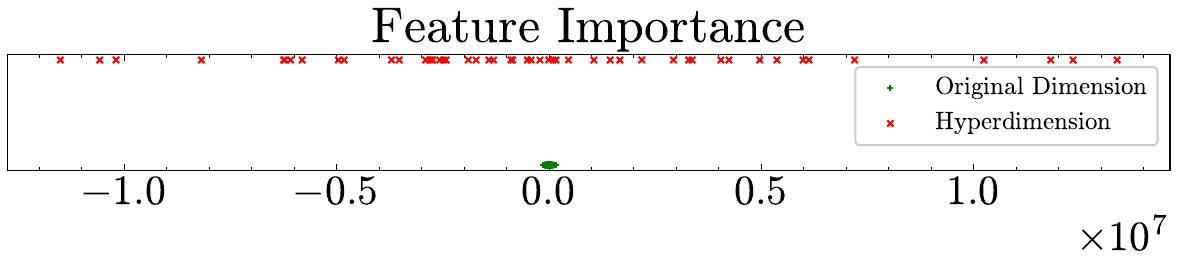}
        \caption{$k=500$}
    \end{subfigure}
    \begin{subfigure}{0.8\textwidth}
        \includegraphics[width=\textwidth,keepaspectratio]{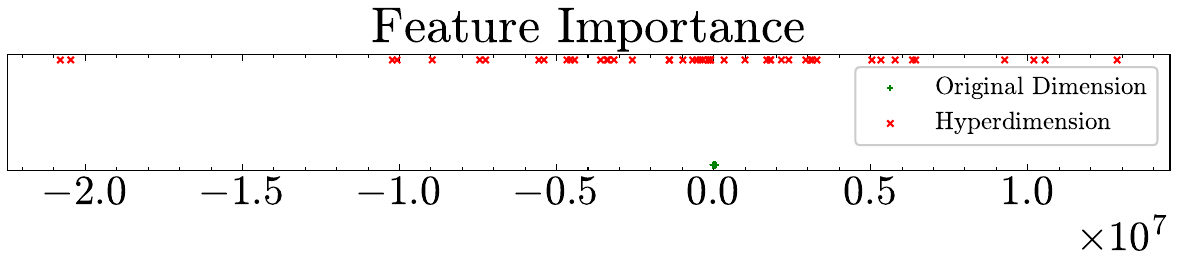}
        \caption{$k=1536$}
    \end{subfigure}
    \caption{Visualisation plots for feature importance of watermarked embedding dimensions in \sst for different values of $k$.}
    \label{fig:feature-importance}
\end{figure*}

We train a linear regression model (as we are dealing with linear transformations) with all the watermarked embeddings (original and hyperdimensions) for the downstream task. We use the weights of the linear regression as the feature importance weights. In \reffig{fig:feature-importance}, we display these plots for different values of $k$. From this, we conclude that we need $k<5$, as for higher values of $k$, hyperdimensions are discernible from the original embedding dimensions, \ie the feature importance values of the original dimensions are close to zero. For higher values of $k$, we have hyperdimensions that have more feature importance, which is unsurprising considering linear combinations used in hyperdimensions will capture all the information in the original dimensions, and so the original dimensions are redundant.

\begin{figure*}[p]
    \centering
    \begin{subfigure}{0.5\textwidth}
        \includegraphics[width=\textwidth,keepaspectratio]{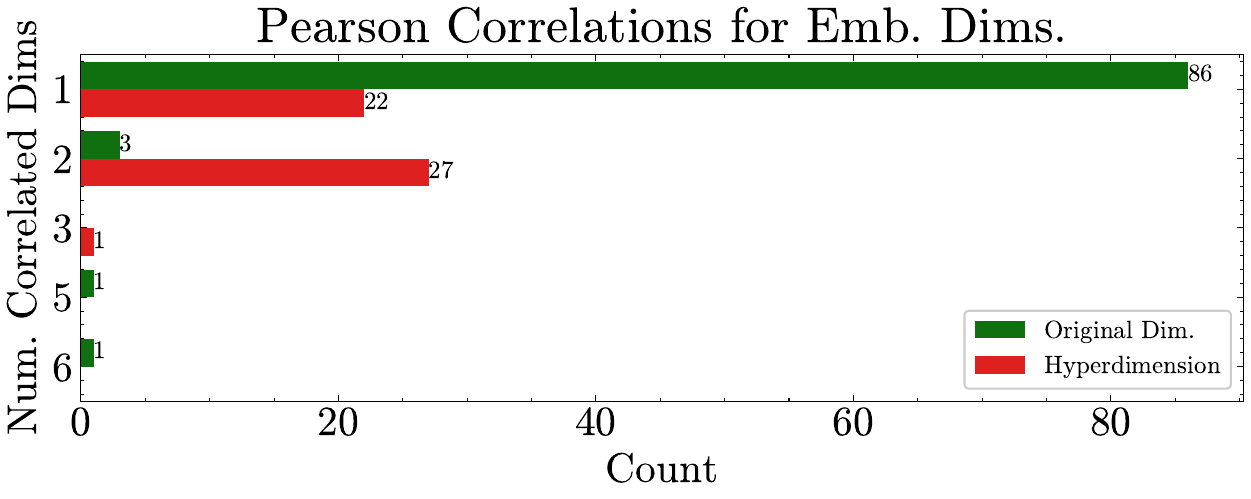}
        \caption{$k=2$}
    \end{subfigure}
    \begin{subfigure}{0.5\textwidth}
        \includegraphics[width=\textwidth,keepaspectratio]{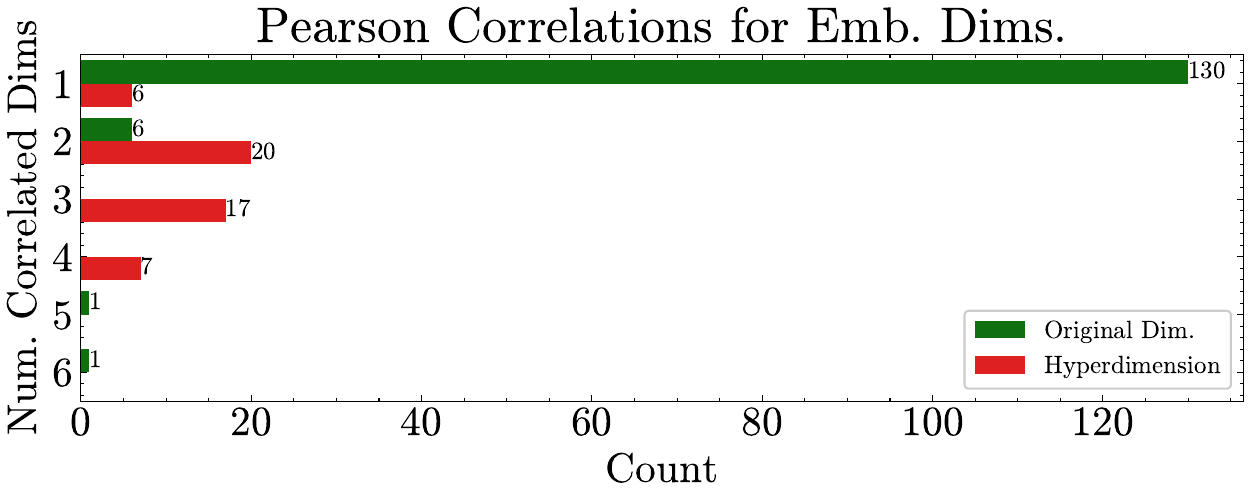}
        \caption{$k=5$}
    \end{subfigure}
    \begin{subfigure}{0.5\textwidth}
        \includegraphics[width=\textwidth,keepaspectratio]{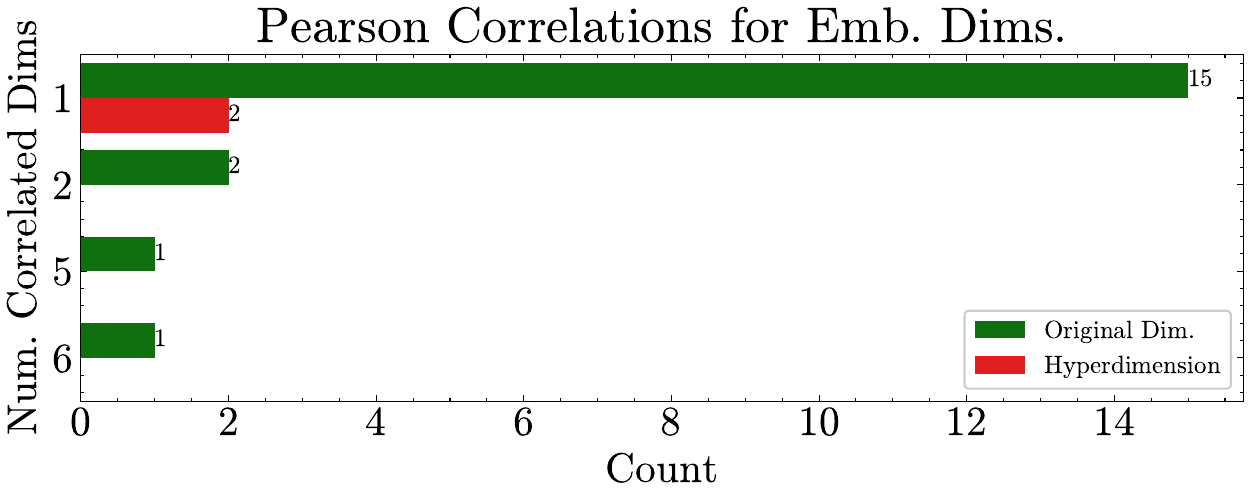}
        \caption{$k=50$}
    \end{subfigure}
    \begin{subfigure}{0.5\textwidth}
        \includegraphics[width=\textwidth,keepaspectratio]{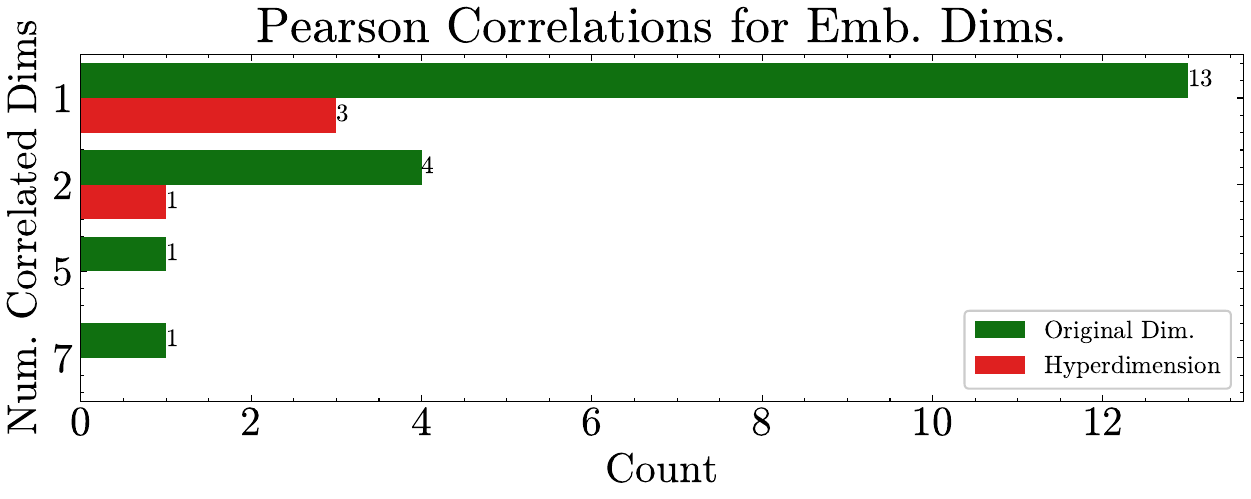}
        \caption{$k=100$}
    \end{subfigure}
    \hspace{.3em}
    \begin{subfigure}{0.5\textwidth}
        \includegraphics[width=\textwidth,keepaspectratio]{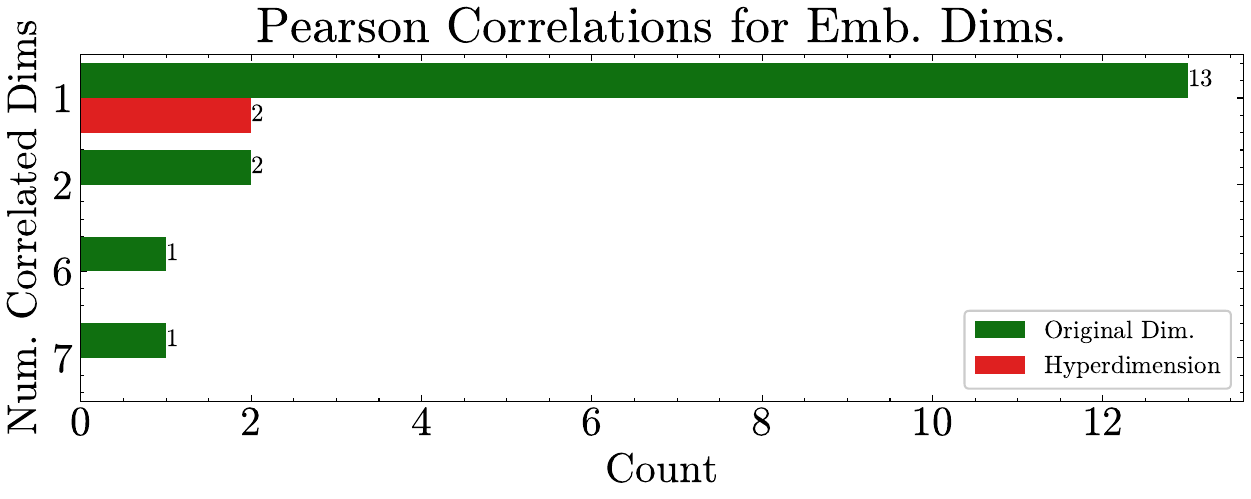}
        \caption{$k=500$}
    \end{subfigure}
    \hspace{.3em}
    \begin{subfigure}{0.5\textwidth}
        \includegraphics[width=\textwidth,keepaspectratio]{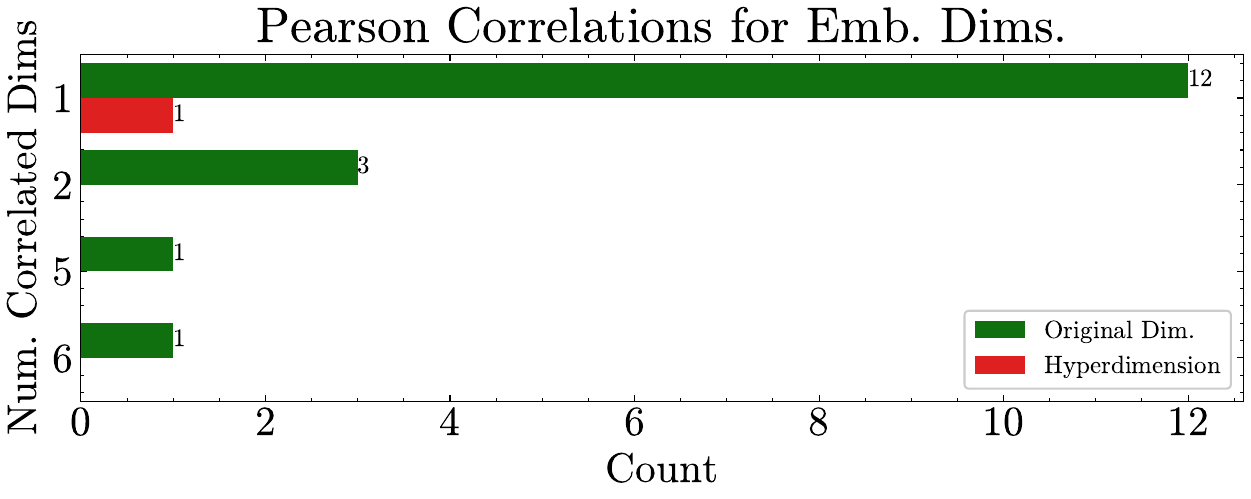}
        \caption{$k=1536$}
    \end{subfigure}
    \caption{Visualisation plots for feature correlations of watermarked embedding dimensions in \sst for different values of $k$.}
    \label{fig:feature-correlations}
\end{figure*}

\paragraph{Feature Correlations.} Pearson's coefficient \citep{sedgwick2012pearson} is a statistical measure capturing the linear relationship (or correlations) between two variables. We apply this for all the dimensions in the embeddings. For stealthiness, hyperdimensions should not be strongly correlated with the original embedding dimensions. Again, we use this statistical measure since we are applying linear transformations. In this analysis, we use Pearson's coefficient with a threshold of 0.4. The plots in \reffig{fig:feature-correlations} indicate that $k$ should be greater than 5, as otherwise, hyperdimensions become very correlated. However, this range conflicts with the values ($k<5$) from the previous feature importance results.

Consequently, these plots (Figures~\ref{fig:feature-importance}~and~\ref{fig:feature-correlations}) lead us to conclude that hyperdimension obfuscation is ineffective as it is detectable, either through feature importance or correlations. These results provide the motivation for WET, which transforms all the original dimensions instead of adding extra hyperdimensions. As explained earlier, the watermarked embeddings in \ourdefence do not have original dimensions used for correlations; therefore, these stealthiness evaluation techniques are ineffective for \ourdefence.

\clearpage
\section{Conclusion}
\label{sec:ch4-conclusion}
In the previous chapter, we showcased the vulnerabilities of existing EaaS watermarks (\secondWM and \firstWM) against paraphrasing in an imitation attack. To address this shortcoming, in this chapter, we devise a simple watermarking technique, \ourdefence, which applies linear transformations to the original embeddings to generate watermarked embeddings. We outline the \ourdefence watermark injection and verification process, followed by a discussion on the transformation matrix construction with required properties. We first theoretically prove that \ourdefence watermarks are preserved after paraphrasing, demonstrating their robustness against paraphrasing attacks. This is further validated by empirical experiments demonstrating near-perfect {verifiability}, even with one sample.
Additionally, the {utility} of embeddings is mostly preserved due to the use of straightforward linear transformations. 
We then justify our choice of \ourdefence hyperparameters controlling the number of correlations used in transformation and dimensions of watermark embeddings. 
Furthermore, we conduct ablation studies to assess the contribution of each component in \ourdefence. We establish why our proposed construction of the transformation matrix is optimal. These ablation studies also cover cases such as resilience to Gaussian noise, watermark transferability, and indifference to the attacker base model size, showcasing \ourdefence's advantages.

In the next chapter, we conclude the thesis, outlining potential future directions and discussing the limitations of this work.

\def\chaptertitle{Conclusion}

\lhead{\emph{\chaptertitle}}
\chapter{\chaptertitle}
\label{chapter5}
\epigraph{\textit{Security is not a product, but a process.}}{---Bruce Schneier\vspace{0.5cm}}

In this chapter, we revisit the research questions proposed in \refsec{sec:RQs}, and summarise the findings and implications in \refsec{sec:concl-overview}. Then, we outline the contributions of this thesis in \refsec{sec:concl-contributions} and discuss the limitations of the current work along with potential future directions in \refsec{sec:limit-and-future}.

\section{Thesis Summary}
\label{sec:concl-overview}
In this thesis, we explored the area of NLP and LLMs, with a particular focus on EaaS. We examined EaaS vulnerabilities, particularly in the context of imitation attacks, and contributed a more effective watermark as a defence. Initially, we highlighted the vulnerabilities of existing EaaS watermarks when paraphrasing input text in an imitation attack. To overcome this shortcoming, we developed a novel watermarking technique, \ourdefence, which applies linear transformations to the original embeddings to generate watermarked embeddings. We conducted comprehensive studies to evaluate the impact of each component involved in the paraphrasing attack and \ourdefence defence.

\refchap{chapter2} presented a detailed literature review, beginning with language models to understand the ubiquity of LLMs. We also established why embeddings are important in NLP due to their generic representations and how the popularity of LLMs has led to the emergence of EaaS. Subsequently, we focused on imitation attack vulnerabilities in these EaaS. We reviewed the literature on watermarking defences for imitation attacks and identified gaps warranting further exploration. We discussed the strengths and weaknesses of the reviewed methods.

In \refchap{chapter3}, we devised a paraphrasing attack that is effective in diluting EaaS watermarks and circumventing copyright verification. Our approach involves generating multiple paraphrases for input text and combining their embeddings, effectively reducing the impact of trigger words (the main component in the watermarks) and thereby diluting the watermark. We also demonstrated the effectiveness of this attack using different paraphrasing setups, such as employing various paraphrasing models and round-trip translation techniques. By exposing flaws and showcasing attacks in the existing watermarking systems, we advocated for better defence.

In \refchap{chapter4}, we next developed a defence mechanism against paraphrasing attacks. We validated the hypothesis presented at the start of the thesis, \ie linearly transforming the original embedding to implant the watermarks, which can then be verified by performing a reverse transformation. We proved theoretically that such linear transformations are retained even with paraphrasing. Empirically, our experiments also verified that \ourdefence is robust against paraphrasing attacks and has a better verifiability performance than previous defence approaches. Additionally, we conducted ablation studies to assess the contribution of each component in \ourdefence.

The overarching theme of this research has been to improve the security of EaaS against imitation attacks, protecting the intellectual property of the models from the service provider's perspective. %
To spur future research, we have open-sourced the code used in this work.\footnote{\url{https://github.com/anudeex/WET.git}} In conclusion, this thesis demonstrated EaaS watermark vulnerabilities to paraphrasing and proposed a countermeasure to the attack with the \ourdefence watermarking technique.

\section{Contributions}
\label{sec:concl-contributions}
In the previous chapters, we have successfully addressed our research questions as outlined in \refsec{sec:RQs}. The main contributions of this thesis are as follows:
\begin{itemize}
    \item \textbf{RQ1:} We introduce and validate that paraphrasing attacks bypass current EaaS watermarking techniques while maintaining the utility of the embeddings. We also conduct extensive experiments to evaluate their effectiveness across various setups and datasets.
    \item \textbf{RQ2:} We design a novel EaaS watermarking method, \ourdefence, and demonstrate its robustness against paraphrasing attacks. \ourdefence maintains the utility of the embeddings post-watermarking with near-perfect AUC verification performance even using a single sample.
\end{itemize}

\subsection*{Social Impact}
We develop a paraphrasing attack that could aid attackers in circumventing EaaS IP infringements. We acknowledge that, with this attack, any existing systems using \secondWM or \firstWM are vulnerable. Nonetheless, we argue that it is critical to show the possibility of such attacks and make users aware of them. The typical first step in security is to first expose the vulnerability. Additionally, to mitigate the aforementioned threat, we contribute an improved watermarking technique, \ourdefence, which could be incorporated with minimal effort.

\section{Limitations and Future Work}
\label{sec:limit-and-future}
We now discuss the limitations of our research and propose some promising directions for future work.

\subsection*{Complex Downstream Tasks for Utility Evaluation}
For the evaluation of watermark \textit{utility}, our focus has been on classification tasks, in line with existing studies \citep{shetty2024warden,peng2023you}. However, these tasks may not fully capture embedding quality. Future studies might explore more complex NLP tasks, such as retrieval and generation \citep{muennighoff-etal-2023-mteb}, to provide a comprehensive understanding of the impact of introducing watermarks into embeddings.

\subsection*{Potential Attacks: Overwriting}
\label{sec:overwriting}
In this thesis, we demonstrate that \ourdefence is a better watermark than both \secondWM \citep{shetty2024warden} and \firstWM \citep{peng2023you}. Unlike existing watermarks, it does not show vulnerability to paraphrasing attacks. Nevertheless, it is important to acknowledge that other potential attacks may exist.

Overwriting watermarks is one such attack \citep{uchida2017embedding,boenisch2021systematic}. Most of the existing watermarks are prone to these overwriting attacks. In this scenario, the watermarked embeddings could undergo further transformation and be overwritten. For example, the attacker could apply a linear transformation to the provided embedding, erasing (overwriting) the original existing watermark, which we discuss next.

\subparagraph{Linear Transformation Overwriting.} Concretely, the attacker might apply another linear transformation to the provided embeddings as follows:
\begin{align}
    \ve_o \overset{\rmT}{\longrightarrow} \ve_p \overset{\rmA}{\longrightarrow} \ve_a, \notag \\
    \ve_o = \rmT^{+} \cdot \rmA^{+} \cdot \ve_a, \notag
\end{align}
where $\ve_o$ is the original embedding, $\rmT$ is the victim's linear transformation matrix, $\ve_p$ is the embedding generated by the victim using $\rmT$, $\rmA$ is the attacker's additional linear transformation matrix, and $\ve_a$ is the overwritten embedding generated by the attacker using $\rmA$.

We now discuss how it may be possible for us (as the victim) to reverse-engineer the attacker's linear transformation matrix $\rmA$. 
By querying the attacker and retrieving a number of samples, we can learn the unknown $\rmA$ by the attacker as follows: 
\[\rmA = \rmE_a \times \rmE_p^{+},\]
where $\rmE$ is matrix stacking corresponding embeddings $\ve$.

Note that, this transformation matrix $\rmA$ could capture a range of transformations like dimension shift, rearrangement (shuffling), and many others.

As future work, it would be interesting to quantify the vulnerability of \ourdefence to these overwriting attacks and how well this proposed approach would recover the attacker transformation matrix to address this weakness.

\subsection*{Transformation Enhancements}
With the current design of the circulant transformation matrix (defined \refsecnum{sec:matrix-construction}), the matrix is compromised if an attacker manages to recover any single row in the matrix. A better approach could be to use different weights for each row in the circulant matrix, but this would mean losing crucial properties such as invertibility and full rank, as observed in \reftab{table:diff-matrix-properties}. Therefore, we opted to retain the current design, but there could be other designs that are stealthy, still invertible, and full rank. 

In this thesis, moreover, we only experimented with linear transformations. Non-linear transformations might also be worth exploring because due to their additional complexity. The watermark will be difficult to reverse-engineer (robust) and imperceptible (stealthy). However, these complex non-linear transformations of original embeddings might lead to significant utility degradation. Also, such complex watermark patterns might not be learned and transferred to the attacker model in the imitation attack. 
Future work can explore this to understand the effectiveness of non-linear transformation as a watermark for embeddings.

\subsection*{New Victim EaaS APIs}
During this research project, OpenAI released a new V3 version of their EaaS APIs \citep{recentopenaiEmbeddingModels}. A prominent enhancement in this version was using the Matryoshka representation learning technique \citep{kusupati2022matryoshka}, providing the ability to slice the initial dimensions of embeddings (essentially another form of dimension reduction). Our preliminary idea of hyperdimension (\refsecnum{sec:hyp-obfuscation}) might run into issues as we could lose some obfuscated dimensions if the attacker slices some initial dimensions. However, given that our final approach involves transforming all dimensions, we can handle attackers slicing initial dimensions. Slicing is a type of linear transformation and should be covered by the defences to overwriting attacks as discussed in \refsec{sec:overwriting}.
We leave support and extension to this new version of EaaS APIs for future work.

\addtocontents{toc}{\vspace{2em}} %

\appendix %

\addtocontents{toc}{\vspace{2em}}  %
\backmatter

\label{Bibliography}
\lhead{\emph{Bibliography}}  %
\bibliographystyle{acl_natbib}  %
\bibliography{Bibliography}  %

@article{shannon1948mathematical,
  title        = {A mathematical theory of communication},
  author       = {Shannon, Claude Elwood},
  year         = {1948},
  journal      = {The Bell system technical journal},
  publisher    = {Nokia Bell Labs},
  volume       = {27},
  number       = {3},
  pages        = {379--423},
}

@article{turing2009computing,
  title        = {Computing Machinery and Intelligence},
  author       = {Alan Turing},
  year         = {1950},
  journal      = {Mind},
  publisher    = {Oxford University Press},
  volume       = {59},
  number       = {October},
  pages        = {433--60},
}

@article{shannon1951prediction,
  title        = {Prediction and entropy of printed English},
  author       = {Shannon, Claude},
  year         = {1951},
  journal      = {Bell system technical journal},
  publisher    = {Wiley Online Library},
  volume       = {30},
  number       = {1},
  pages        = {50--64},
}

@book{Chomsky+1957,
  title        = {Syntactic Structures},
  author       = {Noam Chomsky},
  year         = {1957},
  publisher    = {De Gruyter Mouton},
  address      = {Berlin, Boston},
  isbn         = {9783112316009},
  lastchecked  = {2024-03-28},
}

@article{bahl1983maximum,
  title        = {A maximum likelihood approach to continuous speech recognition},
  author       = {Bahl, Lalit R and Jelinek, Frederick and Mercer, Robert},
  year         = {1983},
  journal      = {IEEE transactions on pattern analysis and machine intelligence},
  publisher    = {IEEE},
  volume       = {PAMI-5},
  number       = {2},
  pages        = {179--190},
}

@article{katz1987estimation,
  title        = {Estimation of probabilities from sparse data for the language model component of a speech recognizer},
  author       = {Katz, Slava},
  year         = {1987},
  journal      = {IEEE transactions on acoustics, speech, and signal processing},
  publisher    = {IEEE},
  volume       = {35},
  number       = {3},
  pages        = {400--401},
}

@inproceedings{cavnar1994n,
  title        = {N-gram-based text categorization},
  author       = {Cavnar, William and Trenkle, John},
  year         = {1994},
  booktitle    = {Proceedings of SDAIR-94, 3rd annual symposium on document analysis and information retrieval},
  volume       = {161175},
  pages        = {14},
  organization = {Ann Arbor, Michigan},
}

@article{gale1995good,
  title        = {Good-turing frequency estimation without tears},
  author       = {Gale, William  and Sampson, Geoffrey},
  year         = {1995},
  journal      = {Journal of quantitative linguistics},
  publisher    = {Taylor \& Francis},
  volume       = {2},
  number       = {3},
  pages        = {217--237},
}

@inproceedings{kneser1995improved,
  title        = {Improved backing-off for m-gram language modeling},
  author       = {Kneser, Reinhard and Ney, Hermann},
  year         = {1995},
  booktitle    = {1995 international conference on acoustics, speech, and signal processing},
  volume       = {1},
  pages        = {181--184},
  organization = {IEEE},
}

@article{bernstein1996middleware,
  title        = {Middleware: a model for distributed system services},
  author       = {Bernstein, Philip},
  year         = {1996},
  journal      = {Communications of the ACM},
  publisher    = {ACM New York, NY, USA},
  volume       = {39},
  number       = {2},
  pages        = {86--98},
}

@misc{cox1997digital,
  title        = {Digital watermarking},
  author       = {Cox, Ingemar and Miller, Matthew and Bloom, Jeffrey},
  year         = {1997},
  publisher    = {Springer},
}

@article{hochreiter1997long,
  title        = {Long short-term memory},
  author       = {Hochreiter, Sepp and Schmidhuber, J{\"u}rgen},
  year         = {1997},
  journal      = {Neural computation},
  publisher    = {MIT press},
  volume       = {9},
  number       = {8},
  pages        = {1735--1780},
}

@book{jelinek1998statistical,
  title        = {Statistical methods for speech recognition},
  author       = {Jelinek, Frederick},
  year         = {1998},
  publisher    = {MIT press},
}

@inproceedings{thede1999second,
  title        = {A second-order hidden Markov model for part-of-speech tagging},
  author       = {Thede, Scott and Harper, Mary},
  year         = {1999},
  booktitle    = {Proceedings of the 37th annual meeting of the Association for Computational Linguistics},
  pages        = {175--182},
}

@phdthesis{hall1999correlation,
  title        = {Correlation-based feature selection for machine learning},
  author       = {Hall, Mark},
  year         = {1999},
  school       = {The University of Waikato},
}

@book{manning1999foundations,
  title        = {Foundations of statistical natural language processing},
  author       = {Manning, Christopher},
  year         = {1999},
  publisher    = {The MIT Press},
}

@article{mahoney1999text,
  title        = {Text compression as a test for artificial intelligence},
  author       = {Mahoney, Matthew},
  year         = {1999},
  journal      = {AAAI/IAAI},
  volume       = {970},
}

@article{bengio2000neural,
  title        = {A neural probabilistic language model},
  author       = {Bengio, Yoshua and Ducharme, R{\'e}jean and Vincent, Pascal},
  year         = {2000},
  journal      = {Advances in neural information processing systems},
  volume       = {13},
}

@article{rosenfeld2000two,
  title        = {Two decades of statistical language modeling: Where do we go from here?},
  author       = {Rosenfeld, Ronald},
  year         = {2000},
  journal      = {Proceedings of the IEEE},
  publisher    = {IEEE},
  volume       = {88},
  number       = {8},
  pages        = {1270--1278},
}

@inproceedings{atallah2001natural,
  title        = {Natural language watermarking: Design, analysis, and a proof-of-concept implementation},
  author       = {Atallah, Mikhail and Raskin, Victor and Crogan, Michael and Hempelmann, Christian and Kerschbaum, Florian and Mohamed, Dina and Naik, Sanket},
  year         = {2001},
  booktitle    = {Information Hiding: 4th International Workshop, IH 2001 Pittsburgh, PA, USA, April 25--27, 2001 Proceedings 4},
  pages        = {185--200},
  organization = {Springer},
}

@inproceedings{atallah2002natural,
  title        = {Natural language watermarking and tamperproofing},
  author       = {Atallah, Mikhail and Raskin, Victor and Hempelmann, Christian and Karahan, Mercan and Sion, Radu and Topkara, Umut and Triezenberg, Katrina E},
  year         = {2002},
  booktitle    = {International workshop on information hiding},
  pages        = {196--212},
  organization = {Springer},
}

@article{hauser2002faculty,
  title        = {The faculty of language: what is it, who has it, and how did it evolve?},
  author       = {Hauser, Marc and Chomsky, Noam and Fitch, W Tecumseh},
  year         = {2002},
  journal      = {science},
  publisher    = {American Association for the Advancement of Science},
  volume       = {298},
  number       = {5598},
  pages        = {1569--1579},
}

@book{barni2004watermarking,
  title        = {Watermarking systems engineering: enabling digital assets security and other applications},
  author       = {Barni, Mauro and Bartolini, Franco},
  year         = {2004},
  publisher    = {Crc Press},
}

@inproceedings{somers-2005-round,
  title        = {Round-trip Translation: What Is It Good For?},
  author       = {Somers, Harold},
  year         = {2005},
  month        = dec,
  booktitle    = {Proceedings of the Australasian Language Technology Workshop 2005},
  address      = {Sydney, Australia},
  pages        = {127--133},
  editor       = {Baldwin, Timothy  and Curran, James  and van Zaanen, Menno},
}

@article{liu2005statistical,
  title        = {Statistical language modeling for information retrieval.},
  author       = {Liu, Xiaoyong and Croft, W Bruce},
  year         = {2005},
  journal      = {Annu. Rev. Inf. Sci. Technol.},
  volume       = {39},
  number       = {1},
  pages        = {1--31},
}

@inproceedings{dwork2006differential,
  title        = {Differential privacy},
  author       = {Dwork, Cynthia},
  year         = {2006},
  booktitle    = {International colloquium on automata, languages, and programming},
  pages        = {1--12},
  organization = {Springer},
}

@inproceedings{hadsell2006dimensionality,
  title        = {Dimensionality reduction by learning an invariant mapping},
  author       = {Hadsell, Raia and Chopra, Sumit and LeCun, Yann},
  year         = {2006},
  booktitle    = {2006 IEEE computer society conference on computer vision and pattern recognition (CVPR'06)},
  volume       = {2},
  pages        = {1735--1742},
  organization = {IEEE},
}

@inproceedings{enron,
  title        = {Spam filtering with naive bayes-which naive bayes?},
  author       = {Metsis, Vangelis and Androutsopoulos, Ion and Paliouras, Georgios},
  year         = {2006},
  booktitle    = {CEAS},
  volume       = {17},
  pages        = {28--69},
  organization = {Mountain View, CA},
}

@inproceedings{topkara2006hiding,
  title        = {The hiding virtues of ambiguity: quantifiably resilient watermarking of natural language text through synonym substitutions},
  author       = {Topkara, Umut and Topkara, Mercan and Atallah, Mikhail},
  year         = {2006},
  booktitle    = {Proceedings of the 8th workshop on Multimedia and security},
  pages        = {164--174},
}

@article{circulant-matrix,
  title        = {Toeplitz and circulant matrices: A review},
  author       = {Gray, Robert},
  year         = {2006},
  journal      = {Foundations and Trends{\textregistered} in Communications and Information Theory},
  publisher    = {Now Publishers, Inc.},
  volume       = {2},
  number       = {3},
  pages        = {155--239},
}

@inproceedings{topkara2006words,
  title        = {Words are not enough: sentence level natural language watermarking},
  author       = {Topkara, Mercan and Topkara, Umut and Atallah, Mikhail},
  year         = {2006},
  booktitle    = {Proceedings of the 4th ACM international workshop on Contents protection and security},
  pages        = {37--46},
}

@book{cox2007digital,
  title        = {Digital watermarking and steganography},
  author       = {Cox, Ingemar and Miller, Matthew and Bloom, Jeffreynd Fridrich, Jessica and Kalker, Ton},
  year         = {2007},
  publisher    = {Morgan kaufmann},
}

@inproceedings{brants2007large,
  title        = {Large language models in machine translation},
  author       = {Brants, Thorsten and Popat, Ashok and Xu, Peng and Och, Franz Josef and Dean, Jeffrey},
  year         = {2007},
  booktitle    = {Proceedings of the 2007 Joint Conference on Empirical Methods in Natural Language Processing and Computational Natural Language Learning (EMNLP-CoNLL)},
  pages        = {858--867},
}

@inproceedings{mikolov2010recurrent,
  title        = {Recurrent neural network based language model.},
  author       = {Mikolov, Tomas and Karafi{\'a}t, Martin and Burget, Lukas and Cernock{\`y}, Jan and Khudanpur, Sanjeev},
  year         = {2010},
  booktitle    = {Interspeech},
  volume       = {2},
  pages        = {1045--1048},
  organization = {Makuhari},
}

@article{collobert2011natural,
  title        = {Natural language processing (almost) from scratch},
  author       = {Collobert, Ronan and Weston, Jason and Bottou, L{\'e}on and Karlen, Michael and Kavukcuoglu, Koray and Kuksa, Pavel},
  year         = {2011},
  journal      = {Journal of machine learning research},
  volume       = {12},
  pages        = {2493--2537},
}

@article{sedgwick2012pearson,
  title        = {Pearson’s correlation coefficient},
  author       = {Sedgwick, Philip},
  year         = {2012},
  journal      = {Bmj},
  publisher    = {British Medical Journal Publishing Group},
  volume       = {345},
}

@article{bhagat-hovy-2013-squibs,
  title        = {{S}quibs: What Is a Paraphrase?},
  author       = {Bhagat, Rahul  and Hovy, Eduard},
  year         = {2013},
  month        = sep,
  journal      = {Computational Linguistics},
  publisher    = {MIT Press},
  address      = {Cambridge, MA},
  volume       = {39},
  number       = {3},
  pages        = {463--472},
}

@article{singh2013survey,
  title        = {A survey of digital watermarking techniques, applications and attacks},
  author       = {Singh, Prabhishek and Chadha, Ramneet Singh},
  year         = {2013},
  journal      = {International Journal of Engineering and Innovative Technology (IJEIT)},
  volume       = {2},
  number       = {9},
  pages        = {165--175},
}

@article{mikolov2013distributed,
  title        = {Distributed representations of words and phrases and their compositionality},
  author       = {Mikolov, Tomas and Sutskever, Ilya and Chen, Kai and Corrado, Greg and Dean, Jeff},
  year         = {2013},
  journal      = {Advances in neural information processing systems},
  volume       = {26},
}

@article{graves2013generating,
  title        = {Generating sequences with recurrent neural networks},
  author       = {Graves, Alex},
  year         = {2013},
  journal      = {arXiv preprint arXiv:1308.0850},
}

@inproceedings{trainingIssueRNN,
  title        = {On the difficulty of training recurrent neural networks},
  author       = {Pascanu, Razvan and Mikolov, Tomas and Bengio, Yoshua},
  year         = {2013},
  booktitle    = {Proceedings of the 30th International Conference on International Conference on Machine Learning - Volume 28},
  location     = {Atlanta, GA, USA},
  publisher    = {JMLR.org},
  series       = {ICML'13},
  pages        = {III–1310–III–1318},
}

@inproceedings{sst2,
  title        = {Recursive Deep Models for Semantic Compositionality Over a Sentiment Treebank},
  author       = {Socher, Richard  and Perelygin, Alex  and Wu, Jean  and Chuang, Jason  and Manning, Christopher  and Ng, Andrew  and Potts, Christopher},
  year         = {2013},
  month        = oct,
  booktitle    = {Proceedings of the 2013 Conference on Empirical Methods in Natural Language Processing},
  publisher    = {Association for Computational Linguistics},
  address      = {Seattle, Washington, USA},
  pages        = {1631--1642},
  editor       = {Yarowsky, David  and Baldwin, Timothy  and Korhonen, Anna  and Livescu, Karen  and Bethard, Steven},
}

@article{bengio2013representation,
  title        = {Representation learning: A review and new perspectives},
  author       = {Bengio, Yoshua and Courville, Aaron and Vincent, Pascal},
  year         = {2013},
  journal      = {IEEE transactions on pattern analysis and machine intelligence},
  publisher    = {IEEE},
  volume       = {35},
  number       = {8},
  pages        = {1798--1828},
}

@inproceedings{pennington2014glove,
  title        = {Glove: Global vectors for word representation},
  author       = {Pennington, Jeffreynd Socher, Richard and Manning, Christopher},
  year         = {2014},
  booktitle    = {Proceedings of the 2014 conference on empirical methods in natural language processing (EMNLP)},
  pages        = {1532--1543},
}

@article{tsai2014software,
  title        = {Software-as-a-service (SaaS): perspectives and challenges},
  author       = {Tsai, WeiTek and Bai, XiaoYing and Huang, Yu},
  year         = {2014},
  journal      = {Science China Information Sciences},
  publisher    = {Springer},
  volume       = {57},
  pages        = {1--15},
}

@article{ag_news,
  title        = {Character-level convolutional networks for text classification},
  author       = {Zhang, Xiang and Zhao, Junbo and LeCun, Yann},
  year         = {2015},
  journal      = {Advances in neural information processing systems},
  volume       = {28},
}

@article{lecun2015deep,
  title        = {Deep learning},
  author       = {LeCun, Yann and Bengio, Yoshua and Hinton, Geoffrey},
  year         = {2015},
  journal      = {nature},
  publisher    = {Nature Publishing Group UK London},
  volume       = {521},
  number       = {7553},
  pages        = {436--444},
}

@inproceedings{ribeiro2015mlaas,
  title        = {Mlaas: Machine learning as a service},
  author       = {Ribeiro, Mauro and Grolinger, Katarina and Capretz, Miriam AM},
  year         = {2015},
  booktitle    = {2015 IEEE 14th international conference on machine learning and applications (ICMLA)},
  pages        = {896--902},
  organization = {IEEE},
}

@book{strang2000linear,
  title        = {Introduction to Linear Algebra},
  author       = {Strang, Gilbert},
  year         = {2016},
  publisher    = {Wellesley},
  isbn         = {9780980232776},
}

@inproceedings{le-zuidema-2016-quantifying,
  title        = {Quantifying the Vanishing Gradient and Long Distance Dependency Problem in Recursive Neural Networks and Recursive {LSTM}s},
  author       = {Le, Phong  and Zuidema, Willem},
  year         = {2016},
  month        = aug,
  booktitle    = {Proceedings of the 1st Workshop on Representation Learning for {NLP}},
  publisher    = {Association for Computational Linguistics},
  address      = {Berlin, Germany},
  pages        = {87--93},
  editor       = {Blunsom, Phil  and Cho, Kyunghyun  and Cohen, Shay  and Grefenstette, Edward  and Hermann, Karl Moritz  and Rimell, Laura  and Weston, Jason  and Yih, Scott Wen-tau},
}

@inproceedings{tramer2016stealing,
  title        = {Stealing machine learning models via prediction {APIs}},
  author       = {Tram{\`e}r, Florian and Zhang, Fan and Juels, Ari and Reiter, Michael K and Ristenpart, Thomas},
  year         = {2016},
  booktitle    = {25th USENIX security symposium (USENIX Security 16)},
  pages        = {601--618},
}

@article{vaswani2017attention,
  title        = {Attention is all you need},
  author       = {Vaswani, Ashish and Shazeer, Noam and Parmar, Niki and Uszkoreit, Jakob and Jones, Llion and Gomez, Aidan N and Kaiser, {\L}ukasz and Polosukhin, Illia},
  year         = {2017},
  journal      = {Advances in neural information processing systems},
  volume       = {30},
}

@inproceedings{yao2017complexity,
  title        = {Complexity vs. performance: empirical analysis of machine learning as a service},
  author       = {Yao, Yuanshun and Xiao, Zhujun and Wang, Bolun and Viswanath, Bimal and Zheng, Haitao and Zhao, Ben Y},
  year         = {2017},
  booktitle    = {Proceedings of the 2017 Internet Measurement Conference},
  pages        = {384--397},
}

@inproceedings{uchida2017embedding,
  title        = {Embedding watermarks into deep neural networks},
  author       = {Uchida, Yusuke and Nagai, Yuki and Sakazawa, Shigeyuki and Satoh, Shin'ichi},
  year         = {2017},
  booktitle    = {Proceedings of the 2017 ACM on international conference on multimedia retrieval},
  pages        = {269--277},
}

@inproceedings{peters-etal-2018-deep,
  title        = {Deep Contextualized Word Representations},
  author       = {Peters, Matthew  and Neumann, Mark  and Iyyer, Mohit  and Gardner, Matt  and Clark, Christopher  and Lee, Kenton  and Zettlemoyer, Luke},
  year         = {2018},
  month        = jun,
  booktitle    = {Proceedings of the 2018 Conference of the North {A}merican Chapter of the Association for Computational Linguistics: Human Language Technologies, Volume 1 (Long Papers)},
  publisher    = {Association for Computational Linguistics},
  address      = {New Orleans, Louisiana},
  pages        = {2227--2237},
  editor       = {Walker, Marilyn  and Ji, Heng  and Stent, Amanda},
}

@article{radford2018improving,
  title        = {Improving Language Understanding by Generative Pre-Training},
  author       = {Alec Radford and Karthik Narasimhan},
  year         = {2018},
  journal      = {OpenAI blog},
}

@inproceedings{devlin-etal-2019-bert,
  title        = {{BERT}: Pre-training of Deep Bidirectional Transformers for Language Understanding},
  author       = {Devlin, Jacob  and Chang, Ming-Wei  and Lee, Kenton  and Toutanova, Kristina},
  year         = {2019},
  month        = jun,
  booktitle    = {Proceedings of the 2019 Conference of the North {A}merican Chapter of the Association for Computational Linguistics: Human Language Technologies, Volume 1 (Long and Short Papers)},
  publisher    = {Association for Computational Linguistics},
  address      = {Minneapolis, Minnesota},
  pages        = {4171--4186},
  editor       = {Burstein, Jill  and Doran, Christy  and Solorio, Thamar},
}

@article{yu2019review,
  title        = {A review of recurrent neural networks: LSTM cells and network architectures},
  author       = {Yu, Yong and Si, Xiaosheng and Hu, Changhua and Zhang, Jianxun},
  year         = {2019},
  journal      = {Neural computation},
  publisher    = {MIT Press One Rogers Street, Cambridge, MA 02142-1209, USA journals-info~…},
  volume       = {31},
  number       = {7},
  pages        = {1235--1270},
}

@inproceedings{wang2019attacks,
  title        = {Attacks on digital watermarks for deep neural networks},
  author       = {Wang, Tianhao and Kerschbaum, Florian},
  year         = {2019},
  booktitle    = {ICASSP 2019-2019 IEEE International Conference on Acoustics, Speech and Signal Processing (ICASSP)},
  pages        = {2622--2626},
  organization = {IEEE},
}

@inproceedings{zheng2019bdpl,
  title        = {BDPL: A boundary differentially private layer against machine learning model extraction attacks},
  author       = {Zheng, Huadi and Ye, Qingqing and Hu, Haibo and Fang, Chengfang and Shi, Jie},
  year         = {2019},
  booktitle    = {Computer Security--ESORICS 2019: 24th European Symposium on Research in Computer Security, Luxembourg, September 23--27, 2019, Proceedings, Part I 24},
  pages        = {66--83},
  organization = {Springer},
}

@inproceedings{darvish2019deepsigns,
  title        = {Deepsigns: An end-to-end watermarking framework for ownership protection of deep neural networks},
  author       = {Darvish Rouhani, Bita and Chen, Huili and Koushanfar, Farinaz},
  year         = {2019},
  booktitle    = {Proceedings of the twenty-fourth international conference on architectural support for programming languages and operating systems},
  pages        = {485--497},
}

@inproceedings{orekondy2019knockoff,
  title        = {Knockoff nets: Stealing functionality of black-box models},
  author       = {Orekondy, Tribhuvanesh and Schiele, Bernt and Fritz, Mario},
  year         = {2019},
  booktitle    = {Proceedings of the IEEE/CVF conference on computer vision and pattern recognition},
  pages        = {4954--4963},
}

@article{radford2019language,
  title        = {Language models are unsupervised multitask learners},
  author       = {Alec Radford and Jeff Wu and Rewon Child and David Luan and Dario Amodei and Ilya Sutskever},
  year         = {2019},
  journal      = {OpenAI blog},
  volume       = {1},
  number       = {8},
  pages        = {9},
}

@inproceedings{juuti2019prada,
  title        = {PRADA: protecting against DNN model stealing attacks},
  author       = {Juuti, Mika and Szyller, Sebastian and Marchal, Samuel and Asokan, N},
  year         = {2019},
  booktitle    = {2019 IEEE European Symposium on Security and Privacy (EuroS\&P)},
  pages        = {512--527},
  organization = {IEEE},
}

@article{liu2019roberta,
  title        = {Roberta: A robustly optimized bert pretraining approach},
  author       = {Liu, Yinhan and Ott, Myle and Goyal, Naman and Du, Jingfei and Joshi, Mandar and Chen, Danqi and Levy, Omer and Lewis, Mike and Zettlemoyer, Luke and Stoyanov, Veselin},
  year         = {2019},
  journal      = {arXiv preprint arXiv:1907.11692},
}

@inproceedings{lewis2019bart,
  title        = {{BART}: Denoising Sequence-to-Sequence Pre-training for Natural Language Generation, Translation, and Comprehension},
  author       = {Lewis, Mike  and Liu, Yinhan  and Goyal, Naman  and Ghazvininejad, Marjan  and Mohamed, Abdelrahman  and Levy, Omer  and Stoyanov, Veselin  and Zettlemoyer, Luke},
  year         = {2020},
  month        = jul,
  booktitle    = {Proceedings of the 58th Annual Meeting of the Association for Computational Linguistics},
  publisher    = {Association for Computational Linguistics},
  address      = {Online},
  pages        = {7871--7880},
  editor       = {Jurafsky, Dan  and Chai, Joyce  and Schluter, Natalie  and Tetreault, Joel},
}

@inproceedings{mind,
  title        = {{MIND}: A Large-scale Dataset for News Recommendation},
  author       = {Wu, Fangzhao  and Qiao, Ying  and Chen, Jiun-Hung  and Wu, Chuhan  and Qi, Tao  and Lian, Jianxun  and Liu, Danyang  and Xie, Xing  and Gao, Jianfeng  and Wu, Winnie  and Zhou, Ming},
  year         = {2020},
  month        = jul,
  booktitle    = {Proceedings of the 58th Annual Meeting of the Association for Computational Linguistics},
  publisher    = {Association for Computational Linguistics},
  address      = {Online},
  pages        = {3597--3606},
  editor       = {Jurafsky, Dan  and Chai, Joyce  and Schluter, Natalie  and Tetreault, Joel},
}

@inproceedings{cheng2020differentially,
  title        = {Differentially private machine learning model against model extraction attack},
  author       = {Cheng, Zelei and Li, Zuotian and Zhang, Jiwei and Zhang, Shuhan},
  year         = {2020},
  booktitle    = {2020 International Conferences on Internet of Things (iThings) and IEEE Green Computing and Communications (GreenCom) and IEEE Cyber, Physical and Social Computing (CPSCom) and IEEE Smart Data (SmartData) and IEEE Congress on Cybermatics (Cybermatics)},
  pages        = {722--728},
  organization = {IEEE},
}

@inproceedings{wallace-etal-2020-imitation,
  title        = {Imitation Attacks and Defenses for Black-box Machine Translation Systems},
  author       = {Wallace, Eric  and Stern, Mitchell  and Song, Dawn},
  year         = {2020},
  month        = nov,
  booktitle    = {Proceedings of the 2020 Conference on Empirical Methods in Natural Language Processing (EMNLP)},
  publisher    = {Association for Computational Linguistics},
  address      = {Online},
  pages        = {5531--5546},
  editor       = {Webber, Bonnie  and Cohn, Trevor  and He, Yulan  and Liu, Yang},
}

@article{raghavan2020impact,
  title        = {Impact of software as a service ({SaaS}) on software acquisition process},
  author       = {Raghavan, Srinivasa and KR, Jayasimha and Nargundkar, Rajendra V},
  year         = {2020},
  journal      = {Journal of Business \& Industrial Marketing},
  publisher    = {Emerald Publishing Limited},
  volume       = {35},
  number       = {4},
  pages        = {757--770},
}

@inproceedings{info-leakage-EaaS,
  title        = {Information Leakage in Embedding Models},
  author       = {Song, Congzheng and Raghunathan, Ananth},
  year         = {2020},
  booktitle    = {Proceedings of the 2020 ACM SIGSAC Conference on Computer and Communications Security},
  location     = {Virtual Event, USA},
  publisher    = {Association for Computing Machinery},
  address      = {New York, NY, USA},
  series       = {CCS '20},
  pages        = {377–390},
  isbn         = {9781450370899},
  numpages     = {14},
}

@article{brown2020language,
  title        = {Language models are few-shot learners},
  author       = {Tom B. Brown and Benjamin Mann and Nick Ryder and Melanie Subbiah and Jared Kaplan and Prafulla Dhariwal and Arvind Neelakantan and Pranav Shyam and Girish Sastry and Amanda Askell and Sandhini Agarwal and Ariel Herbert-Voss and Gretchen Krueger and Tom Henighan and Rewon Child and Aditya Ramesh and Daniel M. Ziegler and Jeffrey Wu and Clemens Winter and Christopher Hesse and Mark Chen and Eric Sigler and Mateusz Litwin and Scott Gray and Benjamin Chess and Jack Clark and Christopher Berner and Sam McCandlish and Alec Radford and Ilya Sutskever and Dario Amodei},
  year         = {2020},
  journal      = {Advances in neural information processing systems},
  volume       = {33},
  pages        = {1877--1901},
}

@inproceedings{li2020protecting,
  title        = {Protecting the intellectual property of deep neural networks with watermarking: The frequency domain approach},
  author       = {Li, Meng and Zhong, Qi and Zhang, Leo Yu and Du, Yajuan and Zhang, Jun and Xiang, Yong},
  year         = {2020},
  booktitle    = {2020 IEEE 19th International Conference on Trust, Security and Privacy in Computing and Communications (TrustCom)},
  pages        = {402--409},
  organization = {IEEE},
}

@article{kaplan2020scaling,
  title        = {Scaling laws for neural language models},
  author       = {Kaplan, Jared and McCandlish, Sam and Henighan, Tom and Brown, Tom B and Chess, Benjamin and Child, Rewon and Gray, Scott and Radford, Alec and Wu, Jeffreynd Amodei, Dario},
  year         = {2020},
  journal      = {arXiv preprint arXiv:2001.08361},
}

@inproceedings{krishna2019thieves,
  title        = {Thieves on Sesame Street! Model Extraction of BERT-based APIs},
  author       = {Kalpesh Krishna and Gaurav Singh Tomar and Ankur P. Parikh and Nicolas Papernot and Mohit Iyyer},
  year         = {2020},
  booktitle    = {International Conference on Learning Representations},
}

@inproceedings{wolf-etal-2020-transformers,
  title        = {Transformers: State-of-the-Art Natural Language Processing},
  author       = {Wolf, Thomas  and Debut, Lysandre  and Sanh, Victor  and Chaumond, Julien  and Delangue, Clement  and Moi, Anthony  and Cistac, Pierric  and Rault, Tim  and Louf, Remi  and Funtowicz, Morgan  and Davison, Joe  and Shleifer, Sam  and von Platen, Patrick  and Ma, Clara  and Jernite, Yacine  and Plu, Julien  and Xu, Canwen  and Le Scao, Teven  and Gugger, Sylvain  and Drame, Mariama  and Lhoest, Quentin  and Rush, Alexander},
  year         = {2020},
  month        = oct,
  booktitle    = {Proceedings of the 2020 Conference on Empirical Methods in Natural Language Processing: System Demonstrations},
  publisher    = {Association for Computational Linguistics},
  address      = {Online},
  pages        = {38--45},
  editor       = {Liu, Qun  and Schlangen, David},
}

@inproceedings{guo2020fine,
  title        = {Fine-tuning Is Not Enough: A Simple yet Effective Watermark Removal Attack for DNN Models},
  author       = {Guo, Shangwei and Zhang, Tianwei and Qiu, Han and Zeng, Yi and Xiang, Tao and Liu, Yang},
  year         = {2021},
  month        = {8},
  booktitle    = {Proceedings of the Thirtieth International Joint Conference on Artificial Intelligence, {IJCAI-21}},
  publisher    = {International Joint Conferences on Artificial Intelligence Organization},
  pages        = {3635--3641},
  note         = {Main Track},
  editor       = {Zhi-Hua Zhou},
}

@article{boenisch2021systematic,
  title        = {A systematic review on model watermarking for neural networks},
  author       = {Boenisch, Franziska},
  year         = {2021},
  journal      = {Frontiers in big Data},
  publisher    = {Frontiers Media SA},
  volume       = {4},
  pages        = {729663},
}

@inproceedings{lhoest-etal-2021-datasets,
  title        = {Datasets: A Community Library for Natural Language Processing},
  author       = {Lhoest, Quentin  and Villanova del Moral, Albert  and Jernite, Yacine  and Thakur, Abhishek  and von Platen, Patrick  and Patil, Suraj  and Chaumond, Julien  and Drame, Mariama  and Plu, Julien  and Tunstall, Lewis  and Davison, Joe  and {\v{S}}a{\v{s}}ko, Mario  and Chhablani, Gunjan  and Malik, Bhavitvya  and Brandeis, Simon  and Le Scao, Teven  and Sanh, Victor  and Xu, Canwen  and Patry, Nicolas  and McMillan-Major, Angelina  and Schmid, Philipp  and Gugger, Sylvain  and Delangue, Cl{\'e}ment  and Matussi{\`e}re, Th{\'e}o  and Debut, Lysandre  and Bekman, Stas  and Cistac, Pierric  and Goehringer, Thibault  and Mustar, Victor  and Lagunas, Fran{\c{c}}ois  and Rush, Alexander  and Wolf, Thomas},
  year         = {2021},
  month        = nov,
  booktitle    = {Proceedings of the 2021 Conference on Empirical Methods in Natural Language Processing: System Demonstrations},
  publisher    = {Association for Computational Linguistics},
  address      = {Online and Punta Cana, Dominican Republic},
  pages        = {175--184},
}

@inproceedings{carlini2021extracting,
  title        = {Extracting training data from large language models},
  author       = {Nicholas Carlini and Florian Tram{\`e}r and Eric Wallace and Matthew Jagielski and Ariel Herbert-Voss and Katherine Lee and Adam Roberts and Tom Brown and Dawn Song and {\'U}lfar Erlingsson and Alina Oprea and Colin Raffel},
  year         = {2021},
  booktitle    = {30th USENIX Security Symposium (USENIX Security 21)},
  pages        = {2633--2650},
}

@inproceedings{he-etal-2021-model,
  title        = {Model Extraction and Adversarial Transferability, Your {BERT} is Vulnerable!},
  author       = {He, Xuanli  and Lyu, Lingjuan  and Sun, Lichao  and Xu, Qiongkai},
  year         = {2021},
  month        = jun,
  booktitle    = {Proceedings of the 2021 Conference of the North American Chapter of the Association for Computational Linguistics: Human Language Technologies},
  publisher    = {Association for Computational Linguistics},
  address      = {Online},
  pages        = {2006--2012},
  editor       = {Toutanova, Kristina  and Rumshisky, Anna  and Zettlemoyer, Luke  and Hakkani-Tur, Dilek  and Beltagy, Iz  and Bethard, Steven  and Cotterell, Ryan  and Chakraborty, Tanmoy  and Zhou, Yichao},
}

@article{sanh2021multitask,
  title        = {Multitask prompted training enables zero-shot task generalization},
  author       = {Victor Sanh and Albert Webson and Colin Raffel and Stephen H. Bach and Lintang Sutawika and Zaid Alyafeai and Antoine Chaffin and Arnaud Stiegler and Teven Le Scao and Arun Raja and Manan Dey and M Saiful Bari and Canwen Xu and Urmish Thakker and Shanya Sharma Sharma and Eliza Szczechla and Taewoon Kim and Gunjan Chhablani and Nihal Nayak and Debajyoti Datta and Jonathan Chang and Mike Tian-Jian Jiang and Han Wang and Matteo Manica and Sheng Shen and Zheng Xin Yong and Harshit Pandey and Rachel Bawden and Thomas Wang and Trishala Neeraj and Jos Rozen and Abheesht Sharma and Andrea Santilli and Thibault Fevry and Jason Alan Fries and Ryan Teehan and Tali Bers and Stella Biderman and Leo Gao and Thomas Wolf and Alexander M. Rush},
  year         = {2021},
  journal      = {arXiv preprint arXiv:2110.08207},
}

@article{bommasani2021opportunities,
  title        = {On the opportunities and risks of foundation models},
  author       = {Rishi Bommasani and Drew A. Hudson and Ehsan Adeli and Russ Altman and Simran Arora and Sydney von Arx and Michael S. Bernstein and Jeannette Bohg and Antoine Bosselut and Emma Brunskill and Erik Brynjolfsson and Shyamal Buch and Dallas Card and Rodrigo Castellon and Niladri Chatterji and Annie Chen and Kathleen Creel and Jared Quincy Davis and Dora Demszky and Chris Donahue and Moussa Doumbouya and Esin Durmus and Stefano Ermon and John Etchemendy and Kawin Ethayarajh and Li Fei-Fei and Chelsea Finn and Trevor Gale and Lauren Gillespie and Karan Goel and Noah Goodman and Shelby Grossman and Neel Guha and Tatsunori Hashimoto and Peter Henderson and John Hewitt and Daniel E. Ho and Jenny Hong and Kyle Hsu and Jing Huang and Thomas Icard and Saahil Jain and Dan Jurafsky and Pratyusha Kalluri and Siddharth Karamcheti and Geoff Keeling and Fereshte Khani and Omar Khattab and Pang Wei Koh and Mark Krass and Ranjay Krishna and Rohith Kuditipudi and Ananya Kumar and Faisal Ladhak and Mina Lee and Tony Lee and Jure Leskovec and Isabelle Levent and Xiang Lisa Li and Xuechen Li and Tengyu Ma and Ali Malik and Christopher D. Manning and Suvir Mirchandani and Eric Mitchell and Zanele Munyikwa and Suraj Nair and Avanika Narayan and Deepak Narayanan and Ben Newman and Allen Nie and Juan Carlos Niebles and Hamed Nilforoshan and Julian Nyarko and Giray Ogut and Laurel Orr and Isabel Papadimitriou and Joon Sung Park and Chris Piech and Eva Portelance and Christopher Potts and Aditi Raghunathan and Rob Reich and Hongyu Ren and Frieda Rong and Yusuf Roohani and Camilo Ruiz and Jack Ryan and Christopher Ré and Dorsa Sadigh and Shiori Sagawa and Keshav Santhanam and Andy Shih and Krishnan Srinivasan and Alex Tamkin and Rohan Taori and Armin W. Thomas and Florian Tramèr and Rose E. Wang and William Wang and Bohan Wu and Jiajun Wu and Yuhuai Wu and Sang Michael Xie and Michihiro Yasunaga and Jiaxuan You and Matei Zaharia and Michael Zhang and Tianyi Zhang and Xikun Zhang and Yuhui Zhang and Lucia Zheng and Kaitlyn Zhou and Percy Liang},
  year         = {2021},
  journal      = {arXiv preprint arXiv:2108.07258},
}

@inproceedings{chen2021refit,
  title        = {Refit: a unified watermark removal framework for deep learning systems with limited data},
  author       = {Chen, Xinyun and Wang, Wenxiao and Bender, Chris and Ding, Yiming and Jia, Ruoxi and Li, Bo and Song, Dawn},
  year         = {2021},
  booktitle    = {Proceedings of the 2021 ACM Asia Conference on Computer and Communications Security},
  pages        = {321--335},
}

@inproceedings{wang2021riga,
  title        = {Riga: Covert and robust white-box watermarking of deep neural networks},
  author       = {Wang, Tianhao and Kerschbaum, Florian},
  year         = {2021},
  booktitle    = {Proceedings of the Web Conference 2021},
  pages        = {993--1004},
}

@inproceedings{zhang2021seat,
  title        = {SEAT: Similarity encoder by adversarial training for detecting model extraction attack queries},
  author       = {Zhang, Zhanyuan and Chen, Yizheng and Wagner, David},
  year         = {2021},
  booktitle    = {Proceedings of the 14th ACM Workshop on artificial intelligence and security},
  pages        = {37--48},
}

@inproceedings{ma2021undistillable,
  title        = {Undistillable: Making A Nasty Teacher That CANNOT teach students},
  author       = {Ma, Haoyu and Chen, Tianlong},
  year         = {2021},
  booktitle    = {International Conference on Learning Representations ICLR 2021},
}

@inproceedings{liu2022poisonedencoder,
  title        = {{PoisonedEncoder}: Poisoning the Unlabeled Pre-training Data in Contrastive Learning},
  author       = {Liu, Hongbin and Jia, Jinyuan and Gong, Neil Zhenqiang},
  year         = {2022},
  booktitle    = {31st USENIX Security Symposium (USENIX Security 22)},
  pages        = {3629--3645},
}

@inproceedings{jia2022badencoder,
  title        = {Badencoder: Backdoor attacks to pre-trained encoders in self-supervised learning},
  author       = {Jia, Jinyuan and Liu, Yupei and Gong, Neil Zhenqiang},
  year         = {2022},
  booktitle    = {2022 IEEE Symposium on Security and Privacy (SP)},
  pages        = {2043--2059},
  organization = {IEEE},
}

@article{he2022cater,
  title        = {Cater: Intellectual property protection on text generation apis via conditional watermarks},
  author       = {He, Xuanli and Xu, Qiongkai and Zeng, Yi and Lyu, Lingjuan and Wu, Fangzhao and Li, Jiwei and Jia, Ruoxi},
  year         = {2022},
  journal      = {Advances in Neural Information Processing Systems},
  volume       = {35},
  pages        = {5431--5445},
}

@article{wei2022chain,
  title        = {Chain-of-thought prompting elicits reasoning in large language models},
  author       = {Wei, Jason and Wang, Xuezhi and Schuurmans, Dale and Bosma, Maarten and ichter, brian and Xia, Fei and Chi, Ed and Le, Quoc V and Zhou, Denny},
  year         = {2022},
  journal      = {Advances in neural information processing systems},
  volume       = {35},
  pages        = {24824--24837},
}

@article{wei2022emergent,
  title        = {Emergent Abilities of Large Language Models},
  author       = {Jason Wei and Yi Tay and Rishi Bommasani and Colin Raffel and Barret Zoph and Sebastian Borgeaud and Dani Yogatama and Maarten Bosma and Denny Zhou and Donald Metzler and Ed H. Chi and Tatsunori Hashimoto and Oriol Vinyals and Percy Liang and Jeff Dean and William Fedus},
  year         = {2022},
  journal      = {Transactions on Machine Learning Research},
}

@inproceedings{he-etal-2022-extracted,
  title        = {Extracted {BERT} Model Leaks More Information than You Think!},
  author       = {He, Xuanli  and Lyu, Lingjuan  and Chen, Chen  and Xu, Qiongkai},
  year         = {2022},
  month        = dec,
  booktitle    = {Proceedings of the 2022 Conference on Empirical Methods in Natural Language Processing},
  publisher    = {Association for Computational Linguistics},
  address      = {Abu Dhabi, United Arab Emirates},
  pages        = {1530--1537},
  editor       = {Goldberg, Yoav  and Kozareva, Zornitsa  and Zhang, Yue},
}

@misc{chatgpt,
  title        = {Introducing {ChatGPT}},
  author       = {OpenAI},
    journal = {OpenAI blog},
  year         = {2022},
  urldate      = {2024-09-18},
  language     = {en-US},
}

@article{kusupati2022matryoshka,
  title        = {Matryoshka representation learning},
  author       = {Kusupati, Aditya and Bhatt, Gantavya and Rege, Aniket and Wallingford, Matthew and Sinha, Aditya and Ramanujan, Vivek and Howard-Snyder, William and Chen, Kaifeng and Kakade, Sham and Jain, Prateek and others},
  year         = {2022},
  journal      = {Advances in Neural Information Processing Systems},
  volume       = {35},
  pages        = {30233--30249},
}

@misc{2022-EaaS-OpenAI,
  title        = {New and improved embedding model},
  author       = {OpenAI},
journal = {OpenAI blog},
  year         = {2022},
  urldate      = {2024-03-31},
  language     = {en-US},
}

@article{NLLB,
  title        = {No language left behind: Scaling human-centered machine translation},
  author       = {NLLB Team and Marta R. Costa-jussà and James Cross and Onur Çelebi and Maha Elbayad and Kenneth Heafield and Kevin Heffernan and Elahe Kalbassi and Janice Lam and Daniel Licht and Jean Maillard and Anna Sun and Skyler Wang and Guillaume Wenzek and Al Youngblood and Bapi Akula and Loic Barrault and Gabriel Mejia Gonzalez and Prangthip Hansanti and John Hoffman and Semarley Jarrett and Kaushik Ram Sadagopan and Dirk Rowe and Shannon Spruit and Chau Tran and Pierre Andrews and Necip Fazil Ayan and Shruti Bhosale and Sergey Edunov and Angela Fan and Cynthia Gao and Vedanuj Goswami and Francisco Guzmán and Philipp Koehn and Alexandre Mourachko and Christophe Ropers and Safiyyah Saleem and Holger Schwenk and Jeff Wang},
  year         = {2022},
  journal      = {arXiv preprint arXiv:2207.04672},
}

@article{lim2022protect,
  title        = {Protect, show, attend and tell: Empowering image captioning models with ownership protection},
  author       = {Lim, Jian Han and Chan, Chee Seng and Ng, Kam Woh and Fan, Lixin and Yang, Qiang},
  year         = {2022},
  journal      = {Pattern Recognition},
  publisher    = {Elsevier},
  volume       = {122},
  pages        = {108285},
}

@inproceedings{he2022protecting,
  title        = {Protecting intellectual property of language generation apis with lexical watermark},
  author       = {He, Xuanli and Xu, Qiongkai and Lyu, Lingjuan and Wu, Fangzhao and Wang, Chenguang},
  year         = {2022},
  booktitle    = {Proceedings of the AAAI Conference on Artificial Intelligence},
  volume       = {36},
  pages        = {10758--10766},
}

@inproceedings{stolen-encoder,
  title        = {StolenEncoder: Stealing Pre-trained Encoders in Self-supervised Learning},
  author       = {Liu, Yupei and Jia, Jinyuan and Liu, Hongbin and Gong, Neil Zhenqiang},
  year         = {2022},
  booktitle    = {Proceedings of the 2022 ACM SIGSAC Conference on Computer and Communications Security},
  location     = {Los Angeles, CA, USA},
  publisher    = {Association for Computing Machinery},
  address      = {New York, NY, USA},
  series       = {CCS '22},
  pages        = {2115–2128},
  isbn         = {9781450394505},
  numpages     = {14},
}

@inproceedings{xu-etal-2022-student,
  title        = {Student Surpasses Teacher: Imitation Attack for Black-Box {NLP} {API}s},
  author       = {Xu, Qiongkai  and He, Xuanli  and Lyu, Lingjuan  and Qu, Lizhen  and Haffari, Gholamreza},
  year         = {2022},
  month        = oct,
  booktitle    = {Proceedings of the 29th International Conference on Computational Linguistics},
  publisher    = {International Committee on Computational Linguistics},
  address      = {Gyeongju, Republic of Korea},
  pages        = {2849--2860},
  editor       = {Calzolari, Nicoletta  and Huang, Chu-Ren  and Kim, Hansaem  and Pustejovsky, James  and Wanner, Leo  and Choi, Key-Sun  and Ryu, Pum-Mo  and Chen, Hsin-Hsi  and Donatelli, Lucia  and Ji, Heng  and Kurohashi, Sadao  and Paggio, Patrizia  and Xue, Nianwen  and Kim, Seokhwan  and Hahm, Younggyun  and He, Zhong  and Lee, Tony Kyungil  and Santus, Enrico  and Bond, Francis  and Na, Seung-Hoon},
}

@article{openAI-EaaS-training,
  title        = {Text and code embeddings by contrastive pre-training},
  author       = {Arvind Neelakantan and Tao Xu and Raul Puri and Alec Radford and Jesse Michael Han and Jerry Tworek and Qiming Yuan and Nikolas Tezak and Jong Wook Kim and Chris Hallacy and Johannes Heidecke and Pranav Shyam and Boris Power and Tyna Eloundou Nekoul and Girish Sastry and Gretchen Krueger and David Schnurr and Felipe Petroski Such and Kenny Hsu and Madeleine Thompson and Tabarak Khan and Toki Sherbakov and Joanne Jang and Peter Welinder and Lilian Weng},
  year         = {2022},
  journal      = {arXiv preprint arXiv:2201.10005},
}

@inproceedings{wang2022language,
  title        = {What language model architecture and pretraining objective works best for zero-shot generalization?},
  author       = {Wang, Thomas and Roberts, Adam and Hesslow, Daniel and Le Scao, Teven and Chung, Hyung Won and Beltagy, Iz and Launay, Julien and Raffel, Colin},
  year         = {2022},
  booktitle    = {International Conference on Machine Learning},
  pages        = {22964--22984},
  organization = {PMLR},
}

@inproceedings{muennighoff-etal-2023-mteb,
  title        = {{MTEB}: Massive Text Embedding Benchmark},
  author       = {Muennighoff, Niklas  and Tazi, Nouamane  and Magne, Loic  and Reimers, Nils},
  year         = {2023},
  month        = may,
  booktitle    = {Proceedings of the 17th Conference of the European Chapter of the Association for Computational Linguistics},
  publisher    = {Association for Computational Linguistics},
  address      = {Dubrovnik, Croatia},
  pages        = {2014--2037},
  editor       = {Vlachos, Andreas  and Augenstein, Isabelle},
}

@article{zhao2023survey-LLM,
  title        = {A survey of large language models},
  author       = {Wayne Xin Zhao and Kun Zhou and Junyi Li and Tianyi Tang and Xiaolei Wang and Yupeng Hou and Yingqian Min and Beichen Zhang and Junjie Zhang and Zican Dong and Yifan Du and Chen Yang and Yushuo Chen and Zhipeng Chen and Jinhao Jiang and Ruiyang Ren and Yifan Li and Xinyu Tang and Zikang Liu and Peiyu Liu and Jian-Yun Nie and Ji-Rong Wen},
  year         = {2024},
  journal      = {arXiv preprint arXiv:2303.18223},
}

@article{ghosal2023survey,
  title        = {A Survey on the Possibilities \& Impossibilities of AI-generated Text Detection},
  author       = {Ghosal, Soumya Suvra and Chakraborty, Souradip and Geiping, Jonas and Huang, Furong and Manocha, Dinesh and Bedi, Amrit},
  year         = {2023},
  journal      = {Transactions on Machine Learning Research},
}

@inproceedings{kirchenbauer2023watermark,
  title        = {A watermark for large language models},
  author       = {Kirchenbauer, John and Geiping, Jonas and Wen, Yuxin and Katz, Jonathan and Miers, Ian and Goldstein, Tom},
  year         = {2023},
  booktitle    = {International Conference on Machine Learning},
  pages        = {17061--17084},
  organization = {PMLR},
}

@article{taori2023alpaca,
  title        = {Alpaca: A strong, replicable instruction-following model},
  author       = {Taori, Rohan and Gulrajani, Ishaan and Zhang, Tianyi and Dubois, Yann and Li, Xuechen and Guestrin, Carlos and Liang, Percy and Hashimoto, Tatsunori B},
  year         = {2023},
  journal      = {Stanford Center for Research on Foundation Models. https://crfm.stanford.edu/2023/03/13/alpaca.html},
  volume       = {3},
  number       = {6},
  pages        = {7},
}

@inproceedings{peng2023you,
  title        = {Are You Copying My Model? Protecting the Copyright of Large Language Models for EaaS via Backdoor Watermark},
  author       = {Peng, Wenjun and Yi, Jingwei and Wu, Fangzhao and Wu, Shangxi and Zhu, Bin Bin and Lyu, Lingjuan and Jiao, Binxing and Xu, Tong and Sun, Guangzhong and Xie, Xing},
  year         = {2023},
  booktitle    = {Proceedings of the 61st Annual Meeting of the Association for Computational Linguistics (Volume 1: Long Papers)},
  pages        = {7653--7668},
}

@article{sadasivan2023can,
  title        = {Can AI-generated text be reliably detected?},
  author       = {Sadasivan, Vinu Sankar and Kumar, Aounon and Balasubramanian, Sriram and Wang, Wenxiao and Feizi, Soheil},
  year         = {2023},
  journal      = {arXiv preprint arXiv:2303.11156},
}

@inproceedings{mitchell2023detectgpt,
  title        = {Detectgpt: Zero-shot machine-generated text detection using probability curvature},
  author       = {Mitchell, Eric and Lee, Yoonho and Khazatsky, Alexander and Manning, Christopher and Finn, Chelsea},
  year         = {2023},
  booktitle    = {International Conference on Machine Learning},
  pages        = {24950--24962},
  organization = {PMLR},
}

@article{house_executive_2023,
  title        = {Executive {Order} {14110}: {Safe}, {Secure}, and {Trustworthy} {Development} and {Use} of {Artificial} {Intelligence}},
  author       = {Biden, Joseph},
  year         = {2023},
  month        = oct,
  journal      = {{Federal Register 88:75191–75226}},
}

@article{team2023gemini,
  title        = {Gemini: a family of highly capable multimodal models},
  author       = {{Gemini Team} and Rohan Anil and Sebastian Borgeaud and Jean-Baptiste Alayrac and Jiahui Yu and Radu Soricut and Johan Schalkwyk and Andrew M. Dai and Anja Hauth and Katie Millican and David Silver and Melvin Johnson and Ioannis Antonoglou and Julian Schrittwieser and Amelia Glaese and Jilin Chen and Emily Pitler and Timothy Lillicrap and Angeliki Lazaridou and Orhan Firat and James Molloy and Michael Isard and Paul R. Barham and Tom Hennigan and Benjamin Lee and Fabio Viola and Malcolm Reynolds and Yuanzhong Xu and Ryan Doherty and Eli Collins and Clemens Meyer and Eliza Rutherford and Erica Moreira and Kareem Ayoub and Megha Goel and Jack Krawczyk and Cosmo Du and Ed Chi and Heng-Tze Cheng and Eric Ni and Purvi Shah and Patrick Kane and Betty Chan and Manaal Faruqui and Aliaksei Severyn and Hanzhao Lin and YaGuang Li and Yong Cheng and Abe Ittycheriah and Mahdis Mahdieh and Mia Chen and Pei Sun and Dustin Tran and Sumit Bagri and Balaji Lakshminarayanan and Jeremiah Liu and Andras Orban and Fabian Güra and Hao Zhou and Xinying Song and Aurelien Boffy and Harish Ganapathy and Steven Zheng and HyunJeong Choe and Ágoston Weisz and Tao Zhu and Yifeng Lu and Siddharth Gopal and Jarrod Kahn and Maciej Kula and Jeff Pitman and Rushin Shah and Emanuel Taropa and Majd Al Merey and Martin Baeuml and Zhifeng Chen and Laurent El Shafey and Yujing Zhang and Olcan Sercinoglu and George Tucker and Enrique Piqueras and Maxim Krikun and Iain Barr and Nikolay Savinov and Ivo Danihelka and Becca Roelofs and Anaïs White and Anders Andreassen and Tamara von Glehn and Lakshman Yagati and Mehran Kazemi and Lucas Gonzalez and Misha Khalman and Jakub Sygnowski and Alexandre Frechette and Charlotte Smith and Laura Culp and Lev Proleev and Yi Luan and Xi Chen and James Lottes and Nathan Schucher and Federico Lebron and Alban Rrustemi and Natalie Clay and Phil Crone and Tomas Kocisky and Jeffrey Zhao and Bartek Perz and Dian Yu and Heidi Howard and Adam Bloniarz and Jack W. Rae and Han Lu and Laurent Sifre and Marcello Maggioni and Fred Alcober and Dan Garrette and Megan Barnes and Shantanu Thakoor and Jacob Austin and Gabriel Barth-Maron and William Wong and Rishabh Joshi and Rahma Chaabouni and Deeni Fatiha and Arun Ahuja and Gaurav Singh Tomar and Evan Senter and Martin Chadwick and Ilya Kornakov and Nithya Attaluri and Iñaki Iturrate and Ruibo Liu and Yunxuan Li and Sarah Cogan and Jeremy Chen and Chao Jia and Chenjie Gu and Qiao Zhang and Jordan Grimstad and Ale Jakse Hartman and Xavier Garcia and Thanumalayan Sankaranarayana Pillai and Jacob Devlin and Michael Laskin and Diego de Las Casas and Dasha Valter and Connie Tao and Lorenzo Blanco and Adrià Puigdomènech Badia and David Reitter and Mianna Chen and Jenny Brennan and Clara Rivera and Sergey Brin and Shariq Iqbal and Gabriela Surita and Jane Labanowski and Abhi Rao and Stephanie Winkler and Emilio Parisotto and Yiming Gu and Kate Olszewska and Ravi Addanki and Antoine Miech and Annie Louis and Denis Teplyashin and Geoff Brown and Elliot Catt and Jan Balaguer and Jackie Xiang and Pidong Wang and Zoe Ashwood and Anton Briukhov and Albert Webson and Sanjay Ganapathy and Smit Sanghavi and Ajay Kannan and Ming-Wei Chang and Axel Stjerngren and Josip Djolonga and Yuting Sun and Ankur Bapna and Matthew Aitchison and Pedram Pejman and Henryk Michalewski and Tianhe Yu and Cindy Wang and Juliette Love and Junwhan Ahn and Dawn Bloxwich and Kehang Han and Peter Humphreys and Thibault Sellam and James Bradbury and Varun Godbole and Sina Samangooei and Bogdan Damoc and Alex Kaskasoli and Sébastien M. R. Arnold and Vijay Vasudevan and Shubham Agrawal and Jason Riesa and Dmitry Lepikhin and Richard Tanburn and Srivatsan Srinivasan and Hyeontaek Lim and Sarah Hodkinson and Pranav Shyam and Johan Ferret and Steven Hand and Ankush Garg and Tom Le Paine and Jian Li and Yujia Li and Minh Giang and Alexander Neitz and Zaheer Abbas and Sarah York and Machel Reid and Elizabeth Cole and Aakanksha Chowdhery and Dipanjan Das and Dominika Rogozińska and Vitaliy Nikolaev and Pablo Sprechmann and Zachary Nado and Lukas Zilka and Flavien Prost and Luheng He and Marianne Monteiro and Gaurav Mishra and Chris Welty and Josh Newlan and Dawei Jia and Miltiadis Allamanis and Clara Huiyi Hu and Raoul de Liedekerke and Justin Gilmer and Carl Saroufim and Shruti Rijhwani and Shaobo Hou and Disha Shrivastava and Anirudh Baddepudi and Alex Goldin and Adnan Ozturel and Albin Cassirer and Yunhan Xu and Daniel Sohn and Devendra Sachan and Reinald Kim Amplayo and Craig Swanson and Dessie Petrova and Shashi Narayan and Arthur Guez and Siddhartha Brahma and Jessica Landon and Miteyan Patel and Ruizhe Zhao and Kevin Villela and Luyu Wang and Wenhao Jia and Matthew Rahtz and Mai Giménez and Legg Yeung and James Keeling and Petko Georgiev and Diana Mincu and Boxi Wu and Salem Haykal and Rachel Saputro and Kiran Vodrahalli and James Qin and Zeynep Cankara and Abhanshu Sharma and Nick Fernando and Will Hawkins and Behnam Neyshabur and Solomon Kim and Adrian Hutter and Priyanka Agrawal and Alex Castro-Ros and George van den Driessche and Tao Wang and Fan Yang and Shuo-yiin Chang and Paul Komarek and Ross McIlroy and Mario Lučić and Guodong Zhang and Wael Farhan and Michael Sharman and Paul Natsev and Paul Michel and Yamini Bansal and Siyuan Qiao and Kris Cao and Siamak Shakeri and Christina Butterfield and Justin Chung and Paul Kishan Rubenstein and Shivani Agrawal and Arthur Mensch and Kedar Soparkar and Karel Lenc and Timothy Chung and Aedan Pope and Loren Maggiore and Jackie Kay and Priya Jhakra and Shibo Wang and Joshua Maynez and Mary Phuong and Taylor Tobin and Andrea Tacchetti and Maja Trebacz and Kevin Robinson and Yash Katariya and Sebastian Riedel and Paige Bailey and Kefan Xiao and Nimesh Ghelani and Lora Aroyo and Ambrose Slone and Neil Houlsby and Xuehan Xiong and Zhen Yang and Elena Gribovskaya and Jonas Adler and Mateo Wirth and Lisa Lee and Music Li and Thais Kagohara and Jay Pavagadhi and Sophie Bridgers and Anna Bortsova and Sanjay Ghemawat and Zafarali Ahmed and Tianqi Liu and Richard Powell and Vijay Bolina and Mariko Iinuma and Polina Zablotskaia and James Besley and Da-Woon Chung and Timothy Dozat and Ramona Comanescu and Xiance Si and Jeremy Greer and Guolong Su and Martin Polacek and Raphaël Lopez Kaufman and Simon Tokumine and Hexiang Hu and Elena Buchatskaya and Yingjie Miao and Mohamed Elhawaty and Aditya Siddhant and Nenad Tomasev and Jinwei Xing and Christina Greer and Helen Miller and Shereen Ashraf and Aurko Roy and Zizhao Zhang and Ada Ma and Angelos Filos and Milos Besta and Rory Blevins and Ted Klimenko and Chih-Kuan Yeh and Soravit Changpinyo and Jiaqi Mu and Oscar Chang and Mantas Pajarskas and Carrie Muir and Vered Cohen and Charline Le Lan and Krishna Haridasan and Amit Marathe and Steven Hansen and Sholto Douglas and Rajkumar Samuel and Mingqiu Wang and Sophia Austin and Chang Lan and Jiepu Jiang and Justin Chiu and Jaime Alonso Lorenzo and Lars Lowe Sjösund and Sébastien Cevey and Zach Gleicher and Thi Avrahami and Anudhyan Boral and Hansa Srinivasan and Vittorio Selo and Rhys May and Konstantinos Aisopos and Léonard Hussenot and Livio Baldini Soares and Kate Baumli and Michael B. Chang and Adrià Recasens and Ben Caine and Alexander Pritzel and Filip Pavetic and Fabio Pardo and Anita Gergely and Justin Frye and Vinay Ramasesh and Dan Horgan and Kartikeya Badola and Nora Kassner and Subhrajit Roy and Ethan Dyer and Víctor Campos Campos and Alex Tomala and Yunhao Tang and Dalia El Badawy and Elspeth White and Basil Mustafa and Oran Lang and Abhishek Jindal and Sharad Vikram and Zhitao Gong and Sergi Caelles and Ross Hemsley and Gregory Thornton and Fangxiaoyu Feng and Wojciech Stokowiec and Ce Zheng and Phoebe Thacker and Çağlar Ünlü and Zhishuai Zhang and Mohammad Saleh and James Svensson and Max Bileschi and Piyush Patil and Ankesh Anand and Roman Ring and Katerina Tsihlas and Arpi Vezer and Marco Selvi and Toby Shevlane and Mikel Rodriguez and Tom Kwiatkowski and Samira Daruki and Keran Rong and Allan Dafoe and Nicholas FitzGerald and Keren Gu-Lemberg and Mina Khan and Lisa Anne Hendricks and Marie Pellat and Vladimir Feinberg and James Cobon-Kerr and Tara Sainath and Maribeth Rauh and Sayed Hadi Hashemi and Richard Ives and Yana Hasson and Eric Noland and Yuan Cao and Nathan Byrd and Le Hou and Qingze Wang and Thibault Sottiaux and Michela Paganini and Jean-Baptiste Lespiau and Alexandre Moufarek and Samer Hassan and Kaushik Shivakumar and Joost van Amersfoort and Amol Mandhane and Pratik Joshi and Anirudh Goyal and Matthew Tung and Andrew Brock and Hannah Sheahan and Vedant Misra and Cheng Li and Nemanja Rakićević and Mostafa Dehghani and Fangyu Liu and Sid Mittal and Junhyuk Oh and Seb Noury and Eren Sezener and Fantine Huot and Matthew Lamm and Nicola De Cao and Charlie Chen and Sidharth Mudgal and Romina Stella and Kevin Brooks and Gautam Vasudevan and Chenxi Liu and Mainak Chain and Nivedita Melinkeri and Aaron Cohen and Venus Wang and Kristie Seymore and Sergey Zubkov and Rahul Goel and Summer Yue and Sai Krishnakumaran and Brian Albert and Nate Hurley and Motoki Sano and Anhad Mohananey and Jonah Joughin and Egor Filonov and Tomasz Kępa and Yomna Eldawy and Jiawern Lim and Rahul Rishi and Shirin Badiezadegan and Taylor Bos and Jerry Chang and Sanil Jain and Sri Gayatri Sundara Padmanabhan and Subha Puttagunta and Kalpesh Krishna and Leslie Baker and Norbert Kalb and Vamsi Bedapudi and Adam Kurzrok and Shuntong Lei and Anthony Yu and Oren Litvin and Xiang Zhou and Zhichun Wu and Sam Sobell and Andrea Siciliano and Alan Papir and Robby Neale and Jonas Bragagnolo and Tej Toor and Tina Chen and Valentin Anklin and Feiran Wang and Richie Feng and Milad Gholami and Kevin Ling and Lijuan Liu and Jules Walter and Hamid Moghaddam and Arun Kishore and Jakub Adamek and Tyler Mercado and Jonathan Mallinson and Siddhinita Wandekar and Stephen Cagle and Eran Ofek and Guillermo Garrido and Clemens Lombriser and Maksim Mukha and Botu Sun and Hafeezul Rahman Mohammad and Josip Matak and Yadi Qian and Vikas Peswani and Pawel Janus and Quan Yuan and Leif Schelin and Oana David and Ankur Garg and Yifan He and Oleksii Duzhyi and Anton Älgmyr and Timothée Lottaz and Qi Li and Vikas Yadav and Luyao Xu and Alex Chinien and Rakesh Shivanna and Aleksandr Chuklin and Josie Li and Carrie Spadine and Travis Wolfe and Kareem Mohamed and Subhabrata Das and Zihang Dai and Kyle He and Daniel von Dincklage and Shyam Upadhyay and Akanksha Maurya and Luyan Chi and Sebastian Krause and Khalid Salama and Pam G Rabinovitch and Pavan Kumar Reddy M and Aarush Selvan and Mikhail Dektiarev and Golnaz Ghiasi and Erdem Guven and Himanshu Gupta and Boyi Liu and Deepak Sharma and Idan Heimlich Shtacher and Shachi Paul and Oscar Akerlund and François-Xavier Aubet and Terry Huang and Chen Zhu and Eric Zhu and Elico Teixeira and Matthew Fritze and Francesco Bertolini and Liana-Eleonora Marinescu and Martin Bölle and Dominik Paulus and Khyatti Gupta and Tejasi Latkar and Max Chang and Jason Sanders and Roopa Wilson and Xuewei Wu and Yi-Xuan Tan and Lam Nguyen Thiet and Tulsee Doshi and Sid Lall and Swaroop Mishra and Wanming Chen and Thang Luong and Seth Benjamin and Jasmine Lee and Ewa Andrejczuk and Dominik Rabiej and Vipul Ranjan and Krzysztof Styrc and Pengcheng Yin and Jon Simon and Malcolm Rose Harriott and Mudit Bansal and Alexei Robsky and Geoff Bacon and David Greene and Daniil Mirylenka and Chen Zhou and Obaid Sarvana and Abhimanyu Goyal and Samuel Andermatt and Patrick Siegler and Ben Horn and Assaf Israel and Francesco Pongetti and Chih-Wei "Louis" Chen and Marco Selvatici and Pedro Silva and Kathie Wang and Jackson Tolins and Kelvin Guu and Roey Yogev and Xiaochen Cai and Alessandro Agostini and Maulik Shah and Hung Nguyen and Noah Ó Donnaile and Sébastien Pereira and Linda Friso and Adam Stambler and Adam Kurzrok and Chenkai Kuang and Yan Romanikhin and Mark Geller and ZJ Yan and Kane Jang and Cheng-Chun Lee and Wojciech Fica and Eric Malmi and Qijun Tan and Dan Banica and Daniel Balle and Ryan Pham and Yanping Huang and Diana Avram and Hongzhi Shi and Jasjot Singh and Chris Hidey and Niharika Ahuja and Pranab Saxena and Dan Dooley and Srividya Pranavi Potharaju and Eileen O'Neill and Anand Gokulchandran and Ryan Foley and Kai Zhao and Mike Dusenberry and Yuan Liu and Pulkit Mehta and Ragha Kotikalapudi and Chalence Safranek-Shrader and Andrew Goodman and Joshua Kessinger and Eran Globen and Prateek Kolhar and Chris Gorgolewski and Ali Ibrahim and Yang Song and Ali Eichenbaum and Thomas Brovelli and Sahitya Potluri and Preethi Lahoti and Cip Baetu and Ali Ghorbani and Charles Chen and Andy Crawford and Shalini Pal and Mukund Sridhar and Petru Gurita and Asier Mujika and Igor Petrovski and Pierre-Louis Cedoz and Chenmei Li and Shiyuan Chen and Niccolò Dal Santo and Siddharth Goyal and Jitesh Punjabi and Karthik Kappaganthu and Chester Kwak and Pallavi LV and Sarmishta Velury and Himadri Choudhury and Jamie Hall and Premal Shah and Ricardo Figueira and Matt Thomas and Minjie Lu and Ting Zhou and Chintu Kumar and Thomas Jurdi and Sharat Chikkerur and Yenai Ma and Adams Yu and Soo Kwak and Victor Ähdel and Sujeevan Rajayogam and Travis Choma and Fei Liu and Aditya Barua and Colin Ji and Ji Ho Park and Vincent Hellendoorn and Alex Bailey and Taylan Bilal and Huanjie Zhou and Mehrdad Khatir and Charles Sutton and Wojciech Rzadkowski and Fiona Macintosh and Konstantin Shagin and Paul Medina and Chen Liang and Jinjing Zhou and Pararth Shah and Yingying Bi and Attila Dankovics and Shipra Banga and Sabine Lehmann and Marissa Bredesen and Zifan Lin and John Eric Hoffmann and Jonathan Lai and Raynald Chung and Kai Yang and Nihal Balani and Arthur Bražinskas and Andrei Sozanschi and Matthew Hayes and Héctor Fernández Alcalde and Peter Makarov and Will Chen and Antonio Stella and Liselotte Snijders and Michael Mandl and Ante Kärrman and Paweł Nowak and Xinyi Wu and Alex Dyck and Krishnan Vaidyanathan and Raghavender R and Jessica Mallet and Mitch Rudominer and Eric Johnston and Sushil Mittal and Akhil Udathu and Janara Christensen and Vishal Verma and Zach Irving and Andreas Santucci and Gamaleldin Elsayed and Elnaz Davoodi and Marin Georgiev and Ian Tenney and Nan Hua and Geoffrey Cideron and Edouard Leurent and Mahmoud Alnahlawi and Ionut Georgescu and Nan Wei and Ivy Zheng and Dylan Scandinaro and Heinrich Jiang and Jasper Snoek and Mukund Sundararajan and Xuezhi Wang and Zack Ontiveros and Itay Karo and Jeremy Cole and Vinu Rajashekhar and Lara Tumeh and Eyal Ben-David and Rishub Jain and Jonathan Uesato and Romina Datta and Oskar Bunyan and Shimu Wu and John Zhang and Piotr Stanczyk and Ye Zhang and David Steiner and Subhajit Naskar and Michael Azzam and Matthew Johnson and Adam Paszke and Chung-Cheng Chiu and Jaume Sanchez Elias and Afroz Mohiuddin and Faizan Muhammad and Jin Miao and Andrew Lee and Nino Vieillard and Jane Park and Jiageng Zhang and Jeff Stanway and Drew Garmon and Abhijit Karmarkar and Zhe Dong and Jong Lee and Aviral Kumar and Luowei Zhou and Jonathan Evens and William Isaac and Geoffrey Irving and Edward Loper and Michael Fink and Isha Arkatkar and Nanxin Chen and Izhak Shafran and Ivan Petrychenko and Zhe Chen and Johnson Jia and Anselm Levskaya and Zhenkai Zhu and Peter Grabowski and Yu Mao and Alberto Magni and Kaisheng Yao and Javier Snaider and Norman Casagrande and Evan Palmer and Paul Suganthan and Alfonso Castaño and Irene Giannoumis and Wooyeol Kim and Mikołaj Rybiński and Ashwin Sreevatsa and Jennifer Prendki and David Soergel and Adrian Goedeckemeyer and Willi Gierke and Mohsen Jafari and Meenu Gaba and Jeremy Wiesner and Diana Gage Wright and Yawen Wei and Harsha Vashisht and Yana Kulizhskaya and Jay Hoover and Maigo Le and Lu Li and Chimezie Iwuanyanwu and Lu Liu and Kevin Ramirez and Andrey Khorlin and Albert Cui and Tian LIN and Marcus Wu and Ricardo Aguilar and Keith Pallo and Abhishek Chakladar and Ginger Perng and Elena Allica Abellan and Mingyang Zhang and Ishita Dasgupta and Nate Kushman and Ivo Penchev and Alena Repina and Xihui Wu and Tom van der Weide and Priya Ponnapalli and Caroline Kaplan and Jiri Simsa and Shuangfeng Li and Olivier Dousse and Fan Yang and Jeff Piper and Nathan Ie and Rama Pasumarthi and Nathan Lintz and Anitha Vijayakumar and Daniel Andor and Pedro Valenzuela and Minnie Lui and Cosmin Paduraru and Daiyi Peng and Katherine Lee and Shuyuan Zhang and Somer Greene and Duc Dung Nguyen and Paula Kurylowicz and Cassidy Hardin and Lucas Dixon and Lili Janzer and Kiam Choo and Ziqiang Feng and Biao Zhang and Achintya Singhal and Dayou Du and Dan McKinnon and Natasha Antropova and Tolga Bolukbasi and Orgad Keller and David Reid and Daniel Finchelstein and Maria Abi Raad and Remi Crocker and Peter Hawkins and Robert Dadashi and Colin Gaffney and Ken Franko and Anna Bulanova and Rémi Leblond and Shirley Chung and Harry Askham and Luis C. Cobo and Kelvin Xu and Felix Fischer and Jun Xu and Christina Sorokin and Chris Alberti and Chu-Cheng Lin and Colin Evans and Alek Dimitriev and Hannah Forbes and Dylan Banarse and Zora Tung and Mark Omernick and Colton Bishop and Rachel Sterneck and Rohan Jain and Jiawei Xia and Ehsan Amid and Francesco Piccinno and Xingyu Wang and Praseem Banzal and Daniel J. Mankowitz and Alex Polozov and Victoria Krakovna and Sasha Brown and MohammadHossein Bateni and Dennis Duan and Vlad Firoiu and Meghana Thotakuri and Tom Natan and Matthieu Geist and Ser tan Girgin and Hui Li and Jiayu Ye and Ofir Roval and Reiko Tojo and Michael Kwong and James Lee-Thorp and Christopher Yew and Danila Sinopalnikov and Sabela Ramos and John Mellor and Abhishek Sharma and Kathy Wu and David Miller and Nicolas Sonnerat and Denis Vnukov and Rory Greig and Jennifer Beattie and Emily Caveness and Libin Bai and Julian Eisenschlos and Alex Korchemniy and Tomy Tsai and Mimi Jasarevic and Weize Kong and Phuong Dao and Zeyu Zheng and Frederick Liu and Fan Yang and Rui Zhu and Tian Huey Teh and Jason Sanmiya and Evgeny Gladchenko and Nejc Trdin and Daniel Toyama and Evan Rosen and Sasan Tavakkol and Linting Xue and Chen Elkind and Oliver Woodman and John Carpenter and George Papamakarios and Rupert Kemp and Sushant Kafle and Tanya Grunina and Rishika Sinha and Alice Talbert and Diane Wu and Denese Owusu-Afriyie and Cosmo Du and Chloe Thornton and Jordi Pont-Tuset and Pradyumna Narayana and Jing Li and Saaber Fatehi and John Wieting and Omar Ajmeri and Benigno Uria and Yeongil Ko and Laura Knight and Amélie Héliou and Ning Niu and Shane Gu and Chenxi Pang and Yeqing Li and Nir Levine and Ariel Stolovich and Rebeca Santamaria-Fernandez and Sonam Goenka and Wenny Yustalim and Robin Strudel and Ali Elqursh and Charlie Deck and Hyo Lee and Zonglin Li and Kyle Levin and Raphael Hoffmann and Dan Holtmann-Rice and Olivier Bachem and Sho Arora and Christy Koh and Soheil Hassas Yeganeh and Siim Põder and Mukarram Tariq and Yanhua Sun and Lucian Ionita and Mojtaba Seyedhosseini and Pouya Tafti and Zhiyu Liu and Anmol Gulati and Jasmine Liu and Xinyu Ye and Bart Chrzaszcz and Lily Wang and Nikhil Sethi and Tianrun Li and Ben Brown and Shreya Singh and Wei Fan and Aaron Parisi and Joe Stanton and Vinod Koverkathu and Christopher A. Choquette-Choo and Yunjie Li and TJ Lu and Abe Ittycheriah and Prakash Shroff and Mani Varadarajan and Sanaz Bahargam and Rob Willoughby and David Gaddy and Guillaume Desjardins and Marco Cornero and Brona Robenek and Bhavishya Mittal and Ben Albrecht and Ashish Shenoy and Fedor Moiseev and Henrik Jacobsson and Alireza Ghaffarkhah and Morgane Rivière and Alanna Walton and Clément Crepy and Alicia Parrish and Zongwei Zhou and Clement Farabet and Carey Radebaugh and Praveen Srinivasan and Claudia van der Salm and Andreas Fidjeland and Salvatore Scellato and Eri Latorre-Chimoto and Hanna Klimczak-Plucińska and David Bridson and Dario de Cesare and Tom Hudson and Piermaria Mendolicchio and Lexi Walker and Alex Morris and Matthew Mauger and Alexey Guseynov and Alison Reid and Seth Odoom and Lucia Loher and Victor Cotruta and Madhavi Yenugula and Dominik Grewe and Anastasia Petrushkina and Tom Duerig and Antonio Sanchez and Steve Yadlowsky and Amy Shen and Amir Globerson and Lynette Webb and Sahil Dua and Dong Li and Surya Bhupatiraju and Dan Hurt and Haroon Qureshi and Ananth Agarwal and Tomer Shani and Matan Eyal and Anuj Khare and Shreyas Rammohan Belle and Lei Wang and Chetan Tekur and Mihir Sanjay Kale and Jinliang Wei and Ruoxin Sang and Brennan Saeta and Tyler Liechty and Yi Sun and Yao Zhao and Stephan Lee and Pandu Nayak and Doug Fritz and Manish Reddy Vuyyuru and John Aslanides and Nidhi Vyas and Martin Wicke and Xiao Ma and Evgenii Eltyshev and Nina Martin and Hardie Cate and James Manyika and Keyvan Amiri and Yelin Kim and Xi Xiong and Kai Kang and Florian Luisier and Nilesh Tripuraneni and David Madras and Mandy Guo and Austin Waters and Oliver Wang and Joshua Ainslie and Jason Baldridge and Han Zhang and Garima Pruthi and Jakob Bauer and Feng Yang and Riham Mansour and Jason Gelman and Yang Xu and George Polovets and Ji Liu and Honglong Cai and Warren Chen and XiangHai Sheng and Emily Xue and Sherjil Ozair and Christof Angermueller and Xiaowei Li and Anoop Sinha and Weiren Wang and Julia Wiesinger and Emmanouil Koukoumidis and Yuan Tian and Anand Iyer and Madhu Gurumurthy and Mark Goldenson and Parashar Shah and MK Blake and Hongkun Yu and Anthony Urbanowicz and Jennimaria Palomaki and Chrisantha Fernando and Ken Durden and Harsh Mehta and Nikola Momchev and Elahe Rahimtoroghi and Maria Georgaki and Amit Raul and Sebastian Ruder and Morgan Redshaw and Jinhyuk Lee and Denny Zhou and Komal Jalan and Dinghua Li and Blake Hechtman and Parker Schuh and Milad Nasr and Kieran Milan and Vladimir Mikulik and Juliana Franco and Tim Green and Nam Nguyen and Joe Kelley and Aroma Mahendru and Andrea Hu and Joshua Howland and Ben Vargas and Jeffrey Hui and Kshitij Bansal and Vikram Rao and Rakesh Ghiya and Emma Wang and Ke Ye and Jean Michel Sarr and Melanie Moranski Preston and Madeleine Elish and Steve Li and Aakash Kaku and Jigar Gupta and Ice Pasupat and Da-Cheng Juan and Milan Someswar and Tejvi M. and Xinyun Chen and Aida Amini and Alex Fabrikant and Eric Chu and Xuanyi Dong and Amruta Muthal and Senaka Buthpitiya and Sarthak Jauhari and Nan Hua and Urvashi Khandelwal and Ayal Hitron and Jie Ren and Larissa Rinaldi and Shahar Drath and Avigail Dabush and Nan-Jiang Jiang and Harshal Godhia and Uli Sachs and Anthony Chen and Yicheng Fan and Hagai Taitelbaum and Hila Noga and Zhuyun Dai and James Wang and Chen Liang and Jenny Hamer and Chun-Sung Ferng and Chenel Elkind and Aviel Atias and Paulina Lee and Vít Listík and Mathias Carlen and Jan van de Kerkhof and Marcin Pikus and Krunoslav Zaher and Paul Müller and Sasha Zykova and Richard Stefanec and Vitaly Gatsko and Christoph Hirnschall and Ashwin Sethi and Xingyu Federico Xu and Chetan Ahuja and Beth Tsai and Anca Stefanoiu and Bo Feng and Keshav Dhandhania and Manish Katyal and Akshay Gupta and Atharva Parulekar and Divya Pitta and Jing Zhao and Vivaan Bhatia and Yashodha Bhavnani and Omar Alhadlaq and Xiaolin Li and Peter Danenberg and Dennis Tu and Alex Pine and Vera Filippova and Abhipso Ghosh and Ben Limonchik and Bhargava Urala and Chaitanya Krishna Lanka and Derik Clive and Yi Sun and Edward Li and Hao Wu and Kevin Hongtongsak and Ianna Li and Kalind Thakkar and Kuanysh Omarov and Kushal Majmundar and Michael Alverson and Michael Kucharski and Mohak Patel and Mudit Jain and Maksim Zabelin and Paolo Pelagatti and Rohan Kohli and Saurabh Kumar and Joseph Kim and Swetha Sankar and Vineet Shah and Lakshmi Ramachandruni and Xiangkai Zeng and Ben Bariach and Laura Weidinger and Tu Vu and Alek Andreev and Antoine He and Kevin Hui and Sheleem Kashem and Amar Subramanya and Sissie Hsiao and Demis Hassabis and Koray Kavukcuoglu and Adam Sadovsky and Quoc Le and Trevor Strohman and Yonghui Wu and Slav Petrov and Jeffrey Dean and Oriol Vinyals},
  year         = {2023},
  journal      = {arXiv preprint arXiv:2312.11805},
}

@article{achiam2023gpt,
  title        = {Gpt-4 technical report},
  author       = {Achiam, Josh and Adler, Steven and Agarwal, Sandhini and Ahmad, Lama and Akkaya, Ilge and Aleman, Florencia Leoni and Almeida, Diogo and Altenschmidt, Janko and Altman, Sam and Anadkat, Shyamal and others},
  year         = {2023},
  journal      = {arXiv preprint arXiv:2303.08774},
}

@article{zhang2023instruction,
  title        = {Instruction tuning for large language models: A survey},
  author={Shengyu Zhang and Linfeng Dong and Xiaoya Li and Sen Zhang and Xiaofei Sun and Shuhe Wang and Jiwei Li and Runyi Hu and Tianwei Zhang and Fei Wu and Guoyin Wang},
  year         = {2023},
  journal      = {arXiv preprint arXiv:2308.10792},
}

@article{touvron2023llama,
  title        = {Llama: Open and efficient foundation language models},
  author       = {Hugo Touvron and Thibaut Lavril and Gautier Izacard and Xavier Martinet and Marie-Anne Lachaux and Timothée Lacroix and Baptiste Rozière and Naman Goyal and Eric Hambro and Faisal Azhar and Aurelien Rodriguez and Armand Joulin and Edouard Grave and Guillaume Lample},
  year         = {2023},
  journal      = {arXiv preprint arXiv:2302.13971},
}

@article{anil2023palm,
  title        = {Palm 2 technical report},
  author       = {Rohan Anil and Andrew M. Dai and Orhan Firat and Melvin Johnson and Dmitry Lepikhin and Alexandre Passos and Siamak Shakeri and Emanuel Taropa and Paige Bailey and Zhifeng Chen and Eric Chu and Jonathan H. Clark and Laurent El Shafey and Yanping Huang and Kathy Meier-Hellstern and Gaurav Mishra and Erica Moreira and Mark Omernick and Kevin Robinson and Sebastian Ruder and Yi Tay and Kefan Xiao and Yuanzhong Xu and Yujing Zhang and Gustavo Hernandez Abrego and Junwhan Ahn and Jacob Austin and Paul Barham and Jan Botha and James Bradbury and Siddhartha Brahma and Kevin Brooks and Michele Catasta and Yong Cheng and Colin Cherry and Christopher A. Choquette-Choo and Aakanksha Chowdhery and Clément Crepy and Shachi Dave and Mostafa Dehghani and Sunipa Dev and Jacob Devlin and Mark Díaz and Nan Du and Ethan Dyer and Vlad Feinberg and Fangxiaoyu Feng and Vlad Fienber and Markus Freitag and Xavier Garcia and Sebastian Gehrmann and Lucas Gonzalez and Guy Gur-Ari and Steven Hand and Hadi Hashemi and Le Hou and Joshua Howland and Andrea Hu and Jeffrey Hui and Jeremy Hurwitz and Michael Isard and Abe Ittycheriah and Matthew Jagielski and Wenhao Jia and Kathleen Kenealy and Maxim Krikun and Sneha Kudugunta and Chang Lan and Katherine Lee and Benjamin Lee and Eric Li and Music Li and Wei Li and YaGuang Li and Jian Li and Hyeontaek Lim and Hanzhao Lin and Zhongtao Liu and Frederick Liu and Marcello Maggioni and Aroma Mahendru and Joshua Maynez and Vedant Misra and Maysam Moussalem and Zachary Nado and John Nham and Eric Ni and Andrew Nystrom and Alicia Parrish and Marie Pellat and Martin Polacek and Alex Polozov and Reiner Pope and Siyuan Qiao and Emily Reif and Bryan Richter and Parker Riley and Alex Castro Ros and Aurko Roy and Brennan Saeta and Rajkumar Samuel and Renee Shelby and Ambrose Slone and Daniel Smilkov and David R. So and Daniel Sohn and Simon Tokumine and Dasha Valter and Vijay Vasudevan and Kiran Vodrahalli and Xuezhi Wang and Pidong Wang and Zirui Wang and Tao Wang and John Wieting and Yuhuai Wu and Kelvin Xu and Yunhan Xu and Linting Xue and Pengcheng Yin and Jiahui Yu and Qiao Zhang and Steven Zheng and Ce Zheng and Weikang Zhou and Denny Zhou and Slav Petrov and Yonghui Wu},
  year         = {2023},
  journal      = {arXiv preprint arXiv:2305.10403},
}

@article{chowdhery2023palm,
  title        = {Palm: Scaling language modeling with pathways},
  author       = {Aakanksha Chowdhery and Sharan Narang and Jacob Devlin and Maarten Bosma and Gaurav Mishra and Adam Roberts and Paul Barham and Hyung Won Chung and Charles Sutton and Sebastian Gehrmann and Parker Schuh and Kensen Shi and Sasha Tsvyashchenko and Joshua Maynez and Abhishek Rao and Parker Barnes and Yi Tay and Noam Shazeer and Vinodkumar Prabhakaran and Emily Reif and Nan Du and Ben Hutchinson and Reiner Pope and James Bradbury and Jacob Austin and Michael Isard and Guy Gur-Ari and Pengcheng Yin and Toju Duke and Anselm Levskaya and Sanjay Ghemawat and Sunipa Dev and Henryk Michalewski and Xavier Garcia and Vedant Misra and Kevin Robinson and Liam Fedus and Denny Zhou and Daphne Ippolito and David Luan and Hyeontaek Lim and Barret Zoph and Alexander Spiridonov and Ryan Sepassi and David Dohan and Shivani Agrawal and Mark Omernick and Andrew M. Dai and Thanumalayan Sankaranarayana Pillai and Marie Pellat and Aitor Lewkowycz and Erica Moreira and Rewon Child and Oleksandr Polozov and Katherine Lee and Zongwei Zhou and Xuezhi Wang and Brennan Saeta and Mark Diaz and Orhan Firat and Michele Catasta and Jason Wei and Kathy Meier-Hellstern and Douglas Eck and Jeff Dean and Slav Petrov and Noah Fiedel},
  year         = {2023},
  journal      = {Journal of Machine Learning Research},
  volume       = {24},
  number       = {240},
  pages        = {1--113},
}

@inproceedings{yan2023rethinking,
  title        = {Rethinking $\{$White-Box$\}$ Watermarks on Deep Learning Models under Neural Structural Obfuscation},
  author       = {Yan, Yifan and Pan, Xudong and Zhang, Mi and Yang, Min},
  year         = {2023},
  booktitle    = {32nd USENIX Security Symposium (USENIX Security 23)},
  pages        = {2347--2364},
}

@inproceedings{xu-he-2023-security,
  title        = {Security Challenges in Natural Language Processing Models},
  author       = {Xu, Qiongkai  and He, Xuanli},
  year         = {2023},
  month        = dec,
  booktitle    = {Proceedings of the 2023 Conference on Empirical Methods in Natural Language Processing: Tutorial Abstracts},
  publisher    = {Association for Computational Linguistics},
  address      = {Singapore},
  pages        = {7--12},
  editor       = {Zhang, Qi  and Sajjad, Hassan},
}

@inproceedings{morris-etal-2023-text,
  title        = {Text Embeddings Reveal (Almost) As Much As Text},
  author       = {Morris, John  and Kuleshov, Volodymyr  and Shmatikov, Vitaly  and Rush, Alexander},
  year         = {2023},
  month        = dec,
  booktitle    = {Proceedings of the 2023 Conference on Empirical Methods in Natural Language Processing},
  publisher    = {Association for Computational Linguistics},
  address      = {Singapore},
  pages        = {12448--12460},
  editor       = {Bouamor, Houda  and Pino, Juan  and Bali, Kalika},
}

@article{christ2023undetectable,
  title        = {Undetectable Watermarks for Language Models},
  author       = {Christ, Miranda and Gunn, Sam and Zamir, Or},
  year         = {2023},
  journal      = {Cryptology ePrint Archive},
}

@misc{openAI-LLM-WM,
  title        = {Watermarking of Large Language Models},
  author       = {Scott Aaronson},
  year         = {2023},
  language     = {en-US},
}

@article{la2023language,
  title        = {Language-Models-as-a-Service: Overview of a New Paradigm and its Challenges},
  author       = {La Malfa, Emanuele and Petrov, Aleksandar and Frieder, Simon and Weinhuber, Christoph and Burnell, Ryan and Nazar, Raza and Cohn, Anthony and Shadbolt, Nigel and Wooldridge, Michael},
  year         = {2024},
  journal      = {Journal of Artificial Intelligence Research},
  volume       = {80},
  pages        = {1497--1523},
}

@inproceedings{chakraborty2023possibilities,
  title        = {Position: On the Possibilities of {AI}-Generated Text Detection},
  author       = {Chakraborty, Souradip and Bedi, Amrit and Zhu, Sicheng and An, Bang and Manocha, Dinesh and Huang, Furong},
  year         = {2024},
  month        = {21--27 Jul},
  booktitle    = {Proceedings of the 41st International Conference on Machine Learning},
  publisher    = {PMLR},
  series       = {Proceedings of Machine Learning Research},
  volume       = {235},
  pages        = {6093--6115},
  editor       = {Salakhutdinov, Ruslan and Kolter, Zico and Heller, Katherine and Weller, Adrian and Oliver, Nuria and Scarlett, Jonathan and Berkenkamp, Felix},
}

@article{kuditipudi2023robust,
  title        = {Robust Distortion-free Watermarks for Language Models},
  author       = {Rohith Kuditipudi and John Thickstun and Tatsunori Hashimoto and Percy Liang},
  year         = {2024},
  journal      = {Transactions on Machine Learning Research},
  issn         = {2835-8856},
}

@misc{recentopenaiEmbeddingModels,
  title        = {{N}ew embedding models and {A}{P}{I} updates},
  author       = {OpenAI},
  year         = {2024},
journal      = {OpenAI blog},
}

@inproceedings{shetty2024warden,
  title        = {{WARDEN}: Multi-Directional Backdoor Watermarks for Embedding-as-a-Service Copyright Protection},
  author       = {Shetty, Anudeex  and Teng, Yue  and He, Ke  and Xu, Qiongkai},
  year         = {2024},
  month        = aug,
  booktitle    = {Proceedings of the 62nd Annual Meeting of the Association for Computational Linguistics (Volume 1: Long Papers)},
  publisher    = {Association for Computational Linguistics},
  address      = {Bangkok, Thailand},
  pages        = {13430--13444},
  editor       = {Ku, Lun-Wei  and Martins, Andre  and Srikumar, Vivek},
}

@article{liu2024survey,
  title        = {A survey of text watermarking in the era of large language models},
  author       = {Liu, Aiwei and Pan, Leyi and Lu, Yijian and Li, Jingjing and Hu, Xuming and Zhang, Xi and Wen, Lijie and King, Irwin and Xiong, Hui and Yu, Philip},
  year         = {2024},
  journal      = {ACM Computing Surveys},
  publisher    = {ACM New York, NY},
}

@inproceedings{he2024can,
  title        = {Can Watermarks Survive Translation? On the Cross-lingual Consistency of Text Watermark for Large Language Models},
  author       = {He, Zhiwei  and Zhou, Binglin  and Hao, Hongkun  and Liu, Aiwei  and Wang, Xing  and Tu, Zhaopeng  and Zhang, Zhuosheng  and Wang, Rui},
  year         = {2024},
  month        = aug,
  booktitle    = {Proceedings of the 62nd Annual Meeting of the Association for Computational Linguistics (Volume 1: Long Papers)},
  publisher    = {Association for Computational Linguistics},
  address      = {Bangkok, Thailand},
  pages        = {4115--4129},
  editor       = {Ku, Lun-Wei  and Martins, Andre  and Srikumar, Vivek},
}

@inproceedings{RTT-wm,
  title        = {Can Watermarks Survive Translation? On the Cross-lingual Consistency of Text Watermark for Large Language Models},
  author       = {He, Zhiwei  and Zhou, Binglin  and Hao, Hongkun  and Liu, Aiwei  and Wang, Xing  and Tu, Zhaopeng  and Zhang, Zhuosheng  and Wang, Rui},
  year         = {2024},
  month        = aug,
  booktitle    = {Proceedings of the 62nd Annual Meeting of the Association for Computational Linguistics (Volume 1: Long Papers)},
  publisher    = {Association for Computational Linguistics},
  address      = {Bangkok, Thailand},
  pages        = {4115--4129},
  editor       = {Ku, Lun-Wei  and Martins, Andre  and Srikumar, Vivek},
}

@inproceedings{zhu-etal-2024-multilingual,
  title        = {Multilingual Machine Translation with Large Language Models: Empirical Results and Analysis},
  author       = {Zhu, Wenhao  and Liu, Hongyi  and Dong, Qingxiu  and Xu, Jingjing  and Huang, Shujian  and Kong, Lingpeng  and Chen, Jiajun  and Li, Lei},
  year         = {2024},
  month        = jun,
  booktitle    = {Findings of the Association for Computational Linguistics: NAACL 2024},
  publisher    = {Association for Computational Linguistics},
  address      = {Mexico City, Mexico},
  pages        = {2765--2781},
  editor       = {Duh, Kevin  and Gomez, Helena  and Bethard, Steven},
}

@inproceedings{kirchenbauer2024on,
  title        = {On the Reliability of Watermarks for Large Language Models},
  author       = {John Kirchenbauer and Jonas Geiping and Yuxin Wen and Manli Shu and Khalid Saifullah and Kezhi Kong and Kasun Fernando and Aniruddha Saha and Micah Goldblum and Tom Goldstein},
  year         = {2024},
  booktitle    = {The Twelfth International Conference on Learning Representations},
}

@article{krishna2024paraphrasing,
  title        = {Paraphrasing evades detectors of ai-generated text, but retrieval is an effective defense},
  author       = {Krishna, Kalpesh and Song, Yixiao and Karpinska, Marzena and Wieting, John and Iyyer, Mohit},
  year         = {2024},
  journal      = {Advances in Neural Information Processing Systems},
  volume       = {36},
}

@inproceedings{zhao2024provable,
  title        = {Provable Robust Watermarking for {AI}-Generated Text},
  author       = {Xuandong Zhao and Prabhanjan Vijendra Ananth and Lei Li and Yu-Xiang Wang},
  year         = {2024},
  booktitle    = {The Twelfth International Conference on Learning Representations},
}

@article{shanahan2024talking,
  title        = {Talking about large language models},
  author       = {Shanahan, Murray},
  year         = {2024},
  journal      = {Communications of the ACM},
  publisher    = {ACM New York, NY, USA},
  volume       = {67},
  number       = {2},
  pages        = {68--79},
}

@inproceedings{chen2024text,
  title        = {Text Embedding Inversion Security for Multilingual Language Models},
  author       = {Chen, Yiyi  and Lent, Heather  and Bjerva, Johannes},
  year         = {2024},
  month        = aug,
  booktitle    = {Proceedings of the 62nd Annual Meeting of the Association for Computational Linguistics (Volume 1: Long Papers)},
  publisher    = {Association for Computational Linguistics},
  address      = {Bangkok, Thailand},
  pages        = {7808--7827},
  editor       = {Ku, Lun-Wei  and Martins, Andre  and Srikumar, Vivek},
}

@article{dubey2024llama,
  title        = {The llama 3 herd of models},
  author       = {Abhimanyu Dubey and Abhinav Jauhri and Abhinav Pandey and Abhishek Kadian and Ahmad Al-Dahle and Aiesha Letman and Akhil Mathur and Alan Schelten and Amy Yang and Angela Fan and Anirudh Goyal and Anthony Hartshorn and Aobo Yang and Archi Mitra and Archie Sravankumar and Artem Korenev and Arthur Hinsvark and Arun Rao and Aston Zhang and Aurelien Rodriguez and Austen Gregerson and Ava Spataru and Baptiste Roziere and Bethany Biron and Binh Tang and Bobbie Chern and Charlotte Caucheteux and Chaya Nayak and Chloe Bi and Chris Marra and Chris McConnell and Christian Keller and Christophe Touret and Chunyang Wu and Corinne Wong and Cristian Canton Ferrer and Cyrus Nikolaidis and Damien Allonsius and Daniel Song and Danielle Pintz and Danny Livshits and David Esiobu and Dhruv Choudhary and Dhruv Mahajan and Diego Garcia-Olano and Diego Perino and Dieuwke Hupkes and Egor Lakomkin and Ehab AlBadawy and Elina Lobanova and Emily Dinan and Eric Michael Smith and Filip Radenovic and Frank Zhang and Gabriel Synnaeve and Gabrielle Lee and Georgia Lewis Anderson and Graeme Nail and Gregoire Mialon and Guan Pang and Guillem Cucurell and Hailey Nguyen and Hannah Korevaar and Hu Xu and Hugo Touvron and Iliyan Zarov and Imanol Arrieta Ibarra and Isabel Kloumann and Ishan Misra and Ivan Evtimov and Jade Copet and Jaewon Lee and Jan Geffert and Jana Vranes and Jason Park and Jay Mahadeokar and Jeet Shah and Jelmer van der Linde and Jennifer Billock and Jenny Hong and Jenya Lee and Jeremy Fu and Jianfeng Chi and Jianyu Huang and Jiawen Liu and Jie Wang and Jiecao Yu and Joanna Bitton and Joe Spisak and Jongsoo Park and Joseph Rocca and Joshua Johnstun and Joshua Saxe and Junteng Jia and Kalyan Vasuden Alwala and Kartikeya Upasani and Kate Plawiak and Ke Li and Kenneth Heafield and Kevin Stone and Khalid El-Arini and Krithika Iyer and Kshitiz Malik and Kuenley Chiu and Kunal Bhalla and Lauren Rantala-Yeary and Laurens van der Maaten and Lawrence Chen and Liang Tan and Liz Jenkins and Louis Martin and Lovish Madaan and Lubo Malo and Lukas Blecher and Lukas Landzaat and Luke de Oliveira and Madeline Muzzi and Mahesh Pasupuleti and Mannat Singh and Manohar Paluri and Marcin Kardas and Mathew Oldham and Mathieu Rita and Maya Pavlova and Melanie Kambadur and Mike Lewis and Min Si and Mitesh Kumar Singh and Mona Hassan and Naman Goyal and Narjes Torabi and Nikolay Bashlykov and Nikolay Bogoychev and Niladri Chatterji and Olivier Duchenne and Onur Çelebi and Patrick Alrassy and Pengchuan Zhang and Pengwei Li and Petar Vasic and Peter Weng and Prajjwal Bhargava and Pratik Dubal and Praveen Krishnan and Punit Singh Koura and Puxin Xu and Qing He and Qingxiao Dong and Ragavan Srinivasan and Raj Ganapathy and Ramon Calderer and Ricardo Silveira Cabral and Robert Stojnic and Roberta Raileanu and Rohit Girdhar and Rohit Patel and Romain Sauvestre and Ronnie Polidoro and Roshan Sumbaly and Ross Taylor and Ruan Silva and Rui Hou and Rui Wang and Saghar Hosseini and Sahana Chennabasappa and Sanjay Singh and Sean Bell and Seohyun Sonia Kim and Sergey Edunov and Shaoliang Nie and Sharan Narang and Sharath Raparthy and Sheng Shen and Shengye Wan and Shruti Bhosale and Shun Zhang and Simon Vandenhende and Soumya Batra and Spencer Whitman and Sten Sootla and Stephane Collot and Suchin Gururangan and Sydney Borodinsky and Tamar Herman and Tara Fowler and Tarek Sheasha and Thomas Georgiou and Thomas Scialom and Tobias Speckbacher and Todor Mihaylov and Tong Xiao and Ujjwal Karn and Vedanuj Goswami and Vibhor Gupta and Vignesh Ramanathan and Viktor Kerkez and Vincent Gonguet and Virginie Do and Vish Vogeti and Vladan Petrovic and Weiwei Chu and Wenhan Xiong and Wenyin Fu and Whitney Meers and Xavier Martinet and Xiaodong Wang and Xiaoqing Ellen Tan and Xinfeng Xie and Xuchao Jia and Xuewei Wang and Yaelle Goldschlag and Yashesh Gaur and Yasmine Babaei and Yi Wen and Yiwen Song and Yuchen Zhang and Yue Li and Yuning Mao and Zacharie Delpierre Coudert and Zheng Yan and Zhengxing Chen and Zoe Papakipos and Aaditya Singh and Aaron Grattafiori and Abha Jain and Adam Kelsey and Adam Shajnfeld and Adithya Gangidi and Adolfo Victoria and Ahuva Goldstand and Ajay Menon and Ajay Sharma and Alex Boesenberg and Alex Vaughan and Alexei Baevski and Allie Feinstein and Amanda Kallet and Amit Sangani and Anam Yunus and Andrei Lupu and Andres Alvarado and Andrew Caples and Andrew Gu and Andrew Ho and Andrew Poulton and Andrew Ryan and Ankit Ramchandani and Annie Franco and Aparajita Saraf and Arkabandhu Chowdhury and Ashley Gabriel and Ashwin Bharambe and Assaf Eisenman and Azadeh Yazdan and Beau James and Ben Maurer and Benjamin Leonhardi and Bernie Huang and Beth Loyd and Beto De Paola and Bhargavi Paranjape and Bing Liu and Bo Wu and Boyu Ni and Braden Hancock and Bram Wasti and Brandon Spence and Brani Stojkovic and Brian Gamido and Britt Montalvo and Carl Parker and Carly Burton and Catalina Mejia and Changhan Wang and Changkyu Kim and Chao Zhou and Chester Hu and Ching-Hsiang Chu and Chris Cai and Chris Tindal and Christoph Feichtenhofer and Damon Civin and Dana Beaty and Daniel Kreymer and Daniel Li and Danny Wyatt and David Adkins and David Xu and Davide Testuggine and Delia David and Devi Parikh and Diana Liskovich and Didem Foss and Dingkang Wang and Duc Le and Dustin Holland and Edward Dowling and Eissa Jamil and Elaine Montgomery and Eleonora Presani and Emily Hahn and Emily Wood and Erik Brinkman and Esteban Arcaute and Evan Dunbar and Evan Smothers and Fei Sun and Felix Kreuk and Feng Tian and Firat Ozgenel and Francesco Caggioni and Francisco Guzmán and Frank Kanayet and Frank Seide and Gabriela Medina Florez and Gabriella Schwarz and Gada Badeer and Georgia Swee and Gil Halpern and Govind Thattai and Grant Herman and Grigory Sizov and Guangyi and Zhang and Guna Lakshminarayanan and Hamid Shojanazeri and Han Zou and Hannah Wang and Hanwen Zha and Haroun Habeeb and Harrison Rudolph and Helen Suk and Henry Aspegren and Hunter Goldman and Ibrahim Damlaj and Igor Molybog and Igor Tufanov and Irina-Elena Veliche and Itai Gat and Jake Weissman and James Geboski and James Kohli and Japhet Asher and Jean-Baptiste Gaya and Jeff Marcus and Jeff Tang and Jennifer Chan and Jenny Zhen and Jeremy Reizenstein and Jeremy Teboul and Jessica Zhong and Jian Jin and Jingyi Yang and Joe Cummings and Jon Carvill and Jon Shepard and Jonathan McPhie and Jonathan Torres and Josh Ginsburg and Junjie Wang and Kai Wu and Kam Hou U and Karan Saxena and Karthik Prasad and Kartikay Khandelwal and Katayoun Zand and Kathy Matosich and Kaushik Veeraraghavan and Kelly Michelena and Keqian Li and Kun Huang and Kunal Chawla and Kushal Lakhotia and Kyle Huang and Lailin Chen and Lakshya Garg and Lavender A and Leandro Silva and Lee Bell and Lei Zhang and Liangpeng Guo and Licheng Yu and Liron Moshkovich and Luca Wehrstedt and Madian Khabsa and Manav Avalani and Manish Bhatt and Maria Tsimpoukelli and Martynas Mankus and Matan Hasson and Matthew Lennie and Matthias Reso and Maxim Groshev and Maxim Naumov and Maya Lathi and Meghan Keneally and Michael L. Seltzer and Michal Valko and Michelle Restrepo and Mihir Patel and Mik Vyatskov and Mikayel Samvelyan and Mike Clark and Mike Macey and Mike Wang and Miquel Jubert Hermoso and Mo Metanat and Mohammad Rastegari and Munish Bansal and Nandhini Santhanam and Natascha Parks and Natasha White and Navyata Bawa and Nayan Singhal and Nick Egebo and Nicolas Usunier and Nikolay Pavlovich Laptev and Ning Dong and Ning Zhang and Norman Cheng and Oleg Chernoguz and Olivia Hart and Omkar Salpekar and Ozlem Kalinli and Parkin Kent and Parth Parekh and Paul Saab and Pavan Balaji and Pedro Rittner and Philip Bontrager and Pierre Roux and Piotr Dollar and Polina Zvyagina and Prashant Ratanchandani and Pritish Yuvraj and Qian Liang and Rachad Alao and Rachel Rodriguez and Rafi Ayub and Raghotham Murthy and Raghu Nayani and Rahul Mitra and Raymond Li and Rebekkah Hogan and Robin Battey and Rocky Wang and Rohan Maheswari and Russ Howes and Ruty Rinott and Sai Jayesh Bondu and Samyak Datta and Sara Chugh and Sara Hunt and Sargun Dhillon and Sasha Sidorov and Satadru Pan and Saurabh Verma and Seiji Yamamoto and Sharadh Ramaswamy and Shaun Lindsay and Shaun Lindsay and Sheng Feng and Shenghao Lin and Shengxin Cindy Zha and Shiva Shankar and Shuqiang Zhang and Shuqiang Zhang and Sinong Wang and Sneha Agarwal and Soji Sajuyigbe and Soumith Chintala and Stephanie Max and Stephen Chen and Steve Kehoe and Steve Satterfield and Sudarshan Govindaprasad and Sumit Gupta and Sungmin Cho and Sunny Virk and Suraj Subramanian and Sy Choudhury and Sydney Goldman and Tal Remez and Tamar Glaser and Tamara Best and Thilo Kohler and Thomas Robinson and Tianhe Li and Tianjun Zhang and Tim Matthews and Timothy Chou and Tzook Shaked and Varun Vontimitta and Victoria Ajayi and Victoria Montanez and Vijai Mohan and Vinay Satish Kumar and Vishal Mangla and Vítor Albiero and Vlad Ionescu and Vlad Poenaru and Vlad Tiberiu Mihailescu and Vladimir Ivanov and Wei Li and Wenchen Wang and Wenwen Jiang and Wes Bouaziz and Will Constable and Xiaocheng Tang and Xiaofang Wang and Xiaojian Wu and Xiaolan Wang and Xide Xia and Xilun Wu and Xinbo Gao and Yanjun Chen and Ye Hu and Ye Jia and Ye Qi and Yenda Li and Yilin Zhang and Ying Zhang and Yossi Adi and Youngjin Nam and Yu and Wang and Yuchen Hao and Yundi Qian and Yuzi He and Zach Rait and Zachary DeVito and Zef Rosnbrick and Zhaoduo Wen and Zhenyu Yang and Zhiwei Zhao},
  year         = {2024},
  journal      = {arXiv preprint arXiv:2407.21783},
}

@article{shetty2024wet,
  title={WET: Overcoming Paraphrasing Vulnerabilities in Embeddings-as-a-Service with Linear Transformation Watermarks},
  author={Shetty, Anudeex and Xu, Qiongkai and Lau, Jey Han},
  journal={arXiv preprint arXiv:2409.04459},
  year={2024}
}

@inproceedings{baumgartner2020pushshift,
  title={The pushshift reddit dataset},
  author={Baumgartner, Jason and Zannettou, Savvas and Keegan, Brian and Squire, Megan and Blackburn, Jeremy},
  booktitle={Proceedings of the international AAAI conference on web and social media},
  volume={14},
  pages={830--839},
  year={2020}
}

@article{laurenccon2022bigscience,
  title={The bigscience roots corpus: A 1.6 tb composite multilingual dataset},
  author={Hugo Laurençon and Lucile Saulnier and Thomas Wang and Christopher Akiki and Albert Villanova del Moral and Teven Le Scao and Leandro Von Werra and Chenghao Mou and Eduardo González Ponferrada and Huu Nguyen and Jörg Frohberg and Mario Šaško and Quentin Lhoest and Angelina McMillan-Major and Gerard Dupont and Stella Biderman and Anna Rogers and Loubna Ben allal and Francesco De Toni and Giada Pistilli and Olivier Nguyen and Somaieh Nikpoor and Maraim Masoud and Pierre Colombo and Javier de la Rosa and Paulo Villegas and Tristan Thrush and Shayne Longpre and Sebastian Nagel and Leon Weber and Manuel Muñoz and Jian Zhu and Daniel Van Strien and Zaid Alyafeai and Khalid Almubarak and Minh Chien Vu and Itziar Gonzalez-Dios and Aitor Soroa and Kyle Lo and Manan Dey and Pedro Ortiz Suarez and Aaron Gokaslan and Shamik Bose and David Adelani and Long Phan and Hieu Tran and Ian Yu and Suhas Pai and Jenny Chim and Violette Lepercq and Suzana Ilic and Margaret Mitchell and Sasha Alexandra Luccioni and Yacine Jernite},
  journal={Advances in Neural Information Processing Systems},
  volume={35},
  pages={31809--31826},
  year={2022}
}

@article{raffel2020exploring,
  title={Exploring the limits of transfer learning with a unified text-to-text transformer},
  author={Raffel, Colin and Shazeer, Noam and Roberts, Adam and Lee, Katherine and Narang, Sharan and Matena, Michael and Zhou, Yanqi and Li, Wei and Liu, Peter},
  journal={Journal of machine learning research},
  volume={21},
  number={140},
  pages={1--67},
  year={2020}
}

\end{document}